%% file: ClassicThesis.tex
\providecommand{\main}{.}
\RequirePackage{silence} % :-\
\RequirePackage{fix-cm} % fix some latex issues see: http://texdoc.net/texmf-dist/doc/latex/base/fixltx2e.pdf
\documentclass[ twoside,openright,titlepage,numbers=noenddot,headinclude,%1headlines,% letterpaper a4paper
                footinclude=true,cleardoublepage=empty,abstractoff, % <--- obsolete, remove (todo)
                BCOR=5mm,paper=letter, fontsize=11pt,%11pt,a4paper,%
                ngerman,american%
                ]{scrreprt}

\input{\main/classicthesis-config}
\usepackage{subfiles}
\usepackage{pdfpages}
\usepackage{biblatex}
\makeatletter
    \newrobustcmd*{\nobibliography}{%
      \@ifnextchar[%]
        {\blx@nobibliography}
        {\blx@nobibliography[]}}
    \def\blx@nobibliography[#1]{}
    \appto{\skip@preamble}{\let\printbibliography\nobibliography}
\makeatother

\begin{document}
\frenchspacing
\raggedbottom
\selectlanguage{american} % american ngerman
%\renewcommand*{\bibname}{new name}
%\setbibpreamble{}
\pagenumbering{roman}
\pagestyle{plain}
%********************************************************************
% Frontmatter
%*******************************************************
\include{FrontBackmatter/DirtyTitlepage}

\include{FrontBackmatter/Titlepage}
\include{FrontBackmatter/Titleback}
%\cleardoublepage\subfile{FrontBackmatter/Declaration}
\includepdf{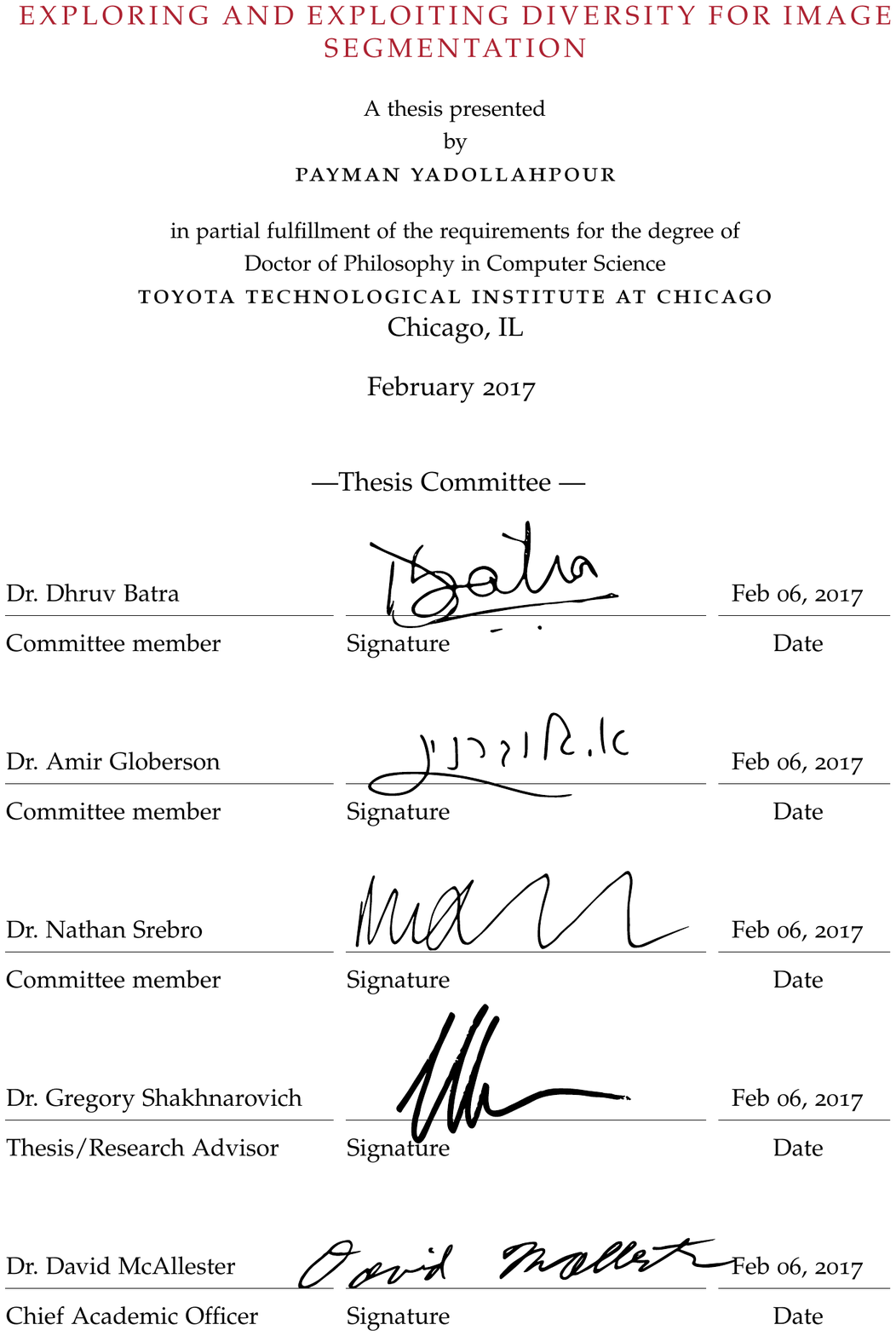}
\cleardoublepage\include{FrontBackmatter/Dedication}
\cleardoublepage\subfile{FrontBackmatter/Abstract}
\cleardoublepage\include{FrontBackmatter/Publications}
\cleardoublepage\include{FrontBackmatter/Acknowledgments}
\pagestyle{scrheadings}
\cleardoublepage\include{FrontBackmatter/Contents}
%********************************************************************
% Mainmatter
%*******************************************************
\cleardoublepage\pagenumbering{arabic}
%\setcounter{page}{90}
% use \cleardoublepage here to avoid problems with pdfbookmark
\cleardoublepage
\subfile{Chapters/Chapter01}

\subfile{Chapters/Chapter02}

\subfile{Chapters/Chapter03}

\subfile{Chapters/Chapter04}

\subfile{Chapters/Chapter05}

\subfile{Chapters/Chapter06}
% ********************************************************************
% Backmatter
%*******************************************************

\printbibliography[heading=bibintoc]
\appendix
\cleardoublepage
\part{Appendix}
\input{Chapters/Chapter04A}
\input{Chapters/Chapter05A}
\end{document}

%% file: classicthesis-config.tex
\PassOptionsToPackage{utf8}{inputenc}
    \usepackage{inputenc}

\PassOptionsToPackage{eulerchapternumbers,listings,%drafting,%
                     pdfspacing,floatperchapter,%linedheaders,%
                     subfig,beramono,eulermath,parts,dottedtoc}{classicthesis}
\newcommand{\myTitle}{Exploring and Exploiting Diversity for Image Segmentation\xspace}

\newcommand{\myName}{Payman Yadollahpour\xspace}

\newcommand{\myFaculty}{Advisor: Gregory Shakhnarovich\xspace}

\newcommand{\myUni}{Toyota Technological Institute\xspace}
\newcommand{\myLocation}{Chicago, IL\xspace}
\newcommand{\myTime}{February 2017\xspace}
\newcommand{\myVersion}{version 4.3c\xspace}

\newcounter{dummy} % necessary for correct hyperlinks (to index, bib, etc.)
\providecommand{\mLyX}{L\kern-.1667em\lower.25em\hbox{Y}\kern-.125emX\@}
\newcommand{\ie}{i.\,e.}

\newcommand{\eg}{e.\,g.}

\PassOptionsToPackage{american}{babel}   % change this to your language(s), main language last
    \usepackage{babel}

\usepackage{csquotes}
\PassOptionsToPackage{%
    backend=bibtex8,bibencoding=ascii,%
    language=auto,%
    style=numeric-comp,%
    sorting=nyt, % name, year, title
    maxbibnames=10, % default: 3, et al.
    natbib=true % natbib compatibility mode (\citep and \citet still work)
}{biblatex}

\PassOptionsToPackage{fleqn}{amsmath}       % math environments and more by the AMS
    \usepackage{amsmath}

\PassOptionsToPackage{T1}{fontenc} % T2A for cyrillics
    \usepackage{fontenc}
\usepackage{textcomp} % fix warning with missing font shapes
\usepackage{scrhack} % fix warnings when using KOMA with listings package
\usepackage{xspace} % to get the spacing after macros right
\usepackage{mparhack} % get marginpar right
\PassOptionsToPackage{printonlyused,smaller}{acronym}
    \usepackage{acronym} % nice macros for handling all acronyms in the thesis
\usepackage{tabularx} % better tables
\usepackage{caption}
\usepackage{subfig}
\usepackage{listings}
\PassOptionsToPackage{pdftex,hyperfootnotes=false,pdfpagelabels}{hyperref}
    \usepackage{hyperref}  % backref linktocpage pagebackref
\PassOptionsToPackage{pdftex}{graphicx}
    \usepackage{graphicx}

\hypersetup{%
    colorlinks=true, linktocpage=true, pdfstartpage=3, pdfstartview=FitV,%
    breaklinks=true, pdfpagemode=UseNone, pageanchor=true, pdfpagemode=UseOutlines,%
    plainpages=false, bookmarksnumbered, bookmarksopen=true, bookmarksopenlevel=1,%
    hypertexnames=true, pdfhighlight=/O,%nesting=true,%frenchlinks,%
    urlcolor=webbrown, linkcolor=RoyalBlue, citecolor=webgreen, %pagecolor=RoyalBlue,%
    pdftitle={\myTitle},%
    pdfauthor={\textcopyright\ \myName, \myUni, \myFaculty},%
    pdfsubject={},%
    pdfkeywords={},%
    pdfcreator={pdfLaTeX},%
    pdfproducer={LaTeX with hyperref and classicthesis}%
}

\makeatletter
\@ifpackageloaded{babel}%
    {%
       \addto\extrasamerican{%
                }%
    }{\relax}
\makeatother

\usepackage{classicthesis}
\usepackage[left=78pt, right=118pt, top=88pt,bottom=84pt]{geometry}
\usepackage{floatrow}
\usepackage{\main/mysymbols}
\usepackage{algpseudocode}
\usepackage{algorithm}
\usepackage{pgfplots}
\pgfplotsset{compat=newest}
\usetikzlibrary{plotmarks}
\usetikzlibrary{arrows.meta}
\usepgfplotslibrary{patchplots}
\usetikzlibrary{decorations.pathmorphing}
\usetikzlibrary{intersections}
\usetikzlibrary{positioning}
\usetikzlibrary{calc}
\usetikzlibrary{fit}
\usepackage[skins]{tcolorbox}
\usepackage{grffile}

\usepackage{mathtools}
\usepackage[mode=buildnew]{standalone}

\usepackage{amsthm}
\newtheorem{theorem}{Theorem}[section]
\newtheorem{prop}{Proposition}[section]
\usepackage{cleveref}
\crefname{section}{§}{§§}
\Crefname{section}{§}{§§}
\usepackage{enumerate}
\usepackage[super]{nth}
\usepackage{import}
\usepackage{sidenotes}
\usepackage{fourier-orns}

%% file: FrontBackmatter/DirtyTitlepage.tex
%*******************************************************
% Little Dirty Titlepage
%*******************************************************
\thispagestyle{empty}
%\pdfbookmark[1]{Titel}{title}
%*******************************************************
\begin{center}
    \spacedlowsmallcaps{\myName} \\ \medskip

    \begingroup
        \color{Maroon}\spacedallcaps{\myTitle}
    \endgroup
\end{center}

%% file: FrontBackmatter/Titlepage.tex
\begin{titlepage}
    \begin{addmargin}[-1cm]{-3cm}
    \begin{center}
        \large

        \hfill

        \vfill

        \begingroup
            \color{Maroon}\spacedallcaps{\myTitle} \\ \bigskip
        \endgroup
        \vfill
        {\small by}\\
        \vfill
        \spacedlowsmallcaps{\myName}

        \vfill

        \includegraphics[width=4cm]{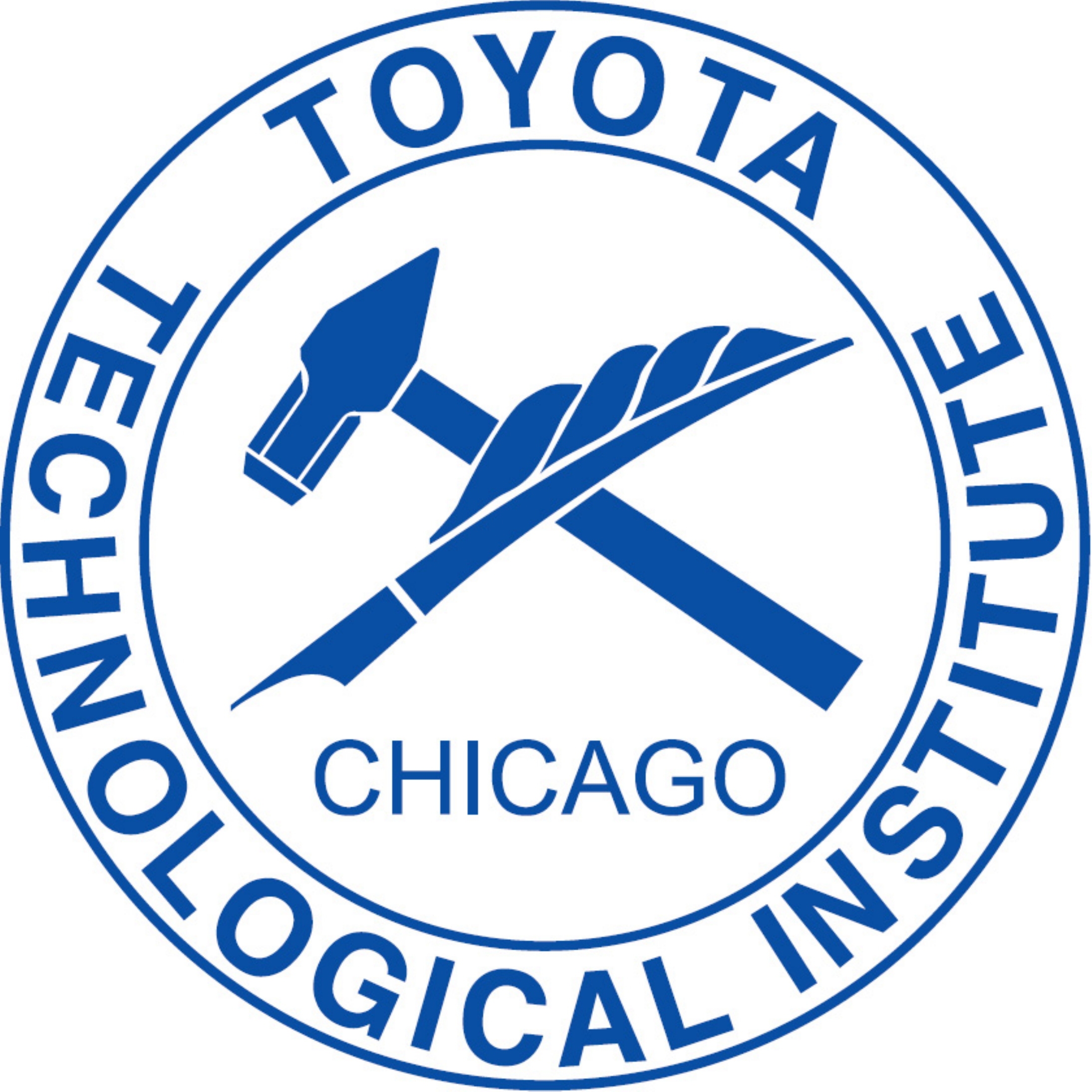} \\ \medskip
        \vfill
    A thesis submitted in partial fulfillment of the requirements for the degree of\\
    \vfill
    Doctor of Philosophy in Computer Science \\
    \vfill
        at the \\
        \vfill
        TOYOTA TECHNOLOGICAL INSTITUTE AT CHICAGO \\
        \myLocation\\
        \bigskip
        \myTime%\ -- \myVersion
        \vspace{2cm}\\
        Thesis committee:\\
        \medskip
        Dr. Gregory Shakhnarovich (Thesis Adivsor) \\
        Dr. Dhruv Batra \\
        Dr. Amir Globerson \\
        Dr. David McAllester \\
        Dr. Nathan Srebro \\

        \vfill

    \end{center}
  \end{addmargin}
\end{titlepage}

%% file: FrontBackmatter/Titleback.tex
\thispagestyle{empty}

\hfill

\vfill

\noindent\myName: \textit{\myTitle,} %\mySubtitle, %\myDegree,
\textcopyright\ \myTime

%% file: FrontBackmatter/Dedication.tex
%*******************************************************
% Dedication
%*******************************************************
\thispagestyle{empty}
%\phantomsection
\refstepcounter{dummy}
\pdfbookmark[1]{Dedication}{Dedication}

\vspace*{3cm}

\begin{center}
    Dedicated to my parents.
\end{center}

%% file: FrontBackmatter/Abstract.tex
\pdfbookmark[1]{Abstract}{Abstract}
\begingroup
\let\clearpage\relax
\let\cleardoublepage\relax
\let\cleardoublepage\relax

\chapter*{Abstract}
Semantic image segmentation is an important computer vision task
that is difficult because it consists of both recognition and segmentation. It is important because it subsumes important aspects of scene understanding such as image classification and object localization. The task is often cast
as a structured output problem on an exponentially large output-space, which is
typically modeled
by a discrete probabilistic model. The best segmentation
is found by inferring the Maximum a-Posteriori (MAP) solution over the output
distribution defined by the model. Due to limitations in
optimization, the model cannot be arbitrarily complex. This leads
to a trade-off: devise a more accurate model that incorporates rich high-order interactions
between image elements at the cost of inaccurate and possibly intractable
optimization OR leverage a tractable model which produces less accurate MAP
solutions but may contain high quality solutions as other modes of its output
distribution.

This thesis investigates the latter and presents a two
stage approach to semantic segmentation akin to cascade models and proposal
generation works. In the first stage a tractable segmentation
model outputs a set of high probability segmentations from the underlying
distribution that are not just minor perturbations of each other. Critically the
output of this stage is a diverse set of plausible solutions and not just a
single one. The first stage reduces the exponential space of solutions to just a
handful of segmentations. In the second
stage, a discriminatively trained re-ranking model selects the best segmentation
from this set. The re-ranking stage can use much more complex features than what
could be tractably used in the segmentation model, allowing a better
exploration of the solution space than possible by simply producing the most
probable solution from the segmentation model. The formulation of the
first stage is agnostic to the
underlying segmentation model (e.g. CRF, CNN, etc.) and optimization algorithm, which makes it applicable
to a wide range of models and inference methods.

Evaluation of the approach on a number of semantic image segmentation
benchmark datasets highlight its superiority over inferring the MAP solution.

\vfill

\endgroup

\vfill

%% file: FrontBackmatter/Publications.tex
%*******************************************************
% Publications
%*******************************************************
\pdfbookmark[1]{Publications}{publications}
\chapter*{Publications}
The ideas and figures in this thesis have appeared previously in the following publications:

\begin{refsection}[ownpubs]
    \small
    \nocite{*} % is local to to the enclosing refsection
    \printbibliography[heading=none]
\end{refsection}

%% file: FrontBackmatter/Acknowledgments.tex
%*******************************************************
% Acknowledgments
%*******************************************************
\pdfbookmark[1]{Acknowledgments}{acknowledgments}

\bigskip

\begingroup
\let\clearpage\relax
\let\cleardoublepage\relax
\let\cleardoublepage\relax
\chapter*{Acknowledgments}
My deepest gratitude goes to my research advisor Dr. Gregory Shakhnarovich for continual guidance and stewardship of my graduate career. He has been a major source of insight and has significantly contributed to my understanding of Computer Vision and Machine Learning disciplines. I would also like to thank Dr. Dhruv Batra for his research collaboration, and much of this thesis owes to the joint collaboration I had with him and Greg. A big thanks to Dr. Ayan Chakrabarti for the many insightful discussions on my thesis.

\bigskip

I would like to mention my appreciation to the entire faculty at Toyota Technological Institute at Chicago for their insistence on research excellence and for making TTIC a leading graduate research institution.

\bigskip

I'd like to give special thanks to my fellow students at Toyota Technological Institute at Chicago, for making the many years enjoyable and the many fruitful discussions. I'd like to especially thank Avleen Bijral, Andrew Cotter, Heejin Choi, Somaye Hashemifar, Taehwan Kim, Gustav Larsson, Mohammadreza Mostajabi, Jian Peng, Karthik Sridharan, Siqi Sun, Hao Tang, Behnam Tavakoli, Shubhendu Trivedi, Zhiyong Wang, and Feng Zhao. I also owe gratitude to Steven Basart, Falcon Dai, Suriya Gunasekar, Nicholas Kolkin, and Mohammadreza Mostajabi, for giving me very useful feedback on my thesis draft.

\bigskip

Lastly, I would like to thank my family for being a source of support and for their patience these many years.

\endgroup

%% file: FrontBackmatter/Contents.tex
%*******************************************************
% Table of Contents
%*******************************************************
%\phantomsection
\refstepcounter{dummy}
\pdfbookmark[1]{\contentsname}{tableofcontents}
\setcounter{tocdepth}{2} % <-- 2 includes up to subsections in the ToC
\setcounter{secnumdepth}{3} % <-- 3 numbers up to subsubsections
\manualmark
\markboth{\spacedlowsmallcaps{\contentsname}}{\spacedlowsmallcaps{\contentsname}}
\tableofcontents
\automark[section]{chapter}
\renewcommand{\chaptermark}[1]{\markboth{\spacedlowsmallcaps{#1}}{\spacedlowsmallcaps{#1}}}
\renewcommand{\sectionmark}[1]{\markright{\thesection\enspace\spacedlowsmallcaps{#1}}}
%*******************************************************
% List of Figures and of the Tables
%*******************************************************
\clearpage

\begingroup
    \let\clearpage\relax
    \let\cleardoublepage\relax
    \let\cleardoublepage\relax
    %*******************************************************
    % List of Figures
    %*******************************************************
    %\phantomsection
    \refstepcounter{dummy}
    %\addcontentsline{toc}{chapter}{\listfigurename}
    \pdfbookmark[1]{\listfigurename}{lof}
    \listoffigures

    \vspace{8ex}

    %*******************************************************
    % List of Tables
    %*******************************************************
    %\phantomsection
    \refstepcounter{dummy}
    %\addcontentsline{toc}{chapter}{\listtablename}
    \pdfbookmark[1]{\listtablename}{lot}
    \listoftables

    \vspace{8ex}
\endgroup

%% file: Chapters/Chapter01.tex
%************************************************
\chapter{Introduction}\label{ch:introduction}
%************************************************
\subfile{\main/Sections/Segmentation}

\subfile{\main/Sections/MAP}

%% file: Sections/Segmentation.tex
The task of automatically labeling every pixel in an image with the category label of the object it covers is an important computer vision problem. Known as full image labelling or semantic segmentation -- because it partitions the image into semantically coherent regions -- it is one valuable proxy for measuring how well a system can reason about what is being depicted in an image. It subsumes important aspects of scene understanding such as image classification and object localization. While its importance as an end task is debatable it is a more refined proxy for measuring a system's discriminative capability on a finite set of object classes than image classification or object detection. This is because the prediction must be made over local regions in the image as opposed to a global prediction over the entire image or simple bounding boxes over objects.

Image segmentation is typically modelled either probabilistically, via
Conditional or Markov Random Fields (CRFs/MRFs) or using discriminative feed
forward approaches. Feed forward approaches include cascade type systems that
first predict region proposals and then predict their most likely labels and
heuristically paste the labelled regions into the image. More recently, neural
network models for segmentation have been proposed, including Convolutional
Neural Networks (CNNs)~\cite{mostajabi2015,hariharan2015,} and Recursive Neural
Networks~\cite{he2016}, which achieve
state-of-the-art accuracy on many difficult image segmentation benchmarks.

Semantic segmentation is a task that has a structured output space; the variables of interest (namely image regions such as pixels or superpixels) are not independent of each other, but rather must be predicted jointly. Given this fact and that the output space of possible labellings of the variables is exponential in size introduces certain limitations on how we can jointly model, train, and infer the variables. For instance, in order to be able to train and run inference, CRF or MRF models often make simplifying independence assumptions over the variables, either by limiting clique sizes, or approximating the partition function. The different sources of error  -- approximation error due to a poor choice of model class, optimization error due to limitations on optimizing over the variables of interest, and estimation error due to a finite training set -- all contribute to the quality of the final predicted segmentation. Because of all these sources of error the predicted probability distribution over the output labeling might be significantly different from the true distribution. Thus the most probable label returned under the model distribution might not be the most probable under the true distribution.

One way to alleviate this is to build more complex models that can capture the complex interactions of the variables, at the cost of making learning and inference (i.e. optimization) more expensive or possibly intractable. In this thesis we explore an alternate approach. Instead of increasing model complexity at the cost of optimization complexity, we propose a framework whereby we can find a small set of highly probable and yet diverse segmentations (``modes'') under the model. By virtue of the fact that this ``mode'' finding algorithm has exponentially reduced the space of segmentations we need to consider, we can evaluate each of them using arbitrarily complex features that can take into account dependencies between variables that would be intractable to capture in the original model. Because of the exponential space of possible segmentations, producing this ``handful'' of highly probable yet diverse segmentations is going to require an approach that is more nuanced than simply enumerating all possible solutions under the model.

We show that combining this mode finding algorithm with an automatic approach to selecting the best segmentation from this smaller set leads to a framework that produces state-of-the-art results on challenging semantic segmentation datasets. It is also general enough to be applicable to a wide variety of problems in vision and elsewhere.

\paragraph{Thesis Outline}{This chapter presents a review of the segmentation
problem, and outlines some common approaches to it, citing related literature -- specifically algorithms for \emph{bottom-up} and \emph{top-down}
segmentation. The chapter closes with presentation of the well known MAP inference problem
and its integer programming formulation which will become relevant in the
formulation of the \divmbest problem. Chapter~\ref{ch:divmbest} reviews a number of
approaches for inferring multiple solutions from a discrete probabilistic model, instead of
just the MAP solution and explains why they are not adequate for improving image segmentation.  Chapter~\ref{ch:divmbest} concludes with presentation
of an alternate approach called the
\divmbest problem --- which leverages existing segmentation models, and algorithms used to do inference over
them, in order to produce sets
of high-quality segmentations that are diverse.
Chapter~\ref{ch:divrank}
presents
a discriminatively trained re-ranking model that selects the best segmentation
from this set. Evaluation of the \divmbest
and \divrank methods on a number of semantic segmentation tasks is presented in
chapter~\ref{ch:divmbestexp} and chapter~\ref{ch:divrankexp} respectively.
}

\section{Segmentation}\label{sec:segmentation}
The task of partitioning all or some of the pixels in an
image into coherent regions is known as image segmentation. When the regions
take on semantic labels, the partitioning is known as semantic segmentation ---
a major topic of this thesis.
The non-semenatic segmentation problem is ill-posed because what we
mean by a segment is not clearly defined --- for example a segment might
belong to a single or multiple connected components
throughout the image. A primary goal
of segmentation is to have pixels within segments share a consistent property
or feature. This is another reason why segmentation is ill-posed because consistent property is problem specific. For example, a common property
we find in the output of most segmentation algorithms is that pixels that fall
within the same segment are all within a local spatial neighborhood in the
image. This is property is not necessarily required however. Other features that do not require it such
as color and texture statistics of the regions around a pixel~\cite{endres2010,dalal2005}, pixel
depth information~\cite{gupta2014}, image contour strength~\cite{arbelaez2011}, can be considered depending on
the segmentation task. A third reason, specific to non-semantic segmentation, is that we do not explicitly associate meaning with the individual segments. The segments could correspond to low-level image cues like regions of
constant color or texture or could be associated with semantic meanings such as
physical objects or parts of objects. Given an image if you were to ask a set of
people to segment the image we would end up getting multiple interpretations of
what is a good segmentation of that image.

Semantic segmentation, however, is a much better posed problem since the output label space is well defined (e.g. object classes). That is to say, one common semantic segmentation task that we care to define is labelling every image region (e.g. pixel or superpixel) with the approriate object class that it is a part of in the image.

As the above examples of segmentation features illustrate, one axis along which
we can define different segmentation algorithms is based on the features used to
capture local information relative to pixels in the image. If the segmentation
task is to partition the full image into spatially coherent segments where
pixels within a segment have similar color, spatial, depth, or boundary
statistics such as curvature --- this is known as low-level, or bottom-up, image
segmentation. On the other hand, in the case of semantic segmentation (also
known as semantic image parsing), the pixels corresponding to a segment share
similar semantic properties (such as a \emph{pixel part of sky in image}).
Additionally, if the semantic categories are limited to foreground objects and
background clutter the partitioning is referred to as figure-ground, or simply,
foreground segmentation. Segmentation can also be with respect to 3D cues of the
objects, such as surface orientation or material properties~\cite{hoiem2005}.

Typically bottom-up image segmentation algorithms partition the image into
disjoint segments. The union of segments is equal to the entire image; in
other words the segmentation covers the entire image. In semantic segmentation
whether the partitioning covers the entire image depends on the semantic
categories considered and how the algorithm partitions the image. For instance,
the algorithm could assign the area in the image not covered by the segments
explicitly labeled with semantic categories to a catch-all category such as
\emph{background}, \emph{don't care}, or \emph{unknown} label. On the other hand
the algorithm might explicitly try to predict ambiguous segments in the image
as a specific category onto itself such as \emph{stuff}, in which case the
partitioning might not cover the entire image.

The size and shape statistics of the segment that we get as the output from
segmentation algorithms also differs depending on the image information used for
segmentation, as well as the algorithm details itself. For example in semantic
segmentation the desired segment shapes and sizes are governed by shapes and
sizes of the objects depicted in the images. On the other hand the output of
low-level segmentation algorithms such as SLIC~\cite{achanta2010} produce
\emph{over-segmentations} of the image, where segments exhibit nearly uniform shape and
size with small spatial support. Usually, low-level segmentations that over-segment an image are a first step towards some other more complex downstream
task, such as semantic segmentation. These segments provide convenient and
predictable objects for downstream processing due to there consistent shape and
size. That is not to say that the output of all low-level algorithms exhibit
this regularity in size and shape. For example hierarchical image segmentation
approaches~\cite{van2011}, which do a bottom-up grouping of image regions, and segmentation based image contour detection~\cite{arbelaez2011}
produce low level image segmentation results where segments can have a variety
of shapes and sizes.

It is common to refer to segments that are the result of low-level image segmentation
algorithms as \emph{superpixels}. Analogously for 3-dimensional segmentation the
3D regions are referred to as \emph{supervoxels}. Generally what is refered to as
a superpixel is the result of an over segmentation of an image, and initially
there is no semantic meaning associated with the superpixel. Most superpixel
algorithms rely on low-level image evidence such as color, intensity, contour,
and texture information. Many but not all of these algorithms produce
superpixels with regular shape and size, that adhere to image contours and some
additionally, roughly, \emph{snap} to a regular grid pattern over the image.
Image contour can be further separated into \emph{internal} and \emph{external}
edges. By internal edges we mean contours that appear due to a marked difference
in intensity, color, or texture between pixels that fall on the \emph{same}
object surface in the image, whereas external edges are those delineating
locations where one object occludes another or of self occlusions. Many of the
low-level algorithms produce superpixels that align to both types of contours.
Indeed a single superpixel boundary can align with one or more internal and
external edges. This is in contrast to the desired output from semantic
segmentation algorithms where the segment boundaries should align to
object-to-object or object-to-background boundaries.

As mentioned earlier it is common for semantic segmentation approaches to rely
on superpixels, generated using low-level segmentation algorithms, as the basic
primitives over which to construct larger segments. This isn't always the case
however -- in fact there are semantic segmentation methods~\cite{carreira2010,carreira2012o2p} that use
complete or partially complete object \emph{proposals} (\ie~segments) as their
basic primitives, and these segments do tend to align better to external edges
in the image. The figure-ground models used to generate the proposals are typically learned by maximizing an objectness score, thereby generating segments that better correspond to objects of interest (i.e. \emph{figure}) than to everything else (i.e. \emph{background}).

As previously mentioned the segments that semantic segmentation methods
generate can also span multiple connected components (in graph parlance). For example if
object \emph{A} is partially occluded by object \emph{B} visually splitting
\emph{A} into two parts in the image, and the two objects are of different
categories, then the correct segmentation component associated with object
\emph{A} is composed of two separate connected components. On the other hand if
both \emph{A} and \emph{B} have the same object category then the correct
semantic segmentation output would be a single component tightly covering
\emph{A} and \emph{B}. Furthermore, if the task is \emph{instance level}
semantic segmentation, and \emph{A} and \emph{B} appear adjacent to each other
in the image but are of the same category label, then the correct output should
be two separate connected components each tightly covering one of the objects
and each assigned a unique instance label.

For downstream computer vision tasks low-level image segmentation, and requisite
superpixel output, provides a nice way to improve computational efficiency.
Compared to working with pixels which number from tens of thousands to millions
in typical images, superpixels tend to number in the dozens or hundreds. That's
a few orders of magnitude reduction in the number of variables that need to be
considered by a semantic segmentation algorithm. Since superpixels are the
results of algorithms designed to align closely with significant image contours,
they provide the added benefit of combining to produce segments that
also align well to significant contours along their boundary. A third reason for
using superpixels instead of pixels is that, in contrast, superpixels provide a
boundary aligned spatial support on which to compute image features. The
segment on which we should compute features for a pixel is less well defined,
and usually local features~\cite{dalal2005,lowe1999,ojala2002} are computed on a
spatial neighborhood around the pixel that is grid aligned. In conjunction, the
fact that superpixels can span many pixels and cover a large pixel neighborhood,
including neighbors that are more than one pixel apart, provides useful long-range dependencies between areas in the image. As we will elaborate on later in
this chapter, these long-range dependencies allow short-range dependency (\ie
dependency between adjacent elements in a neighborhood) graph
based semantic segmentation algorithms to incorporate implicit long-range
information for local prediction of superpixel labels --- producing
segmentations that are more consistent with the image --- at the same time
bypassing the complexities involved with incorporating explicit long-range edges
in the graph.

So far we have talked about a few different segmentation tasks: low-level,
semantic, instance level, and figure-ground segmentation. This is by no means an
exhaustive list of segmentation tasks. Some other common segmentation tasks that
we'll mention here include interactive segmentation, cosegmentation,
object-proposals, and holistic scene segmentation.

Interactive segmentation is an approach where the user is in the loop. In this
task the goal is to have a system that, given an image, asks the user to input
exemplars for the types of regions that the user would like the system to
segment. The exemplars could be pixels, superpixels, or other regions, in the
image and the user interacts with the system via
scribbles~\cite{boykov2001}, bounding boxes~\cite{rother2004}, or polygons, etc on the
image indicating the regions corresponding to different categories they'd like
to segment. Given the user \emph{annotation} an initial segmentation of the
image (be it a multi-category, figure-ground, or low-level segmentation) is
produced by the system and offered to the user. Depending on the quality of the
segmentation the user has the option to refine or provide more annotations as
before and have the system refine the segmentation. This iterative process
continues until the user is happy with the segmentation at which point the
process terminates.

In cosegmentation~\cite{rother2006,batra2010,schnitman2006} the task, usually, is to jointly segment different instances
of the same object category that appear in a set of images. Alternately the
images could contain the same object instance under different views or
deformations. If the task is to segment frames in video sequences this approach
to segmentation has clear advantages because it leverages more information in
learning the segmentation model for object categories and instances.

A major advance in semantic segmentation came with the use of object
proposals~\cite{carreira2010,endres2010}. The idea here is to produce multiple object proposals for the
image. Each object proposal is either a figurie-ground segment or bounding box in
the image, and the proposals are allowed to overlap. The semantic segmentation
task shifts from labeling pixels/superpixels in the image to selecting a subset
of the top ranking object proposals and assigning them semantic labels. Using
object proposals makes the problem much simpler because the set of object
proposals is much smaller than the number of pixels/superpixels in the image.
Object proposals also provide much larger spatial support for computing features
useful in determining objectness likelihood of the underlying image region. It's
also more likely that one of the object proposals is a good candidate segment
for an object. The object proposals are usually generated using
class-independent methods. For example the object proposals can be bottom-up
segmentations computed over the image which are ranked according to an \emph{objectness} score that takes
into account cues like color, texture, location, saliency,
etc~\cite{endres2010}. Alternatively,
a bottom-up approach can be taken to produce multiple figure-ground masks using
a graph-based model initialized with different random seeds~\cite{carreira2010}. The masks
are then ranked according to class specific regressors trained to maximizes the
likelihood that the mask tightly covers the underlying object of that category.
One of the most successful approaches for building object proposals that is very
fast and gives high recall on objects present in the image is Selective
Search~\cite{van2011}. Here multiple hierarchical segmentations over superpixels are
computed. The object proposals consist of either segments within this hierarchy
or bounding boxes around them.

Finally, there are approaches~\cite{yao2012} that try to reason about
multiple tasks over the image in order to come up with a \emph{holistic}
interpretation of what is being depicted. The task, then, becomes to jointly
reason about both the segmentation of the image into semantically meaningful
segments while simultaneously predicting the scene classification and
detecting what objects are in the image along with their locations and extents.
Allowing for joint prediction of multiple tasks has the added benefit of
incorporating multiple compatibility measures. Each of these compatibility
features is an added source of rich information that the model can use in order
to improve the segmentation accuracy.

In the next section we'll dive a little deeper into some of the most popular
bottom-up and top-down segmentation methods and explain in more detail how they
work. We'll also describe the specific segmentation models we used in the
experiments of subsequent chapters.

\subsection{Methods}\label{sec:segmethods}
Segmentation has a long and rich history and we will not try to enumerate all
the different segmentation methods. Instead we'll highlight a few of the most
popular methods for both low-level and semantic segmentation.

One class of segmentation methods is based on algorithms that try to find the
maxima, or modes, of a data distribution given a discrete set of points. They
are clustering methods because they assign all points within a \emph{basin of
attraction} of a mode to the same cluster. Another nice property is that these
methods are non-parametric. That means that the space of data points can be
viewed as the empirical probability density function of the parameter the data
points represent. The modes of the density function will correspond to dense
regions, or clusters, in the data space. That's why these methods are also
referred to as hill-climbing or gradient based methods because they find the
maxima of the data distribution. This is nice because we don't need to known
apriori the number and shape of the clusters.

\subsubsection{Mean Shift} One of the most popular such methods is based on the
mean shift algorithm~\cite{comaniciu2002}.

In the mean shift algorithm the unknown data density is estimated using the
kernel density estimator,
\begin{flalign}
    \hat{f}(\bx) =\frac{1}{nh^d}\sum\limits_{i=1}^{n}K\left(\frac{\bx-\bx_i}{h}\right),
\end{flalign}
where $\bx$, $\bx_i$ are $\emph{d}$-dimensional data points, and
$K$ here is assumed to be a multivariate normal kernel with diagonal bandwidth
matrix $h^2\mathbf{I}$ for simplicity ($h$ is the bandwidth parameter), though any
radially symmetric kernel that satisfies some mild assumptions would suffice.
The modes of the density are locations where $\hat{\nabla}f(\bx)=0$. The density
gradient when assuming normal kernel $K$ is,
\begin{flalign}
    \hat{\nabla}f(\bx) = \frac{1}{nh^{d}}\sum\limits_{i=1}^nK\left(\frac{\bx-\bx_i}{h}\right) \
    \left(\frac{\sum_{i=1}^n \bx_i K\left(\frac{\bx-\bx_i}{h}\right)}{
\sum_{i=1}^n K\left(\frac{\bx-\bx_i}{h}\right)}- \bx \right).
\end{flalign}
The second term is the mean shift,
\begin{flalign}
    \label{eqn:meanshift} m(\bx) =
    \frac{\sum_{i=1}^n \bx_i K\left(\frac{\bx-\bx_i}{h}\right)}{ \sum_{i=1}^n
K\left(\frac{\bx-\bx_i}{h}\right)}- \bx ,
\end{flalign}
which is the difference between the weighted average of the points and $\bx$. It
can be shown~\cite{fukunaga1975,comaniciu2002,cheng1995} that the mean shift is proportional to,
\begin{flalign}
    m(\bx) \propto\frac{\hat{\nabla} f(\bx)}{\hat{f}(\bx)},
\end{flalign} or in other words the
mean shift points along the direction of steepest ascent of the empirical
density at point $\bx$. This is a nice property because it provides a natural
algorithm for mode finding:
\begin{itemize}
    \item Start at a data point $\bx$,
    \item Repeat the following steps till convergence (i.e.
    $m(\bx)\approx 0$):
        \begin{itemize}
            \item Compute the mean shift vector $m(\bx)$ (\ref{eqn:meanshift}),
            \item Update the location of the kernel window using $m(\bx)$,
        \end{itemize}
\end{itemize}
The fact that the mean shift
is normalized by the density makes the mean shift algorithm an adaptive gradient
ascent algorithm that takes large steps in areas of low density and takes
increasingly smaller steps as it approaches high density areas where the modes
are.

All points that converge to the same stationary point are within the \emph{basin
of attraction} of a mode. These points can all be considered as one cluster and
assigned the same cluster label.

The mean shift algorithm has been applied to the task of image
segmentation~\cite{comaniciu2002} where, normally, the $\bx_i$ --- $i=1,\dots,n$, where $n$
is the number of pixels in an image --- are $\emph{d}$-dimensional feature
vectors containing the pixel location and LAB (or LUV) color or intensity
information. Typically $\emph{d}$ is small because mean shift suffers from the
\emph{curse of dimensionality}. A higher dimensional space will be sparsely
populated with data points and the  density concentrated in a very small part of
the space making the kernel density estimator a poor estimate of the true
density. Another issue with mean shift is that the feature space is assumed to
be a Euclidean space, or some other space where an inner product or Riemannian
metric is defined, which might not generally hold. The mean shift algorithm is
also pretty slow with $\bigO{n^2}$ running time.

\subsubsection{Quick Shift} An alternative, simpler, strategy for mode seeking
is quick shift~\cite{vedaldi2008}. Whereas in mean shift we had an iterative algorithm and
had to compute the gradient, in quick shift we only need to take one step for
each data point and no gradients are needed. In quick shift each data point is
moved to the location of a neighboring point that increases the probability
density. We can write the probability density estimate at $\bx_j$ as,
\begin{flalign}
    \hat{f}(\bx_j) = \frac{c}{n}\sum\limits_{l=1}^n k\left(d(\bx_l,\bx_j)\right),
\end{flalign}
where $k$ is some radially symmetric
kernel function, $d(\bx_i,\bx_j)$ is a metric on $x$, and $c$ is a normalization
constant. For data point $\bx_i$ we assign it the data point $y_i$ such that,
\begin{flalign} y_i \leftarrow \argmin\limits_{j: \hat{f}(\bx_j) >
\hat{f}(\bx_i)} d(\bx_i,\bx_j), \end{flalign} guaranteeing that we move up the
hill toward a mode. Doing this procedure for every point $\bx_i$ connects all
the points into a tree, with edge weights set to $d(\bx_i,\bx_j)$. Cutting edges
with weight larger than some threshold $\emph{t}$ breaks the tree into subtrees
that cluster the points with the root nodes as possible modes of the empirical
distribution. Adjusting the threshold controls how much fragmentation of the
modes there is which affects the number of segments you get. The method is still
rather slow with $\bigO{\emph{d}n^2}$ complexity, where $\emph{d}$ is a small
constant.

\subsubsection{Watershed Transform} Another approach to segmentation is based on
the watershed transform. There are a number of different watershed transforms,
such as watershed by immersion or by topographical distance~\cite{roerdink2000}. The basic
idea is simple though, you can view an intensity or grey level image as a
landscape with \emph{catchment basins} or wells in the topography and
\emph{watersheds} where multiple basins meet. The general approach is to start
off by assigning local minima in the intensity or grey level image to distinct
basins. In the watershed by immersion approach (cf.~\cite{roerdink2000}) the basins are
recursively grown by iteratively increasing the level set and assigning
unlabeled pixels that have intensity value no greater than the level set value
to the catchment basin that is closest. If the pixel is equidistant to two or
more catchment basins then it is not assigned to any basin and is reconsidered
in the next iteration. The process continues until all level sets (\ie image
intensity values) have been considered, at which point all unlabeled pixels are
assigned as watershed (\ie boundary). The basins are the resultant segmentation
of the image.

The watershed transform by topographical distance approach
(cf.~\cite{roerdink2000}) assumes a
cost between neighboring pixels $\emph{p}$ and $\emph{q}$ that takes into
account the slope between $\emph{p}$ and $\emph{q}$. The topographical distance
along a path is defined as the sum of costs between neighboring pixels along the
path. The topographical distance between two points is then just the minimum
topographical distance of any path connecting them. A catchment basin around a
local minimum is then defined as all the pixels that are closer to that minimum
in terms of topographical distance than to any other local minimum in the image.
The watershed boundaries are the set difference of the image with the pixels in
all the catchment basins.

\subsubsection{Graph Based - Normalized Cuts} There are also graph based
approaches for low-level segmentation that let each pixel be a node in a graph,
with some edge connectivity between pixels, and partition the graph to produce a
segmentation of disjoint components. Of these there is a subtype of algorithms
that are based on spectral partitioning of the graph --- normalized cuts being
one such method. In the normalized cuts algorithm~\cite{shi2000} we assume a graph
$G=(V,E)$ where each vertex in $V$ is associated with unique pixel in the image
(vertices in $V$ cover the image) and $E$ contains edges between all pairs of
pixels. For each edge $(v_i,v_j)\in E$ we assign a weight $w_{ij}$ capturing how
likely it is that $v_i$ and $v_j$ belong to the same object in the image.
Usually $w_{ij}$ is a similarity measure between feature vectors computed at $i$
and $j$. A \emph{cut} of $G$ given two disjoint components $C_1$ and $C_2$ is
defined as,
\begin{flalign}
    \emph{cut}(C_1,C_2) = \sum\limits_{i\in C_1,j\in
C_2} w_{ij}.
\end{flalign}
The minimum cut of $G$, \ie the subset of $E$ that
minimizes the total edge weight crossing the cut, is the optimal partition of
the image into two components. Using the total edge weight crossing the cut is
not ideal for segmentation because it tends to favor partitioning small
components since the cut value grows as the number of edges across the
bipartition grows. To account for this in normalized cuts they use the
\emph{normalized cut} (cf.\cite{shi2000}),
\begin{flalign}\label{eq:ncut} \emph{ncut}(C_1,C_2)=
    \frac{\emph{cut}(C_1,C_2)}{\emph{assoc}(C_1,V)} +
    \frac{\emph{cut}(C_1,C_2)}{\emph{assoc}(C_2,V)},
\end{flalign} where
$\emph{assoc}(C_i,V) = \sum_{i\in C_i,j\in V} w_{ij}$ is the total weight of edges
from pixels in $C_i$ to all pixels in the graph. If $C_1$ is small,
$\emph{assoc}(C_1,V)$ will tend to also be small, increasing the cut value,
$\emph{ncut}(C_1,C_2)$. Consequently it prevents cuts that favor producing
small components.

It turns out that computing a bipartition of the graph into $C_1$ and $C_2$ that
minimizes~\ref{eq:ncut} amounts to solving the following generalized eigenvalue
problem \begin{flalign}\label{eqn:ncuteig} (\bD - \bW)\by=\lambda\bD\by,
\end{flalign} where $\bD$ is an $n\times n$ diagonal matrix with
$\bD_{ii}=\sum_j w_{ij}$ along the diagonal, and $\bW$ is an $n\times n$
symmetric matrix with $\bW_{ij}=w_{ij}$. It can be shown~\cite{shi2000} that the
eigenvector $\by$ corresponding to the second smallest eigenvalue
of~\ref{eqn:ncuteig} gives the assignment that partitions $G$ into components
$C_1$ and $C_2$ that minimize the normalized cut. The eigenvector $\by$ ideally
will have two discrete points $+1,-1$ indicating whether or not a pixel $i$ is
assigned to component $C_1$. But in order to solve \ref{eqn:ncuteig}, $\by$ is
relaxed to take on real values. Therefore the final assignment can be done by
either using $0$ as the threshold on the values of $\by$ --- where all elements
with value $>0$ are assigned to $C_1$ and $C_2$ otherwise --- or we can search
over different thresholds and pick the partitioning the minimizes the
$\emph{ncut}$.

This leads to a simple normalized cut algorithm for segmenting the graph called
\emph{two-way ncut}~\cite{shi2000},
\begin{itemize}
    \item Construct a fully connected graph $G=(V,E)$ over pixels in the image, with edge weights
    $w_{ij}$ measuring similarity between pairs of pixels.
    \item Solve the generalized eigenvalue problem $(\bD - \bW)\by=\lambda\bD\by$ for
    eigenvectors with smallest eigenvalues.
    \item Bipartition the graph using
    the eigenvector corresponding to the second largest eigenvalue. If $\by$
    contains more than two discrete values, search for the splitting point that
    gives the minimum $\emph{ncut}$ value.
    \item  For each component created
    after the bipartition we can decide whether to recursively apply the same
    procedure again to partition the component into two separate components
    based on the $\emph{ncut}$ value.
    \item Stop partitioning when the
    $\emph{ncut}$ value is below a certain threshold.
\end{itemize}
\begin{figure}[t!]
        \makebox[\textwidth][c]{
            \subfloat[][$k=5$]{\includegraphics{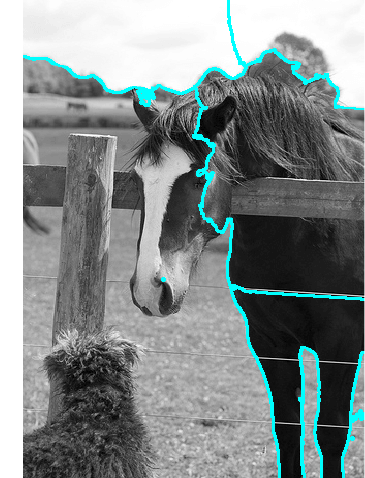}}\hspace{1pt}
            \subfloat[][$k=15$]{\includegraphics{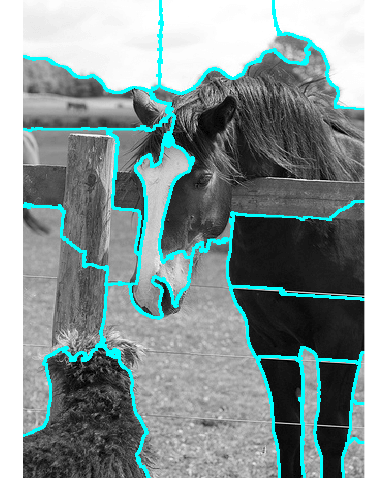}}\hspace{1pt}
            \subfloat[][$k=30$]{\includegraphics{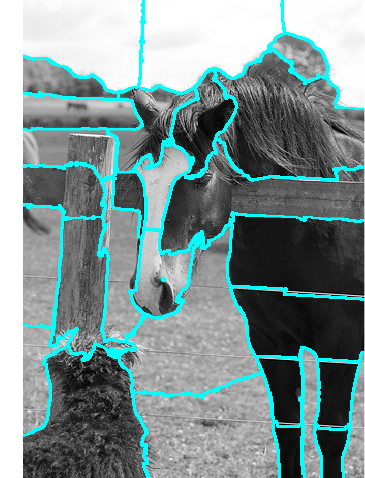}}\hspace{1pt}
            \subfloat[][$k=45$]{\includegraphics{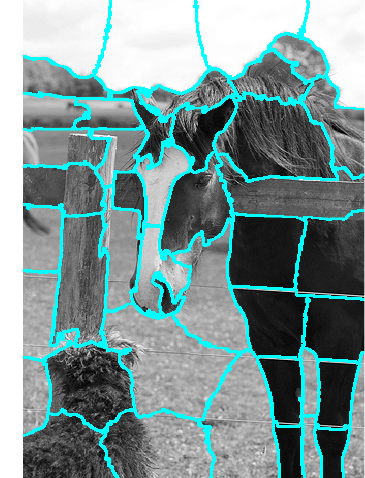}}
%\subfloat[][$k=5$]{\setlength\figurewidth{0.4\textwidth}
            %\setlength\figureheight{0.33\textwidth}\includestandalone{gfx/ncut/ncut5seg}}
            %\subfloat[][$k=30$]{\setlength\figurewidth{0.4\textwidth}
            %\setlength\figureheight{0.33\textwidth}\includestandalone{gfx/ncut/ncut30seg.tex}}
            %\subfloat[][$k=45$]{\setlength\figurewidth{0.4\textwidth}
%    \setlength\figureheight{0.33\textwidth}\includestandalone{gfx/ncut/ncut45seg.tex}}
} \caption{Bottom-up segmentation using normalized \emph{k-way cut} algorithm as
a function of the number of segments (top eigenvectors), $k$. Note that a
segment can consists of multiple disconnected components (\eg segment on nose of
horse).} \label{fig:ncut}
\end{figure}
The resulting segmentations have some nice properties. Because the cut is over a
fully connected graph the solutions take global image information in producing
the segmentation. The components in the segmentation also need not be connected
components in the graph --- a component can consist of pixels found in multiple
disjoint regions in the image. The normalized cut algorithm is relatively slow
with a running time complexity of $\bigO{n^{3/2}}$, where $n$ is the number of
pixels in the image.

Results of the normalized cuts algorithm are shown in figure~\ref{fig:ncut}.
These are base on the alternate \emph{$k$-way} cut algorithm
(cf.~\cite{shi2000}) that
produces a simultaneous segmentations into $k$ regions. In the $k$-way cut
approach they associate an $n$-dimensional vector with each pixel in the image
by stacking the top $n$ eigenvectors. An over-segmentation of $p\geq k$ segments
is produced by clustering the $n$-dimensional vectors using k-means. Next,
either a greedy merging strategy can be used or a global recursive cut is
performed. In the greedy merging approach regions are merged iteratively until
$k$ segments are left. In each iteration the two regions that minimize the
following $k$-way normalized cut are merged,
\begin{flalign}
Ncut_k=\frac{\emph{cut}(A_1,V-A_1)}{\emph{assoc}(A_1,V)} +
\frac{\emph{cut}(A_2,V-A_2)}{\emph{assoc}(A_2,V)} + \dots
+\frac{\emph{cut}(A_k,V-A_k)}{\emph{assoc}(A_k,V)}
\end{flalign} where the
$A_i$'s are the segments. Alternatively, a new graph can be constructed with the
$A_i$'s as nodes and edge weights, $w_{ij}$, corresponding to
$\emph{assoc}(A_i,A_j)$, capturing the total weight between pixels in $A_i$ and
$A_j$. Recursively bipartitioning this new graph --- by either solving the
eigensystem defined earlier for the $\emph{ncut}$ criterion or exhaustively
minimizing the \emph{ncut} criterion when $k$ is small results in the final
$k$-way partition. For further details of the approach, refer to Shi and
Malik~\cite{shi2000}.

\subsubsection{Graph Based - Felzenswalb \& Huttenlocher} A similar approach to
the normalized cuts method is that of Felzenswalb and Huttenlocher~\cite{felzenszwalb2004}. Here
again they assume a graph $G=(V,E)$ where for every pixel $p_i$ in the image
there is a corresponding node $v_i$ in $V$. The edge set $E$ is assumed to be
locally connected however, where the local neighborhood could be 4 or
8-connected neighborhood of a pixel or any other local neighborhood
connectivity. For each edge $(v_i,v_j)$ there is an associated weight $w_{ij}$,
similar to normalized cuts, that measures the dissimilarity between $p_i$ and
$p_j$. For example for an intensity image $I$ we could define
$w_{ij}=|I_{p_i} - I_{p_j}|$.

To determine whether two components should be connected they define a boundary
predicate,
\begin{flalign}\label{eqn:boundpred}
    D(C_1,C_2)=\left\{\begin{array}{ll} {\texttt true} & \mbox{if
            $\emph{Dif}(C_1,C_2) > \emph{MInt}(C_1)$} \\ {\texttt false} &
            \mbox{otherwise} \end{array}\right.
\end{flalign} where $\emph{Dif}(C_1,C_2)$ is defined as,
\begin{flalign} \emph{Dif}(C_1,C_2)
        = \min\limits_{v_i\in C_1, v_j\in C_2} w_{ij}
\end{flalign} and $\emph{MInt}(C_1,C_2)$ as,
\begin{flalign}
    \emph{MInt}(C_1,C_2) = \min(\emph{Int}(C_1)+\tau(C_1), \emph{Int}(C_2)+ \tau(C_2)),
\end{flalign}
where $\emph{Int}(C)$ is the value of the maximum weight edge
in the minimum spanning tree of component $C$ in graph $G$. Intuitively
$\emph{Dif}(C_1,C_2)$ captures the difference between two connected
components. For two connected components that don't have a connecting edge
the value is set to $\infty$. The value $\emph{MInt}(C_1,C_2)$ captures the
minimum internal difference of either $C_1$ or $C_2$. When the difference
between two components is larger than the minimum difference in at least one
component then we want a boundary between the two components (\ie
$D(C_1,C_2)={\texttt true}$), hence the inequality in eqn.~\ref{eqn:boundpred}.
The threshold $\tau(C)$ controls how much larger the inter-component
difference needs to be relative to the intra-component difference in order
to have a boundary, and they set it to, \begin{flalign}
\tau(C)=\frac{k}{|C|} \end{flalign} for some constance $k$, where $|C|$ is
the size of $C$. Increasing $k$ results in larger components.

The Felzenswalb and Huttenlocher algorithm~\cite{felzenszwalb2004} for partitioning the graph
using the predicate is straight forward: \begin{enumerate} \item Order the $m$
    edges in $E$ according to decreasing edge weights $w_{ij}$. \item Set each
    node $v_i \in V$ as a distinct component in the initial segmentation, $S_0$.
\item Repeat the following from $k=1,\dots,m$: \begin{itemize} \item For edge
        $e_k=(v_i,v_j)$ in the ordering if $v_i$ and $v_j$ are in separate
        components of $S_{k-1}$ and if $w_{ij}<\emph{MInt}(C^{k-1}_i,C^{k-1}_j)$
then merge $C^{k-1}_i$ and $C^{k-1}_j$ into a single component in $S_k$.
Otherwise $S_k=S_{k-1}$. \end{itemize} \end{enumerate} This simple algorithm can
be applied to different neighborhood relations between vertices in the graph ---
for example 8-pixel neighborhoods on grid graphs over the image, or nearest
neighbor graphs over feature space. In the nearest neighbor graph edges connect
vertices that are neighbors in feature space, with weights equal to the distance
between features. The neighborhood can be all vertices falling within a
euclidean ball or simply a fixed number of nearest neighbors returned by an
approximate nearest neighbor method. In either the grid or nearest neighbor
graphs, using approximate nearest neighbor methods, the running time is shown to
be $\bigO{n\log n}$. The results when partitioning in feature space using the
nearest neighbor graph tend to contain higher level information because they
capture more global image information. The resulting segmentations tend to align
well with image boundaries.

Ratio cut~\cite{wang2003} is another well known such method. \subsubsection{SLIC}
%\newlength\figurewidth \newlength\figureheight
\begin{figure}[h!]
    \makebox[\textwidth][c]{
        \subfloat[][$k=100,\; m=10$]{\includegraphics{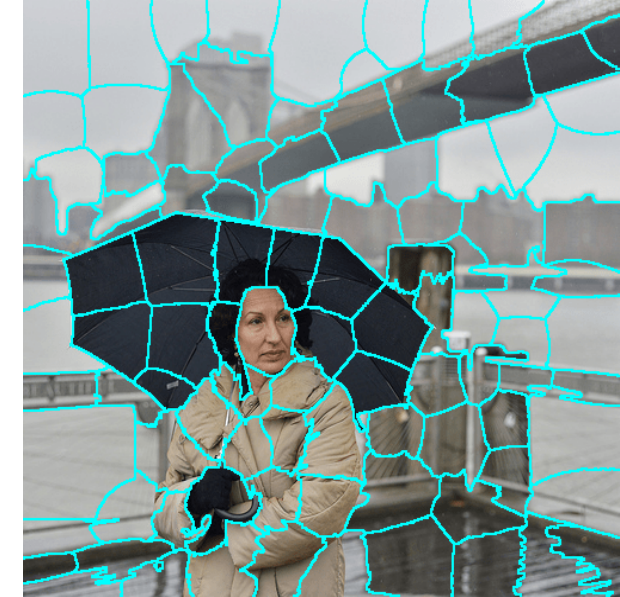}}\hspace{1pt}
        \subfloat[][$k=100,\; m=50$]{\includegraphics{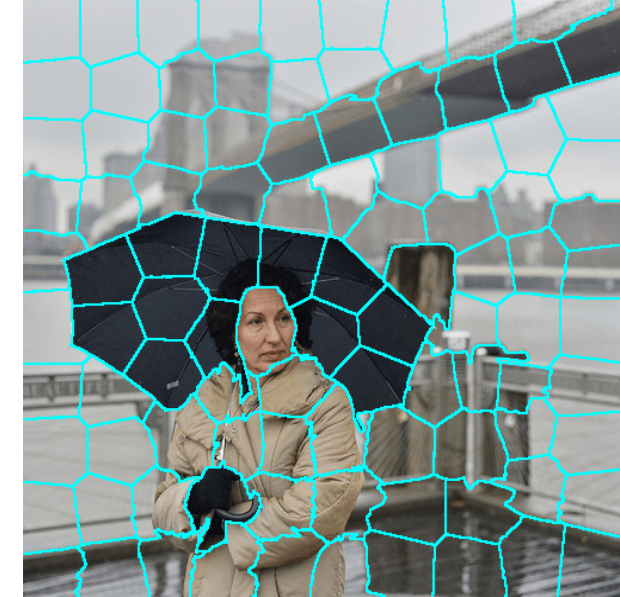}}
%\subfloat[][$k=100,\; m=10$]{\setlength\figurewidth{0.48\textwidth}
        %\setlength\figureheight{0.48\textwidth}\input{gfx/slic/slic100k10m.tex}}
        %\subfloat[][$k=100,\; m=50$]{\setlength\figurewidth{0.48\textwidth}
        %\setlength\figureheight{0.48\textwidth}\input{gfx/slic/slic100k50m.tex}}
}
    \makebox[\textwidth][c]{
        \subfloat[][$k=500,\; m=10$]{\includegraphics{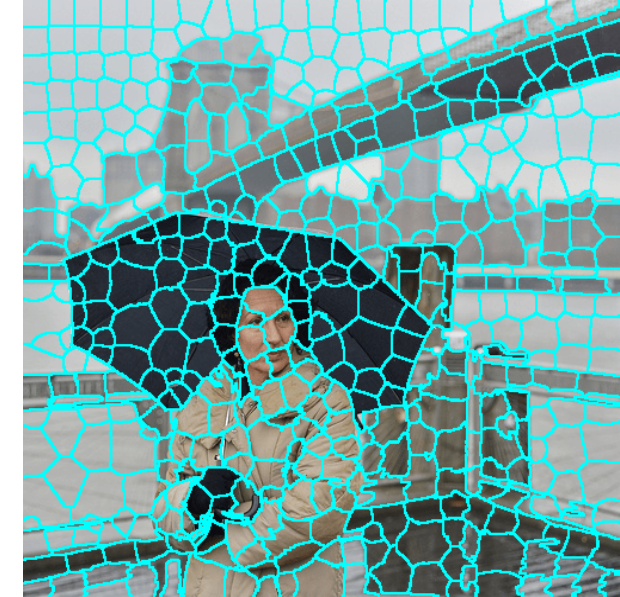}}\hspace{1pt}
        \subfloat[][$k=500,\; m=50$]{\includegraphics{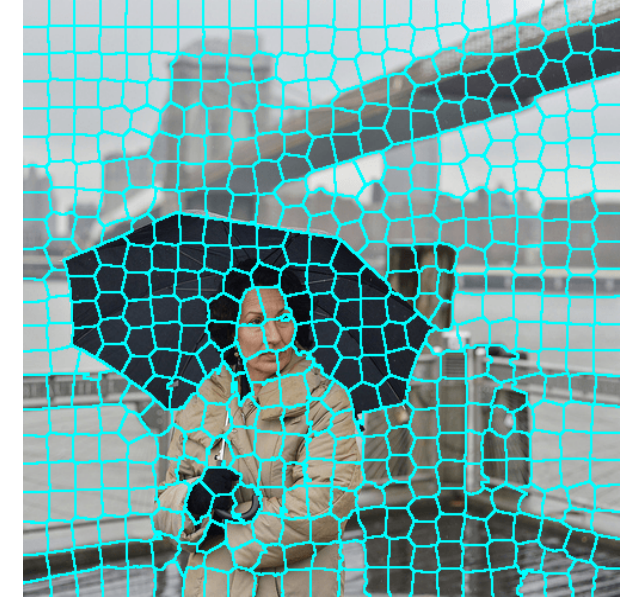}}
%\subfloat[][$k=500,\; m=10$]{\setlength\figurewidth{0.48\textwidth}
%\setlength\figureheight{0.48\textwidth}\input{gfx/slic/slic500k10m.tex}}
%\subfloat[][$k=500,\; m=50$]{\setlength\figurewidth{0.48\textwidth}
%\setlength\figureheight{0.48\textwidth}\input{gfx/slic/slic500k50m.tex}}
}
%\subfloat[][$k=100,\; m=10$]{\setlength\figurewidth{0.48\textwidth}
%\setlength\figureheight{0.48\textwidth}\input{gfx/slic/slic500k10m.tex}}
%\subfloat[][$k=500,\; m=10$]{\setlength\figurewidth{0.48\textwidth}
%\setlength\figureheight{0.48\textwidth}\input{gfx/slic/slic500k10m.tex}}
%\subfloat[][$k=500,\; m=10$]{\setlength\figurewidth{0.48\textwidth}
%\input{./gfx/slic/slic500k10m.tex}}
        \caption{SLIC superpixel results as a function of the desired number,
    $k$, and compactness, $m$. As $m$ is increased the superpixels exhibit more
regular appearance, aligning less with image contours and more with spatial
grid.}
\end{figure}

Arguably one of the best low-level segmentation algorithms
that produces compact superpixels with very good boundary adherence is
\emph{Simple Linear Iterative Clustering} (SLIC)~\cite{achanta2010}. It is remarkably
simple, and one of the fastest algorithms with a linear runtime complexity in
the number of pixels, $\bigO{n}$. It is also the superpixel algorithm we chose
for a number of our experiments in chapters~\ref{ch:divmbestexp}
and~\ref{ch:divrankexp}. SLIC relies on a
$k$-means approach to clustering that assigns pixels with similar color features
and spatial locations to the same cluster. The standard $k$-means approach of
considering all pixels in the image when finding nearest neighbors of a pixel is
prohibitively slow so in SLIC a different approach is taken. Initially a set of
cluster centers, $c_i$, are assigned along a regular spatial grid over the
entire image, corresponding to the centers of the superpixels. The cluster
centers are usually 5-dimensional vectors containing the color intensity and
spatial location. They are initialized with the pixel color and location falling
under the grid locations. The grid spacing, $S=\sqrt{n/k}$ is proportional to
the number of superpixels, $k$, that the user would like to have in the image.
These initial centers are first adjusted to the lowest gradient locations within
a $3\times3$ neighborhood of their initial locations so that they don't fall on
boundaries or noisy pixel locations. The next step in the algorithm is to assign
pixels to cluster centers, which means computing the feature distance between
each $c_i$ and the pixels that are within a spatial neighborhood. Limiting the
search space to pixels within a local neighborhood of a cluster center is why this
algorithm is so fast. Since the desired superpixel size is $S\times S$, the SLIC
algorithm searches within a $2S\times 2S$ neighborhood of $c_i$. To each pixel
the algorithm assigns the label of the cluster $c_i$ that is closest to it in
feature space. There are no more that eight possible cluster labels for a pixel
to be assigned due to the limited search region during assignment. Once each
pixel is assigned a cluster center the cluster centers are updated. The process
is repeated until the residual error between the current and previous iteration
cluster centers reduces below some threshold. Algorithm~\ref{alg:slic} is the
same as that found in the SLIC paper~\cite{achanta2010}.
\begin{algorithm} \caption{SLIC superpixel segmentation~\cite{achanta2010}}\label{alg:slic} \begin{algorithmic}[1]
%    \Procedure{SLIC}{}
    \State Initialize cluster centers $c_i$ by sampling pixels at regular grid
    interval $S$.
    \State Move cluster centers to lowest gradient location within
    $3\times 3$ neighborhood.
    \State label $l(p) \gets -1$ for each pixel $p$.
    \State distance $d(p) \gets \infty$ for each pixel $p$.
    \Repeat
    \Comment{Assignement}
    \For{each cluster center $c_i$}
        \For{each pixel $p$ in $2S\times 2S$ neighborhood around $c_i$}
            \State Compute distance $D$ between $c_i$ and $p$.
            \If{ $D < d(p)$ }
                \State $l(p) \gets i$
                \State $d(p) \gets D$
            \EndIf
        \EndFor
    \EndFor
    \State update cluster centers using assignments $l(p)$.\Comment{Update}
    \State compute residual error $E$ between current and previous cluster centers.
    \Until{ $E \leq$ threshold }

    \Comment{Post-Porcess}
    \State enforce connectivity of superpixels by assigning orphaned pixels to nearest superpixels.
%    \EndProcedure
\end{algorithmic} \end{algorithm} A critical component of why SLIC superpixels
tend to be compact is the distance function $D$ that it employs. In order to
balance between spatial compactness and color consistency of a superpixel, in
SLIC they devise the following distance,
\begin{flalign} D=\sqrt{d_c^2 +
\left(\frac{d_s}{S}\right)^2 m^2},
\end{flalign}
where $d_c$ and $d_s$ are the
Euclidean color and spatial distances between two points respectively, and
constant $m$ controls the relative importance of color similarity versus spatial
compactness of the superpixels. Normalizing $d_s$ by $S$ above balances the
spatial distance relative to color, which is important, otherwise compact
superpixels would be favored by the distance measure. Other distance functions
such as geodesic distance can also been considered. One issue that needs to be
handled at the end of running algorithm~\ref{alg:slic} is that some pixels are
orphaned from the superpixel they belong to. A post-processing step is done to
reassign pixels to nearby superpixels so that they have a connected structure.

\subsubsection{Object Proposals - CPMC} Another class of segmentation algorithms
is based on producing a large set of region proposals that would provide good
overlap with foreground objects. Object proposals have the benefit of providing
a larger image support for tasks where we want to do category detection or semantic
segmentation. Compared to the space of possible solutions over pixel/superpixels
the set of proposals provides orders of magnitude fewer candidates and more
efficient search. Given that the vast majority of possible solutions over pixels
or superpixels do not conform to general appearances of objects in images ---
such as spatial connectivity --- considering a much smaller subset of proposals
that adhere to image cues like boundary, and color and spatial uniformity, is
appealing. One popular region proposal approach is \emph{constrained parametric
min-cut} (CPMC)~\cite{carreira2010}. In CPMC a set of initial candidate figure-ground
proposals are generated very efficiently which are subsequently pruned and
ranked according to how likely the proposals are to tightly cover a foreground
object. The top ranking proposals can be retained for a higher level recognition
task.

The initial set of proposal in CPMC is constructed by solving multiple
parametric min-cut~\cite{carreira2010} problems on a submodular grid graph over the image,
with multiple different initializations. Given a graph $G=(V,E)$ over pixels
where adjacent pixels in a 4-neighborhood share edges, a figure-ground
segmentation is performed by minimizing the following objective,
\begin{flalign}\label{eqn:cpmcenergy}
    E(X,\lambda)=\sum\limits_{i\in V} D(x_i,\lambda) + \sum\limits_{(i,j)\in E} V_{ij}(x_i,x_j),
\end{flalign} where
the data term is defined to be,
\begin{flalign} D(x_i,\lambda)
    =\left\{\begin{array}{ll} 0 & \mbox{if $x_i=1, \; i \not\in \mathcal{V}_b$}
            \\ \infty & \mbox{if $x_i=1,\; i \in \mathcal{V}_b$} \\ \infty &
            \mbox{if $x_i=0,\; i \in \mathcal{V}_f$} \\ f(x_i)+\lambda &
    \mbox{if $x_i=0,\; i \notin \mathcal{V}_f$} \end{array} \right.
\end{flalign}
where $\mathcal{V}_f$ and $\mathcal{V}_b$ are the pixels in a seed
region for the foreground and background respectively. The foreground seed
regions are groupings of pixels forming small squares a few pixels wide, sampled
regularly along a grid over the image, and the background seed regions are
horizontal or vertical edges of the image. The pixel labeling takes on either
foreground ($x_i=1$) or background ($x_i=0$). The cost of assigning pixels not
in the foreground seed region to background is set by $f(x_i) + \lambda$. The
first term is either uniformly set to $0$ or is the log ratio of probabilities
of pixel $i$ belonging to foreground versus background, where the probability of
foreground is, \begin{flalign} p_f(i) = e^{-\gamma \min_j
(||\emph{I}(i)-\emph{I}(j)||)}, \end{flalign} where $j$ is over pixels in the
seed region --- and similarly for the background probability. The parameter
$\lambda$ is a foreground bias that can be adjusted, where for each setting a
different solution is computed.

The pairwise term in eqn.~\ref{eqn:cpmcenergy} penalizes adjacent pixels that
cross an image boundary,
\begin{flalign} V_{ij}(x_i,x_j) =
    \left\{\begin{array}{ll} 0 & \mbox{if $x_i=x_j$} \\
            e^{-\frac{\max(\emph{gPb}(i),\emph{gPb}(j))}{\sigma^2}} & \mbox{if
            $x_i \neq x_j$} \end{array}\right. ,
\end{flalign} where
$\emph{gPb}(i)$ is the contour strength at pixel $i$ computed using
globalPb~\cite{arbelaez2011}. Using a parametric min-cut solver minimization of
eqn.~\ref{eqn:cpmcenergy} can be done for all setting of $\lambda$ in
the same time complexity as doing a single min-cut. The complexity
of computing the initial set of regions using $k$ different
combinations of foreground/background seeds and choice of $f(x_i)$
is $\bigO{kmn\log n}$, where $m$ is the number of edges and $n$ is
the number of pixels. Successively increasing values of $\lambda$
given the same seeds and $f(x_i)$ results in nested regions that
progressively get larger.

Given this initial bag of figure-ground proposal regions pruning is done by
throwing away very small regions and sorting the remaining segments using ratio
cut~\cite{wang2003} value. The top sorted $2000$ proposals are kept.

Ranking of the proposals is done by training a regressor (random forest) that
takes region, Gestalt, and graph partition features over the proposal (details
in~\cite{carreira2010}) to regress onto the intersection-over-union score of the proposal
with the best overlapping ground-truth region. The objective is to retain the
minimal set of ranked region proposals that maximize a covering of the
ground-truth regions, and discard the rest. To do this CMPC uses the following
covering score,
\begin{flalign}\label{eqn:cpmccover} C(G,S(r)) =
\frac{1}{n}\sum\limits_{g\in G} |g| \max\limits_{s\in S(r)} O(g,s),
\end{flalign}
where $G$ and $S(r)$ are the ground-truth segments and region
proposals with rank higher than $r$ respectively, $|g|$ is the number of pixels
in the ground-truth segment and $O(g,s)$ in the intersection-over-union score
between ground-truth and region proposal segments. As the authors of CPMC note,
many of the segments are similar in shape size and location. Similar segments
end up having similar features which mean the regressor ranks them to similar
scores, so the sorted list of region proposals will have many cases where
sequentially ranked segments have the same quality of coverage for a
ground-truth segment. Using the covering measure in eqn.~\ref{eqn:cpmccover} to
pick a cut-off rank $r$ would result in a bag of segments with many redundant
ones. To aleviate this they propose to diversify the final bag using the
\emph{Maximal Marginal Relevance} (MMR) measure~\cite{carbonell1998},
\begin{flalign}
    \emph{MMR}= \argmax\limits_{s_i\in S\setminus S_k} \theta\cdot {\texttt
    score}(s_i) - (1-\theta)\cdot\max\limits_{s_j\in S_k}o(s_i,s_j),
\end{flalign}
where $S$ is the set of all proposals, $S_k$ is the set of
selected proposals in round $k$, $\texttt{score}(s_i)$ is the regressor score for
proposal $s_i$, and $o(s_i,s_j)$ is the overlap between the two proposal. The
$\emph{MMR}$ measure is applied in an iterative fashion. Starting with the
highest scoring proposal which is placed in $S_1$, the next proposal, $s_i$,
picked is one that has the best trade-off in maximizing the regressor score at
the same time minimizing its overlap with any of the previously picked
proposals. This selection procedure can continue until the covering score using
the current set of region proposals reaches some threshold.

This method produces very good quality region proposals and, at the time of our
experiments in later chapters, was a state of the art region proposal method. It
shows up as the underlying region-proposal method for the semantic segmentation
model \otwop~\cite{carreira2012o2p} which we use in Chapters~\ref{ch:divmbest} \& \ref{ch:divrank}.  \subsubsection{Object
Proposals - continued} There are a number of other noteworthy object proposal
methods which we'll briefly outline here.

Endres and Hoiem~\cite{endres2010,endres2014} introduced a region proposal method that generates an
initial bag of region proposals and ranks them to produce a set of diverse
region proposals that maximally cover foreground objects in the image. The
initial set of region proposals is generated using the occlusion boundary
algorithm~\cite{hoiem2007} which constructs a hierarchical segmentation of the image.

This hierarchical segmentation algorithm uses cues on both regions, boundaries,
and 3D surface and 3D depth to predict occlusion boundary probability (\ie
boundaries between different objects) as well as figure-ground probability at
each pixel. After predicting the occlusion boundary probabilities, agglomerative
clustering by iteratively merging regions with minimum boundary strength up to a
threshold produces the hierarchical segmentation.

Given the segmentation hierarchy, initial seeds are picked from the hierarchy as
starting points from which to construct object proposals. These seed regions are
used to label superpixels as either belonging to the object or background
depending on their affinity with the seed region (\ie likelihood of belonging to
the same foreground object as the seed). This problem is formulated as a CRF
over superpixels consisting of two terms, an affinity measure between
superpixels and the seed region, and an edge cost for two adjacent superpixels
to take on a different label (\ie foreground/background), which is proportional
to the probability of the occlusion boundary between the two superpixels. The
region affinity term uses features such as layout prediction of the seed and
superpixel on the object (e.g. right+left of object) to capture their layout
agreement, and their layout location (e.g. center, top, bottom, etc) on the
object. Maximizing the CRF energy infers a labeling over the superpixels
indicating which of them are part of the same foreground region as the seed.

Having generated a set of candidate proposals by considering multiple seed
regions their method ranks the proposals so that higher ranked proposals will be
more likely to tightly cover a foreground region while at the same time have
minimal overlap with any higher ranked proposal. Given a ranking $\br$ over a
set of proposals $\bx$ they define the following score,
\begin{flalign}\label{eqn:hoiemscore} S(\bx,\br;\bw) = \sum\limits_i \alpha(r_i)
    \cdot (w^T_a \bs{\psi}(x_i) - \bw^T_p\bs{\phi}(r_i)),
\end{flalign} where
$\bs{\psi}(x)$ are appearance features and $\bs{\phi}(r)$ is an overlap
penalty incurring a cost for a proposal to overlap with the set of higher
ranked proposals.  The monotonical decreasing function of rank,
$\alpha(r_i)$, encourages high ranked proposals to have higher score. The
appearance features captures how likely the proposal is to be an object
region. Therefore they use occluding boundary probabilities, interior and
exterior boundary probabilities, likelihood of region being background from
a background predictor, and statistical differences in color and texture
between the region and the surrounding area to capture the appearance. It's
not possible to maximize eqn.~\ref{eqn:hoiemscore} with respect to $\br$
exactly so a greedy maximization strategy that incrementally selects
proposals based on which one maximizes the marginal gain. To optimize over
$\bw$ a latent max-margin structure learning approach is used that minimizes
the score of the highest scoring incorrect ranking order while
simultaneously maximizing the score of correct ranking. They use a margin
that encourages the best region proposals for each object to have the
highest rank. More details can be found in~\cite{endres2010,endres2014}.

The resulting region proposals from this method are competitive with CPMC. The
CPMC region proposals tend to be a bit less diverse and as the overlap threshold
for removing redundant regions is lowered the recall of object regions is worse
than Endres adn Hoiem's method~\cite{endres2010,endres2014}.
\subsubsection{ALE and the Associative Hierarchical CRF} A prime example of an object
segmentation method that can achieve accurate segmentation results by solving an
inference problem on a Conditional Random Field (CRF) over the image is the
\emph{Automatic Labelling Environment}~\cite{ladicky2012}. It is a segmentation system that
is the culmination of a number of papers by Ladicky et al~\cite{ladicky2010,ladicky2010detect,ladicky2012joint,russell2012exact,kohli2009robust,ladicky2009,sturgess2009,kohli2009p3}. The underlying model
is a hierarchical CRF on pixels, segments, and super segments over the image.

A main contribution of this segmentation model is the observation that the image
quantization level that one chooses is critical in producing good segmentations.
It is a common observation that inference on CRFs defined over random variables
corresponding to pixels in the image often produce segmentations that do not
align well with object boundaries. Conversely, the assumption that superpixels
from bottom-up algorithms align well with object boundaries is often wrong. That
is why CRFs over superpixels often yield segmentations that also do not align
well with boundaries. Superpixels do offer advantages though --- they provide
both larger and specific spatial support (context) to compute features and as
primitives they allow for more efficient inference over graphical models. Also,
the assumption that all pixels falling within the same superpixel should take on
the same label, though often incorrect, is nevertheless a strong prior that
often holds true. So the natural question is, what is the right level of image
quantization? The answer seems to be that it depends on image and the objects in
it. The major technical contribution of ALE, the \emph{associative hierarchical
CRF}~\cite{ladicky2009}, tries to tackle this issue by considering multiple quantizations.
The structure is a three-level hierarchical CRF (the model and algorithm have no
constraint on the number of levels), where the bottom-most level consist of
random variables over pixels in the image. At the pixel level the random field
consists of a grid graph over pixels in the image with a 4-pixel neighborhood
for each pixel. The next level up consists of random variables over segments.
Each segment's node has an edge (conditional dependence) between it and the
pixels that fall under it in the image. Edges connect adjacent segments in the
image. The top-most layer consists of random variables corresponding to
super-segments that are composed of the segments. Once again, the super-segment
nodes share an edge to the segment nodes beneath if the segment is contained by
the super-segment. Super-segments share edges if they are adjacent to each other
in the image. The initial (super)segmentations used at the different levels in
the CRF hierarchy are produced by running the bottom-up mean shift segmentation
algorithm with varying bandwidth parameters for the color and spatial channels
to produce progressively coarser superpixels.

To summarize the model we can write the energy of the hierarchical model as
presented in ~\cite{ladicky2009},
\begin{flalign}\label{eqn:alecrf} E^{(0)} =
    \sum\limits_{i\in \mathcal{S}^{(0)}} \psi_i(x_i^{(0)}) + \sum\limits_{ij\in
    \mathcal{N}^{(0)}} \psi_{ij}(x_i^{(0)},x_j^{(0)}) + \min\limits_{\bx^{(1)}}
E^{(1)}(\bx^{(0)},\bx^{(1)}),
\end{flalign} where
$\bx=\{\bx^{(0)},\dots,\bx^{(K)}\}$ is the vector of random variables, called
the labeling, taking on values from the label set $\mathcal{L}^n$. The term
$\psi_i(x_i^{(0)})$ is the pixel-wise unary potential, and
$\psi_{ij}(x_i^{(0)},x_j^{(0)})$ is the pixel-wise label consistency term
between neighboring pixels, $\mathcal{S}^{(k)}$ is the set of pixels or
segments, and $\mathcal{N}^{(k)}$ is the set of neighbors of a pixel or segment
at level $k$. The last term in eqn.~\ref{eqn:alecrf} can be recursively written
as,
\begin{flalign}\label{eqn:hierenergy} E^{(k)}(\bx^{(k-1)},\bx^{(k)}) = &
    \sum\limits_{c\in\mathcal{S}^k} \psi_c^p(\bx_c^{(k-1)},x_c^{(k)}) +
\sum\limits_{cd\in\mathcal{N}^{(k)}} \psi_{cd}(x_c^{(k)},x_d^{(k)}) \\
&+\min\limits_{\bx^{(k+1)}} E^{(k+1)}(\bx^{(k)},\bx^{(k+1)}).\nonumber
\end{flalign}

The unary pixel-wise potentials are based on classifiers trained on color, shape
and texture features (\ie textons of \emph{TextonBoost}~\cite{shotton2006}), \emph{historgrams of
oriented gradients} (HOG~\cite{dalal2005}), and pixel location. The classifiers are
applied at each pixel to estimate the probability of the pixel to take on a
particular label. The pairwise pixel terms are \emph{contrast sensitive}
potentials~\cite{boykov2001} that encourage neighboring pixels to take on the same label,
\begin{flalign} \psi_{ij}(x_i,x_j) = \left\{\begin{array}{ll} 0 & \mbox{if $x_i
= x_j$} \\ g(i,j) & \mbox{otherwise} \end{array}\right.
\end{flalign} where
$g(i,j) =|c|^{\theta_\alpha}(\theta_p + \theta_v
\exp(-\theta_\beta||f_i-f_j||^2))$~\cite{kohli2009,boykov2001}, where $I_i$ and
$I_j$ are the color vectors at pixels $i$ and $j$, and $\theta_p$, $\theta_v$,
and $\theta_B$ are learned parameters.

The higher order potentials $\psi_c^p(\bx_c^{(k-1)},x_c^{(k)})$ are robust $P^n$
potentials~\cite{kohli2009} that are equivalent to minimizing a pairwise graph over
$\bx_c$. Here $x_c^{(k)}$ is the random variable associate with a segment, or
super-segment, $c$ at level $k$ in the hierarchy. The variable $c$ also stands
for the \emph{clique} consisting of variables (\ie segments or pixels) at level
$k-1$ that fall within the (super)segment $c$ at level $k$. Therefore,
$\bx_c^{(k-1)}$ are the random variables associated with the clique, $c$, of
pixels or segments at level $k-1$ that fall under segment $c$ at level $k$. The
robust $P^n$ potential~\cite{ladicky2009} is thus,
\begin{flalign}\label{eqn:pnunary}
    \psi_c^p(\bx_c^{(k-1)},x_c^{(k)}) = \phi_c(x_c^{(k)}) + \sum\limits_{i\in c}
\phi_c(x_c^{(k)},x_i^{(k-1)}),
\end{flalign} where $x_c^{(k)}$ takes on labels
from $\mathcal{L} \cup\{L_F\}$, and $\phi_c(x_c^{(k)})$ is the (super)segment
unary potential that has a cost of $\gamma_c^l$, if $x_c^{(k)}=l$, or
$\gamma_c^{\max}$, if $x_c^{(k)}$ is assigned the free label $L_F$, where
$\gamma_c^l \leq \gamma_c^{\max}$, $\forall l\in \mathcal{L}$. The pairwise
potential~\cite{ladicky2009} is defined as,
\begin{flalign}\label{eqn:pnpairwise}
\phi_c(x_c^{(k)},x_i^{(k-1)})= \left\{\begin{array}{ll} 0 & \mbox{if
    $x_c^{(k)}=L_F$ or $x_c^{(k)}=x_i^{(k-1)}$} \\ w_ik_c^{x_i^{(k-1)}} &
\mbox{otherwise}, \end{array}\right.
\end{flalign}
where $w_i$ are learned
weights and $k_c^l$ are costs associated with labeling a variable clique $c$ in
level $k-1$ (\ie child node) with a label $l$ that is different than the label
for (super)segment $c$ in level $k$ (\ie parent node). Therefore, the pairwise
potential in eqn.~\ref{eqn:pnpairwise} encourages that all child variables in
the lower level take on the same label as the parent node in the higher level.
Otherwise there is a cost incurred on each and every variable in the clique
taking on a different label. Combined with the unary
potential~\ref{eqn:pnunary}, the robust $P^n$ potential encourages child
variables to take on the same label as the parent variables, but allows the
possibility of heterogeneous labeling of the child nodes. More specifically, by
ensuring that the constraint $\sum_i w_ik_c^l \geq 2\phi_c(l)$, $\forall l\in
\mathcal{L}$, is satisfied, the parent variable will take a label
$l\in\mathcal{L}$ if and only if the (weighted) majority of child nodes takes on
the same label. Otherwise, if the parent node takes on label $L_F$, then the
child nodes are free to take on any label that minimizes lower level unary and
pairwise costs defined by the CRF, with an added cost for a heterogeneous
labeling of $\gamma_c^{\max}$. In summary, the higher order robust $P^n$
potentials favor homogeneous labelings of (super)segments but allow for the
possibility that regions within the (super)segments take on different labels.

The unary potentials over (super)segments are the responses of classifiers
trained on normalized histograms of clustered dense (pixel-level) features. The
dense features include color, textons, HOG, and pixel location. The classifiers
are multiple week learners trained via AdaBoost~\cite{freund1995,freund1999short}. The variables, as define
in~\cite{ladicky2009}, are set to, \begin{flalign} \gamma_c^l = \lambda_s|c|
\min(-H_l(c)+K,\alpha^h), \end{flalign} where the log probability of clique (aka
(super)segment) $c$ taking on label $l$ is $H_l(c)$ (given by classifier),
$\alpha^h$ is a truncation threshold, and $K=\log(\sum_{l\in\mathcal{L}}
e^{H_l(c)})$. The other variables in the robust $P^n$ potential are set to
$\gamma_c^{\max}= |c|(\lambda_p+\lambda_s\alpha^h)$, and $k_c^l =
(\gamma_c^{\max}-\gamma_c^l)/0.1|c|$ (up to 10\% of the pixels in a segment can
be assigned a different label than the segment variable before the variable is
assigned $L_F$). The pairwise (super)segment potentials
$\psi_{cd}(x_c^{(k)},x_d^{(k)})$ are the Euclidean distance between normalized
color histograms over pixels within the (super)segments.

What's hidden in eqn.~\ref{eqn:hierenergy} for every layer are weight constants,
$\lambda_1^{(k)}$ and $\lambda_2^{(k)}$ for the unary and pairwise potentials.
In order to learn these parameters the approach taken is to do a layer by layer
search for the optimal parameter settings, on a validation set, that minimizes
the error between the dominant ground truth label for a clique according to the
ground-truth labeling and the label $x_c^{(k)}$ assigned by the \emph{maximum
a-posteriori estimate} (MAP) over the CRF.

Inference over the CRF, by computing the most probable label assignment
(otherwise known as the MAP estimate) for the energy defined by
eqn.~\ref{eqn:hierenergy} has been shown to run in polynomial time~\cite{russell2012} using
graph cut move making algorithms ($\alpha$-expansion, $\alpha\beta$-swap).

The ALE system also combines object detection with semantic segmentation by
including potentials over detections into the CRF energy. Ladicky et al.
argue~\cite{ladicky2010detect} that since object detectors are good at localizing
\emph{things}, which have describable size and shape (as opposed to \emph{stuff}
which are shapeless), the detections can be used to improve segmentation
accuracy. Object detections also provide object instance level information that
semantic segmentation does not; and coupled with the bounding box size, shape,
and location information provide rich information for scene understanding.

To incorporate detections from object detectors into the CRF, an additional term
is added to the energy in eqn.~\ref{eqn:alecrf}. Denoting $E_{hier}$ as the
energy in eqn.~\ref{eqn:alecrf}, the new energy is defined
as~\cite{ladicky2010detect},
\begin{flalign}\label{eqn:aleenergycomb}
E(\bx)=E_{hier}(\bx) + \sum\limits_{d\in\mathcal{D}} \psi_d(\bx_d,H_d,l_d),
\end{flalign}
where $\mathcal{D}$ are the set of detections given by a detection
algorithm such as~\cite{felzenszwalb2010}. Each detection has associated with it the
bounding box (\ie spatial extent) surrounding the object, the predicted object
label $l_d$, and corresponding label probability $H_d$. The detector potential
introduces an auxiliary variable $y_d\in\{0,1\}$ that indicates whether the
detector prediction is used or not. The form of the detector potential as define
by Ladicky et al.~\cite{ladicky2010detect} is, \begin{flalign} \psi_d(\bx_d,H_d,l_d) =
\min\limits_{y_d\in \{0,1\}} (g(N_d,H_d)y_d - f(\bx_d,H_d)y_d), \end{flalign}
where the first term is a cost for having pixels inside the detector bounding
box take on different values that the detector label $l_d$, and the second term
is the likelihood of an object being present inside the detector bounding box.
The likelihood term is, \begin{flalign} f(\bx_d,H_d) = w_d|\bx_d|
\max(0,H_d-H_t), \end{flalign} with $H_t$ a threshold controlling the number of
detections. The label inconsistency term, $g(N_d,H_d)$, is defined to be,
\begin{flalign} g(N_d,H_d) = \frac{f(\bx_d,H_d)}{p_d|x_d|}N_d, \end{flalign}
    where $N_d$ is the number of pixels inside the detector bounding box having
    a different label from $l_d$, and $p_d$ is set to a threshold percentage of
    inconsistent pixels.

Recall that the first energy term in eqn.~\ref{eqn:aleenergycomb} can be
minimized efficiently using graph cut move making algorithms~\cite{russell2012}. It
turns out that the detector potentials can also be minimized with respect to
$\bx_d$ using $\alpha\beta$-swap and $\alpha$-expansion
algorithms~\cite{ladicky2010detect}, so inference on the CRF defined by the energy
in eqn.~\ref{eqn:aleenergycomb} can be computed in polynomial time.

The addition of the detector potentials help further disambiguate and correct
false labellings.

\subsubsection{Zoom-out convolutional neural network}
\begin{figure}[!h] \centering
    \includegraphics{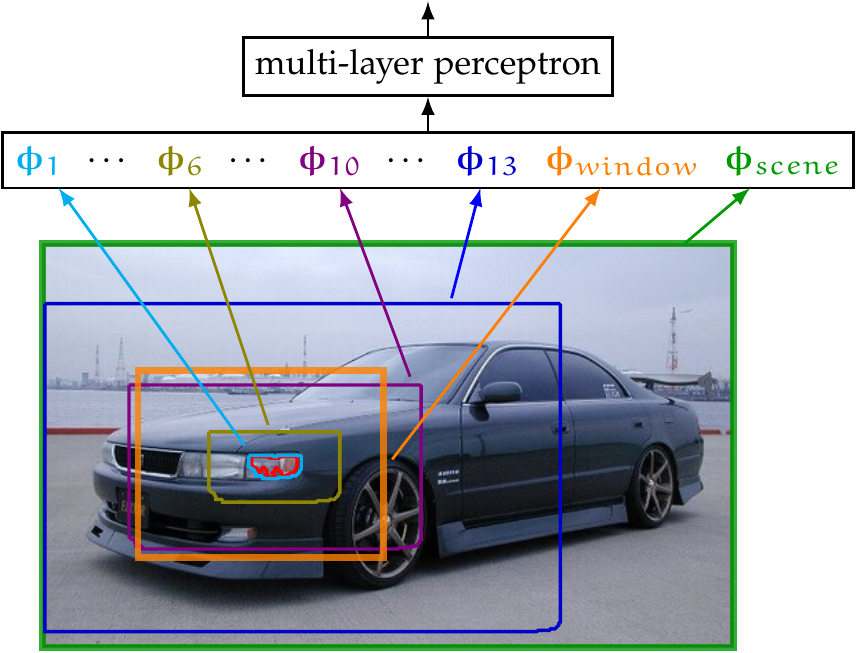}
\caption{For a given superpixel (red) zoom-out features are computed at multiple zoom-out levels (6 levels shown). The features from each level are stacked into a column vector representation of the superpixel, and a multi-layer perceptron is used to predict the superpixel class probabilities. }\label{fig:zoomoutmain}
\end{figure}

Semantic segmentation is often viewed as a structured prediction task,
because of the relationship between variables in the output space --- for
example pixels appearing in a similar context take the same segmentation label
and the likelihood of pixels taking a particular label are conditionally
dependent on what type of scene the image depicts. As the hierarchical CRF model
in ALE demonstrated, one common way to do structured prediction is to model the
variables and their conditional dependencies using a graphical model and running
inference over the graph to compute the most likely labeling. Here we present
another approach to doing structured prediction for semantic segmentation that
side steps the issue of explicitly imposing conditional dependencies between
variables via a graph structure and higher order potentials. The advantage is a
model that avoids the hard or intractable inference and learning that often
plagues conventional structured prediction problems, while simultaneously
incorporating higher order clique interactions between regions in the image, in
an implicit way.

The idea by Mostajabi et al.~\cite{mostajabi2015} proposes to label each region in the image
by classifying it using features computed on the image. We will assume that the
regions are the result of some bottom-up image segmentation algorithm such as
SLIC. The task is to semantically label each superpixel by classifying it such
that the majority ground-truth label over pixels falling in that superpixel
agree with the semantic label predicted by the classifier. This approach has
been taken before where the classification is based on features computed over
the superpixel. What is new in the approach by Mostajabi is the spatial extent
over which features are computed. Instead of computing features limited to the
spatial extent of the superpixel, multiple spatial scales of influence or
context around the superpixel are also considered. The increasing spatial
scales, or \emph{zoom-out levels}, can be broadly categorized as local,
proximal, distant, and scene. They can be described as follows,
\begin{itemize}
    \item {\emph Local} --- the spatial extent defined by the superpixel itself.
        Features computed in this region capture local color, texture pattern,
        and gradient cues specific to the superpixel. Neighboring superpixels
        can have very different features, for example if they appear on
        different objects in the image.
    \item {\emph Proximal} --- regions
        centered on the superpixel extending over a one or two superpixel
        neighborhood. Neighboring superpixels will have overlapping proximal
        regions and their corresponding features will have some similarity. As
        the distance between superpixels grows beyond the one or two superpixel
        neighborhood the features computed in their respective proximal regions
        will capture increasingly different image statistics. Therefore the
        proximal region features implicitly encode local conditional
        dependencies between superpixels that are in an approximate neighborhood
        of each other.
    \item {\emph Distant} --- further along the scale, distant
        regions are centered on the superpixels and capture a much larger
        portion of the image than proximal regions. Superpixels that are
        adjacent have very similar distant level features due to the large
        overlap in the regions. Distant level features implicitly capture long
        range dependencies between superpixels in the image. As the superpixels
        drift further apart their distant level features gradually differ. These
        higher-order interactions are difficult to incorporate into standard
        structured prediction models.
    \item {\emph Scene} --- this is at the level
        of the entire image. Features computed at the scene level capture what
        the scene is depicting as a whole and provide strong cues as to what
        object categories might be present in the scene. As such all superpixels
        share the same scene level features, which provides soft global
constraints to the classifier that impact local predictions. \end{itemize}
\begin{figure*}[!th] \centering \begin{tabular}{ccc}
    \includegraphics[height=1.6in]{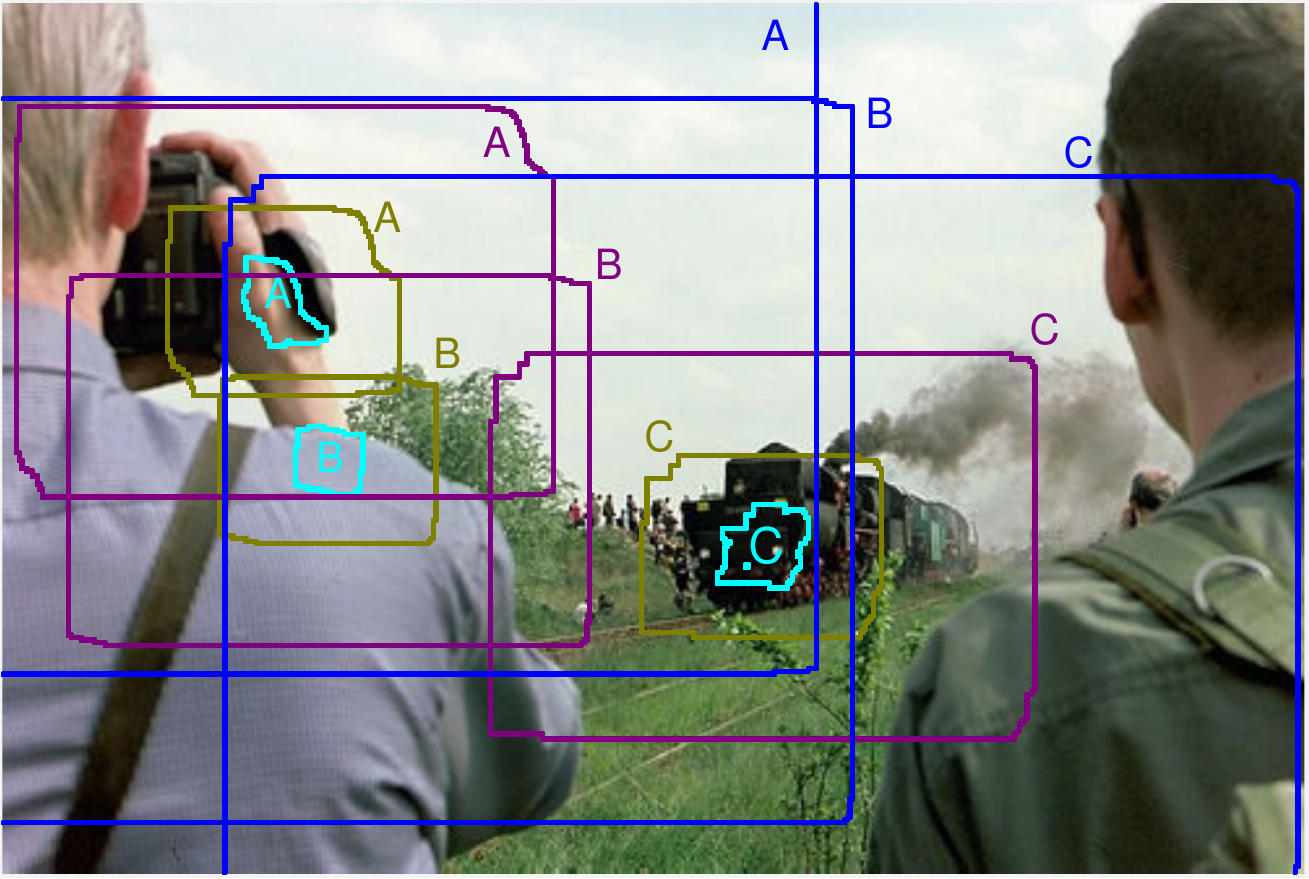} &
\includegraphics[height=1.6in]{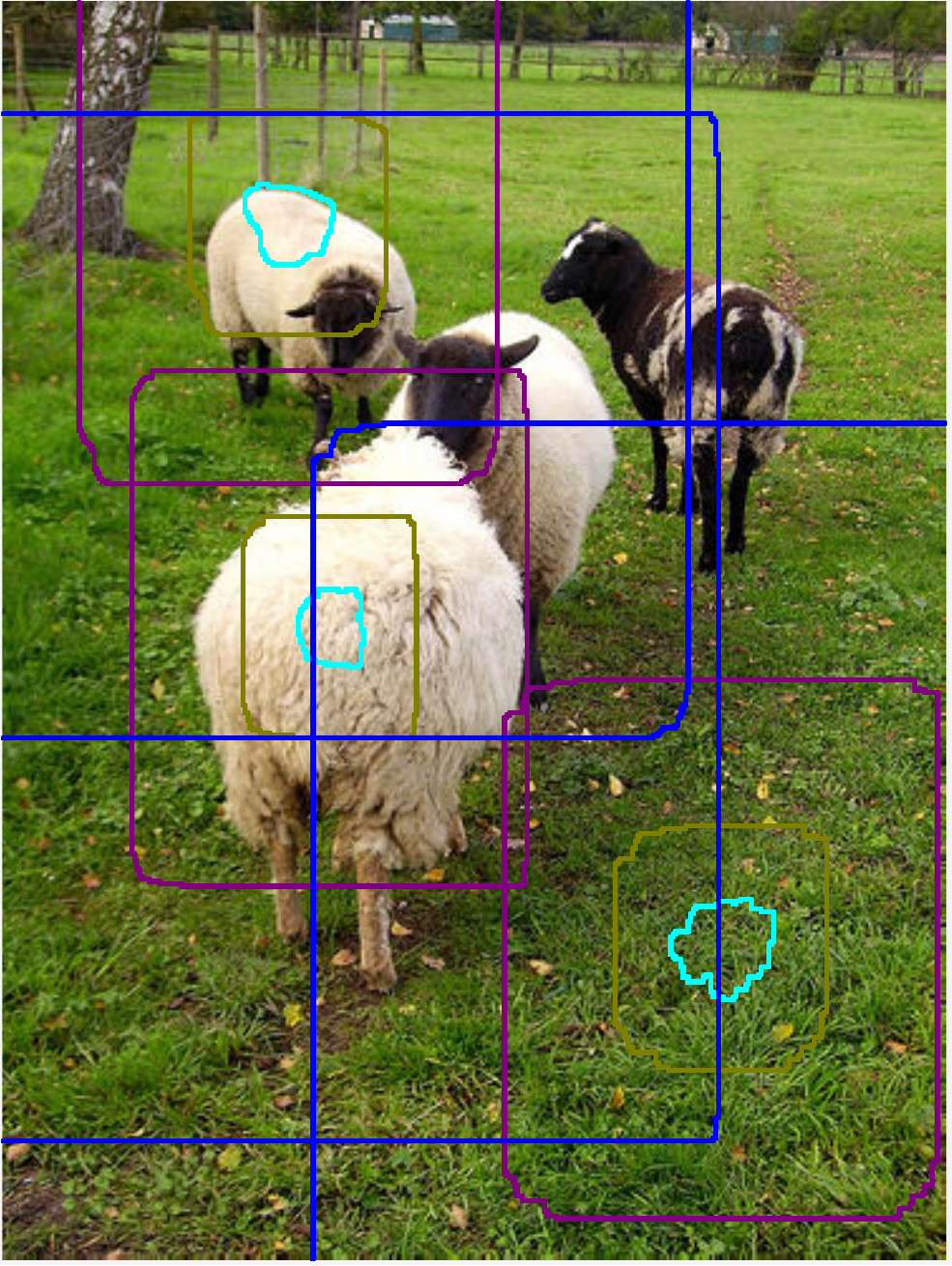} &
\includegraphics[height=1.6in]{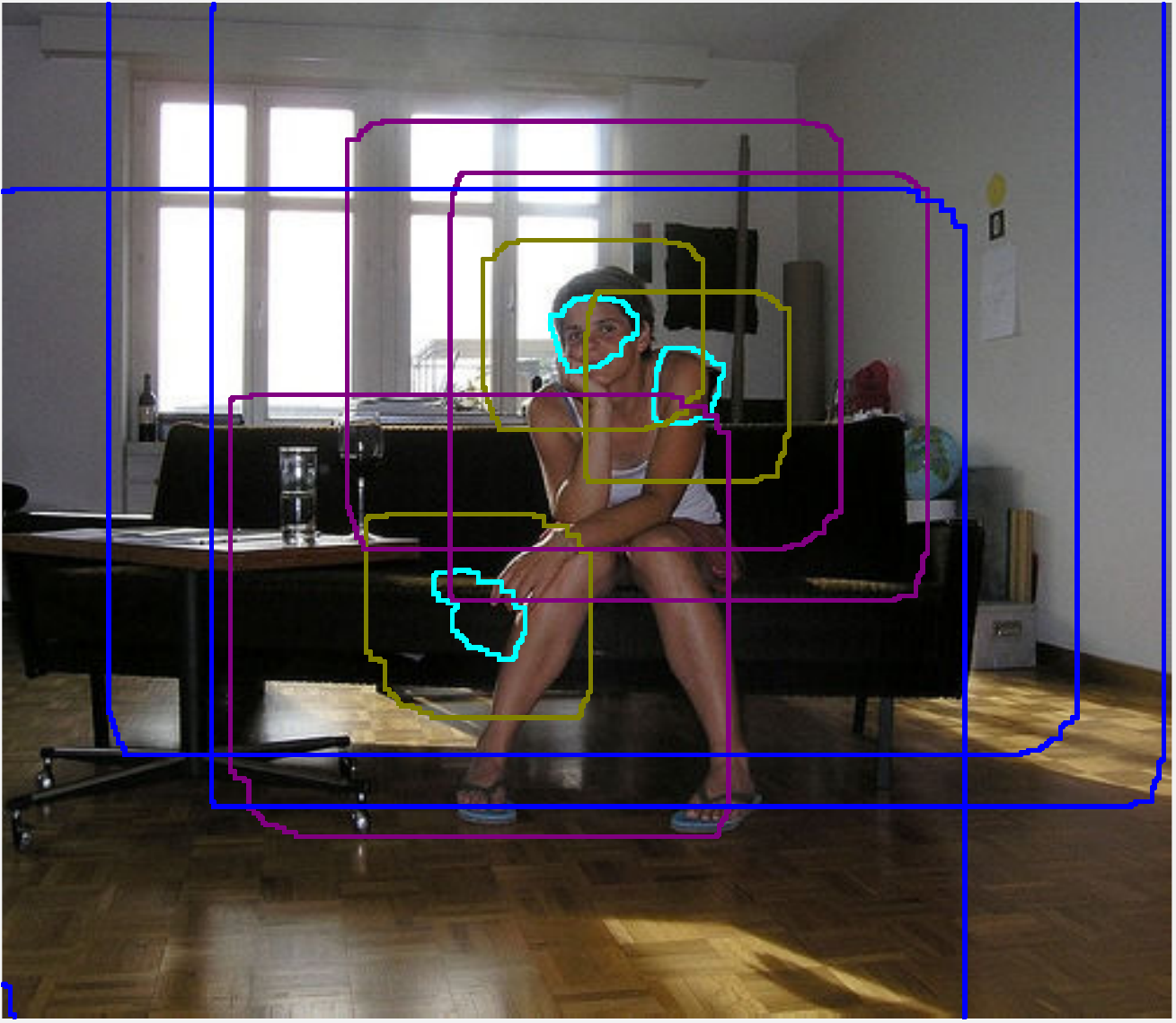}
\end{tabular}

  \caption{Examples of zoom-out regions. We show four out of fifteen levels:
  \textcolor{cyan}{1}(\textcolor{cyan}{cyan}, nearly matching the superpixel
boundaries), \textcolor{olive}{6 (olive)}, \textcolor{blue!50!red}{10 (purple)}
and \textcolor{blue}{13 (blue)}.} \label{fig:zoomoutregions}
\end{figure*}

The spatial categories can be further refined along a more fine-grained scale
pyramid. A depiction of the features used by the local region classifier is
shown in figure~\ref{fig:zoomoutmain}. The features across the spatial regions
are concatenated and used as input to the classifier. The classifier, in turn,
is trained to predict the likelihood of the superpixel taking on a semantic
label. Examples of the different zoom-out regions for three superpixels in
various images is shown in~\ref{fig:zoomoutregions}. You will notice superpixels
that are closely spaced in the image have overlapping proximal regions but as
they are spaced further apart their proximal regions no longer overlap but their
distant level regions do. Thus superpixels that are close share much of the same
image statistics whereas distant superpixels do not. Note from the figures that
if two nearby superpixels are on the same object but with very different local
image features there's still a high likelihood that the classifier will assign
them the same semantic label, given the fact that they share proximal and
distant features. Conversely, if two nearby superpixels are on different objects
their respective features computed over local regions will hopefully be
different enough to bias the classifier to label them as different semantic
classes.

\begin{figure*}[!t] \centering \includegraphics{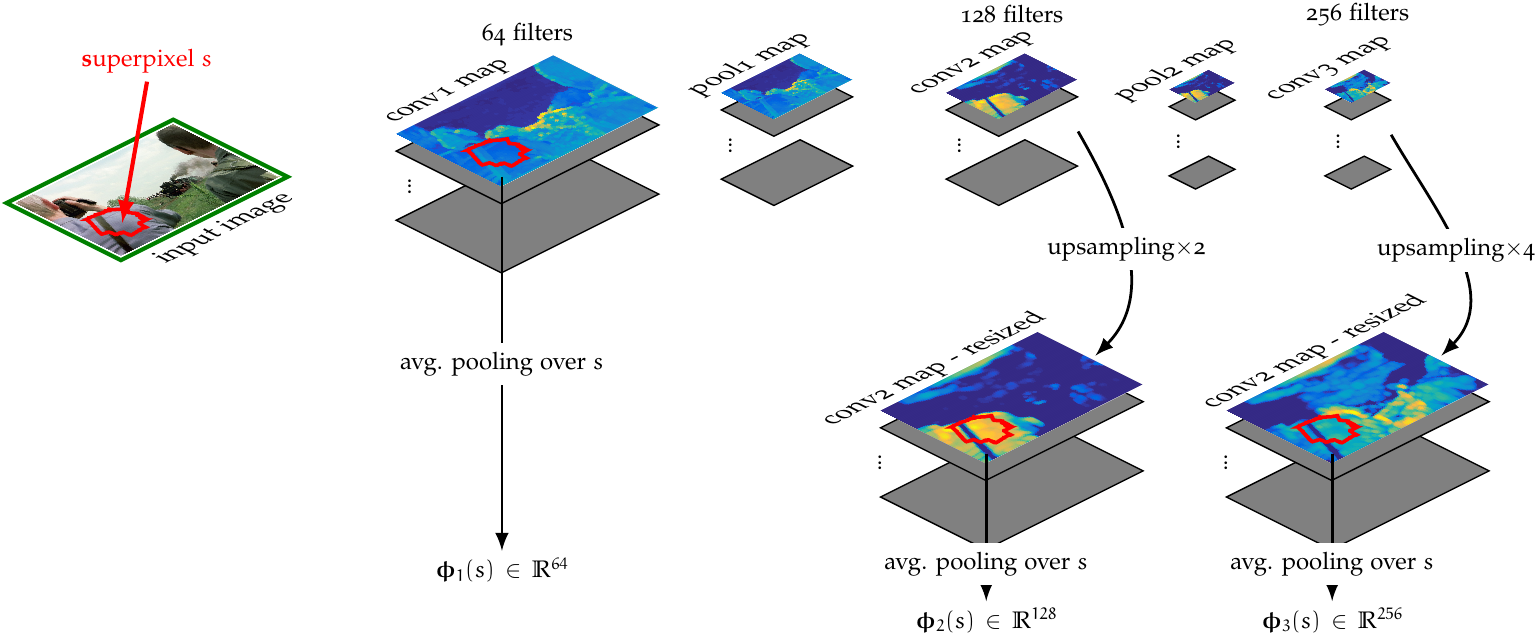}
    \caption{Zoom-out network architecture using an image classification CNN backbone, computed over superpixels. The output response from the convolutional layers plus the softmax output (scene level) are up-sampled to the image size and stacked into the final feature map representation over the image. For each superpixel a feature vector representation is constructed by pooling the feature map over the superpixel.
} \label{fig:zoomoutCNN}
\end{figure*}

In order to incorporate the concept of multi-scale feature pooling, including
scene-level features, with state-of-the-art learned features, Mostajabi et al.
use a \emph{convolutional neural network} (CNN) architecture trained on scene
level classification. The feature computation at multiple scales centered on a
specific location in the image can be mapped directly to the filter response of
different convolutional layers in the CNN, corresponding to the different
spatial extents. Each filter response in a layer corresponds to a receptive
field in the image centered at a particular location. The layer response is a
three dimensional \emph{feature map}, that is $w\times h\times d$ dimensional.
Responses from different convolutional layers have different feature dimension,
$d$. As you move further up the CNN the feature maps have progressively smaller
spatial extents, $w\times h$, due to convolutional kernel stride and feature
pooling layers. Inversely, feature map locations from further layers in the CNN
have larger receptive field (region of influence) in the image. In order to
replicate the zoom-out idea with a CNN the responses of the convolutional layers
are upsampled so that their spatial extents match the image size, and
subsequently pooled over superpixels to produce scale-space features.
Figure~\ref{fig:zoomoutCNN} illustrates how the features for a superpixel are
extracted from a CNN. Scene level features are also extracted from the CNN as
the final softmax probabilities for each semantic category from the last layer
of the CNN.

The CNN can be arbitrary, though deeper networks provide more
spatial scales to consider. Mostajabi et al used the propular VGG-16
convolutional neural network~\cite{simonyan2014} as the backbone, that is initially trained on the
scene classification task.

The superpixel zoom-out features are next classified. Mostajabi et al.
experiment with both linear and non-linear shallow \emph{multilayer perceptron}
classifiers, which are trained on Zoom-out features extracted from the
training set. The classifier loss they minimize is the standard category
classification cross-entropy loss. Note that the zoom-out feature classifier MLP
and CNN used for extracting the superpixel features can be combined and trained
in an end-to-end fashion so that the CNN feature extraction layers can benefit
from supervision on the task of semantic segmentation over superpixels (as
opposed to just scene level classifaction supervision). This segmentation task
specific supervision further improves the accuracy (cf.~\cite{mostajabi2015}).

\begin{figure*}[!t] \centering \begin{tabular}{cccccc}
    \includegraphics[width=.15\textwidth]{\main/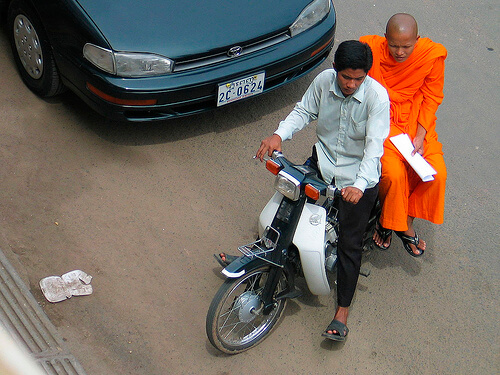} &
    \includegraphics[width=.15\textwidth]{\main/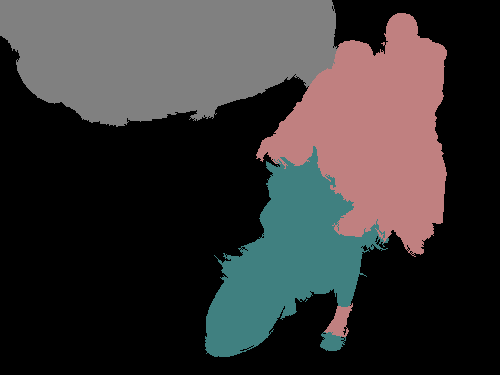}&
    \includegraphics[width=.15\textwidth]{\main/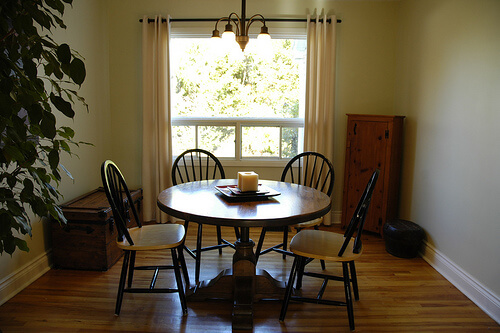} &
    \includegraphics[width=.15\textwidth]{\main/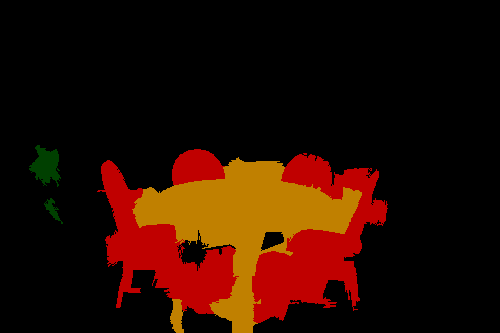}&
    \includegraphics[width=.15\textwidth]{\main/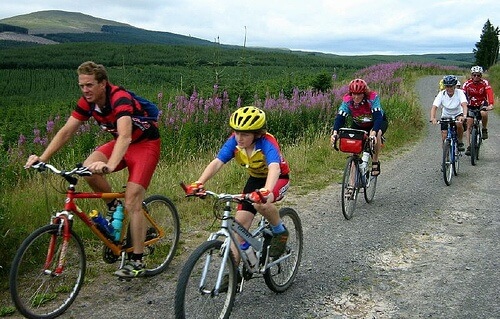} &
    \includegraphics[width=.15\textwidth]{\main/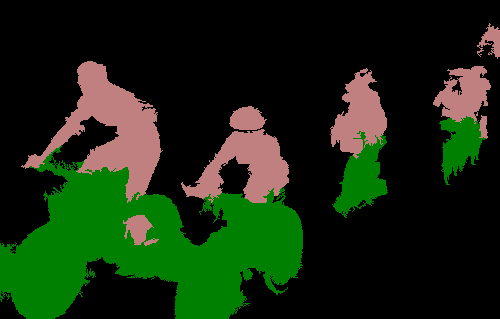}\\
    \includegraphics[width=.15\textwidth]{\main/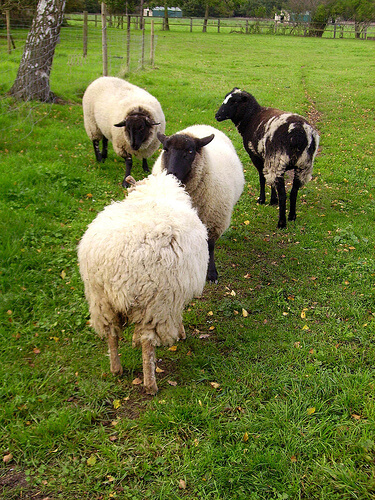} &
    \includegraphics[width=.15\textwidth]{\main/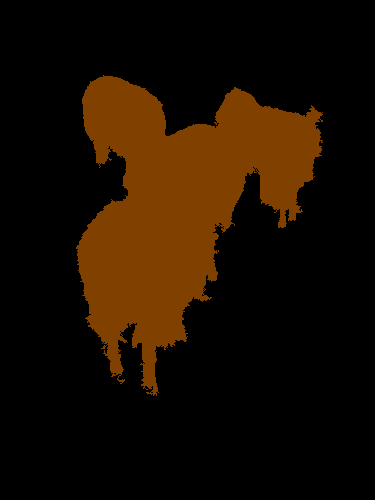}&
    \includegraphics[width=.15\textwidth]{\main/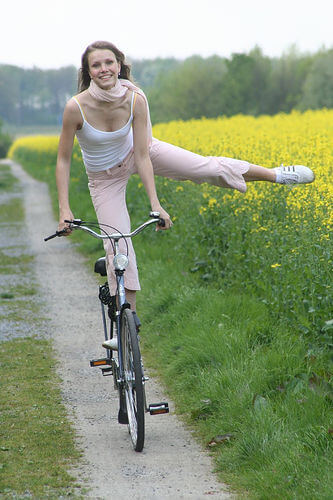} &
    \includegraphics[width=.15\textwidth]{\main/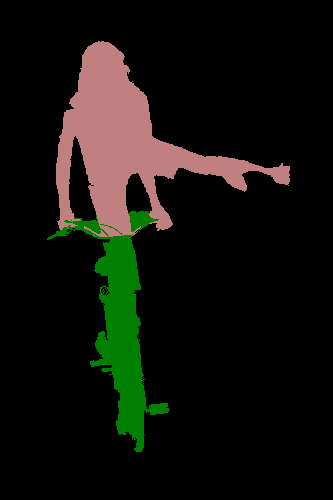}&
    \includegraphics[width=.15\textwidth]{\main/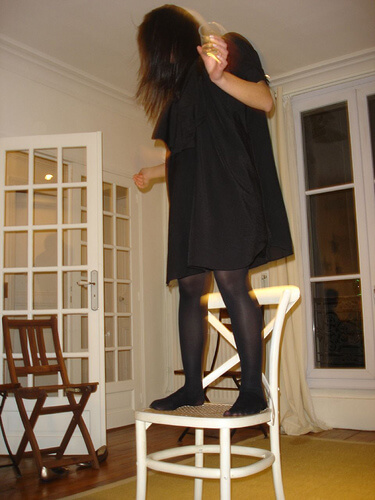} &
    \includegraphics[width=.15\textwidth]{\main/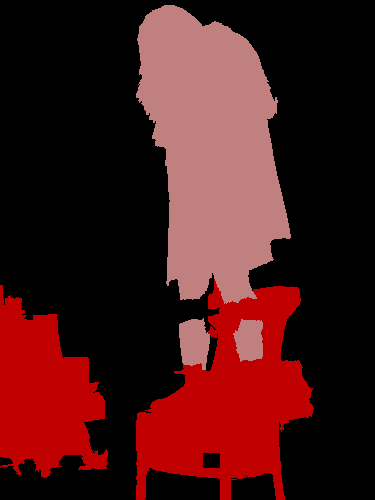}
\end{tabular}
\caption{Example semantic segmentation on VOC2012 val images using
a 3-layer perceptron classifier used to classify zoom-out features over
superpixels across 15 zoom-out levels of a CNN originally trained for scene
classification.} \label{fig:zoomoutresults}
\end{figure*}

Experimental analysis of the relative importance of the features extracted at
different scales shows that they all contribute a non-negligible signal toward
the prediction of semantic labels~\cite{mostajabi2015}. The competitive performance
(relative to state-of-the-art semantic segmentation algorithms) of the Zoom-out
network coupled with its relative simplicity --- and computational efficiency of
a feed-forward model (relative to alternative structured prediction models) ---
make it an attractive semantic segmentation model.
Figure~\ref{fig:zoomoutresults} displays some typical segmentation results from
the Zoom-out network.

The Zoom-out network of Mostajabi~\etal can also be applied densely (to every pixel) instead of over superpixels. This is done by first converting the CNN backbone into a fully-convolutional CNN where the final fully-connected classification layers are converted to $1\times 1$-convolutional layers. Additional \emph{skip-connections} are introduced that take the output response of the intermediate convolutional layers, and pass them through a concatenation layer that stacks the intermediate feature maps (after up-sampling to be the same size as the image using bilinear interpolation) into a \emph{hypercolumn}~\cite{hariharan2015} representation for every pixel in the output. This feature map is then fed into the final fully-convolutional classification layers of the network followed by softmax activation to make predictions at every pixel. Further dense refinement of the label predictions can then be made by applying efficient approximate inference on a special dense (\ie fully-connected) CRF~\cite{krahenbuhl2011,adams2010} over the pixels. This CRF uses the pixel label probabilities as unary potentials. The fully-convolutional CNN backbone (minus the CRF) can then be learned in an end-to-end fashion. Typically the CNN is initially trained on a complimentary task (and dataset) such as image classification and then converted into a fully-convolutional network with skip-connections which is then finetuned on the semantic segmentation task (and corresponding dataset). A similar approach is taken by contemporary works such as~\cite{hariharan2015,long2015,chen2014,chen2016}.

%% file: Sections/MAP.tex
\section{MAP Problem} \label{sec:mapproblem}
A \emph{Markov network} --- also known as a \emph{Markov random field} ---  is defined an undirected graph, $G=(V,E)$, with a set of vertices, $V$, associated with
$n=|V|$ random variables $\bX=\{X_v | v\in V\}$, and a set of edges, $E$,
between variables associated with the probabilistic relationships between those
variables. A Markov network structure encodes the dependence assumptions
associated with the random variables. The variables in the Markov network have
the following \emph{Markov properties}:
\begin{itemize}
    \item any two non-neighboring variables are conditionally independent given all other variables in $G$,
    \begin{flalign}
        X_s \independent X_t\; |\; X_{V\setminus\{s,t\}}, \quad \forall (s,t)\not\in E   \end{flalign}
    \item any variable is conditionally independent of all other random variables
in $G$ given the random variables that are its immediate neighbors (\ie those it shares an edge with),
\begin{flalign}
    X_s \independent X_{V\setminus \{s \cup ne(s)\}} \;|\; X_{ne(s)}
\end{flalign} This neighborhood of a node is referred to as its \emph{Markov blanket}.
    \item any two subsets of variables, $S_1$ and $S_2$, in $G$ are
        conditionally independent given the subset of variable that connect
    them,
\begin{flalign}
    X_{S_1} \independent X_{S_2}\; | \; X_C,
\end{flalign}
where every path from $X_{S_1}$ to $X_{S_2}$ in $G$ pass through some
variable(s) in $X_C$. \end{itemize} We will restrict ourselves to
discrete Markov random fields where each random variable $X_s$ takes
values from the finite label set $\mathcal{X}_s \coloneqq
\{0,\dots,r-1\}$, so that the random vector $\bX \in \mathcal{X}^n
\coloneqq \mathcal{X}_1 \times \cdots \times \mathcal{X}_n$, encodes the
joint configuration over all the variables.

According to the \emph{Hammersley-Clifford theorem} any probability distribution
that is strictly positive satisfies the above Markov properties if and only if
it can be factorized according to cliques of the graph. Therefore a Markov
random field where $p(\bx)>0$ for all $\bx\in \mathcal{X}^n$ can be written as,
\begin{flalign}\label{eqn:gibbs}
    p(\bx)=\frac{1}{Z}\prod\limits_{c\in\calC} \psi_c(\bx_c),
\end{flalign}
where $\calC$ is the set of cliques in $G$, $\bx_c=\{x_s\;|\; s\in c\}$, and $\psi_c(\bx_c)$ are functions from $X_c \rightarrow \mathbb{R}$ called \emph{factors} or $\emph{clique potentials}$. A
\emph{clique} is a maximal subgraph of $G$. A clique, $\bX_c$, is \emph{maximal}
if any superset containing $X_c$ is not a clique. Equation~\ref{eqn:gibbs} is
referred to as either the Gibbs distribution, Gibb random field, or Markov
random field.  The normalizing constant $Z$ is called the partition function,
\begin{flalign}
    Z=\sum\limits_{\bX \in \calX^n} \tilde{p}(\bX),
\end{flalign}
with unnormalized value,
\begin{flalign}
    \tilde{p}(\bX)=\prod\limits_{c\in\calC} \psi_c(\bx_c)
\end{flalign}
Equivalently, the set $\mathcal{C}$ in the factorization of
eqn.~\ref{eqn:gibbs} can be restricted to only contain maximal cliques.
This is because any factorization over complete subgraphs can be
equivalently written as a factorization over maximal cliques with
corresponding clique potentials that are products over all factors who's
scope is covered by the maximal clique.

The structure of the Markov network generally does not capture the factorization
according to the Gibbs distribution, because the factorization could be over
maximal or non-maximal cliques and the Markov network does not show this
explicitly. An alternate parameterization of the Markov network is via a
\emph{factor graph}. In a factor graph additional factor nodes are introduced.
Edges are only between variable nodes and factor nodes. Each factor, $\psi$, in
the factorization is associated with a corresponding factor node in the graph.
Variables are connected with a factor node with an edge if the variables are
found in the scope of the factor corresponding to the factor node.

An even more explicit parameterization of a Markov network is via
\emph{log-linear models}. In this case each factor is written in an equivalent
\emph{energy function} form,
\begin{flalign}
    \psi(\bx) = \exp(-\gamma(\bx)),
\end{flalign}
with energy function $\gamma(\bx)=-\ln\psi(\bx)$. The
joint probability can then be written as,
\begin{flalign}
    p(\bx)=\frac{1}{Z}\exp\left(-\sum\limits_{c\in \calC} \gamma_c(\bx_c)\right).
\end{flalign}
To get the log-linear representation we can associate one or more
\emph{potential functions} with each clique $C$. More specifically assume a set
of potential functions $\{\phi_\alpha | \alpha \in \calI(C)\}$. A potential
function, $\phi : \calX^n \rightarrow \mathbb{R}$, associated with a clique $C$,
has $\bx_C$ as its scope, and $\calI(C)$ is some index set over $C$. Let $\calI
= \cup_C \calI(C)$ be the union over the index sets of all cliques. The log
linear model can be written as,
\begin{flalign} \label{eqn:loglinear}
    p(\bx; \theta) = \frac{1}{Z}
\exp\left(\sum\limits_{\alpha\in\calI} \theta_\alpha\phi_\alpha(\bx)\right),
\end{flalign}
with parameters $\theta_\alpha$. Notice that any energy
function over discrete variables, $\gamma_C(\bx_C)$, can be written as a weighted sum,
$\sum_{\alpha\in\calI(C)} \theta_\alpha \phi_\alpha(\bx)$, with appropriate
choice of potential functions and $\alpha$. The above definitions and characterizations of Markov networks can be found in more detail in~\cite{koller2009}.

In what follows we will review a re-characterization of the MAP inference problem
as an integer programming problem attributed to Wainwright \etal~\cite{wainwright2005,wainwright2008}.

If we let $\bfgreek{theta}=[\theta_1,\cdots,\theta_d]$, where $d=|\calI|$, for
the collection $\{\theta_\alpha\;|\; \alpha\in\calI\}$ and define the
mapping $\bfgreek{phi} : \calX^n \rightarrow \mathbb{R}^d$, for the collection
$\{\phi_\alpha\; |\; \alpha \in \calI\}$, we can write eqn.~\ref{eqn:loglinear} more
compactly as,
\begin{flalign}\label{eqn:expfamily}
    p(\bx;\bfgreek{theta})=\frac{1}{Z} \exp \langle
        \bfgreek{theta},\bfgreek{phi}\rangle.
\end{flalign}
Equation~\ref{eqn:expfamily} defines a \emph{linear exponential family} of
distributions --- therefore discrete MRFs are linear exponential families. Each $\bfgreek{theta}$ defines a different MRF.

The class of random fields that we will focus on are \emph{metric} MRFs with discrete random
variables and at most pairwise factors. In fact any Markov random field with
discrete random variables and higher order factors can be turned into an
equivalent MRF with only pairwise factors (see~\cite{wainwright2008}). For the pairwise MRF the index set
is,
\begin{flalign}\label{eqn:pairindexset}
    \calI\coloneqq\left\{(s;j)\; |\; s\in V,\; j\in X_s \right\}\; \cup\;
    \left\{(st;jk)\; | \; (s,t)\in V,\; (j,k)\in X_s\times X_t \right\}
\end{flalign}

The metric pairwise MRF potential functions, $\bfgreek{phi}$, take specific form of indicator functions.
Specifically, the node and parwise interaction potentials are,
\begin{flalign}\label{eqn:nodepot}
    \mathbb{I}_{s;j}(x_s) \coloneqq \left\{\begin{array}{ll}
            1 & \mbox{if $x_s=j$,}\\
            0 & \mbox{otherwise}
        \end{array}\right. \quad \forall s\in V,\; j\in \calX_s ,
\end{flalign}
\begin{flalign}\label{eqn:pairpot}
    \mathbb{I}_{st;jk}(x_s,x_t) \coloneqq \left\{\begin{array}{ll}
            1 & \mbox{if $x_s=j$ and $x_t=k$,}\\
            0 & \mbox{otherwise}
    \end{array}\right. \quad \forall (s,t)\in E,\; (j,k)\in
    \calX_s\times\calX_t,
\end{flalign}
and are referred to as the \emph{canonical overcomplete representation}, with
the corresponding $\bfgreek{theta}$ called the \emph{canonical parameters}~\cite{wainwright2005}.
The representation is overcomplete because they satisfy certain linear
constraints, namely,
\begin{subequations}
\begin{flalign}
    \sum\limits_{j\in \calX_s} \mathbb{I}_{s;j}(x_s) &= 1 \qquad \forall s\in V, \label{eqn:canonlocal1}\\
    \sum\limits_{(j,k)\in \calX_s\times \calX_t} \mathbb{I}_{st;jk}(x_s,x_t) &= 1
    \qquad \forall (s,t) \in E, \label{eqn:canonlocal2}\\
    \sum\limits_{j\in \calX_s} \mathbb{I}_{st;jk}(x_s,x_t) &= \mathbb{I}_{t;k}(x_t)
    \qquad \forall (s,t) \in E \quad \forall k \in \calX_t, \label{eqn:canonlocal3}
\end{flalign}
\end{subequations}
Plugging eqns.~\ref{eqn:nodepot} and~\ref{eqn:pairpot} into
eqn.~\ref{eqn:expfamily} and using the index set in eqn.~\ref{eqn:pairindexset}
the joint probability distribution for the pairwise MRF can be written as,
\begin{flalign}\label{eqn:pairmrfprob}
    p(\bx; \bfgreek{theta})=
    \exp\left(\sum\limits_{s\in V}\sum\limits_{j\in \calX_s} \theta_{s;j}
        \mathbb{I}_{s;j}(x_s) + \sum\limits_{(s,t)\in E} \sum\limits_{(j,k)\in
        \calX_s\times\calX_t} \theta_{st;jk}\mathbb{I}_{st;jk}(x_s,x_t) -
    A(\bfgreek{theta})\right),
\end{flalign}
where $A(\bfgreek{theta}) \coloneqq \ln Z(\bfgreek{theta})$. A more compact
representation can be written if the following substitutions are made,
\begin{flalign}
    \theta_s(x_s) \coloneqq \sum\limits_{j\in \calX_s} \theta_{s;j}\mathbb{I}_{s;j}(x_s),
\end{flalign}
\begin{flalign} \theta_{st}(x_s,x_t) \coloneqq \sum\limits_{(j,k)\in
    \calX_s\times\calX_t}
    \theta_{st;jk}\mathbb{I}_{st;jk}(x_st,x_t),
\end{flalign}
into eqn.~\ref{eqn:pairmrfprob} resulting in the pairwise MRF joint probability
distribution,
\begin{flalign}\label{eqn:mrfprobcompact}
    p(\bx;\bfgreek{theta})=\exp\left(\sum\limits_{s\in V} \theta_s(x_s) +
    \sum\limits_{(s,t)\in E} \theta_{st}(x_s,x_t) - A(\bfgreek{theta})\right).
\end{flalign}
In this thesis we will only be concerned with a specific type of inference
problem --- namely finding the joint configuration $\bx$ that maximizes the
distribution $p(\bx;\bfgreek{theta})$ specified by a particular
$\bfgreek{theta}$. This is known as the \emph{maximum a posterior} or MAP
assignment problem. Note that there can be multiple maximizing assignments. Formally we want to find an assignment $\bx^*$ such that,
\begin{flalign}
    \bx^* \in \{\bx \in \calX^n \;|\; p(\bx;\bfgreek{theta}) \geq
    p(\by;\bfgreek{theta}),\; \forall \by \in \calX^n\}
\end{flalign}
Notice that in eqn.~\ref{eqn:mrfprobcompact},  $A(\bfgreek{theta})$ is
independent of $\bx$, so the value of the assignment that maximizes
the joint probability is equivalent to,
\begin{flalign}\label{eqn:pairmap}
    \max\limits_{\bx \in \calX^n} \langle
    \bfgreek{theta},\bfgreek{phi}(\bx)\rangle \coloneqq
    \max\limits_{\bx \in \calX^n} \sum\limits_{s\in V} \theta_s(x_s) +
    \sum\limits_{(s,t)\in E} \theta_{st}(x_s,x_t).
\end{flalign}
A maximizing assignment is thus,
\begin{flalign}
    x^*\coloneqq \argmax\limits_{\bx\in\calX^n}\langle
        \bfgreek{theta},\bfgreek{phi}(\bx)\rangle.
\end{flalign}

Equation~\ref{eqn:pairmap} is an linear integer program (IP) because $\bx\in\calX^n$ take
on integer values and both the constraint set and the objective function are
linear. This integer program is a discrete combinatorial optimization problem
that is known to be NP-hard to solve for general graphs~\cite{wainwright2008}. To overcome this
problem the integer program can be relaxed into a continuous \emph{linear
program}.

Take the set  $\calP\coloneqq \{p(\bx) \;|\; p(\bx)\geq 0,\; \sum_{\bx}
p(\bx)=1\}$ of all probability distributions on $\bx$. It is easy to see that
the following equality holds,
\begin{flalign}\label{eqn:maxeq}
    \max\limits_{\bx\in\calX^n}\langle \btheta, \bphi(\bx)\rangle =
    \max\limits_{p\in \calP} \sum\limits_{\bx\in \calX^n} p(\bx)\langle
        \btheta,\bphi(\bx)\rangle,
\end{flalign}
because for any $\bx$ that satisfies the LHS there exists a probability
distribution $p(\cdot)$ that puts all probability mass on $\bx$ so the value of the RHS is
at least as large as the LHS. The RHS is also a convex combination of $\langle
\btheta,\phi(\bx)\rangle$ terms so cannot be any larger than
$\max\limits_{\bx\in\calX^n} \langle \btheta,\bphi(\bx)\rangle$.

Expanding the RHS term of eqn.~\ref{eqn:maxeq} we get,
\begin{flalign}
    \max\limits_{\bx\in\calX^n}\langle \btheta, \bphi(\bx)\rangle &=
    \max\limits_{p\in \calP} \sum\limits_{\bx\in \calX^n} p(\bx)
    \left(\sum\limits_{s\in V} \theta_s(x_s) +
    \sum\limits_{(s,t)\in E}\theta_{(s,t)}(x_s,x_t)\right)\\
    &=\max\limits_{p\in \calP} \sum\limits_{s\in V} \sum\limits_{j\in\calX_s}
    \theta_{s;j}\sum\limits_{\bx\in \calX^n}p(\bx)\mathbb{I}_{s;j}(x_s) +
    \label{eqn:margprobexpand}
    \\\nonumber
    &\qquad+ \sum\limits_{(s,t)\in E}\sum\limits_{(j,k)\in \calX_s\times
    \calX_t}
    \theta_{st;jk}\sum\limits_{\bx\in\calX^n}p(\bx)\mathbb{I}_{st;jk}(x_s,x_t),
\end{flalign}
We can define the following quantities,
\begin{flalign}
    \mu_{s;j} &\coloneqq \sum\limits_{\bx\in
    \calX^n}p(\bx)\mathbb{I}_{s;j}(x_s)= \mathbb{E}_p[\mathbb{I}_{s;j}(x_s)] =
    \mathbb{P}[X_s=j], \label{eqn:unarymarg}\\
    \mu_{st;jk} &\coloneqq
    \sum\limits_{\bx\in\calX^n}p(\bx)\mathbb{I}_{st;jk}(x_s,x_t) =
    \mathbb{E}_p[\mathbb{I}_{st;jk}(x_s,x_t)]=\mathbb{P}[X_s=j \wedge
    X_t=k],\label{eqn:pairmarg}
\end{flalign}
called mean parameters which have intuitive meaning --- namely
$\mu_{s;j}$ is the node marginal probability that random variable $x_s$ takes label
$j$, and $\mu_{st;jk}$ is the edge marginal probability of the joint assignment
$(x_s=j, x_t=k)$. Plugging definitions~\ref{eqn:unarymarg}
and~\ref{eqn:pairmarg} into eqn.~\ref{eqn:margprobexpand} we get,
\begin{flalign}\label{eqn:meanmap}
    \max\limits_{\bx\in\calX^n}\langle \btheta, \bphi(\bx)\rangle &=
    \max\limits_{p\in \calP} \sum\limits_{s\in V} \sum\limits_{j\in\calX_s}
    \theta_{s;j} \mu_{s;j} + \sum\limits_{(s,t)\in E}\sum\limits_{(j,k)\in \calX_s\times
\calX_t}\theta_{st;jk}\mu_{st;jk}.
\end{flalign}
Similar to $\bphi$ we let $\bmu=\{\mu_\alpha\;|\; \alpha \in \calI\}$.
We can define the set of all possible marginal probabilities on graph $G$ as,
\begin{flalign}\label{eqn:marginalpolytope}
    \mathbb{M}(G) \coloneqq \{\bmu \in \mathbb{R}^d \;|\; \exists p(\cdot)\in
        \calP \; \st \; \mu_{s;j}
    = \mathbb{E}_p[\mathbb{I}_{s;j}(x_s)],\; \mu_{st;jk}=
\mathbb{E}_p[\mathbb{I}_{st;jk}(x_s,x_t)]  \}.
\end{flalign}
$\mathbb{M}(G)$ is called a \emph{marginal polytope}~\cite{wainwright2005}.
From $\mathbb{M}(G)$ we
see that searching over probability distributions $\calP$ maps to searching over
$\bmu$, which means that the RHS of eqn.~\ref{eqn:meanmap} can be written as,
\begin{flalign}\label{eqn:meanmap2}
    \max\limits_{\bx\in\calX^n}\langle \btheta, \bphi(\bx)\rangle &=
    \max\limits_{\mu\in \mathbb{M}(G)} \sum\limits_{s\in V} \sum\limits_{j\in\calX_s}
    \theta_{s;j} \mu_{s;j} + \sum\limits_{(s,t)\in E}\sum\limits_{(j,k)\in \calX_s\times
\calX_t}\theta_{st;jk}\mu_{st;jk}\\
&=\max\limits_{\bmu\in \mathbb{M}(G)} \langle \btheta, \bmu \rangle
\label{eqn:maplinearprog}.
\end{flalign}
Equation~\ref{eqn:maplinearprog} is the linear programming (LP) relaxation of the
original MAP integer program.

Note that $\mathbb{M}(G)$ is the convex hull of the overcomplete representation
defined in eqn.~\ref{eqn:nodepot} and eqn.~\ref{eqn:pairpot} over the finite index set
in eqn.~\ref{eqn:pairindexset}. These indicator functions define the extreme points of
$\mathbb{M}(G)$. Since the extreme points are indicator functions they take
$\{0,1\}$ values, which means they are all integral. From standard linear
programming optimization theory the optimal solution of an LP always lies at an
extreme point of the feasible set (\ie one of the vertices of $\mathbb{M}(G)$).

For each $\bx\in \calX^n$ the canonical overcomplete
representation $\bphi(\bx)$ corresponds to an extreme point $\bmu_{\bx}$ of
$\mathbb{M}(G)$, thus the optimal solution is integral and in one-to-one
correspondence with assignments $\bx$. Moreover this means that $\mathbb{M}(G)$ has $|\calX^n|$ extreme points, which is
exponential in $n$. Optimization over an exponential number of constraints is
not feasible so a simpler (read fewer constraints) outer bound on the marginal polytope is desired.

According to the \emph{Minkowski-Weyl theorem} any convex hull over a finite set of vectors can be represented
equivalently by the intersection of a finite number of linear half-spaces
(\ie of the form $\{\bmu: \ba^T\bmu \leq b\}$ for some $\ba \in \mathbb{R}^d$ and $b\in
\mathbb{R}$). Included in the half-space representation of $\mathbb{M}(G)$ the linear inequality (half-space)
constraints also include the equality constraints that are a consequence of the
overcomplete representation. They are analogous to the consistency constraints in
eqns.~\ref{eqn:canonlocal1}--\ref{eqn:canonlocal3}, namely,
\begin{subequations}
\begin{flalign}
    \sum\limits_{j\in \calX_s} \mu_{s;j}(x_s) &= 1 \qquad \forall s\in V, \label{eqn:marginconsist1}\\
    \sum\limits_{(j,k)\in \calX_s\times \calX_t} \mu_{st;jk}(x_s,x_t) &= 1 \qquad \forall
    (s,t)\in E,
    \label{eqn:marginconsist2}\\
    \sum\limits_{j\in \calX_s} \mu_{st;jk}(x_s,x_t) &= \mu_{t;k}(x_t) \qquad \forall
    (s,t) \in E \quad \forall k\in \calX_t,
    \label{eqn:marginconsist3}
\end{flalign}
\end{subequations}
as well as non-negativity constraints on the marginal probabilities (\ie
$\mu_\alpha \geq 0$ for all $\alpha \in \calI$). Note that each of the above equality
constraints can be written as two inequality constraints --- \ie the constraint
$a^T\mu=b$ is equivalent to maintaining the following two constraints: $a^T\mu \leq b$ and $-a^T\mu \leq -b$.

It turns out that for general graphs with cycles representing the marginal
polytope $\mathbb{M}(G)$ as an intersection of half-spaces, or $\emph{facets}$,
becomes difficult because the number of half-spaces becomes exponential. Other
than for tree structured graphs the number of facets of $\mathbb{M}(G)$ in a general graph are not known. Instead a simpler outer
bound on $\mathbb{M}(G)$ can be constructed by simply considering the
intersection of a subset of the half-space constraints, namely those in
eqns.~\ref{eqn:marginconsist1}--~\ref{eqn:marginconsist3}, along with the
non-negativity constraint on $\bmu$. This gives the \emph{local polytope} which
is set of locally consistent marginal distributionsi~\cite{wainwright2005,wainwright2008},
\begin{flalign}
    \mathbb{L}(G) \coloneqq \{ \bmu \in \mathbb{R}^d_+ \;|\;
        \mbox{eqns.~\ref{eqn:marginconsist1}--~\ref{eqn:marginconsist3}
    hold}\}\label{eqn:localpolytope}
\end{flalign}
$\mathbb{L}(G)$ is the intersection of a subset of the half-space
constraints required to represent $\mathbb{M}(G)$, thus $\mathbb{M}(G)$ is a
subset of $\mathbb{L}(G)$. It can be shown~\cite{wainwright2005} that for trees (\ie any
acyclic connected graph) $\mathbb{L}(G)=\mathbb{M}(G)$. For general graphs with cycles though
$\mathbb{M}(G)$ will be a strict subset of $\mathbb{L}(G)$. The number of facets of
$\mathbb{L}(G)$ is polynomial in graph size. The number of extreme points of $\mathbb{L}(G)$
is larger than $\mathbb{M}(G)$ for general graphs, which include the integral
extreme points, $\{\bmu_{\bx} \;|\; \bx \in \calX^n\}$, plus a set of fractional
extreme points that lie outside of $\mathbb{M}(G)$, the total number of which is
unknown for general graphs. But that fact that $\mathbb{L}(G)$ can be
represented as polynomial number of facets (\ie inequality constraints) means
that the following alternate problem,
\begin{flalign}
    \max\limits_{\bmu\in \mathbb{L}(G)}\langle \btheta, \bmu \rangle
\end{flalign}
can be efficiently solved. From eqns.~\ref{eqn:meanmap2}--\ref{eqn:maplinearprog} and the definition of the
local polytope in eqn.~\ref{eqn:localpolytope} we have the following relations,
\begin{flalign}\label{eqn:mapeqns}
    \max\limits_{\bx\in\calX^n}\langle \btheta, \bphi(\bx)\rangle =
    \max\limits_{\bmu\in \mathbb{M}(G)}\langle \btheta, \bmu \rangle \leq
    \max\limits_{\bmu\in \mathbb{L}(G)}\langle \btheta, \bmu \rangle
\end{flalign}
The relaxation on the RHS of eqn.~\ref{eqn:mapeqns} is tight for tree structured
graphs, but is not guaranteed to be tight for general graphs with cycles. Solutions to the
RHS are optimal (\ie the relaxation is tight) if they lie at one of the integral
vertices but on general graphs with cycles the solutions are often at one of the
fractional vertices. Much work has been done to develop algorithms that give the
tightest upper bound on the solution, and the relationship between the
the RHS relaxation (and its dual) and various efficient approximate MAP
inference algorithms such as \emph{tree-reweighted max-product
message-passing}~\cite{wainwright2005} and \emph{dual decomposition}~\cite{komodakis2007}
have been established.

The MAP integer program and its linear relaxation formulation described above can be credited to Wainwright~\etal and a more detailed exposition can be found in the respective material~\cite{wainwright2005,wainwright2008}.

%% file: Chapters/Chapter02.tex
\chapter{DivMBest}\label{ch:divmbest}
%************************************************
\subfile{\main/Sections/DivmbestRW}
\printbibliography

%% file: Sections/DivmbestRW.tex
The primary objective of the research efforts described in this thesis is to improve semantic segmentation.
Typically, improving on an existing segmentation model means devising a new
model that produces more accurate segmentations. There are a number of sources
of error in any model that need to be addressed in order to improve upon it.
\emph{Approximation error} --- the error due to limitations imposed by the choice of model
class --- is addressed by devising more accurate, and often more complex, models
for semantic segmentation. Unfortunately, as is often the case, more complex
models exhibit more optimization and estimation error. When models
become too complex for exact inference the approximate inference surrogate
methods introduce \emph{optimization error}. More complex models often incorporate
higher order interactions between variables and typically have more free
variables --- all of which require more examples to train. The limitation of a
finite training set to train the more complex model leads to larger
\emph{estimation error}. Worse yet, it may not even be clear how to incorporate
certain higher-order information into a model for segmentation. Even if we are
able to, we may end up with models that are intractable to train or do inference
on.

For all of the above reasons coming up with a new model can be difficult. Suppose we would like to
improve upon an existing discrete semantic segmentation model which can be
trained efficiently on a finite training set and on which inference is
tractable. Without loss of generality, given an image, the model is trained to
minimize the average error over the training set between the segmentation it
produces and the ground-truth for the image. At test time, given an image, we
use the trained model to infer the most likely (read probable) semantic
segmentation of the image. When we are working with a probabilistic model this
segmentation is the \emph{maximum a-posteriori} (MAP) solution (or MAP
assignment). Without loss of generality we'll use MAP solution to mean the most
likely segmentation returned by the model irrespective of it being a
probabilistic model or not. We can assume that the model assigns a score to
every possible labelling (\ie assignment to all the variables over the image), indicating how likely
the labeling is a correct segmentation of the image. Alternatively, we can associate a
probability with the likelihood of the image returning a certain segmentation for
an image.

When we devise a more accurate segmentation model in effect what we want to
achieve is a model that produces a MAP solution that is at least as close to the
ground-truth segmentation for the image as the MAP solution produced by a less
accurate model. However, the more accurate model might be intractable or at best
comes at a cost of higher estimation and approximation error.

Instead we can consider using the less accurate model to output multiple highly
probable solutions, not just the MAP. One of these other solutions might be a
more accurate segmentation of the image. We can then consider returning a set of
segmentation for the image or pick a single one from the set. If we consider
that we are reasonably confident in our sub-optimal segmentation model to return to give high probablity to solutions that "do
the right thing" in many areas of a typical image, then by producing multiple
high probability segmentations from the model we are considering alternate
explanations that the model exhibits for the same image. This approach is
analogous to cascade models~\cite{sapp2010,viola2004,weiss2010} where successive stages of the cascade refine
the output of previous stages. Since the space of segmentations (\ie labelings
over (super)pixels) is exponential, by producing an initial set of
segmentations, instead of just the MAP, and then refining it simplifies the
inference problem by reducing it from a 1-out-of-$|\calL|^n$ to a 1-out-of-$M$
inference task (where $\calL$ is the cardinality of the label set on each of the
$n$ variables, and $M$ is the number of high probability segmentations in the
set, where $M \ll |\calL|^n$). By producing not just one but multiple high
probability segmentations we are simultaneously providing an explicit way to
manage the uncertainty in the model. That is to say, compared to a single MAP solution, we are better summarizing the uncertanties that the model has about the output space of labellings.

The problem of producing the $M$ most probable solutions, which are different
from MAP and each other, is called the \mbest MAP problem. We will see that in
fact this idea has been studied in the context of problems outside segmentation
and vision. We will review the most well known approaches.

We will show that for the semantic segmentation task the \mbest segmentations
are not an ideal set of segmentations of an image. This is because the \mbest
formulation only enforces the segmentations not to be the same --- there is no
explicit control on how diverse the segmentations in the set are. Contrary to
problems in other domains, generating a set of segmentations that are
simultaneously \emph{highly probable} and \emph{diverse} is a better way of
managing uncertainty in the segmentation task. To this end, this chapter
presents the \divmbest problem that produces a set of highly probable
segmentations that are different from the MAP solution (and each other), with explicit control
over the amount of diversity between solutions. Intuitively the goal of the
\divmbest approach is to construct a set of segmentations that correspond to the
modes of the output distribution of the underlying segmentation model.
We will show that \divmbest is a very general framework that can be applied to
virtually any tractable segmentation model, and it can be particularly efficient
when we consider special forms of dissimilarity between solutions. In fact
\divmbest, like the \mbest algorithm, can be applied to any problem that can
benefit from inferring more than just the MAP solution. We also show that
\divmbest generalized the \mbest method and more generally contains other
related formulations as special cases.

For the case of a probabilistic segmentation model, a simple alternate way to generate multiple segmentations is to sample from it, such as with Markov chain Monte Carlo (MCMC) sampling. There have been a number of works~\cite{barbu2005,park2011,porway2011} that take this strategy. It could, however, be a prohibitively time consuming approach, because of the time required to return samples from modes with small support. Additionaly, in contrast to the \divmbest approach, there isn't any explicit control over diversity in the set of sampled solutions, which would necessitate sampling a larger set to cover the space of alternate explanations of the image. Related to \divmbest, Papandreou and Yuille~\cite{papandreou2011} present an approach (perturb-and-MAP) to sample from a discrete probabilistic model (e.g. random field) by perturbing the model parameters with random noise and solving for the MAP solution using existing discrete optimization algorithms. This extends deterministic MAP inference to non-deterministic iid sampling of the model distribution. In contrast to perturb-and-MAP the \divmbest approach modifies the model parameters in a deterministic way resulting in a set of highly probable diverse solutions.

In this chapter we review the related \mbest algorithms and present the
\divmbest formulation, limiting the discussion to the case of discrete
probabilistic models for ease of exposition. We formulate the \divmbest problem
as an integer program (see~\cref{sec:mapproblem}) minimizing a discrete energy (probability distribution) model over a set
of random variables subject to diversity constraints on the solutions and
consider a linear programming relaxation of it. We present a greedy iterative
algorithm to efficiently compute the \divmbest solutions, as well as a gradient
ascent method to set search over the diversity parameters. We show that for
certain measures of diversity the LP dual of the \divmbest problem enjoys some
nice theoretical guarantees. In subsequent chapters we show the superiority of
\divmbest over MAP, and \mbest MAP inference, for various segmentation tasks. In
order to handle the 1-out-of-M inference task we also introduce an approach to
rank the \divmbest segmentation sets in order to return a single segmentation.

\section{Related Work}

In the next section we begin with a review of the \mbest MAP problem and related literature.

\subsection{\mbest}\label{sec:lawlermbest}
One of the earliest methods to address the \mbest problem was by
Lawler~\cite{lawler1972}.
It is a simple and general method for computing the $M$ optimal solutions of
discrete optimization problems, and is agnostic to the optimization algorithm
used to compute the solution of any specific problem. It is a divide-and-conquer
method that solves multiple independent discrete optimization problems that are
created by iteratively partitioning the assignment space. We'll outline the
basic method here (cf. ~\cite{lawler1972}).

Without loss of generality the method assumes a set of binary variables,
$x_1,\dots,x_n\in\{0,1\}$. Note that in problems where the discrete
variables take on more than two values one can make a straightforward
transformation to $\{0,1\}$-valued variables --- \eg  $\forall x_i\in
\{0,\dots,\ell-1\}$ replace with $\ell$ variables: $x_{i;j}\in \{0,1\}$, where
$x_{i;j}=[\![x_i==j]\!]$, with an implicit constraint that $\sum_{j}
x_{i;j}=1$.

The task is to return the top-M solutions to a discrete
optimization problem (\wlg assume a minimization) over the $x_i$'s -- that is the $M$ solutions that best minimize the problem. The method starts by computing the optimal solution to the original problem. It
then iteratively partitions the assignment space into disjoint sets in a way that
removes the previous top $m-1$ solutions from consideration, and for each set solves a new
optimization problem returning a candidate solution, whereby
the next best solution is the one in the set of candidates with lowest value.

\begin{algorithm}[!h] \caption{\mbest-Lawler~\cite{lawler1972}}\label{alg:lawlermbest}
\begin{algorithmic}[1]
    \State $\emph{OPTS}=\emptyset$, $\emph{TOP-M}=\emptyset$.
    \State $m\gets 1$: compute optimal solution $\bx^{(1)}\gets\argmax_{\bx} f(\bx)$,
    without fixing any variables in $\bx$
    \State $\emph{OPTS}\gets \emph{OPTS} \cup (\bx^{(1)},\emptyset)$.
    \Repeat
    \State $(\bx^{(m)},P^{(m)}) \gets \texttt{bestsol}(\emph{OPTS})$,\quad $\emph{TOP-M}\gets
        \bx^{(m)}$
    \Comment{{\tiny {\texttt bestsol} returns the minimum value solution along with the
    set of fixed variables in the corresponding problem $P^{(m)}$.}}
        \If{ $m=M$ }
            \State break;
        \EndIf
        \State \parbox[t]{\dimexpr\linewidth-\algorithmicindent\relax}{%
                \setlength{\hangindent}{\algorithmicindent}%
            Let $x_1,\dots, x_s$ be the variables that were fixed
            in the problem solved to get $\bx^{(m)}$, \ie $P^{(m)}$. Construct $(n-s)$ new problems by fixing additional variables:
            \begin{itemize}
                \item [($P_1$):]
                    \begin{scriptsize}$x_1=x_1^{(m)},\dots,x_s=x_s^{(m)},\;x_{s+1}=1-x_{s+1}^{(m)}$
                    \end{scriptsize}
                \item [($P_2$):]
                    \begin{scriptsize}$x_1=x_1^{(m)},\dots,x_s=x_s^{(m)},\;x_{s+1}=x_{s+1}^{(m)},\;x_{s+2}=1-x_{s+2}^{(m)}$\end{scriptsize}
                \item [($P_3$):]
                \begin{scriptsize}$x_1=x_1^{(m)},\dots,x_s=x_s^{(m)},\;x_{s+1}=x_{s+1}^{(m)},\;x_{s+2}=x_{s+2}^{(m)},\;x_{s+3}=1-x_{s+3}^{(m)}$\end{scriptsize}
                \item [\vdots]\hspace{5em}\vdots
                \item [($P_{n-s}$):]
                \begin{scriptsize}$x_1=x_1^{(m)},\dots,x_s=x_s^{(m)},\;x_{s+1}=x_{s+1}^{(m)},\;x_{s+2}=x_{s+2}^{(m)},\dots,x_{n-1}=x_{n-1}^{(m)},x_n=1-x_n^{(m)}$\end{scriptsize}
            \end{itemize}
        }\strut
        \For {$j=1,\dots,n-s$}
            \State Solve $\bx^{P_j}\gets\argmin\limits_{\mbox{$\bx$ \st $P_j$ satisfied}} f(\bx)$
            \State $\emph{OPTS}\gets \emph{OPTS} \cup (\bx^{P_j},P_j)$.
        \EndFor
        \State $m \gets m+1$
    \Until{$m=M$}
\end{algorithmic}
\end{algorithm}
The algorithm is reproduced in alg.~\ref{alg:lawlermbest}. More specifically, in each iteration $m$, of alg.~\ref{alg:lawlermbest}, $n-s$ new problems are created,
where $s$ is the number of variables that were fixed in the optimization problem
that produced the previous solution $x^{(m-1)}$. The key, in line 9 of
alg.~\ref{alg:lawlermbest}, is to partition the assignment space into $n-s$ disjoint sets, $P_1,\dots,P_{n-s}$. Each problem $P_j$
has the first $s$ variables plus an additional $j$ variables fixed. The $j$
additional variables are fixed in such a way to remove $x^{(m-1)}$ from the set of
feasible solutions to $P_j$. Note that if for any $j$, $x_{s+j}$ is set to $1-x^{(m-1)}_{s+j}$, then $x^{(m-1)}$ has been removed from the set of feasable solutions for $P_j$. Additionally the set of feasible solutions
for $P_j$ is disjoint from the rest of the problems $\{P_i\; |\; i\neq j\}$. Moreover, $\bigcup\limits_{j=1}^{n-s} P_j= X - \{x^{(m-1)}\}$.

Since in each
iteration the $s$ variables from the problem used to produce solution $x^{(m-1)}$
remain fixed all the previous optimal solutions are also removed from
consideration in each problem, \ie in iteration $m$ we have $\bigcup\limits_{j=1}^{n-s} P_j= X -\{x^{1},\dots,x^{(m-1)}\}$.

Effectively alg.~\ref{alg:lawlermbest} recursively partitions the assignment
space. Each new partitioning occurs on the set of feasible solutions used to
constraint the problem that produced the optimal solution in the previous round.
For example in the first iteration the partitioning is over the entire
assignment space because no variables were fixed in the optimization problem that
compute the initial best assignment $x^{(1)}$.

The computational complexity of computing the top-M solutions using
alg.~\ref{alg:lawlermbest}  is $\bigO{Mn\rho(n)}$, where $\rho(n)$ is the cost of
solving a single optimization problem over $n$ variables. Since it is a general
\mbest algorithm it is not tailored to any specific discrete optimization
problem so it cannot simultaneously solve for the top $M$ solutions. This makes
the algorithm less efficient than specialized inference methods and in each
iteration of the algorithm $n-s$ separate optimization problems need to be solved.

The space required for the algorithm is $M(n-1)$ because at most that many items
are in $\emph{OPTS}$.

\subsection{\mbest and Max-Flow Propagation}
Dawid~\cite{dawid1992} and later Nilsson~\cite{nilsson1998} extended the \mbest task
to the problem of computing optimal configurations over directed and undirected
graphical models with cycles. Their approach relies on the ability to exactly
and efficiently compute the maximizing assignment to a joint distribution, over a
set of random variables, which factorizes according to cliques in the graphical model. In order to achieve this, the approach is based on constructing a higher order
structure, called a \emph{junction tree}, from the graph. We'll denote the
junction tree with $\calT$. We won't explain the
junction tree construction here (details can be found in~\cite{lauritzen1988,jensen1990,koller2009}) but instead mention important properties. In a
junction tree nodes correspond to cliques of random variables from the set
$\{C_i\; :\; i\in \calC\}$, where $\calC$ is an index set over cliques. Edges in the
set $\{S_{ij}\; : \; (i,j) \in \calS\}$, connect adjacent nodes $C_i$ and
$C_j$, and are associated with the
variables shared between the two cliques, \ie $S_{ij}=C_i \cap C_j$. The edge
sets $S_{ij}$ are called \emph{separators}. For any
variable $x_u$, if it appears in any two cliques $C_i$ and $C_j$ of the junction
tree, then it must also appear in all the cliques in the unique path from $C_i$
to $C_j$. This is known as the \emph{junction tree property~\cite{nilsson1998,koller2009}}.

The joint probability function, $f$, over  random variables $X$, taking values
$\bx\in\calX$, factorizes according to
the cliques and edges in the tree as follows,
\begin{flalign}\label{eqn:jtreef}
    f(\bx) =
    \frac{\prod_{i\in\calC}f_{C_i}(\bx_{C_i})}{\prod_{(i,j)\in\calS}f_{S_{ij}}(\bx_{S_{ij}})},
\end{flalign}
where $f_C$ and $f_S$ are non-negative real functions on cliques and edges
respectively. As before these functions are referred to as potentials. Since
computing the probabilities of $f$ over the space of configurations $\calX$ can
often be exponential in the number of variables, a factorization of $f$ over
cliques can be computationally advantageous if the clique sizes are limited.
Using a message-passing algorithm that limits computation of (max) probabilities over
just the cliques allows for efficient inference, as long as the \emph{tree
width} (\ie the maximum size of any clique in the tree) is small.

To compute the assignment of variables $\bx \in \calX$ that maximizes $f$ a
max-flow message passing algorithm over the junction tree is used.
\subsubsection{Max-flows over Junction Tree}
Assume that we are given an initial factorization of $f$ over a set of clique and separator potentials,
\begin{flalign}
    (\{f_{C_i}\;:\;i\in \calC\},\;\{f_{S_{ij}}\;:\;(i,j)\in \calS\}),
\end{flalign}
A message from node $C_i$ to an adjacent node $C_j$ is defined as the
normalized max-flow from $C_i$ to $C_j$:
\begin{flalign}\label{eqn:maxflowmessage}
    \delta_{i\rightarrow j} = \frac{f'_{S_{ij}}}{f_{S_{ij}}},
\end{flalign}
where,
\begin{flalign}
    f'_{S_{ij}}=\max\limits_{C_i\setminus S_{ij}} f_{C_i}.
\end{flalign}
Given two sets $B\subset A$, and function $g$ on $\calX_B$,  the above $\max$ notation means,
\begin{flalign}
    \max_{A\setminus B} g(\bx_B) = \max\limits_{\bz\in \calX_A}\;\{g(\bz)\;:\;
    \bz_B=\bx_B\},
\end{flalign}
where $\bx_B \in \calX_B \doteq\times_{u\in B} \calX_u$. The update to clique
potential $f_{C_j}$ is then,
\begin{flalign}\label{eqn:maxflowupdate}
    f'_{C_j} = f_{C_j} \cdot \delta_{i\rightarrow j}.
\end{flalign}
Message-passing proceeds with the following update schedule: pick any node in
$\calT$, say $C_1$, as the root node. Starting from the leaves of $\calT$ pass
max-flow messages up to $C_1$ and back down to the leaves. A clique sends a
message to its neighbor $C_j$ once it has received all messages from its
neighbors with possible exception of $C_j$, such a message is called an
\emph{active} max-flow. After this two-phase propagation of
messages the potentials are guaranteed to have reached equilibrium resulting in
\emph{max-marginal} potentials ~\cite{dawid1992},
\begin{flalign}\label{eqn:maxmarginalpot}
    (\{\widehat{f}_{C_i} \;:\; i\in \calC\},\;\{\widehat{f}_{S_{ij}}\;:\; (i,j)\in
    \calS\}),
\end{flalign}
where the max-marginal potential over set $A$ is defined to be,
\begin{flalign}
    \widehat{f}_{A}(\bx_{A}) = \max\limits_{\bz\in\calX}\; \{f(\bz)\;:\;
    \bz_A=\bx_A\}.
\end{flalign}
An important property of the update rule is that $f$ in eqn.~\ref{eqn:jtreef} is
invariant to max flow updates. To see this consider adjacent cliques $C_i$,
$C_j$ and the separator, $S_{ij}$ between them. A max-flow update gives,
\begin{flalign}
    \begin{array}{ll}
        f'_{S_{ij}}=\max\limits_{C_i\setminus S_{ij}} f_{C_i},  &\quad
        \delta_{i\rightarrow j} =\frac{f'_{S_{ij}}}{f_{S_{ij}}} \\
        f'_{C_j} = f_{C_j}\cdot\delta_{i\rightarrow j} = f_{C_j}\cdot
        \frac{f'_{S_{ij}}}{f_{S_{ij}}} &\quad
        \frac{f'_{C_j}}{f'_{S_{ij}}}=\frac{f_{C_j}\cdot
        f'_{S_{ij}}}{f'_{S_{ij}}\cdot f_{S_{ij}}} = \frac{f_{C_j}}{f_{S_{ij}}},
    \end{array}
\end{flalign}
where the RHS of the bottom row shows the invariance in max-flow update to $f$
in the contribution by $C_j$ and $S_{ij}$.

Also note the following property~\cite{nilsson1998},
\begin{flalign}\label{eqn:maxconsistency}
    \max\limits_{C_i\setminus S_{ij}} \widehat{f}_{C_i} = \widehat{f}_{S_{ij}} =
    \max\limits_{C_j\setminus S_{ij}} \widehat{f}_{C_j},
\end{flalign}
known as the \emph{max-consistency} property, which holds after computing the
max-marginal potentials. Also note the following theorem,

\begin{theorem}[Max-marginal theorem~\cite{nilsson1998}]\label{thm:theorem_maxf}
The joint distribution $f$ and the marginals agree on the
maximimum value,
\begin{flalign}
    \max\limits_{\bx\in \calX} f(\bx) = \max\limits_{\bx_{C_k}\in \calX_{C_k}}
    \widehat{f}_{C_k}(\bx_{C_k}).
    \quad \forall k\in \calC\
\end{flalign}
\end{theorem}
This is a direct result of the definition of $\widehat{f}_{C_k}$,
\begin{flalign}
    \max_{\bx_{C_k} \in \calX_{C_k}} \widehat{f}_{C_k}(\bx_{C_k}) =
    \max\limits_{\bx_{C_k}\in \calX_{C_k}}\left\{
    \max\limits_{\bz\in \calX} \{ f(\bz)\;:\; \bz_{C_k} = \bx_{C_k}\}\right\} =
    \max_{\bx\in\calX} f.
\end{flalign}
\subsubsection{Maximizing assignment and traceback}\label{sec:traceback}
To compute the maximizing assignment, $\bx^*$, given the max-marginal potentials
(eqn.~\ref{eqn:maxmarginalpot}) the algorithm starts at the root of $\calT$, say $C_i$, and
picks the assignment $\bx^*_{C_i}$ that maximizes $\widehat{f}_{C_i}(\bx_{C_i})$. It
then propagates \emph{simple max-flows}~\cite{nilsson1998}.
From thm.~\ref{thm:theorem_maxf} we have that
$\widehat{f}_{C_i}(\bx^*_{C_i})=\max_{\bx\in\calX} f(\bx)$. Next the algorithms takes an incident
separator $S_{ij}$ and assigns the variables $X_{S_{ij}}$ the corresponding
values in $\bx^*_{C_i}$ to get $\bx^*_{S_{ij}}$. Because of max-consistency
(eqn.~\ref{eqn:maxconsistency}) we have that
$\widehat{f}_{S_{ij}}(\bx^*_{S_{ij}})=\max_{x\in\calX}f(\bx)$. The algorithm now moves to
$C_j$ and assigns values from $\bx^*_{S_{ij}}$ to variables in $X_{C_j}$ that
coincide with $X_{S_{ij}}$. Then the algorithm finds the maximizing assignment, $\bx^*_{C_j}$, to
$\widehat{f}_{C_j}$ over the remaining variables in $C_j$ -- such that
$\widehat{f}_{C_j}(\bx_{C_j}^*)=\max_{\bx\in \calX} f(\bx)$ due to
max-consistency. The algorithm proceeds until it's processed the leaves of $\calT$.

In a tree with $m$ nodes there's $m-1$ edges. Thus, given that
    $\widehat{f}_{C_i} = \max f$,  for all $i\in\calC$, and
$\widehat{f}_{S_{ij}} = \max f$, for all $(i,j)\in\calS$, then we are assured~\cite{nilsson1998},
\begin{flalign}
    f(\bx^*) = \frac{\prod_{i\in\calC
    \widehat{f}_{C_i}(\bx_{C_i}^*)}}{\prod_{(i,j)\in\calS}\widehat{f}_{S_{ij}}(\bx^*_{S_{ij}})}
    = \frac{(\max\limits_{\bx\in\calX}
    f(\bx))^{|\calC|}}{(\max\limits_{\bx\in\calX}
f(\bx))^{|\calS|}}=\max\limits_{\bx\in\calX} f(\bx).
\end{flalign}

\subsubsection{Simplified max-flow propagation algorithm~\cite{nilsson1998}}\label{sec:smfp}
A rather simple but inefficient approach of producing \mbest solutions uses
Lawler's \mbest algorithm of~\cref{sec:lawlermbest}. This algorithm is referred to as the simplified max-flow propogation algorithm or SMFP~\cite{nilsson1998}. Assume the vector of random
variable assignments $\bx^{(1)}=(x^{(1)}_1,\dots,x^{(1)}_n)$, represents the maximizing
assignment to the joint probability distribution $f(\bx)$ on $\calT$ that we get
after running the max-flow propagation algorithm of the previous section. In order to compute the next highest assignment SMFP
partitions the space into $n$ subsets that cover $\calX\setminus\{x^{(1)}\}$,
\begin{addmargin}[2em]{2em}
\begin{itemize}
    \item[($P_1$):] $\{x\in \calX \;:\; x_1 \neq x_1^{(1)}\}$
    \item[$\vdots$] \qquad $\dots$
    \item[($P_i$):] $\{x\in \calX \;:\; x_1
    =x_1^{(1)},\dots,\;x_{i-1}=x_{i-1}^{(1)},\;x_i\neq x_i^{(1)}\}$
    \item[($P_n$):] $\{x\in \calX \;:\; x_1
    =x_1^{(1)},\dots,\;x_{n-1}=x_{n-1}^{(1)},\;x_n\neq x_n^{(1)}\}$
\end{itemize}
\end{addmargin}
Note that each assignment space $P_i$ constrains one of the variables to take on a different value than it did in $x^(1)$, thereby removing $x^(1)$ from the space of assigments.
 In order to encode the constraints for an assignment space $P_i$, SMFP defines a series of functions as follows,
 \begin{flalign}
     \bar{f}_i(\bx) = \left \{\begin{array}{ll}
         f(\bx) & \mbox{if $x_1= x_1^{(1)}\;,\dots,\; x_{i-1}=x_{i-1}^{(1)},\;
         x_i\neq x_i^{(1)}$}, \\
        0 & \mbox{otherwise.}
     \end{array}  \right.
 \end{flalign}
 Then it's easy to see that,
 \begin{flalign}\label{eqn:maxpartition}
     \max\limits_{i\in [n]} \max\limits_{\bx\in\calX} \bar{f}_i(\bx) =
     \max\limits_{\bx \in \calX\setminus \{\bx^{(1)}\}} f(\bx).
 \end{flalign}
 To compute $\max_{\bx\in\calX} \bar{f}_i(\bx)$ (\ie the maximum assignment to
 $f(\bx)$ constrained to $P_i$) a subset of variables are fixed,
 \begin{flalign}
    X_1=x_1^{(1)},\dots,\;X_{i-1}=x_{i-1}^{(1)},\;X_i\neq x_i^{(1)}\},
 \end{flalign}
by introducing the following representation that modifies the potential
functions: given,
\begin{flalign}
    \delta_{1}(x_{1};C) &= \left\{\begin{array}{ll}
            1 & \mbox{if $(X_{1}\in C \wedge x_1= x^{(1)}_1)$ or
            $X_1\not\in C$},\\
            0 & \mbox{otherwise}
        \end{array}\right.\\
    \vdots \nonumber\\
    \delta_{i-1}(x_{i-1};C) &= \left\{\begin{array}{ll}
            1 & \mbox{if $(X_{i-1}\in C \wedge x_{i-1}= x^{(1)}_{i-1})$ or
            $X_{i-1}\not\in C$},\\
            0 & \mbox{otherwise}
        \end{array}\right.\\
    \delta_{i}(x_{i};C) &= \left\{\begin{array}{ll}
            1 & \mbox{if $(X_{i}\in C \wedge x_{i}\neq x^{(1)}_{i})$ or
            $X_{i}\not\in C$},\\
            0 & \mbox{otherwise}
        \end{array}\right. ,
\end{flalign}
SMFP modifies the clique and separator potentials as follows,
\begin{flalign}
    f_{C_k}(\bx_{C_k})=f_{C_k}(\bx_{C_k})\prod\limits_{q=1}^i \delta_q(x_q;C_k)
\end{flalign}
Max-flows are then propagated in $\calT$ until equilibrium. Given the
max-consistency property (eqn.~\ref{eqn:maxconsistency}) and
thm.~\ref{thm:theorem_maxf} the maximum of
$\bar{f}_i(\bx)$ can be computed, and the corresponding maximizing assignment $\bx^{(2)}$.

To find the third highest assignment to $f(\bx)$, the partitioning is as
follows. If $\bx^{(2)}$ belongs to subset $P_i$, then it is refined by partitioning it
into the following subsets,
\begin{addmargin}[4em]{2em}
\begin{itemize}
    \item[($P_{n+1}$):] $\{x\in \calX \;:\; x_1
        =x_1^{(2)},\dots,\;x_{i-1}=x_{i-1}^{(2)},\;x_i\neq
    \{x_i^{(1)},x_i^{(2)}\}\}$
    \item[($P_{n+2}$):] $\{x\in \calX \;:\; x_1
            =x_1^{(2)},\dots,\;x_{i}=x_{i}^{(2)},\;x_{i+1}\neq
        x_{i+1}^{(2)}\}$
    \item[$\vdots$] \qquad $\dots$
    \item[($P_{2n-i+1}$):] $\{x\in \calX \;:\; x_1
    =x_1^{(2)},\dots,\;x_{n-1}=x_{n-1}^{(2)},\;x_n\neq x_n^{(2)}\}$.
\end{itemize}
\end{addmargin}
Together with $\calP^0=\{P_1,\dots,P_{n}\}\setminus P_i$ the new partitioning
$\calP^1=\{P_{n+1},\dots,P_{2n+1-i}\}$ covers $\calX \setminus \{\bx^{(1)},\bx^{(2)}\}$.
Repeating the above procedure for computing eqn.~\ref{eqn:maxpartition}, $\bx^{(3)}$ can be found. Continuing in this way SMFP finds the \mbest solutions~\cite{nilsson1998}.

The down side of the approach is that many max-flow operations over $\calT$ have
to be done in order to compute $\max\limits_{i\in [n]} \max\limits_{\bx\in\calX}
\bar{f}_i(\bx)$ each round -- a two-pass max flow propagation thru the entire tree for each assignment set.

Nilsson~\cite{nilsson1998} also presents an alternate partitioning strategy that's much more efficient which relies on the \emph{running intersection property} of junction trees. This improved partitioning strategy allows for the max of $f$ over partitions to be found with a single root to leaf propogation of max flows.

\subsection{\mbest solutions for loopy graphs and the BMMF
algorithm}\label{sec:mbestmap}
So far we have discussed \mbest algorithms when exact inference is tractable.
This included inference over general graphs with small tree-width that could be
converted to junction trees in order to carry out exact inference. When the
tree-width of the graph becomes large however inference over the junction tree
becomes infeasible because the clique sizes are too large, so alternate \mbest algorithms are needed.

Recall that when the graph is a tree (\eg junction tree) exact inference can be
carried out using the max-product message-passing algorithm we reviewed
earlier. Also, recall from theorem~\ref{thm:theorem_maxf} that the max-marginals
and joint posterior distribution over the variables agree on the maximizing
value. When the graph is a tree the max-marginals can be computed exactly and a
traceback operation can be subsequently carried out to find the most probable
variable assignments.

If we have a loopy graph, using the junction tree representation for inference
becomes inefficient and approximate inference methods are needed for computing
the \emph{approximate} max-marginals over the graph. Moreover using traceback
operation
over the max-marginals on a loopy graph isn't guaranteed to return the
maximizing assignment (see ~\cite{yanover2004} for an example). Independently picking
the variable assignments that maximize each individual max-marginal will not
work either because ties can exist in the max-marginal tables (\ie max-marginal
has more than one maximizing label for a variable) which means that
theorem~\ref{thm:theorem_maxf} will not hold (cf.~\cite{yanover2004}). Since ties
can exist and traceback over a loopy graph will not work the alternative is to
have multiple rounds of computing the max-marginals, where in each round
additional tied variables are constrained to take on a single maximizing
label. This process is continued until no more ties exist and we can get the
maximizing assignment by independently maximizing over individual
max-marginals.

Nilsson's SMFP algorithm that we discussed in~\ref{sec:smfp} is an example of an
\mbest algorithm that computes the max-marginals in a junction tree by using
max-product message passing algorithm and subsequently uses the max-consistency
property (eqn.~\ref{eqn:maxconsistency}) and thm.~\ref{thm:theorem_maxf} to find
the maximizing assignment. It needs $\bigO{Mn}$ computations of the max-marginals
which is very expensive, where $M$ is the number of
\mbest solutions and $n$ is the number of variables in the graph.

The following algorithm by Yanover and Weiss~\cite{yanover2004} can find the
\mbest solutions in loopy-graphs with only $2M$ computations of max-marginals
($M$ is the number of \mbest solutions), and no trace-back operations (relying
only on thm.~\ref{thm:theorem_maxf}).

\begin{algorithm}[!h]\caption{Best Max-Marginal First (BMMF) algorithm for
    \mbest solutions~\cite{yanover2004}}\label{alg:bmmf}
    \begin{algorithmic}[1]
    %\Comment{initialize}
        \State $SCORE_1(i,j) \gets \max\limits_{\bx\;:\;\bx_i=j}
        f(\bx)$\label{bmmf:l0}
    \State $\bx^{(1)}_i \gets \argmax\limits_j SCORE_1(i,j)$ \label{bmmf:l1}
    \State $CONSTRS_1\gets \emptyset$
    \State $USED_2 \gets \emptyset$
    \For{$m=2,\dots,M$}
        \State $SEARCH_m \gets (i,\;j,\;k<m\;:\; \bx^{(k)}_i\neq j,\; (i,j,k)\not\in
        USED_m)$ \label{bmmf:l2}
        \State $(i_m\;,j_m\;,k_m\;)\gets \argmax\limits_{(i,j,k)\in SEARCH_m}
        SCORE_k(i,j)$ \label{bmmf:l3}
        \State $CONSTRS_m \gets CONSTRS_{k_m} \cup \{\bx_{i_m} = j_m\}$
        \label{bmmf:l4}
        \State $SCORE_m(i,j) \gets \max\limits_{\bx\;|\; \bx_i=j,\; CONSTRS_m}
        f(\bx)$\label{bmmf:l5}
        \State $\bx^{(m)}_i \gets \argmax_j SCORE_m(i,j)$ \label{bmmf:l6}
        \State $USED_{m+1} \gets USED_m \cup \{(i_m,j_m,k_m)\}$
        \State $CONSTRS_{k_m} \gets CONSTRS_{k_m}\cup \{\bx_{i_m} \neq j_m\}$
        \label{bmmf:l7}
        \State $SCORE_{k_m}(i,j) \gets \max\limits_{\bx\;|\;
        \bx_i=j,CONSTRS_{k_m}} f(\bx)$\label{bmmf:l8}
    \EndFor
    \State \Return $\{\bx^{(m)}\}_{m=1}^M$
\end{algorithmic}
\end{algorithm}
In alg.~\ref{alg:bmmf} the joint probability over all variables of interest (\eg posterior probability) is
represented as $f(\bx)$. The algorithm start by inferring the max-marginals in
line~\ref{bmmf:l0}. In
line~\ref{bmmf:l1} the maximizing MAP assignment is found using the max-marginal
theorem. To compute the remaining $M-1$ solutions the algorithm repeats the
following operations: the max-marginal tables are searched to find the variable
with next best max-marginal value, (cf. lines~\ref{bmmf:l2}-\ref{bmmf:l3}). The
variable is fixed to the label corresponding to this value (\ie $\bx_{i_t} =
j_t$) and added as a constraint for the next round of max-marginal computations (see
cf. lines~\ref{bmmf:l4}-\ref{bmmf:l5}). Using th max-marginal theorem the next
best solution is computed (cf. line~\ref{bmmf:l6}). The complementary
constraint (\ie $\bx_{i_t}\neq j_t$) is added to the constraint set used to
produce the max-marginals that gave the highest value earlier and the
max-marginals are recomputed with this augmented set of constraints (cf.
lines~\ref{bmmf:l7}-\ref{bmmf:l8}). In each iteration, $t$, a new set of max-marginals
is added (\ie $SCORE_t(i,j)$) that is the result of a running inference on the
graph with some of the variables fixed. This fixing of variables successively
refines the partitioning of the assignment space in such a way that the
previous best solutions are removed from consideration.

It turns out that for exact max-marginal computation, the assignment $x^{(m)}$
produced by the BMMF algorithm~\ref{alg:bmmf} is the $m$-th most probable assignment under
$f(\bx)$ (cf.~\cite{yanover2004}).

For loopy graphs where approximate inference algorithms have to be used for
computing the max-marginals (\eg loopy max-product belief propagation), the
solutions produced by BMMF (so called loopy-BMMF) are not guaranteed to correspond to the \mbest
solutions but tend to be quite good in practice, compared to the top M
assignments produced by Gibbs sampling.

\subsection{\mbest MAP and its linear programming formulation}
We've seen that when computing MAP assignments is not tractable approximate
methods can be used to compute the approximate \mbest solutions. Yanover and
Weiss' \mbest MAP method (loopy-BMMF)~\cite{yanover2004}, that we reviewed earlier, is
one such method. The downside of approaches such as loopy max-product is that
they do not provide bounds on the optimal values of the solutions. However, LP
approximations to MAP do provide bounds on the optimal value and Fromer and
Globerson~\cite{fromer2009} provide an extension of the LP MAP formulation to the
\mbest setting. This section provides an overview of their approach.

To start, recall from the review in chapter 1 that the MAP problem can be formulated as the
following LP,
\begin{flalign}\label{eqn:maplp2}
    \max\limits_{\bx} f(\bx) = \max\limits_{\bmu \in \mathbb{M}(G)} \bmu \cdot
    \btheta
\end{flalign}
and that the maximizing $\bmu^{(*)}$ is integral and found at a vertex of
$\mathbb{M}(G)$ --- where $\mathbb{M}(G)$ is the marginal polytope defined in
eqn.~\ref{eqn:marginalpolytope}. Moreover, $\bmu^{(*))}$ corresponds to the MAP
assignment $\bx^{(1)}$. For general graphs representing $\mathbb{M}(G)$ requires
an exponential number of inequalities so recall that the LP is relaxed by using an
outer bound on $\mathbb{M}(G)$, called the local polytope (cf.
eqn.~\ref{eqn:localpolytope}), $\mathbb{L}(G)$ which can be represented by far
fewer inequality constraints over the variables (\ie half-spaces). As we
mentioned in \cref{sec:mapproblem}, it has been shown that $\mathbb{M}(G)=\mathbb{L}(G)$ for tree
structured graphs, so solving the LP-relaxation yields the exact MAP assignment.

\subsection{\mbest MAP LP when $G$ is a tree}
First consider tree-structured graphs. In order to extend the MAP LP formulation
in eqn.~\ref{eqn:maplp2} to the \nth{2} best MAP problem Fromer and
Globerson~\cite{fromer2009} propose to swap $\mathbb{M}(G)$ for the following
\emph{assignment-excluding marginal polytope},
\begin{flalign}\label{eqn:aempolytope}
    \widehat{\mathbb{M}}(G,\bx^{(1)}) = \{\bmu \;|\; \exists p(\bx)\in \calP\;
        \mbox{s.t.}\; p(\bx^{(1)}) = 0,\; p(\bx_s,\bx_t)=\bmu_{st}(\bx_s,\bx_t),\;
    p(\bx_s)=\bmu_s(\bx_s)\},
\end{flalign}
where $\widehat{\mathbb{M}}(G,\bx^{(1)})$ is the convex hull of a set of integral
vectors corresponding to the different assignments, excluding only $\bx^{(1)}$.
They show that,
\begin{flalign}
    \max\limits_{\bx\neq\bx^{(1)}} f(\bx) = \max\limits_{\bmu\in
    \widehat{\mathbb{M}}(G,\bx^{(1)})} \bmu\cdot\btheta.
\end{flalign}
In order to represent $\widehat{\mathbb{M}}(G,\bx^{(1)})$ as inequalities in the MAP
LP, Fromer and Globerson propose the following: when $G$ is a tree they show
that adding the single inequality $I(\bmu,\bx^{(1)})\leq 0$ to $\mathbb{M}(G)$ will
result in $\widehat{\mathbb{M}}(G,\bx^{(1)})$, \ie
\begin{flalign}\label{eqn:aempolytopetree}
    \widehat{\mathbb{M}}(G,\bx^{(1)}) = \{\bmu\;|\; \bmu\in \mathbb{M}(G),\;
    I(\bmu,\bx^{(1)})\leq 0\}
\end{flalign}
where,
\begin{flalign}
    I(\bmu,\bx^{(1)}) = \sum\limits_{s\in V} (1-d_s)\bmu_s(\bx^{(1)}_s) +
    \sum\limits_{(s,t)\in E} \bmu_{st}(\bx^{(1)}_s,\bx^{(1)}_t),
\end{flalign}
and $d_s$ is the degree of the nodes $s$ in the tree (cf.~\cite{fromer2009}).
They show that when $G$ is a tree the polytope $\widehat{\mathbb{M}}(G,\bx^{(1)})$
will remove only the integral solutions $\bx^{(1)}$ and will not introduce
fractional solutions. They point out that for general graphs however, adding $I(\bmu,\bx^{(1)})\leq 0$ to $\mathbb{G}$ removes some other integral vertices and may introduce fractional vertices.

\subsection{\mbest MAP LP when $G$ is a general graph}
Recall from chapter 1 that when $G$ is a general graph the polytope of feasible
solutions, $\mathbb{M}(G)$, for the MAP LP needs an exponential number of
constraints, so the simpler outer-bound approximation, $\mathbb{L}(G)$ is used.
Analogously, for the \mbest MAP problem Fromer and Globerson~\cite{fromer2009} propose an
outer-bound approximation to $\widehat{\mathbb{M}}(G,\bx^{(1)})$. The approach they
takes is to add inequalities to $\mathbb{L}(G)$ to separate $\bx^{(1)}$ from the
other integral vertices. Each new constraint also removes some fractional
vertices. If enough such constraints are added then maybe only an intergral
solution is left. The type of constraints they add are inequalities over
\emph{spanning trees} on the graph,
\begin{flalign}
    I^T(\bmu,\bx^{(1)}) = \sum\limits_{s\in V} (1-d^T_s)\bmu_s(\bx^{(1)}_s) +
    \sum\limits_{(s,t)\in E}\bmu_{st}(\bx^{(1)}_s,\bx^{(1)}_t),
\end{flalign}
where $d^T_s$ is the degree of node $s$ in spanning-tree $T$. Analogous to when
$G$ is a tree (eqn.~\ref{eqn:aempolytopetree}), for general graphs they propose
an assignment-excluding marginal polytope that incopropates all spanning-tree
inequalities of the graph,
\begin{flalign}
    \widehat{\mathbb{L}}^{ST}(G,\bx^{(1)}) = \{\bmu\;|\;\bmu\in \mathbb{L}(G), \forall
    \mbox{ tree } T \subseteq E\quad I^T(\bmu,\bx^{(1)})\leq 0\}.
\end{flalign}
The \nth{2} best MAP LP for general graphs is thus,
\begin{flalign}\label{eqn:2ndbestmaplp}
    \max\limits_{\bmu\in \widehat{\mathbb{L}}^{ST}(G,\bx^{(1)})} \bmu\cdot\btheta,
\end{flalign}
which is an approximation to solving over the feasible set
$\widehat{\mathbb{M}}(G,\bx^{(1)})$. They note that maximizing over
$\widehat{\mathbb{M}}(G,\bx^{(1)})$ is guaranteed to give an integral solution whereas
maximizing over $\widehat{\mathbb{L}}^{ST}(G,\bx^{(1)})$ does not.

The number of spanning trees over $G$ is exponential in $n$ but Fromer and
Globerson use an efficient approach to consider all spanning trees. They first
note that given $\bmu$ and spanning tree $T$, the quantity $I^T(\bmu,\bx^{(1)})$ can
be decompose over edges,
\begin{flalign}
    I^T(\bmu,\bx^{(1)})=\sum\limits_{(s,t)\in E} (\bmu_{st}(\bx^{(1)}_s,\bx^{(1)}_t) -
    \bmu_s(\bx^{(1)}_s) - \bmu_t(\bx^{(1)}_t)) +\sum\limits_{s\in V} \bmu_s(\bx^{(1)}_s),
\end{flalign}
therefore to find the tree that maximizes $I^T(\bmu,\bx^{(1)})$ is equivalent to
computing the max-weight spanning-tree over $G$ where the edge weights are set
to,
\begin{flalign}
    w_{st} = \bmu_{st}(\bx^{(1)}_s,\bx^{(1)}_t) - \bmu_s(\bx^{(1)}_s) - \bmu_t(\bx^{(1)}_t).
\end{flalign}
They then rely on existing efficient algorithms for computing the max-weight spanning-tree of a graph.

To solve the LP-relaxation in eqn.~\ref{eqn:2ndbestmaplp} they use a
cutting-plane algorithm that adds the most violated constraint
$I^T(\bmu,\bx^{(1)})>0$ to the LP. Starting with any spanning-tree of $G$ the most
violated spanning-tree inequality for the current setting of $\bmu$ is found and
added to the LP. This inequality removes $\bmu$ from the polytope of feasible
solutions. The LP is solved again for a new setting of $\bmu$. The process
continues until a non-fractional $\bmu$ is found or all the constraints are
satisfied. If there are no violated constraints and $\bmu$ is still fractional
Fromer and Globerson propose additional constraints that can be added but they
note that typically only a few iterations of the cuntting-plane algorithm are
required to give integral solutions.

To extend the \nth{2} best MAP problem to the \mbest MAP problem they propose an
algorithm that recursively partitions the assignment space, similar to that of
Nilsson~\cite{nilsson1998} and Weiss~\cite{yanover2004} which we reviewed earlier. Their Partioning for Enumerated Solutions (PES) algorithm is shown in
alg.~\ref{alg:pes}.
\begin{algorithm}[!h]\caption{PES Algorithm
    (cf.~\cite{fromer2009})}\label{alg:pes}
\begin{algorithmic}[1]
    \For {$m=1,\dots,M$}
        \If {$m=1$}
            \State $\bx^{(1)} \gets \arg\max_{\bx} f(\bx)$
            \Comment{MAP assignment}
            \State $CONSTRS_1 \gets \emptyset$
        \Else
            \State $k \gets \argmax\limits_{k\in \{1,\dots,m-1\}} f(y^{(k)})$
            \Comment{{\tiny find assignment space containing highest valued
            assignment}}
            \State $\bx^{(m)} \gets \by^{(k)}$
            \Comment{{\tiny next best assignment}}
            \State $(v,a) \gets \mbox{ any member of the set } \{(s,\bx_s^{(m)})\;|\; \bx_s^{(m)}\neq
        \bx_s^{(k)}\}$
            \State $CONSTRS_m \gets CONSTRS_k \cup \{\bx_v=a\}$
            \Comment{{\tiny remove $x^{(k)}$ from assignment space $m$}}
            \State $CONSTRS_k \gets CONSTRS_k \cup \{\bx_v\neq a\}$
            \Comment{{\tiny remove $x^{(m)}$ from assignment space $k$}}
            \State $\by^{(k)} \gets \texttt{NextBestSolution}(CONSTRS_k,\; \bx^{(k)})$
        \EndIf
        \State $\by^{(m)} \gets \texttt{NextBestSolution}(CONSTRS_m,\bx^{(m)})$
    \EndFor

    \State \Return $\{\bx^{(m)}\}_{m=1}^M$
    \Statex
    \Procedure{\texttt{NextBestSolution}}{CONSTRS,\;$\bx^{(*)}$}
        \State \Return $\by \gets \argmax\limits_{\bx\neq \bx^{(*)},\; CONTRS}
        f(\bx)$\label{alg:lineinference}
    \EndProcedure
\end{algorithmic}
\end{algorithm}

The most computationally expensive part of the algorithm is the inference on
line~\ref{alg:lineinference}. The LP's are solved using general LP solvers such as
CPLEX~\cite{cplex2017}. When the \mbest inference in line~\ref{alg:lineinference} is the
LP-relaxation of eqn.~\ref{eqn:2ndbestmaplp}, Fromer and Globerson refer to the algorithm as Spanning Tree Inequalities and Partitioning for Enumerated Solutions (STRIPES)~\cite{fromer2009}.

%\section[DivMBest Alogirthm]{DivMBest Algorithm\protect\footnote{\protect\text{}\protect\cite{batra2012}.}}
\section{DivMBest Algorithm}\footnotemark\footnotetext{The contributions to the thesis presented in this section are found in \protect\cite{batra2012}, and are in collaboration with Gregory Shakhnarovich and Dhruv Batra.}
\mbest algorithms only constrain the $m$-th solution to be different than the
previous $m-1$ high probability solutions. For each of the previous solutions
the current one needs to have a different value for at least one variable.
While the set of \mbest solutions is a more diverse set to pick from than the
MAP assignment, the amount of diversity in the \mbest set is not a
parameter that can be adjusted and the minimum amount of diversity between
solutions is not a-priori quantifiable. This is why applying \mbest methods to discrete
probabilistic models for image segmentation tend to produce \mbest
segmentations that are very similar to the MAP solution and each other. The
number of possible segmentations for a typical image is $|\calL|^n$,
where $n$ is the number of pixels (
between tens of thousands to millions), and the number of
labels per pixel, $\calL$ (two or more). The number of segmentations is
exponential in $n$. If the discrete distributions over the space of assignments, that
our probabilistic models learn, had spiky modes
around very different solutions with nearly equal probability then the exact \mbest
solutions would indeed be diverse. Generally though the learned distributions
contain modes that are smooth around neighborhoods of very similar solutions giving them
nearly equal probability, which results in \mbest solutions that are very
similar. Given that the space of segmentations is large, these neighborhoods
around modes can contain a large number of very similar segmentations, each with
nearly the same high probability. Having a set of segmentations that are very similar to one
another and the MAP segmentation, both qualitatively and quantitatively, doesn't
provide an advantage over choosing the MAP segmentation. Instead we want to
produce a set of segmentations that meet certain criteria.

The key criteria of the set of segmentations produced with an
\mbest-like method include,
\begin{enumerate}
    \item the set contains highly probable segmentations,
    \item the segmentations are sufficiently different from one another and the
        MAP segmentation,
    \item the set is as small as possible
\end{enumerate}

The last property is important because we would like to reduce the assignment
space to a set small enough on which more complex inference methods can be
applied to pick a single high probability segmentation. This includes having a
user in the loop to pick from the set. Clearly the first two properties are opposing
--- the more diverse the segmentations are the more likely that the
set contains low probability ones, and inversely, higher probability
segmentations tend to come from the same mode, hence are very similar.

The ideal set containing segmentations corresponding to the
\mbest-\emph{modes} of the distribution learned by the probabilistic model
satisfy the three properties above.

In this section we introduce an \mbest-like approach that tries to ensure the
above properties, called \divmbest --- in contrast to \mbest MAP, the  \divmbest
approach emphasizes diversity between solutions. We will show that the \mbest
MAP problem is a special case of the \divmbest formulation.

\begin{figure}[!th]
  \centering
  \begin{tabular}{cccccc}
input & MAP & mode & input & MAP & mode\\
\includegraphics[width=.15\textwidth]{\main/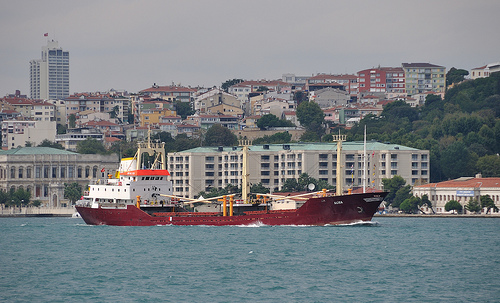} &\hspace{-6pt}
\includegraphics[width=.15\textwidth]{\main/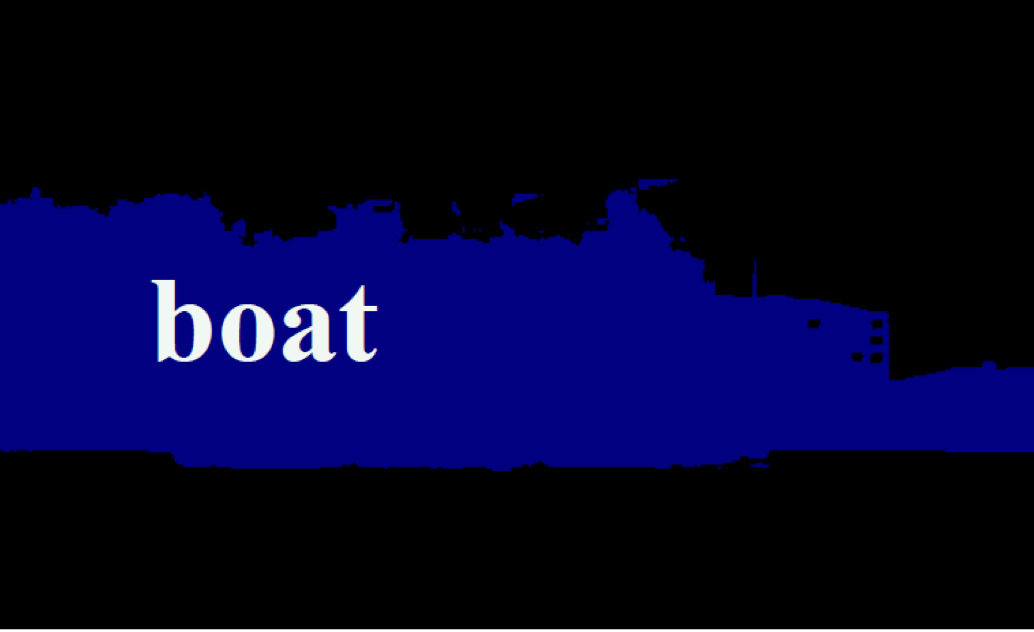} &\hspace{-6pt}
\includegraphics[width=.15\textwidth]{\main/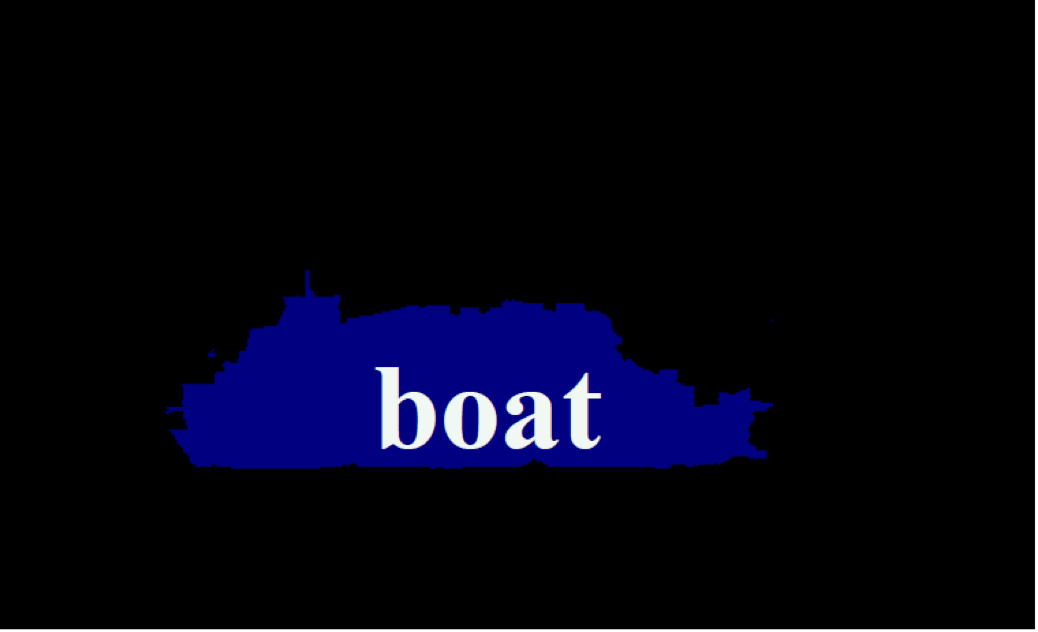} &\quad
\includegraphics[width=.15\textwidth]{\main/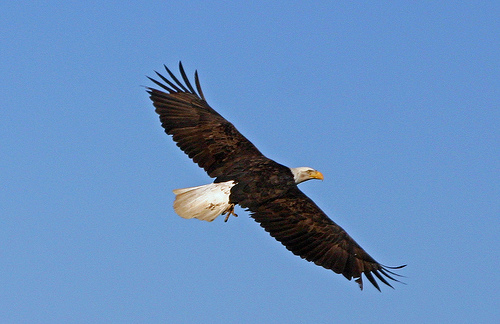} &\hspace{-6pt}
\includegraphics[width=.15\textwidth]{\main/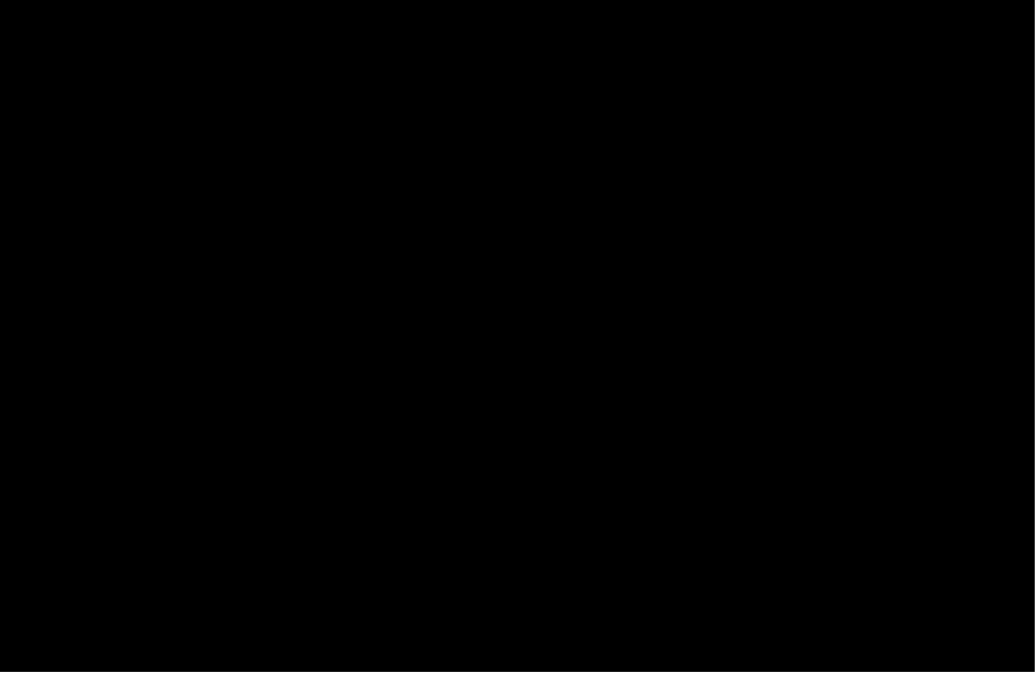} &\hspace{-6pt}
\includegraphics[width=.15\textwidth]{\main/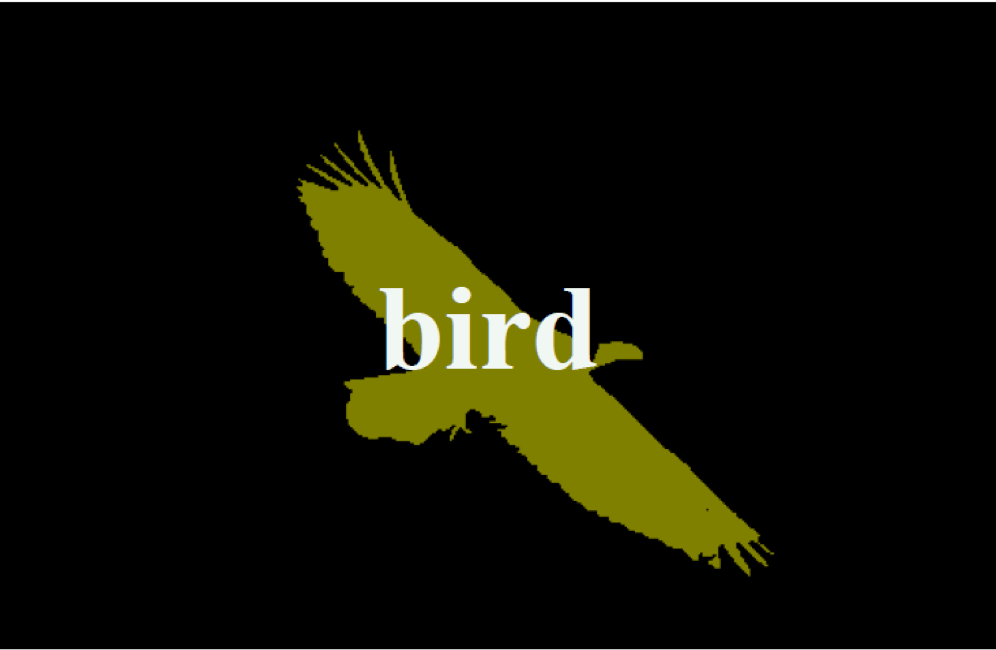}\\
\includegraphics[width=.15\textwidth]{\main/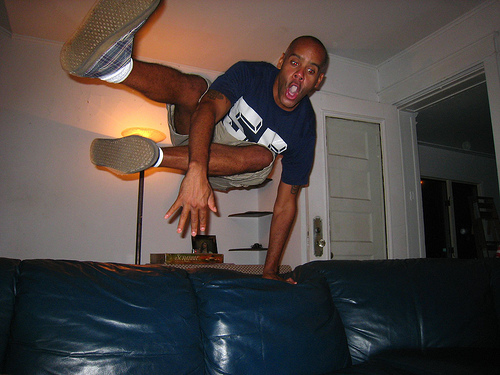} &\hspace{-6pt}
\includegraphics[width=.15\textwidth]{\main/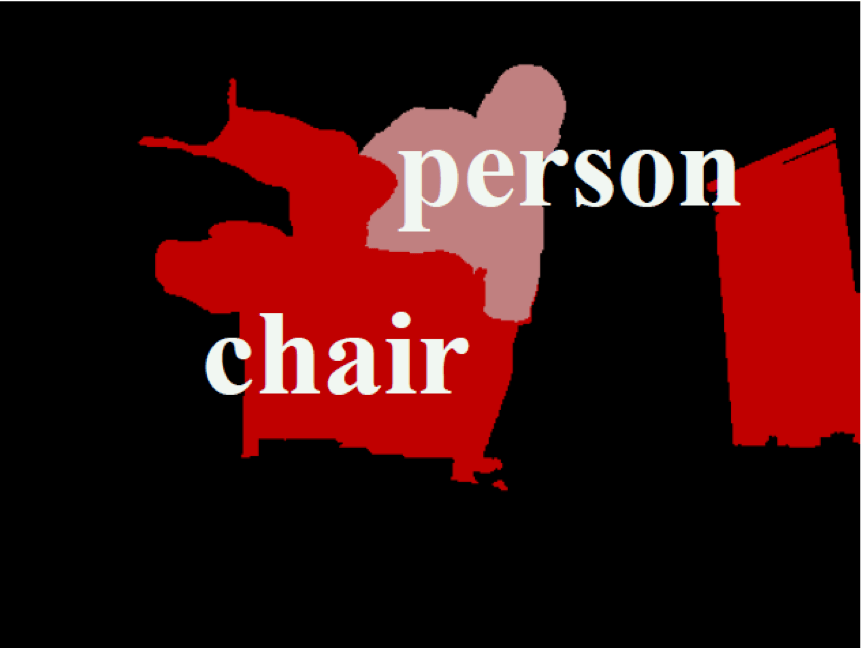} &\hspace{-6pt}
\includegraphics[width=.15\textwidth]{\main/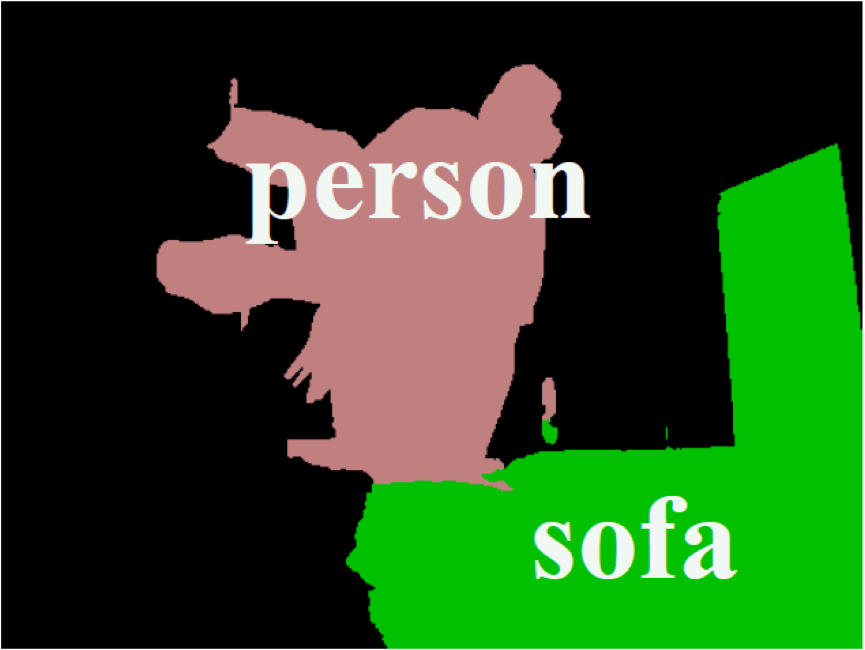} &\quad
\includegraphics[width=.15\textwidth]{\main/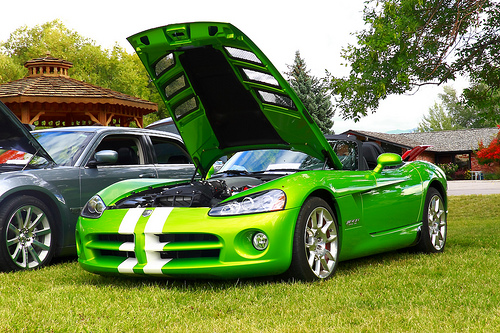} &\hspace{-6pt}
\includegraphics[width=.15\textwidth]{\main/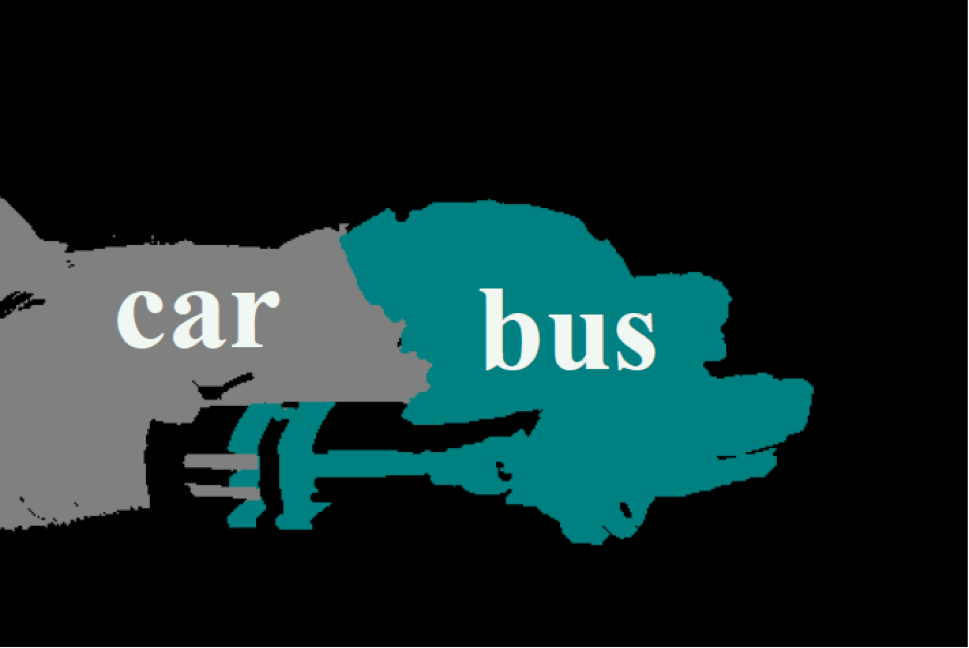} &\hspace{-6pt}
\includegraphics[width=.15\textwidth]{\main/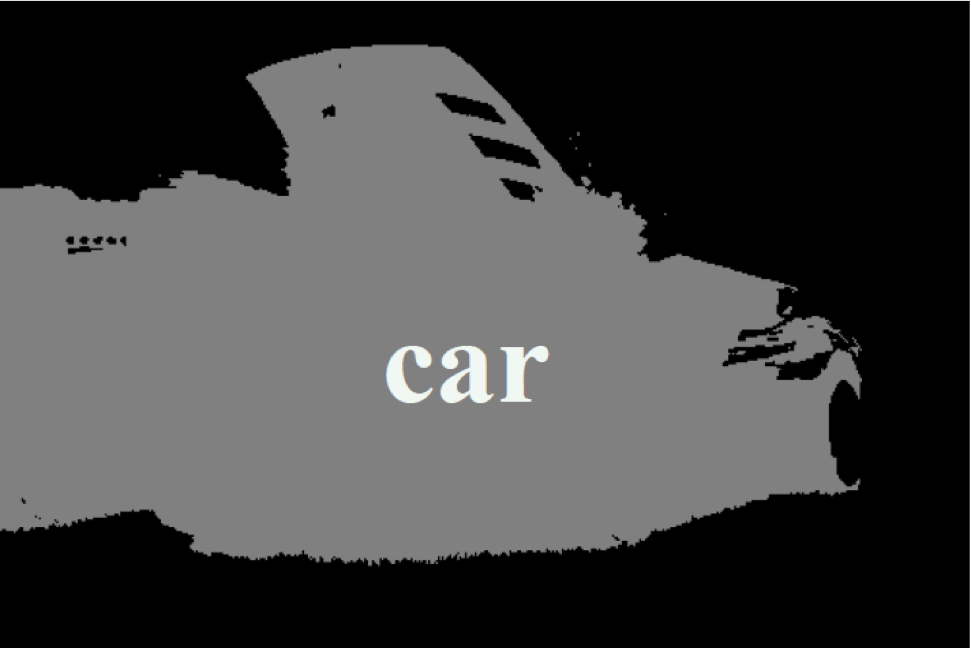}\\
\includegraphics[width=.15\textwidth]{\main/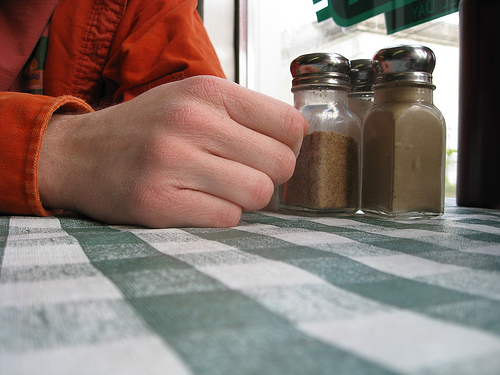} &\hspace{-6pt}
\includegraphics[width=.15\textwidth]{\main/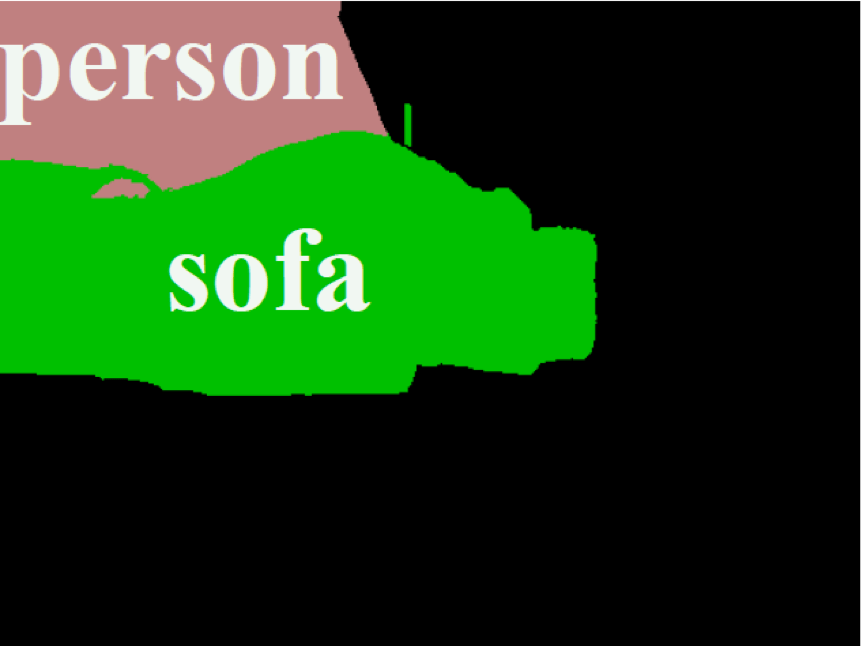} &\hspace{-6pt}
\includegraphics[width=.15\textwidth]{\main/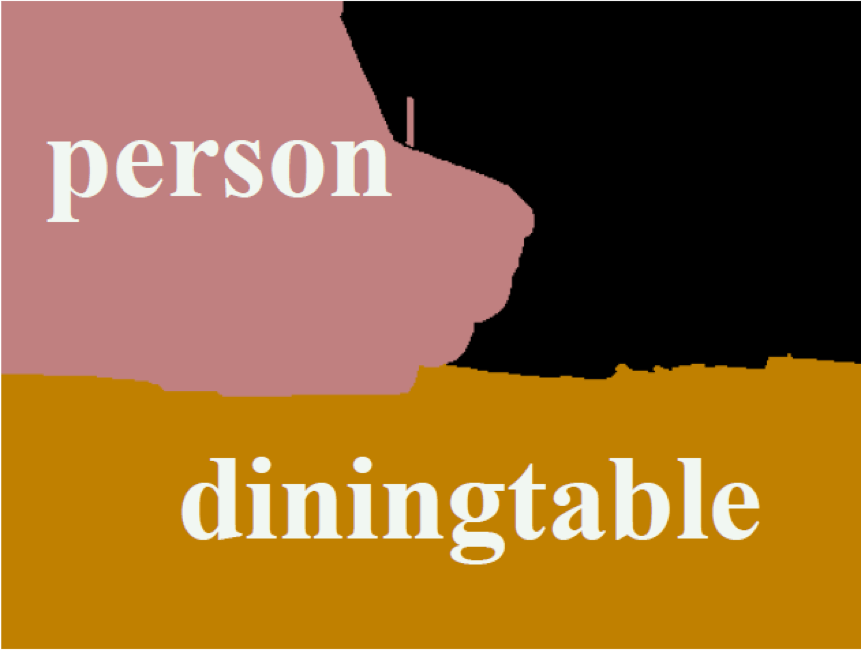} &\quad
\includegraphics[width=.15\textwidth]{\main/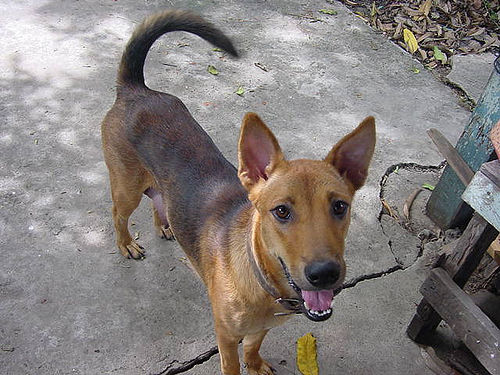} &
\includegraphics[width=.15\textwidth]{\main/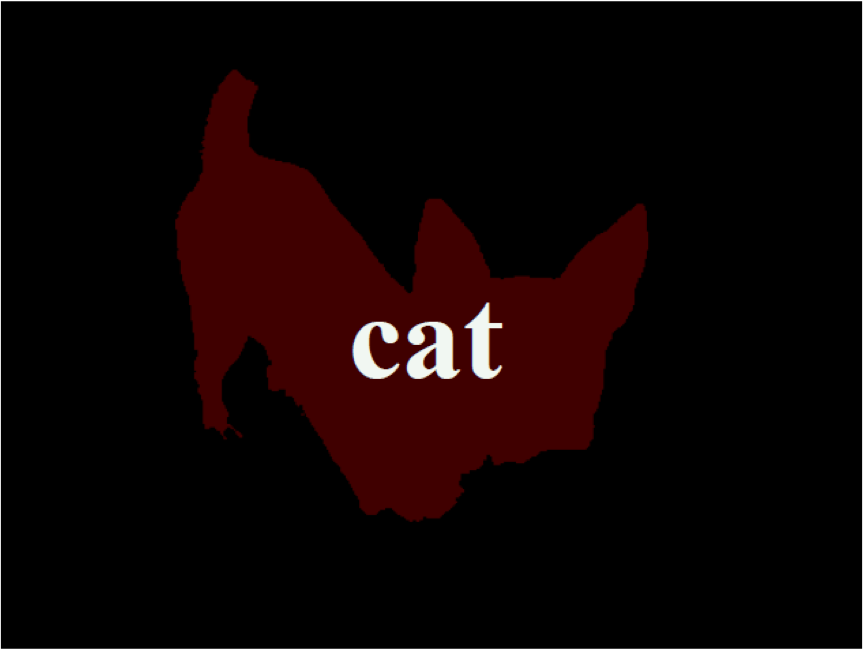} &
\includegraphics[width=.15\textwidth]{\main/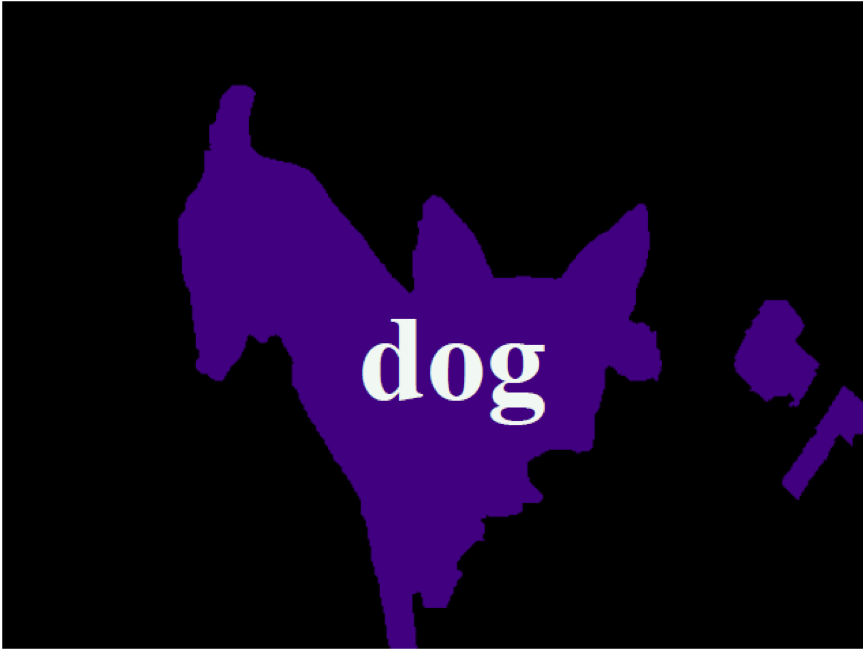}
  \end{tabular}
  %\vspace{\captionReduceTop}
  \caption{{\small Semantic segmentations on test images from PASCAL VOC 2010.
  For each image, from left: input image, MAP segmentation, best out of 10 modes
  obtained with \divmbest.}} %\vspace{\captionReduceBot}\vspace{-5pt}
  \label{fig:pascalExamples}
\end{figure}

\begin{figure}[!th]
\vspace{3pt}
\centering
%\subfloat[]
%{
\begin{tabular}{cccc}
Input & MAP & $2^{nd}$ MAP & $2^{nd}$ Mode\\
\includegraphics[width=0.2\textwidth]{\main/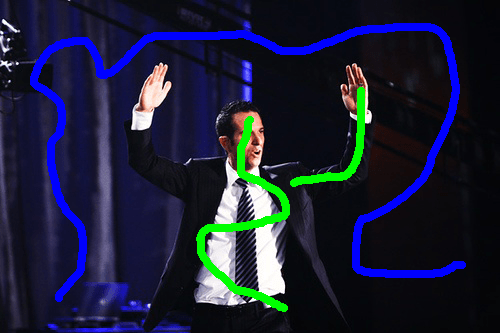} & \hspace{-4pt}
\includegraphics[width=0.2\textwidth]{\main/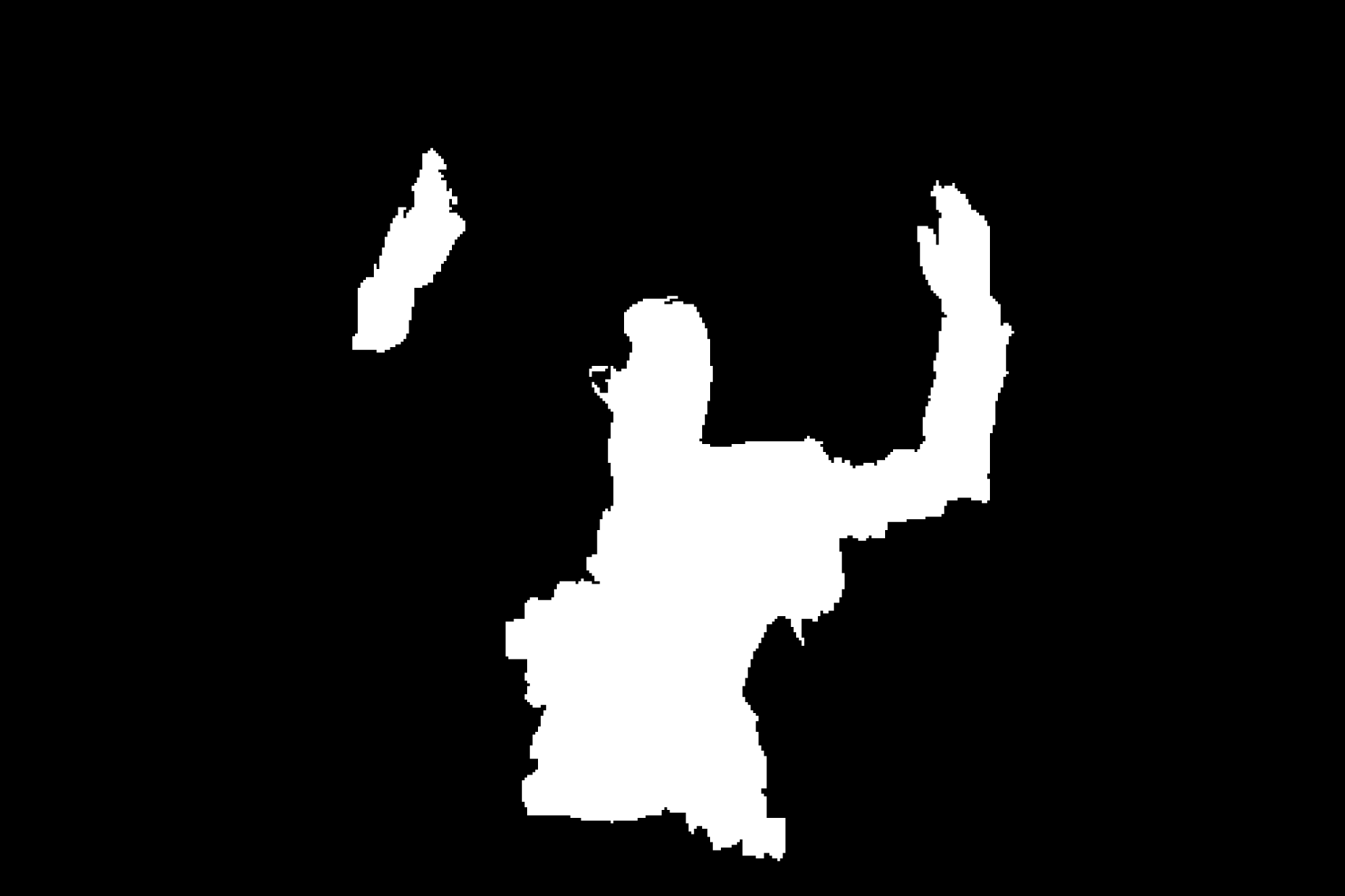} & \hspace{-4pt}
\includegraphics[width=0.2\textwidth]{\main/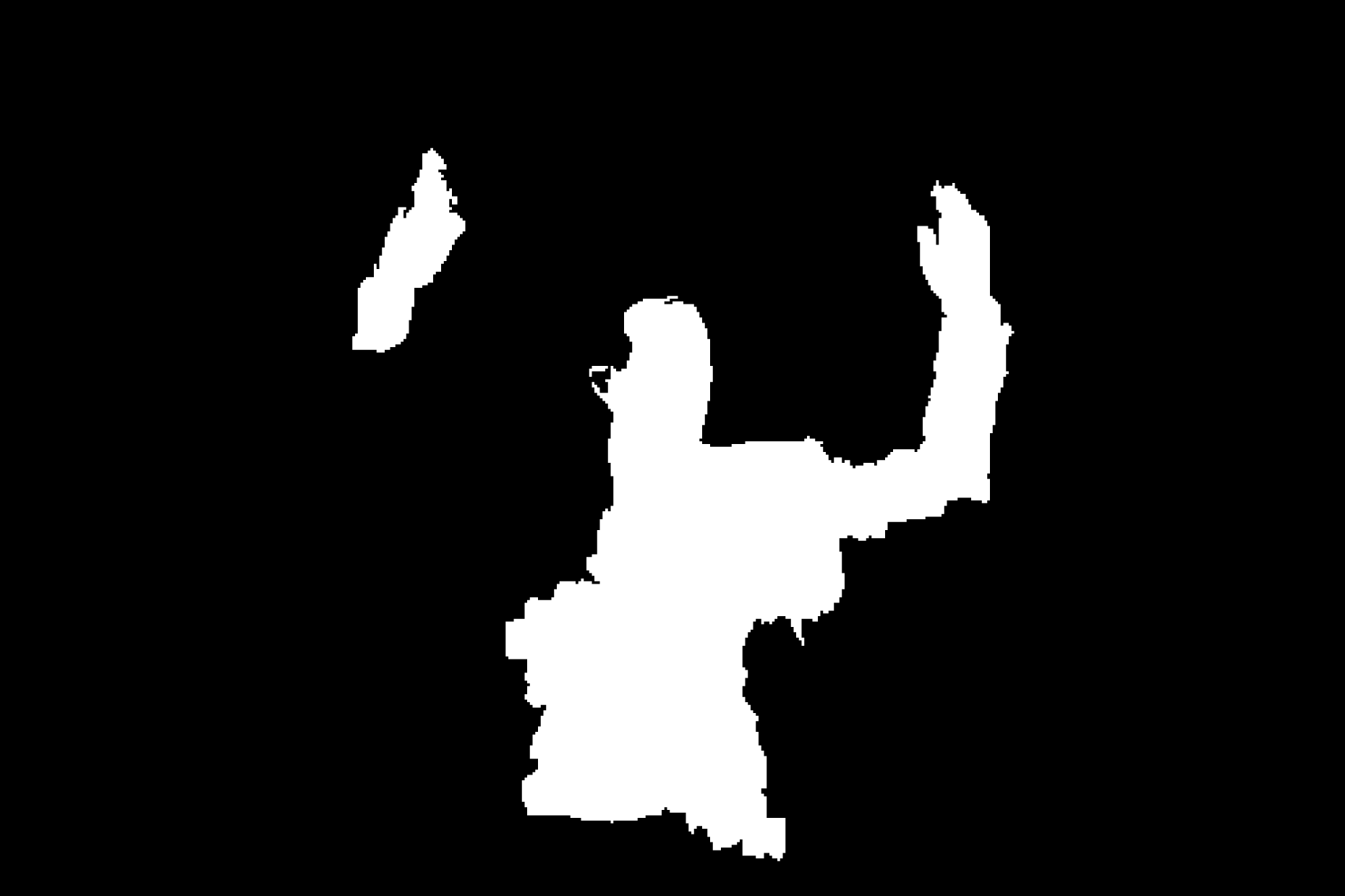} & \hspace{-4pt}
\includegraphics[width=0.2\textwidth]{\main/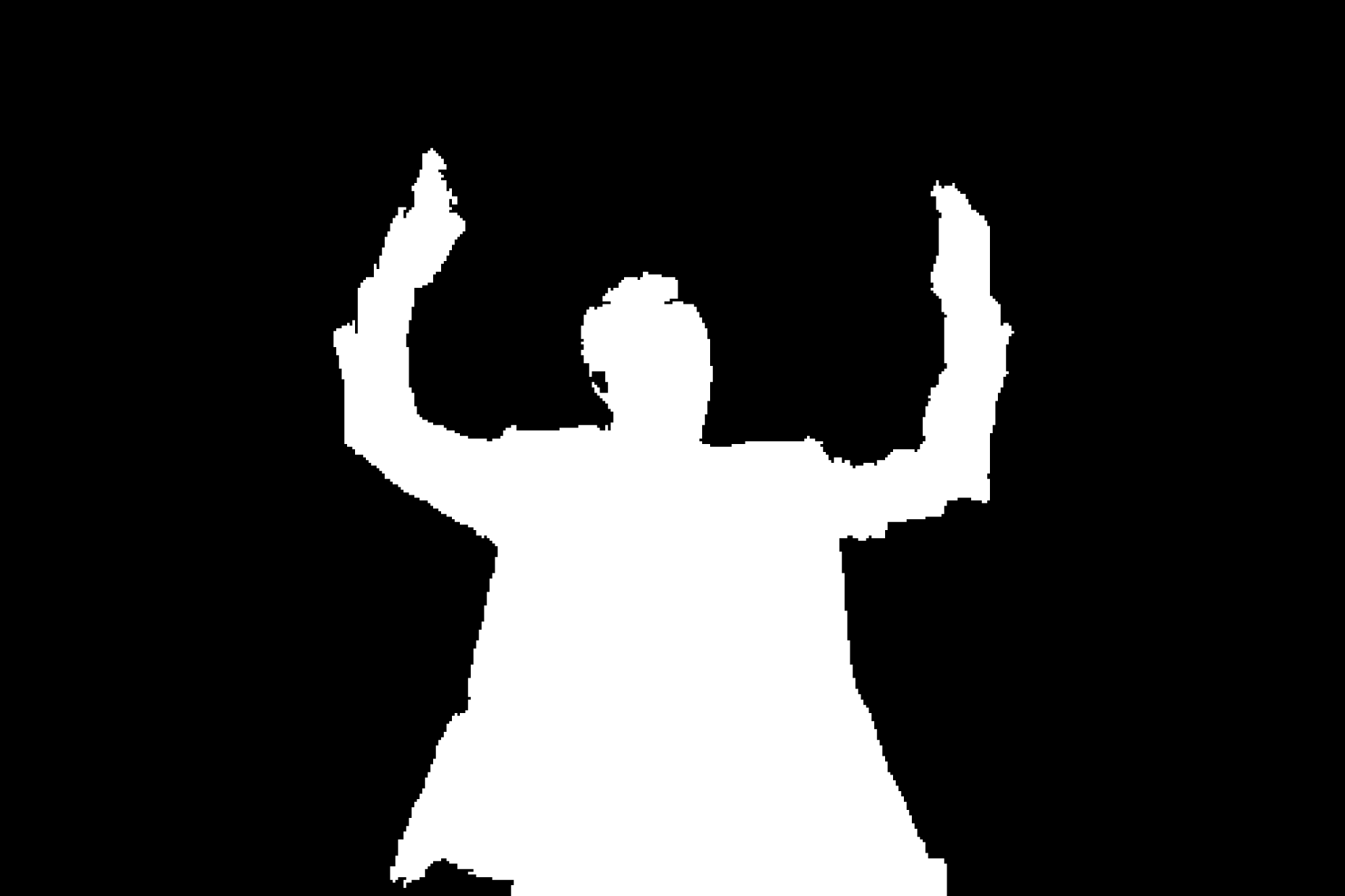}
\\%\quad
\includegraphics[width=0.2\textwidth]{\main/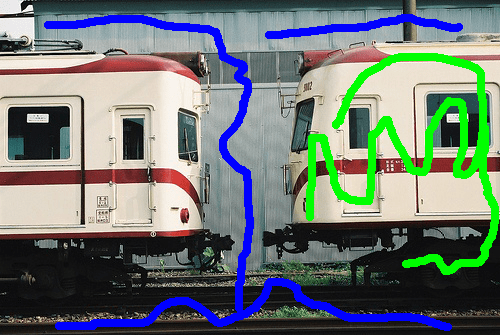} & \hspace{-4pt}
\includegraphics[width=0.2\textwidth]{\main/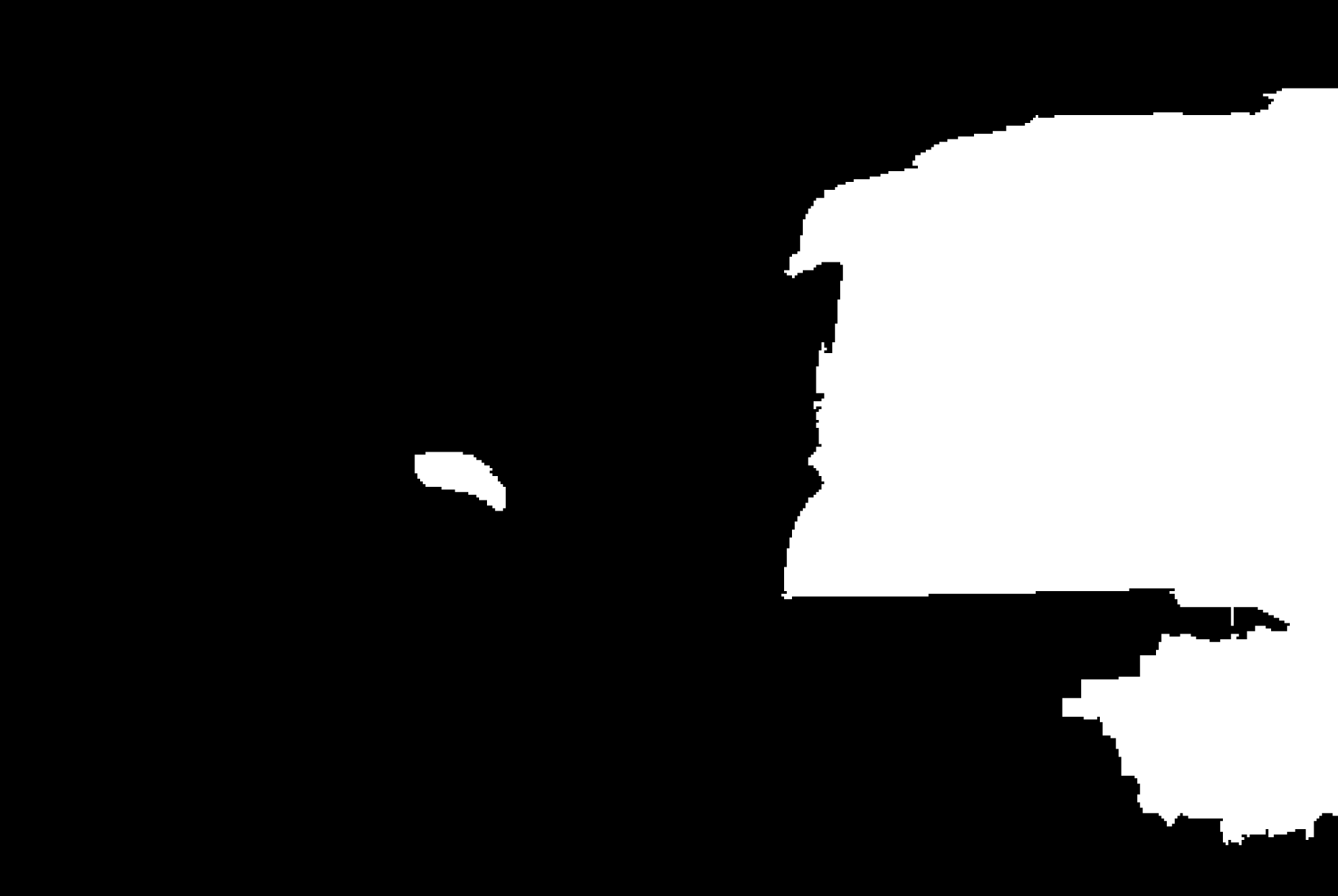} & \hspace{-4pt}
\includegraphics[width=0.2\textwidth]{\main/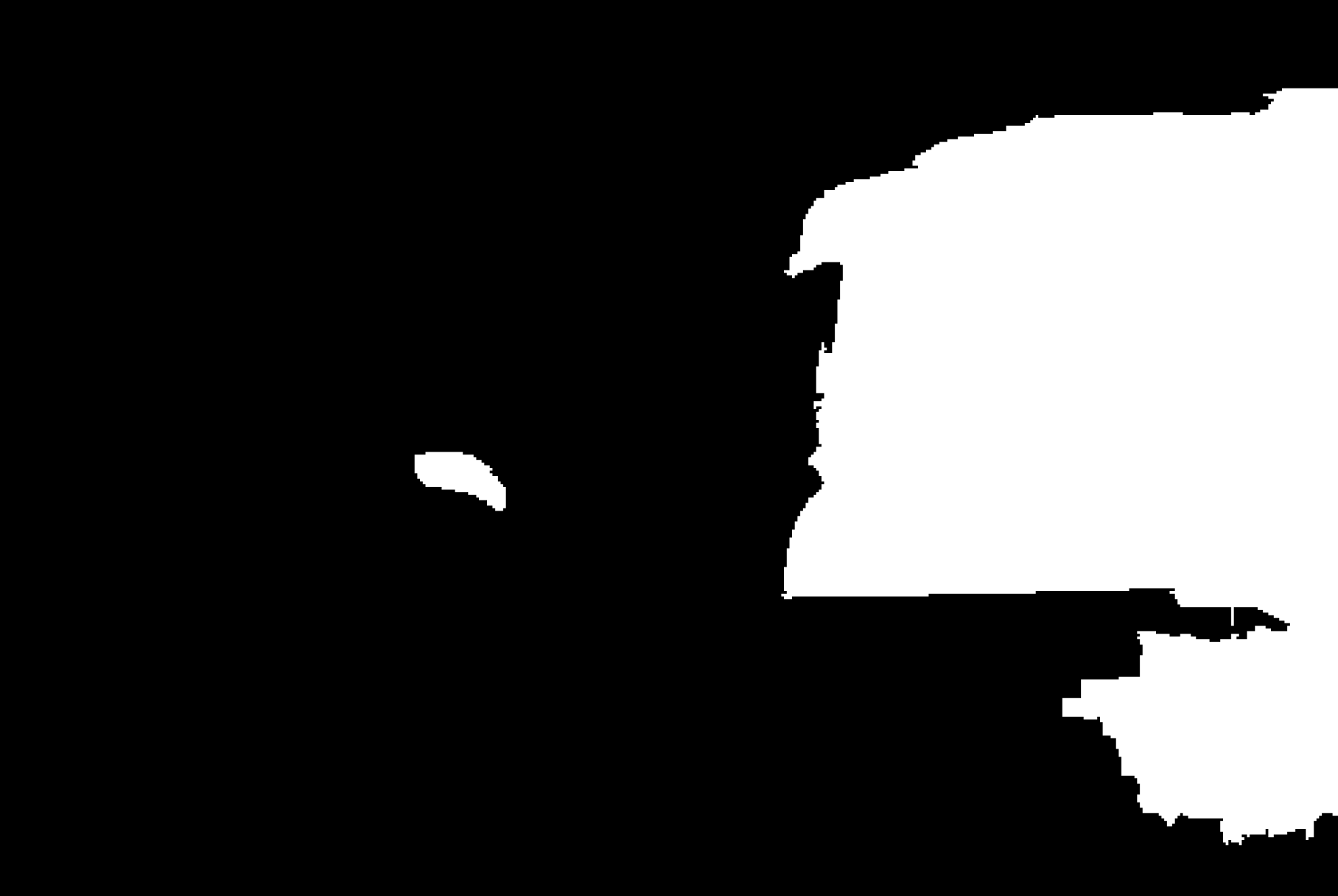} & \hspace{-4pt}
\includegraphics[width=0.2\textwidth]{\main/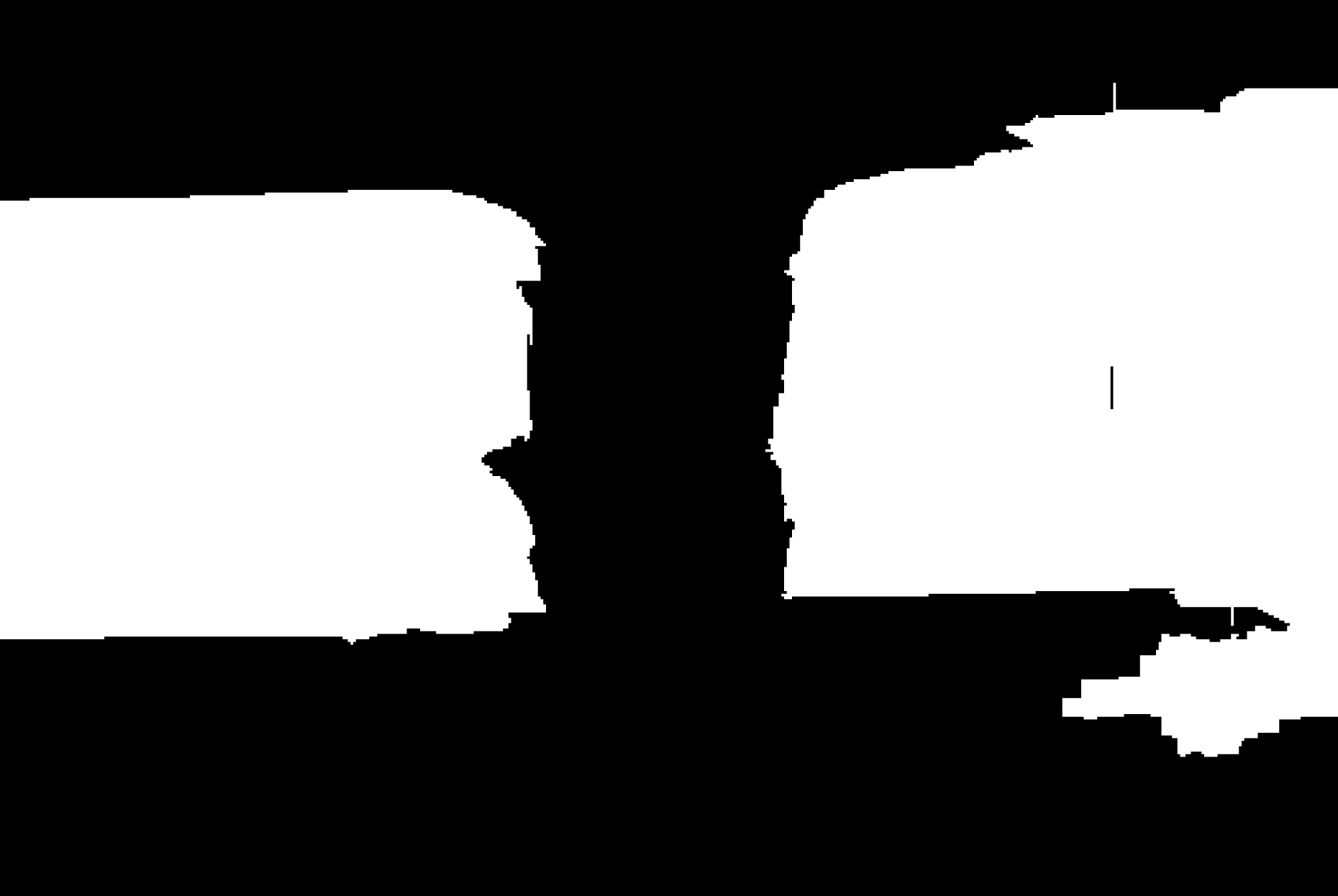}
\end{tabular}
%}
%\vspace{\captionReduceTop}
\caption{{\small Interactive segmentations. For each image, from left: input
image, MAP solution, $2^{nd}$ best MAP, and the $2^{nd}$ best mode obtained with
\divmbest.}}%\vspace{\captionReduceBot}\vspace{-10pt}
  \label{fig:interactiveExamples}
\end{figure}

To ensure that the set of segmentations contains sufficiently diverse
segmentations the \divmbest formulation incorporates a measure on dissimilarity
between two segmentations. The formulation maximizes a linear combination of the
probability of solution and dissimilarity to previous solutions. In
fig.~\ref{fig:pascalExamples}
and fig.~\ref{fig:interactiveExamples} we illustrate, qualitatively, the differences between the MAP
segmentation, and various alternate segmentations returned by \mbest and
\divmbest methods, for two segmentation tasks.

Figure~\ref{fig:pascalExamples}
provides evidence that a diverse set of highly probably segmentations under the
learned model can contain significantly more accurate explanations of the image compared
to the MAP assignment. Figure~\ref{fig:interactiveExamples} compares
segmentations corresponding to the MAP,
2nd best MAP, and 2nd best mode (second assignment of \divmbest algorithm) for an interactive segmentation task. The discrete probabilistic model
is learned over bottom-up superpixels computed over the image. The 2nd best MAP
segmentation is almost identical to the MAP estimate, and one must look closely
to see the minor difference. In contrast, the 2nd best mode of the \divmbest
approach recovers a large portion of the object in one case and gives a
drastically different explanation of the image in the other image. The advantage
of using a \divmbest approach to segmentation over \mbest MAP is apparent from these examples.
\subsection{Overview}
This section presents the  \divmbest approach which is a generalization of
the \mbest MAP problem. It borrows a similar formulation as
that of the \mbest MAP integer programming problem~\cite{fromer2009}. The
\divmbest problem has access to, and so assumes is available, a dissimilarity
function,
$\Delta(\cdot,\cdot)$, measuring the difference between two
solutions. The Lagrangian relaxation of the \divmbest integer program yields a
problem that minimizes a linear combination of the energy and similarity to previous
solutions. We conclude the section by presenting some nice properties of this
linear programming relaxation of the original \divmbest problem.

\subsection{Contributions}
The main contributions of the thesis in this section include,
\begin{itemize}
    \item the first principled formulation for extracting a set of diverse highly
        probable solutions in discrete MRFs. The \mbest MAP problem is a special
        case of this.
    \item  For certain families of diversity functions between solutions, we
        show that the Lagrangian relaxation to the \divmbest integer program is
        no more difficult to solve than the MAP problem. This makes it an
        attractive approach for inference because the same exact or provable
        approximate algorithms used to compute the MAP solution can be used to
        compute the \divmbest solutions.
\end{itemize}

\subsection{Notation}
To refresh the notation we gave in Chapter~\ref{ch:introduction}, recall that we are given a set of
discrete random variables $\bX=\{X_s\;|\;s\in [n]\}$ (where $[n]\doteq {1,2,\dots,n}$),
each takes a value from a finite set, $x_s \in \calX_s$. Give a clique
$C\subseteq [n]$, from a set of cliques $C\in \calC$, let $\bx_C$ denote
$\{x_s\;|\; s\in C\}$, and the label space $\calX_C$ be the cartesian product of
the label spaces in the clique, $\times_{s\in C} \calX_s$.

\subsection{MAP problem}
Let $G=(V,E)$ be a graph defined over the random variables $\bX$, and
$\theta_C \;:\; \calX_C \rightarrow \mathbb{R}$ be functions defining the energy
over cliques in the graph. Let $\calI(C)$ be some index set over $C$ and let
$\calI \doteq \cup_C \calI(C)$ be the union over index sets of all cliques in
the graph. The maximum a-postiriori (MAP) problem is to find
the assignment $\bx\in \calX^n$
that minimizes the following energy function:
\begin{flalign}
    \min\limits_{\bx\in \calX^n} \sum\limits_{\alpha \in \calI}
    \theta_\alpha(\bx_\alpha) = \min\limits_{\bx\in\calX^n} \sum\limits_{s\in V}
    \theta_s(x_s) + \sum\limits_{(s,t)\in E} \theta_{st}(x_s,x_t),
\end{flalign}
where we've restricted the cliques to be over nodes and edges of the graph for
ease of exposition, but the method developed here apply to higher-order MRFs as
well.

\subsection{MAP integer program and its LP relaxation}
Representing the energy in exponential form, the MAP problem can be written
using the canonical overcomplete representation~\cite{wainwright2005} where the node and edge energy
functions can be defined as,
\begin{flalign}
    \theta_s(x_s) \doteq \sum\limits_{j\in \calX_s} \theta_{s;j}\mathbb{I}_{s;j}(x_s),
\end{flalign}
\begin{flalign} \theta_{st}(x_s,x_t) \doteq \sum\limits_{(j,k)\in
    \calX_s\times\calX_t}
    \theta_{st;jk}\mathbb{I}_{st;jk}(x_s,x_t),
\end{flalign}
using the following node and edge potential functions,
\begin{flalign}\label{eqn:nodepot}
    \mathbb{I}_{s;j}(x_s) \doteq \left\{\begin{array}{ll}
            1 & \mbox{if $x_s=j$,}\\
            0 & \mbox{otherwise}
        \end{array}\right. \quad \forall s\in V,\; j\in \calX_s ,
\end{flalign}
\begin{flalign}\label{eqn:pairpot}
    \mathbb{I}_{st;jk}(x_s,x_t) \doteq \left\{\begin{array}{ll}
            1 & \mbox{if $x_s=j$ and $x_t=k$,}\\
            0 & \mbox{otherwise}
    \end{array}\right. \quad \forall (s,t)\in E,\; (j,k)\in
    \calX_s\times\calX_t,
\end{flalign}
yielding the re-written MAP inference problem,
\begin{flalign}\label{eqn:mapproblem}
    \min\limits_{\bx\in \calX^n} \quad\sum\limits_{s\in V}\btheta_s \cdot
    \mathbb{I}_s + \sum\limits_{(s,t)\in E} \btheta_{st}\cdot
        \mathbb{I}_{st}
\end{flalign}
where for each clique $C$ we have the set of energies for all possible
configurations of $\bx_C$, $\btheta_C\doteq\{\theta_{C;\rho} \;|\;
\rho \in \calX_C\}$, and corresponding potential functions $\mathbb{I}_C \doteq
\{\mathbb{I}_{C;\rho}(\bx_C) \;|\; \rho \in \calX_C \}$.

Instead of minimizing over $\bx$ we could alternatively assume that there are
unknown variables $\bmu$ such that the above MAP inference problem is equivalent
to,
\begin{subequations}\label{eqn:mapip}
\begin{flalign}
    \min\limits_{\bmu_s,\bmu_{st}} &\quad\sum\limits_{s\in V}\btheta_s \cdot
    \bmu_s + \sum\limits_{(s,t)\in E} \btheta_{st}\cdot
    \bmu_{st} \hfill\label{eqn:mapconstr1}\\
        \mbox{s.t.} &\quad \sum\limits_{j\in \calX_s} \bmu_{s;j}(x_s) = 1
         &\forall s\in V, \label{eqn:mapconstr2}\\
        & \quad \sum\limits_{(j,k)\in \calX_s\times \calX_t}
        \bmu_{st;jk}(x_s,x_t) = 1 &\forall (s,t) \in E,\label{eqn:mapconstr3}\\
        & \quad \sum\limits_{j\in \calX_s} \bmu_{st;jk}(x_s,x_t) =
        \bmu_{t;k}(x_t) & \forall (s,t) \in E \quad \forall k \in
        \calX_t,\label{eqn:mapconstr4} \\
        & \quad \sum\limits_{k\in \calX_t} \bmu_{st;jk}(x_s,x_t) =
        \bmu_{t;k}(x_s) & \forall (s,t) \in E \quad \forall j \in
        \calX_s,\label{eqn:mapconstr5}\\
        & \quad \bmu_{s;j}(x_s),\;\bmu_{st;jk}(x_s,x_t)\in
        \{0,1\}\label{eqn:mapconstr6}
\end{flalign}
\end{subequations}
where the constraints enforce that each variable is assigned a single label and
the assignments agree across edges. Note that when indicator variable
$\bmu_A(v)$ is set to 1, this corresponds to $x_A$ taking label $v$.
If we let $\calL(G)$ denote the set of constraints in
(~\ref{eqn:mapconstr1})-(~\ref{eqn:mapconstr5}) then we can write
eqn.~\ref{eqn:mapip} more concisely as,
\begin{flalign}\label{eqn:mapipdiv}
    \min\limits_{\bmu\in\calL(G),\; \bmu_A(x_A)\in\{0,1\}}
    \sum\limits_{A\in V\cup E} \btheta_A \cdot \bmu_A.
\end{flalign}
The above integer program is equivalent to
the MAP problem in eqn.~\ref{eqn:mapproblem}, and is known to be NP-hard in
general. We describe a Linear Programming relaxation of this problem in Chapter~\ref{ch:introduction}. A good review is also found in~\cite{werner2007}.

\subsection{\divmbest: Formulation}\label{sec:divmbest}
This section presents the \divmbest formulation. The goal is to generate a
diverse set of low-energy (high-probability) solutions. The approach is an
iterative greedy algorithm --- in each iteration we find the lowest energy
solution that is at least some minimum dissimilarity from the previously
generated solutions. To measure the dissimilarity between solutions the
algorithm has access to a dissimilarity function $\Delta(\bmu^{(1)},\bmu^{(2)})$
between two solutions. Suppose that we have already computed the MAP solution,
which we denote as $\bmu^{(1)}$. In order to compute the second best
\emph{mode} (we use the term \emph{mode} loosely to mean low-energy diverse
solutions) we propose to solve the following general problem,
\begin{subequations}
\begin{flalign}
    \bmu^{(2)} = \argmin\limits_{\bmu\in\calL(G),\;
    \bmu_A(\bx_A)\in\{0,1\}} &\quad \sum\limits_{A\in V\cup U} \btheta_A \cdot \bmu_A
    \hfill\\
    \mbox{s.t.} &\quad \Delta(\bmu,\bmu^{(1)}) \geq k
    \label{eqn:deltaconstr},\hfill
\end{flalign}
\end{subequations}
which we call 2\textsc{Modes}$(\Delta,k)$. The constraint in (\ref{eqn:deltaconstr})
ensures that the next solution is at least $k$-units away from $\bmu^{(1)}$
according to $\Delta(\cdot,\cdot)$.
The choice of $\Delta$ and $k$ are design choices which we'll describe in
greater detail later.

Since \divmbest is an iterative greedy approach we can extend the 2\textsc{Modes}$(\Delta,k)$
problem to the \textsc{MModes}$(\Delta,\bk)$ problem ($\bk=\{k_i\;|\; i\in
[m-1]\}$) in a straightforward manner
by searching for the lowest energy solution that is $k_m$-units away from each
of the previous $m-1$ solutions,
\begin{subequations}\label{eqn:mmodes}
    \begin{flalign}
    \bmu^{(m)} = \argmin\limits_{\bmu\in\calL(G),\;
    \bmu_A(\bx_A)\in\{0,1\}} &\quad \sum\limits_{A\in V\cup U} \btheta_A \cdot \bmu_A
    \hfill\\
    \mbox{s.t.} &\quad \Delta(\bmu,\bmu^{(1)}) \geq k_1
    \label{eqn:deltaconstr1},\hfill\\
    &\quad \Delta(\bmu,\bmu^{(2)}) \geq k_2 \label{eqn:deltaconstr2},\hfill\\
    &\quad \vdots \hfill \\
    &\quad \Delta(\bmu,\bmu^{(m-1)}) \geq k_{m-1} \label{eqn:deltaconstrm}\hfill
\end{flalign}
\end{subequations}
\subsection{\divmbest: Lagrangian Relaxation and the Lagrangian dual
function}\label{sec:divmbestrelax}
%It should be clear that the integer program of eqn.~\ref{eqn:mmodes} is as
%difficult to solve as the MAP inference problem in eqn.~\ref{eqn:mapipdiv}, thus
%it is also NP-hard in general.
Given the extra inequality constraints of
(\ref{eqn:deltaconstr1})-({\ref{eqn:deltaconstrm}}) it's not clear how this
problem relates to common MAP inference approximations to the MAP
IP in eqn.~\ref{eqn:mapipdiv}. The Lagrangian
relaxation of \textsc{MModes}$(\Delta,\bk)$ that we get by \emph{dualizing} the constraints can be written as,
\begin{flalign}\label{eqn:lagrangian}
    f(\bfgreek{lambda}) = \min\limits_{\bmu\in\calL(G),\;
    \bmu_A(\bx_A)\in\{0,1\}} \sum\limits_{A\in V\cup E} \btheta_A\cdot \bmu_A -
    \sum\limits_{i=1}^{m-1} \lambda_i(\Delta(\bmu,\bmu^{(i)})-k_i),
\end{flalign}
where $\bfgreek{lambda}$ are the dual variables, also referred to as Lagrangian multipliers.
The Lagrange dual, $f(\bfgreek{lambda})$, minimizes a linear combination of the
solution energy and similarity to the previous solutions. The Lagrange
multipliers, $\bfgreek{lambda}$, are non-negative and control the amount of penalty incurred for
violating the minimum dissimilarity constraints. The following theorem holds for
the Lagrange dual,

\begin{prop}{}\label{thm:prop1}
    The Lagrangian dual function, $f(\blambda)$, is a piece-wise linear function that
    is concave in $\blambda$ and is a lower-bound on the optimal value
    $\bmu^{(m)^*}$ of the
    primal problem, \textsc{MModes}$(\Delta,\bk)$, for all values of
    $\blambda\ge 0$.
\end{prop}

\begin{proof}
    The proof can be found in~\cite{boyd2004}, and is reproduced here
    for completeness.
    \begin{enumerate}[(i)]
        \item
    The Lagrange dual can be written in the following form,
    $f(\blambda)=\min_{\bmu} \ba_{\bmu}\cdot \blambda + b_{\bmu}$, which shows
    that $f(\blambda)$ is a piece-wise linear function. Assume only two different
    values of $\bmu$, namely $\bmu_1$ and $\bmu_2$, and corresponding linear
    functions, $f_{\bmu_1}$ and $f_{\bmu_2}$, where
    $f_{\bmu_j}=\ba_{\bmu_j}\cdot \blambda + b_{\bmu_j}$.  We can then write the
    pointwise minimum $f$ as,
    \begin{flalign}
        f(\blambda) = \min \{ f_{\bmu_1}(\blambda),\; f_{\bmu_2}(\blambda) \}
    \end{flalign}
    Let $0\leq \theta \leq 1$ and consider $\blambda_1$, $\blambda_2\in \bs{dom}\;
    f$. Then we have,
    \begin{flalign}
        f(\theta \blambda_1 + (1-\theta)\blambda_2) =& \min\; \{f_{\bmu_1}(\theta
            \blambda_1 + (1-\theta)\blambda_2),\; f_{\bmu_2}(\theta \blambda_1
    +(1-\theta) \blambda_2) \}\nonumber\\
    =&\min\; \{\theta f_{\bmu_1}(\blambda_1)+
    (1-\theta)f_{\bmu_2}(\blambda_2),\; \theta f_{\bmu_2}(\blambda_1) +
(1-\theta)f_{\bmu_2}(\blambda_2) \} \nonumber\\
\geq & \theta \min\; \{f_{\bmu_1}(\blambda_1),\; f_{\bmu_2}(\blambda_1) \} +
(1-\theta)\min\; \{f_{\bmu_1}(\blambda_2),\; f_{\bmu_2}(\blambda_2)\}\nonumber\\
=&\theta f(\blambda_1) + (1-\theta) f(\blambda_2),
    \end{flalign}
    which shows that $f(\blambda)$ is concave in $\blambda$.
    We can easily extend this to $k$ functions $f_{\bmu_1},\dots,f_{\bmu_k}$,
    for $k$ finite to show the general result.
        \item
            To see that $f(\blambda)$ is a lower-bound on the optimal value
            $\bmu^{(m)^*}$ of the primal problem, \textsc{MModes}$(\Delta,\bk)$,
            consider a feasible point $\tilde{\bmu}$ of the primal problem. Then
            $\Delta(\tilde{\bmu},\bmu^{(i)})-k_m \geq 0$ for all $i\in[m-1]$.
            This gives,
            \begin{flalign}
                L(\tilde{\bmu},\blambda)
                \leq \sum\limits_{A\in V\cup E} \btheta_A\cdot \tilde{\bmu}_A,
            \end{flalign}
            where $L(\bmu,\blambda)$ is the Lagrangian,
            \begin{flalign}
                L(\bmu,\blambda)=
                \sum\limits_{A\in V\cup E} \btheta_A\cdot \bmu_A
                -\sum\limits_{i=1}^{m-1} \lambda_i(\Delta(\bmu,\bmu^{(i)})-k_i).
            \end{flalign}
            Therefore we have,
            \begin{flalign}
                f(\blambda) = \min\limits_{\bmu\in\calL(G),\;
                \bmu_A(\bx_a)\in\{0,1\}} L(\bmu,\blambda) \leq
                L(\tilde{\bmu},\blambda)
                \leq \sum\limits_{A\in V\cup E} \btheta_A\cdot \tilde{\bmu}_A.
            \end{flalign}
            Since $f(\blambda)\leq \sum\limits_{A\in V\cup E}
            \btheta_A\cdot\tilde{\bmu}_A$ holds for \emph{all} feasible points
            $\tilde{\bmu}$ it also holds for $\bmu^{(m)^*}$.
    \end{enumerate}
\end{proof}
\subsection{Diversity Functions}
The \divmbest formulation in eqn.~\ref{eqn:mmodes} relies on defining a
dissimilarity function $\Delta(\cdot,\cdot)$ between two solutions. It turns out
that the \divmbest formulation is general enough to include other methods as
special cases, through the right choice of dissimilarity function. Below are two
such cases.

\subsubsection{0-1 dissimilarity and M-Best MAP}
If we let $\Delta(\bmu,\bmu^{(i)})=[[\bmu\neq \bmu^{(i)}]]$, where $[[\cdot]]$ is an
indicator function which is 1 if the predicate is true and 0 otherwise, and set $k_i=1$, such
that the constraints in (\ref{eqn:deltaconstr1})-(\ref{eqn:deltaconstrm})
are of the form $\Delta(\bmu,\bmu^{(i)}) \geq 1$ for $i\in[m-1]$, then we
recover the $M$-Best MAP problem. These constraints force the $m$-th
solution to be different from each of the previous $(m-1)$ solutions in at
least one location.

\subsubsection{Local dissimilarity and N-Best Maximal Decoding of Park and Ramanan}
If we let $\Delta(\bmu,\bmu^{(i)})=\max_{s\in V} \Delta_n(\bmu_s,\bmu_s^{(i)})$,
where there's potentially a different dissimilarity function, $\Delta_n$, defined for each node
in $V$, and set $k_i=1$, then we recover the N-Best maximal decoding of Park and
Ramanan~\cite{park2011}. In terms of the local measure of dissimlarity, $\Delta_n$, setting $k_i=1$ forces the $m$-th solution to be
different from each of the previous $(m-1)$ solutions at least one node.

\vspace{2em}
Some dissimilarity functions can be decomposed according to the structure of the
graph which provide some nice properties. A specific class of decomposable
dissimilarity functions that are used extensively in the experiments are
dot-product dissimilarity functions.

\subsubsection{Dot-product dissimilarity}
If we let $\Delta(\bmu,\bmu^{(i)})=-\sum_{s\in V} \bmu_s^TW\bmu_s^{(i)}$ and if
the solution vectors, $\bmu$ and $\bmu^{(i)}$ are discrete, then
$\Delta(\bmu,\bmu^{(i)})$ encodes the weighted Hamming distance between the two
solutions, where the weights $W$ capture the importance of various pairwise
labellings across the two solutions. If $W=I$ (\ie $W$ is set to the identity
matrix) then $\Delta$ is the straight-forward Hamming distance between the two
solutions. It's interesting to see the form of the Lagrangian when $\Delta$ is
the general dot-product dissimilarity between two solutions,
\begin{multline}
    \sum\limits_{A\in V\cup E} \btheta_A\cdot \bmu_A -
    \sum\limits_{i=1}^{m-1} \lambda_i(\Delta(\bmu,\bmu^{(i)})-k_i) = \\
    \sum\limits_{s\in V}\left(\btheta_s + \sum\limits_{i=1}^{m-1}
    \lambda_i W\bmu_s^{(i)}\right)\cdot\bmu_s + \sum\limits_{(s,t)\in
    E}\btheta_{st}\cdot\bmu_{st} + \sum\limits_{i=1}^{m-1} \lambda_i k_i.
    \label{eqn:dotproddis}
\end{multline}
Note that $f(\blambda)$ is now comprised of the three terms in
eqn.~\ref{eqn:dotproddis} where the first two terms are simply the MAP problem
of eqn.~\ref{eqn:mapipdiv}, with modified unary energies, and the last term is
independent of $\bmu$. When $W=I$ there is a cost paid, proportional to
$\lambda_i$, for setting local parts of the current solution, $\bmu_s$, equal to
$\bmu_s^{(i)}$ of each of the previous $i\in [m-1]$ solutions. When $W$ is non-identity the cost is spread
over larger parts of the assignment.

Thus in the case where the $\Delta$-function is a Hamming distance, since the
first two terms of $f(\blambda)$ are the same as for the$MAP$ problem, and
the last term is independent of the minimization variables, we can compute $f(\blambda)$
using any MAP inference machinery (exact or approximate) that was
used to compute the first solution $\bmu^{(1)}$. Moreover, the edge energies are
left \emph{unchanged} which means that certain classes of efficient MAP inference --- such
graph-cut algorithms that require submodular edge potentials --- remain viable options for computing subsequent solutions.

\subsubsection{Higher-order dissimilarity}
Consider higher order dissimilarity functions of the form
$\Delta(\bmu,\bmu^{(i)})=\sum_{C\in \calI} \Delta_{C}(\bmu_C,\bmu_C^{(i)})$,
where $C$ are subsets of variables, $\calI$ is an index set on subsets of
variables, and $\Delta_C(\cdot,\cdot)$ has some structure allowing for efficient
message passing. Unlike dot-product dissimilarity the higher-order dissimilarity
does not decompose over nodes in the graph. Examples include cardinality
potentials~\cite{tarlow2010}, pattern-based
sparse higher order potentials~\cite{komodakis2009,rother2009}, and lower
linear-envelope potentials~\cite{kohli2010}.
We'll describe how efficient inference on the $\Delta$-augmented energy of the
Lagrangian dual function can be performed in the next section.

\subsection{Supergradient Ascent on Lagrangian dual function}
As previously mentioned the Lagrangian relaxation, $f(\blambda)$, is a lower-bound on
the value of the primal \divmbest problem. We would like to find the $\blambda^*$ such
that $f(\blambda^*)$ provides the tightest lower bound on the value of the
primal problem. If there is no duality gap between the primal and dual problem
values then \emph{strong duality} holds and solving the primal problem is equivalent
to solving the Lagrangian dual relaxation. To find the tightest lower bound on
the primal problem we need to solve the following \emph{Lagrange dual problem},
\begin{subequations}\label{eqn:lagdual}
\begin{flalign}
    \max\limits_{\blambda} &\quad f(\blambda)\hfill\\
    \mbox{s.t.}&\quad \blambda \geq 0\hfill
\end{flalign}
\end{subequations}
Recall that $f(\blambda)$ is a piece-wise linear function that is concave in
$\blambda$ (prop.~\ref{thm:prop1}). We can solve problem~\ref{eqn:lagdual} using a projected
\emph{supergradient ascent} algorithm (alg.~\ref{alg:supergrad}) on
$\blambda$~\cite{shor2012}.

\begin{algorithm}[!h]\caption{Projected Supergradient Ascent
    (cf.~\cite{shor2012})}\label{alg:supergrad}
\begin{algorithmic}[1]
    \State $t \gets 1$
    \State $\{\alpha_t\; |\; \alpha_t \geq 0,\; \lim_{t\rightarrow \infty}
    \alpha_t = 0,\; \sum_{t=0}^{\infty} \alpha_t = \infty\}$.
    \Comment{define sequence of step-sizes}
    \State Initialize $\blambda^{(0)}$
    \State $f_{best}^{(0)} \gets f(\blambda^{(0)})$
    \Repeat
    \State $\blambda^{(t)} \gets \blambda^{(t-1)} + \alpha_t\nabla
    \label{alg:update}
        f(\blambda^{(t-1)})$
        \State $\blambda^{(t)} \gets [\blambda^{(t)}]_+$
        \Comment{project onto positive orthant}
        \State $f^{(t)}_{best} \gets \min\{f^{(t-1)}_{best},f(\blambda^{(t)})\}$
        \Comment{keep track of best point found thus far}
        \State $t \gets t+1$
        \Until {$\lim\limits_{t\rightarrow \infty} |f_{best}^{(t)}-f^*| \leq \epsilon$}
    \Comment{stopping criteria}
\end{algorithmic}
\end{algorithm}
In alg.~\ref{alg:supergrad} the supergradient of $f$ at $\blambda^{(t)}$ is
denoted by $\nabla f(\blambda^{(t)})$. In order to guarantee convergence of the
algorithm a convergent sequence of non-negative step-sizes, $\{\alpha_t\}$, has to be chosen such that $\lim_{t\rightarrow \infty} \alpha_t=0$, and $\sum_{t=0}^{\infty} \alpha_t
= \infty$ (\eg $\alpha_t = \Gamma/\sqrt{t}$, where $\Gamma >0$). In practice the stopping criteria on the last line of alg.~\ref{alg:supergrad} is such that if the value of $f^{(t)}_{best}$ does not improve the algorithm terminates.

Recall that $f(\blambda)$ is a point-wise minimum of a set of linear functions,
\begin{flalign}
    f(\blambda) = \min\limits_{\bmu} \ba_{\bmu}\cdot \blambda + b_{\bmu},
\end{flalign}
where the supergradient of $f$ is $\nabla f(\blambda) =
\ba_{\widehat{\bmu}(\blambda)}$, with $\widehat{\bmu}(\blambda)\doteq \argmin_{\bmu}
\ba_{\bmu}\cdot \blambda + b_{\bmu}$.
\begin{proof}
    To see this consider the definition of the supergradient, namely $g$ is a
    supergradient of a concave function $f$ at $x\in \bs{dom}\; f$ if
    \begin{flalign}
        f(y) \leq f(x) + g^T(y-x), \quad \forall y\in \bs{dom}\; f.
    \end{flalign}
    Consider $f(\blambda')$, for some $\blambda'\in \bs{dom}\; f$, which equals
    $\ba_{\widehat{\bmu}(\blambda')}\cdot \blambda'
    +b_{\widehat{\bmu}(\blambda')}$, where
    $\widehat{\bmu}(\blambda')\doteq \argmin_{\bmu}\ba_{\bmu}\cdot
    \blambda'+b_{\bmu}$, by definition. Clearly the following inequality holds for all $\blambda$,
    \begin{flalign}
        f(\blambda')=\ba_{\widehat{\bmu}(\blambda')}\cdot \blambda'+b_{\widehat{\bmu}(\blambda')} \leq
    \ba_{\widehat{\bmu}(\blambda)}\cdot \blambda' +b_{\widehat{\bmu}(\blambda)}.
    \end{flalign}
    We can add and subtract the quantity $\ba_{\widehat{\bmu}(\blambda)}\cdot
    \blambda$ to the RHS to get,
    \begin{flalign}
    f(\blambda') =\ba_{\widehat{\bmu}(\blambda')}\cdot \blambda'+b_{\widehat{\bmu}(\blambda')}
    &\leq
    \ba_{\widehat{\bmu}(\blambda)}\cdot \blambda' +b_{\widehat{\bmu}(\blambda)}
    +\ba_{\widehat{\bmu}(\blambda)}\cdot \blambda -
    \ba_{\widehat{\bmu}(\blambda)}\cdot \blambda \\
    &=f(\blambda) + \ba_{\widehat{\bmu}(\blambda)}(\blambda'-\blambda),
    \end{flalign}
    thus $\nabla f(\blambda)\doteq
    \ba_{\widehat{\bmu}(\blambda)}$ is a supergradient of
    $f$ at $\blambda$.
\end{proof}
The supergradient of the Lagrangian dual function at $\blambda$ is thus,
\begin{flalign}
    \nabla f(\blambda) = -\left[ \begin{array}{c}
            \Delta(\widehat{\bmu}(\blambda),\bmu^{(1)})- k_1 \\
            \vdots\\
            \Delta(\widehat{\bmu}(\blambda),\bmu^{(m-1)})- k_{m-1}
        \end{array}
    \right],
\end{flalign}
where $\widehat{\bmu}(\blambda)$ is optimal solution to
problem~\ref{eqn:lagrangian} for the current setting of $\blambda$. The
supergradient has an intuitive meaning in relation to the projected
supergradient descent algorithm presented in alg.~\ref{alg:supergrad}. If at
time $t$, a constraint is violated, say,
$\Delta(\widehat{\bmu}(\blambda^{(t)}),\bmu^{(i)}) - k_i < 0$, for some $i$ ---
then the supergradient vector with respect to $\blambda^{(t)}$ will be positive
at index $i$ and the update in alg.~\ref{alg:supergrad}-line~\ref{alg:update}
will increase the cost, $\lambda^{(t+1)}_i$, for violating the $i$-th constraint.
Conversely, $\nabla f(\blambda)$ is negative for constraints that are
satisfied, thus reducing the corresponding costs, $\lambda_j^{(t+1)}$, for
violating those constraints (because the constraints are probably not active)
and thus allowing for lower energy solutions.

One nice property is that in each iteration of alg.~\ref{alg:supergrad}, the inference problem that is
needed to be solved for $\blambda^{(t)}$ is very similar to that used to solve
$\blambda^{(t-1)}$. Thus warm-starting the solver for $\blambda^{(t)}$ with the
solutions $\blambda^{(t-1)}$ can be beneficial (\eg re-using search trees in
graph-cuts~\cite{kohli2005}, or reusing messages in dual-decomposition).

\subsection{How tight is the Lagrange relaxation?}\label{sec:lagrangetight}\footnotemark\footnotetext{Results in  this section are due to Dhruv Batra~\cite{batra2012}}
As mentioned earlier we'd like to find the tightest lower bound on the primal
problem, \textsc{MModes}. We gave a Lagrange relaxation of
\textsc{MModes}$(\Delta,\bk)$,
termed $f(\blambda)$ which we showed to be a lower bound on the value of
\textsc{MModes}$(\Delta,\bk)$ for all feasible $\bu$, and all $\blambda\geq 0$.
We know state the following result on the Lagrangian dual problem,
$\max_{\blambda\geq 0}f(\blambda)$, which is the tightest lower bound on
\textsc{MModes}$(\Delta,\bk)$.
\begin{theorem}
    \begin{enumerate}[(i)]
        \item The Langrangian dual problem, $\max_{\blambda\geq 0} f(\blambda)$
            is equivalent to solving the following relaxation of
            \textsc{MModes}$(\Delta,\bk)$,
            \begin{subequations}\label{eqn:mmodeslagr}
            \begin{flalign}
                \min\limits_{\bmu} &\quad \sum\limits_{A\in V\cup E} \btheta_A\cdot
                \bmu_A\\
                \mbox{s.t.} &\quad \bmu\in co\{\bmu_A(x_A) \in \{0,1\}\;|\; \bmu
                \in \calL(G)\}\\
                &\quad \Delta(\bmu,\bmu^{(i)})\geq k_i &\forall i\in [m-1]
            \end{flalign}
            \end{subequations}
            where $co\{\cdot\}$ is the convex hull of a set of discrete
            solutions.
        \item Generally the Lagrangian relaxation is not guaranteed to be tight,
            but, for some dissimilarity functions $\Delta(\cdot,\cdot)$, the convex
            hull cat be replaced with the discrete solutions
            $\mu_A(x_A)\in\{0,1\}$, $\bmu\in\calL(G)$ themselves resulting
            in a tight Lagrangian relaxation.
    \end{enumerate}
\end{theorem}
\begin{proof}
    \begin{enumerate}[(i)]
        \item The result follows directly from the following equivalent LP-dual
            problems shown by Geoffrion~\cite{geoffrion1974},
            \begin{subequations}
                \begin{flalign}
                    \mbox{\emph{(dual)}} \qquad &\max\limits_{\blambda\geq 0}\;
                    \min\limits_{\bx\geq 0} \bc^T\bx - \blambda^T(A\bx-\bb)\\
                    \mbox{s.t.}\qquad & B\bx \geq \bd\\
                                      & x_j\in \mathbb{I},\; j\in \calI,
                \end{flalign}
            \end{subequations}
            and,
            \begin{subequations}
                \begin{flalign}
                    \mbox{\emph{(primal)}} \qquad &\min\limits_{\bx} \bc^T\bx\\
                    \mbox{s.t.}\qquad & A\bx\geq \bb\\
                                      & \bx\in co\{\bx\geq 0,\; B\bx\geq \bd,\;
                x_j\in \mathbb{I},\; j\in\calI\},
                \end{flalign}
            \end{subequations}
            where $\calI$ is an index set over variables. Making appropriate
            substitutions gives the desired result.
        \item Recall from Chapter~\ref{ch:introduction} that $\mathbb{M}(G)$ is the set of
            realizable marginal distributions over graph $G$. Moreover in
            ~\cref{sec:mapproblem} we mentioned the result that
            $\mathbb{M}(G)$ is the convex hull of the overcomplete
            representation defined in eqn.~\ref{eqn:nodepot} and
            eqn.~\ref{eqn:pairpot} over the finite index set in
            eqn.~\ref{eqn:pairindexset}, where the indicator functions take on
            $\{0,1\}$ values and are the extreme points of the polytope.
            Therefore $\mathbb{M}(G)$ is exactly $co\{\bmu_A(\bx_A \in\{0,1\} \;
            | \; \bmu \in \calL(G)\}$.

            It's also a well known fact that minimizing a linear objective over
            a convex hull has the optimal solution at some extreme point of the
            convex hull, therefore it's equivalent to minimizing over the
            discrete solutions (which are the extreme points). However, we also
            have the diversity constraints. The set of feasible solutions for
            problem~\ref{eqn:mmodeslagr} are
            those in the set,
            \begin{flalign}
                P\doteq \left\{\bmu\;|\;\Delta(\bmu,\bmu^{(i)})\geq k_i\ \forall
                    i\; \cap\;
            co\{\bmu_A(x_A)\in \{0,1\} \; | \; \bmu\in \calL(G)\}\right\}.
            \end{flalign}
            Therefore when $P$ is a polytope with integral vertices we can remove
            $co\{.\}$ from the constraints. Since $co{.}$ has integral vertices
            we need to check whether $\Delta(\bmu,\bmu^(i))\geq k_i$ introduce
            fractional vertices. When we have the M-Best MAP dissimilarity
            function, Fromer and Globerson~\cite{fromer2009} presented spanning-tree
            inequalities that are guaranteed not to introduce fractional
            vertices when $G$ is a tree.

            In general, though, when no assumption on $\Delta(\cdot,\cdot)$ are
            made, the Lagrangian relaxation is not guaranteed to be tight.
            Consider the dot-product (Hamming distance) dissimlarity,
            $\Delta(\bmu,\bmu^{(i)})=-\sum_{s\in V} \bmu_s^T \bmu_s^{(i)}$. This
            dissimilarity introduces fractional vertices which was described as
            a counter-example by Fromer and Globerson~\cite{fromer2009},

            \emph{Counter-example:} Suppose we have a graph consisting of two
            nodes with an edge between them, and each node takes on two labels.
            Let the node energies be $\btheta_1=\btheta_2=(0,0)$, and the edge
            energy be $\theta_{12}=(0,10,10,10)$. The MAP solution (minimizing
            solution over this graph) is $(0,0)$. To find the second best
            solution --- which is constrained to be different from the MAP solution
            with $k=1$ --- we introduce the constraint, $-\bmu_1(0)-\bmu_2(0)\geq -1
            \Longrightarrow \bmu_1(0)+\bmu_2(0) \leq 1$. The solution
            minimizing the energy over the graph with this new constraint is
            $(0.5,0.5)$ with energy value $5$, wheres the other non-MAP
            solutions have energy value $10$. Since the solution is fractional
            the Lagrangian relaxation is not tight.
    \end{enumerate}

\end{proof}
\subsection{Computing Supergradient under different diversity functions}
Recall that for certain diversity functions such as the Hamming distance
dissimilarity functions we can compute $f(\blambda)$ using the same MAP
inference machinery (exact or provably approximate) that was used to compute the
first solution $\bmu^{(1)}$ because imposing the dissimilarity function between
solutions amounts to only modifying the node energies but leaving the edge
energies unaffected.

Not all dissimilarity functions share this decomposability property, especially
when dissimlarity is measured between subsets of variables, which we term
higher-order dissimilarity functions. However, there are some higher-order
dissimilarity functions where the individual terms over subset of variables,
$\Delta_C(\cdot,\cdot)$ (where $C$ is a subset of variables), have some
structure that can be exploited in order to carry on efficient inference over
the $\Delta$-augmented energy. Here we mention how for such $\Delta$-augmented
energies, where the higher-order dissimilarities contain specific structure,
efficient energy minimization can be performed via dual-decomposition based
message-passing algorithms.

For simplicity of exposition, consider the Lagrangian relaxation to the
\textsc{2MModes} problem,
\begin{flalign}
    \min\limits_{\bmu\in\calL(G),\;
    \bmu_A(\bx_A)\in\{0,1\}} \sum\limits_{A\in V\cup E} \btheta_A\cdot \bmu_A -
    \overline{\lambda}_1\Delta(\bmu,\bmu^{(1)})- \overline{\lambda}_1 k_1.
\end{flalign}
and suppose $\Delta(\bmu,\bmu^{(1)})$ is a higher-order dissimilarity function
that does not decompose according to nodes in the graph. Assume $\overline{\lambda}_1$ to
be a \emph{fixed} variable and let
$\theta_{hop}^{(1)}(\bmu)\doteq-\overline{\lambda}_1\Delta(\bmu,\bmu^{(1)})$. Since
$\overline{\lambda}_1 k_1$ is independent of $\bmu$ the problem is reduced to,
\begin{flalign}\label{eqn:hopprob}
    \min\limits_{\bmu\in\calL(G),\;
    \bmu_A(\bx_A)\in\{0,1\}} \sum\limits_{A\in V\cup E} \btheta_A\cdot \bmu_A +
    \theta^{(1)}_{hop}(\bmu).
\end{flalign}
In contrast to dissimilarity functions that do decompose over nodes, even if the
MAP problem (eqn.~\ref{eqn:mapipdiv}) could be solved efficiently,  this
$\Delta$-augmented energy function is difficult to solve because of the higher
order potential, $\theta_{hop}^{(1)}$. However, for certain higher-order
potentials with structure where messages can be efficiently computed
dual-decomposition based message-passing algorithms can be used to approximate the
supergradient.

\subsection{Dual-Decompostition and the approximate supergradient for higher
order potentials}
In order to minimize energy in problem~\ref{eqn:hopprob} we apply the
dual-decomposition approach~\cite{bertsekas1999,guignard2003,komodakis2007}. We introduce auxiliary variables
for each of the optimization variables, $\bmu_s$ in problem~\ref{eqn:hopprob}
and write the following \emph{equivalent} problem,
\begin{subequations}\label{eqn:dualdecomp}
\begin{flalign}
    \min\limits_{\bmu,\;
    \bmu^{hop}} &\quad\sum\limits_{A\in V\cup E} \btheta_A\cdot \bmu_A +
    \theta^{(1)}_{hop}(\bmu^{hop})\hfill\\
    \mbox{s.t.} &\quad \bmu\in\calL(G) \\
                &\quad \bmu_s^{hop} = \bmu_s &\forall s\in V,\label{eqn:ddconstr}\\
                &\quad \bmu_A,\; \bmu_A^{hop}\in \{0,1\}
\end{flalign}
\end{subequations}
where we are now minimizing over two sets of variables $\bmu$, and $\bmu^{hop}$
which are constrained to agree. Introducing Lagrange multiplies $\nu_s$ for each
constraint in line~\ref{eqn:ddconstr}, we can write the Lagrangian relaxation of
problem~\ref{eqn:dualdecomp} as,
\begin{subequations}\label{eqn:ddlagrange}
\begin{flalign}
    g(\bnu)=\min\limits_{\bmu,\;
    \bmu^{hop}} &\quad\sum\limits_{A\in V\cup E} \btheta_A\cdot \bmu_A +
    \theta^{(1)}_{hop}(\bmu^{hop}) - \sum\limits_{s\in V} \bnu_s (\bmu_s^{hop
    } - \bmu_s)\hfill\\
    \mbox{s.t.} &\quad \bmu\in\calL(G) \\
                &\quad \bmu_A,\; \bmu_A^{hop}\in \{0,1\}
\end{flalign}
\end{subequations}
We can rewrite the above objective as a sum of two separate minimizations, one
over the variables $\bmu$ and the other over $\bmu^{hop}$,
\begin{subequations}\label{eqn:ddindep}
\begin{flalign}
    g(\bnu) = &\min\limits_{\bmu\in\calL(G),\; \bmu_{s},\bmu_{st}\in \{0,1\}}
    \sum\limits_{s\in V} (\btheta_s + \bnu_s)\bmu_s + \sum\limits_{(s,t)\in E}
    \btheta_{st} \bmu_{st} \label{eqn:ddmap}\\
    +&\min\limits_{\bmu^{hop}_A\in\{0,1\}} \theta_{hop}^{(1)}(\bmu^{hop}) -
    \sum\limits_{s\in V}\bnu_s \bmu_s^{hop} \label{eqn:ddhop}
\end{flalign}
\end{subequations}
To find the tightest Lagrangian relaxation  we want to maximize $g(\bnu)$,
\begin{flalign}
    \max\limits_{\bnu\in\mathbb{R}^n} g(\bnu)
\end{flalign}
which we can do using the supergradient method of alg.~\ref{alg:supergrad}.
The term in line~\ref{eqn:ddmap} is the original MAP problem with perturbed
unary potentials, so the minimization over $\bmu$ can be carried out using the
same efficient inference machinery used to compute the MAP solution. The term in
line~\ref{eqn:ddhop} is a minimization over $\bmu_A^{hop}$ which is efficiently
computable for higher order potentials that have structure such as cardinality
potentials~\cite{gupta2010,tarlow2010}, lower linear-envelope
potentials~\cite{kohli2010} or sparse
higher-order potentials~\cite{komodakis2009,rother2009}. For example, in the case of cardinality
potentials Gupta et al.~\cite{gupta2010} and Tarlow et al.~\cite{tarlow2010} message-passing
algorithms to compute them.

\subsection{Setting k: the amount of diversity}
The Lagrangian relaxation to the \textsc{MModes}$(\Delta,\bk)$ problem provides a trade-off
between minimizing the energy and the amount of diversity between solutions.
Choosing the value $k$ relates to the minimum amount of diversity we want between
solutions. Choosing the right value for $k$ is important because if the value of
$k$ is too small then the next solution might not be outside the energy valley
of one of the previous solutions. On the other hand, too large a value for $k$
could mean than several valid modes would be ignored. Also note that for each value of $k$ there is a different value of $\blambda$ that minimizes the Lagrangian relaxation, $\hat{\blambda}(\bk)=\argmin_{\blambda(\bk) \geq 0} f(\blambda(\bk))$. This means that
we would have to search for the optimal value of $k$, where for each value we'd
have to run the supergradient ascent algorithm, which is expensive.
Alternatively we can directly do grid search over values of $\blambda$. This is
analogous to tuning the regularization parameter in learning. Since directly
tuning $\blambda$ is more efficient in practice, the amount of diversity is
tuned in the experiments found in later chapters using cross-validation on
$\blambda$, instead of directly searching over $\bk$.

\subsection{Summary}
To summarize, this chapter has presented the \divmbest problem which finds a
diverse set of highly probable solutions under a discrete probabilistic model.
The \divmbest problem is a generalization of the M-best MAP problem. The
\divmbest problem is formulated as a Lagrangian relaxation of an integer linear
program that involves solving the $\Delta$-augmented energy minimization
problem which minimizes a linear combination of the energy and similarity to
previous solutions. For certain classes of $\Delta$-function, the \emph{modes}
of the underlying distribution can be computed using the same inference
algorithms that are used to compute the MAP solution.

The \divmbest method provides an alternative approach to image segmentation ---
instead of devising complex models with higher-order terms that are hard to
optimize over one can use simpler models in which exact or approximate MAP
inference is tractable. With proper choice of $\Delta$-function the same
inference machinery can be used by the \divmbest algorithm to obtain a set of
diverse solutions. This small set of segmentations can then simply be evaluated by a more complex
model in order to rank them. We introduce the ranking mechanism in
chapter~\ref{ch:divrank}.

The \divmbest approach is a greedy approximate strategy to finding a set of highly probable and yet diverse solutions under the model. In contrast Kirillov~\etal~\cite{kirillov2015} present the joint \divmbest problem which simultaneously finds all $M$ segmentations using an approximate solver that minimizes a single joint energy. In contrast to the \divmbest formulation in this chapter their approach gives better quality results at the cost of significantly slower run time. For submodular energies Kirillov~\etal~\cite{kirillov2015m} later propose an exact solver which is efficient albeit slower than the sequential approach. This is extended in~\cite{kirillov2016joint}, specific to binary submodular energies, to give a solver that is faster than the sequential approach presented in this chapter.

%% file: Chapters/Chapter03.tex
\chapter{DivMBest+ReRank}\label{ch:divrank}
%************************************************
\subfile{\main/Sections/Divrank}
\printbibliography

%% file: Sections/Divrank.tex
There are many confounding factors that make semantic segmentation an inherently
difficult task --- from inter and intra object occlusion to lighting and varying
appearance and pose. A segmentation algorithm will confront all these sources of
uncertainty. However, devising fully probabilistic models that can incorporate
all confounding factors in order to reason about the distribution over all
possible segmentations jointly is usually intractable. This leads to two separate
approaches to devising segmentation models. We can either build
\begin{enumerate}
    \item \emph{Restrictive Probabilistic Models} that can make efficient joint
        predictions over a posterior distribution of all variables of interest
        at the cost of limited prediction capacity due to simplifying
        independence assumptions, or
    \item \emph{Expressive Feed-Forward Models} that can incorporate more complex
        interaction of variables by using simple feed-forward predictions but
        propagate uncertainty by not modelling all the variables in a
        probabilistic joint-prediction framework.
\end{enumerate}
Semantic segmentation models that fall into the first approach include
Conditional Random Field (CRF) models such
as~\cite{boix2012,kohli2009,ladicky2009}. To make joint prediction
on all variables in a CRF tractable simplifying independence assumptions are
usually made such as only local variable interactions that are associative or
attractive~\cite{ladicky2009}. The second approach includes feed-forward pipelines
like~\cite{arbelaez2012,carreira2012,gu2009} that find regions that are scored and then combined into a
segmentation. The feed-forward approach can incorporate rich dependencies
between regions that are difficult to capture in a tractable CRF, but errors
propagate and accumulate in the pipeline.

\begin{figure}[!t]%{I}{.5\textwidth}
    \floatbox[{\capbeside\thisfloatsetup{capbesideposition={right,top},capbesidewidth=0.5\textwidth}}]{figure}%[\FBwidth]
    {\caption{{An overview of the \divrank approach. In \emph{Stage 1} diverse segmentations are computed
%using the \divmbest algorithm
from a tractable probabilistic model. These are fed to a large-margin \reranker in \emph{Stage 2}.
The top \reranked segmentation is returned as the final solution. Even though the most probable
segmentation from Stage 1 is incorrect, the set of segmentations does contain an accurate solution, which the \reranker
is able to score to the top.}}
    \label{fig:divrankteaser}}
    {\includegraphics[width=0.5\textwidth]{\main/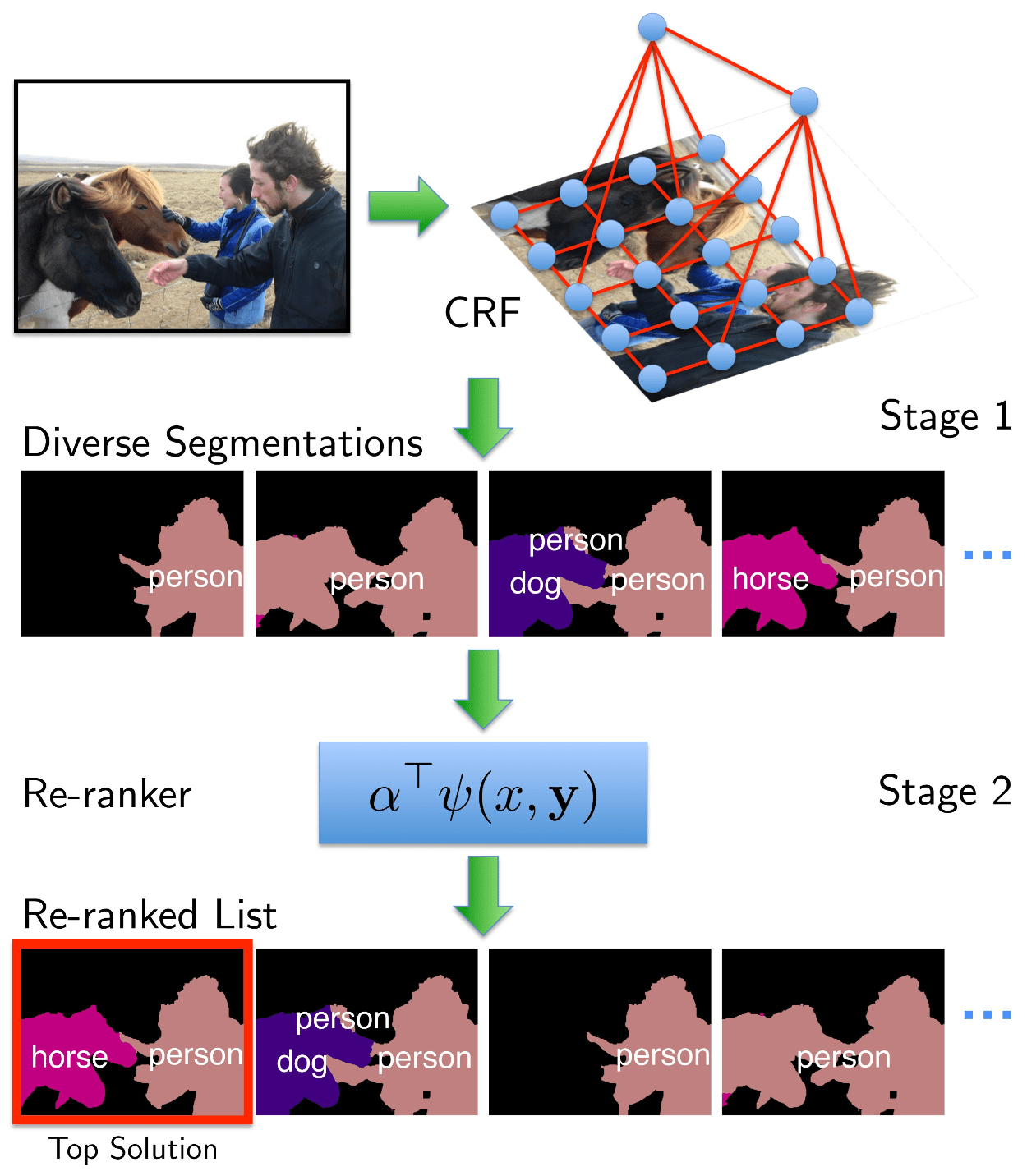}}
\end{figure}

This chapter introduces a two-stage hybrid approach called \divrank that
leverages both approaches. The first stage consists of a tractable
probabilistic model that reasons about an exponentially large output state-space
and makes joint predictions --- but crucially outputs a diverse set of plausible
segmentations not just a single one. The second stage of the approach is a
discriminative re-ranker that is free to use arbitrarily complex features, and
attempts to pick out the best segmentation from this set.
Figure~\ref{fig:divrankteaser} gives an
illustration of this approach.

\divrank approach to semantic segmentation has several key advantages:
\begin{itemize}
    \item \emph{Global optimization over a simple model.} The first stage of
        this approach is able to perform global optimization over all the
        variables of interest, in a tractable albeit imperfect model to find a
        small set ($\approx 10-30$) of plausible hypotheses. Experimentally we
        find that typically at least one of these solutions is highly accurate.
    \item \emph{Rich (higher-order) features in re-ranker.} Since the number of
        segmentations that the re-ranker needs to consider is small we do not
        have to worry about tractability issues when designing re-ranker
        features. The re-ranker is free to use arbitrarily complex features that
        would be intractable to add to the probabilistic model in the first
        stage. This is because the re-ranker does not need to \emph{optimize} over all
        possible segmentations but merely \emph{evaluate} these features on a small set
        of solutions.
    \item \emph{Discrimination only within the set.} The re-ranker features need
        not be \emph{globally} discriminative over all possible
        segmentations, rather only \emph{locally} discriminative within the set
        returned by the first stage. Specifically, for the re-ranker the goal is
        not to identify generic good segmentations but use features that can
        help it discriminate good solutions from bad ones \emph{within} a small set.
\end{itemize}

\subsection{Contributions}
The main contribution presented in this chapter is a discriminative re-ranking
formulation for semantic segmentation. Our algorithm takes as input a set of
labellings $\{\by^{(1)},\dots,\by^{(m)}\}$~\graffito{note the change of notation from $\bx^{(i)}$ to $\by^{(i)}$.}  for an image and predicts the most
accurate labelling from this list. The learning task is formulated as a
Structured SVM (SSVM)~\cite{tsochantaridis2005}, where the task loss penalizes the re-ranker for
deviating form the most accurate solution in this set.

\section{Related Work}
The \divrank approach to segmentation that is presented in this chapter is
similar in spirit to the Constrained Parametric Min-Cuts (CPMC) approach
of Carreira \etal~\cite{carreira2010}, that was reviewed in~\cref{sec:segmethods}. CPMC produces a small set of
 high quality object segmentation proposals over an image that are scored
 according to how likely they are to be of an object. In the first stage of CPMC an
large set of overlapping candidate figure-ground segmentation proposals are
generated (using a bottom-up segmentation method) that are further pruned to
remove redundant segmentations and finally ranked and culled in the second stage. The
ranking is done by a regressor that is trained to map mid-level features, computed
over the image and the figure-ground segment proposal, to the largest overlap the segment has with a
object in the image (measured against ground-truth), quantifying the "objectness" of the proposal. Key to the
approach is the concept of reducing the solution space (space of generated
figure-ground regions) in the first stage. As opposed to building a model of
object segmentation that includes both local and global interaction terms to
capture the large scope of dependencies between regions in the image, the
approach uses simple models in the first stage that incorporate only local
interactions. This avoids the intractable nature of complex models with
high-order terms and serves as a filter that reduces the state space to a much
smaller set. The elements of this set have good alignment with image contours
--- a feature that can be captured using local interactions alone. In this restricted
solution space the ranker in stage two can use features that capture more global
properties that would potentially be intractable or at best inefficient to
compute over the original exponential state space --- global "objectness"
features such as convexity and smoothness of region boundaries, eccentricity,
and other gestalt features. An important property of the filtering stage is that
it maintains high-recall, preserving high-quality segmentations, but reduces the
state-space enough (\ie reduces false-positive rates) so that higher-order
processing is tractable on the remaining solutions.

Given that adjacent ranked segments tend to be very similar to each other a diversifying model is also incorporated in the second stage of~\cite{carreira2010} using
Maximal Marginal Relevance (MMR)~\cite{carbonell1998}. MMR is a sequential
procedure ---
starting with the top-scoring segment the next
segment is chosen by MMR that maximizes the original ranking score minus a score for
having a redundant segment that is based on amount of overlap with the previously
selected segments.

\subsection{Relation to cascade approaches}
Related to the \divrank approach are cascade models that consist of multiple
stages of successively more complex inference models. Starting with a relatively
simple model the stages progressively prune the output state-space in order to
speed up inference and increase prediction accuracy. Cascade models accomplish
this by leveraging the computational efficiency of simple models in the initial
stages to filter out the majority of examples that are easy to discriminate. In
further stages the cascades benefit from the predictive power of much more complex models, that though
expensive to compute on the original state space, become tractable for a
small set of solutions. The learnable model in each cascade stage is trained on the
filtered output from the previous stage. Therefore negative examples that reach
later stages (\ie examples that pass the filtering of all the earlier stages) tend to
be harder, and training the complex models in further stages to focus on
discriminating these examples can lead to lower false-positive rates and
improved performance. Key to any cascade approach is the balance between the
following two criteria,
\begin{enumerate}
    \item \emph{Accuracy:} Minimize the number of errors made by each stage of
        the cascade to ensure accurate inference in subsequent models. This is so that the errors propogated to later stages is minimized.
    \item \emph{Efficiency:} Reduce the output state space of each stage so that inference in subsequent models can be done more efficiently.
\end{enumerate}

This section reviews a number of relevant cascade models for vision tasks that
are learned to balance these two criteria.

\subsubsection{Face detection cascade}
One of the early works to show effective use of cascade models was Viola and
Jones' face detector~\cite{viola2004}. There a classifier cascade consisting of multiple
consecutive stages of simple to more complex classifiers is used to increase the
efficiency of the detector (compared to using a single complex classifier) while
simultaneously increasing the performance. Each classifier in the cascade takes
as input features computed within a sub-window of the
image, and predicts whether or not the sub-window contains a face. The
dictionary of features can be very large and so too can the number of features
computed within any sub-window. In each stage the classifier  is a combination
of a set
of weak classifiers that has been trained using AdaBoost~\cite{freund1995}.
In~\cite{viola2004} they
restrict the weak learners to be classifiers that each depend on a single feature. Thus boosting amounts to a selection mechanism over the dictionary of features,
retaining only the most informative features for classification.

Although using
boosted classifiers itself reduces classification time compared to classifiers that
depend on many more features the detection performance tends not to be as good.
To improve on this Viola and Jones introduce the face detection cascade (\ie classifier cascade)
which is a simple approach to improve detection performance while giving low
run-time complexity. The idea is to construct a cascade of progressively more
complex classifiers, where complexity is in terms of number of features used by the
classifier. The simpler classifiers at the early stages of the cascade are
trained to detect all positive sub-windows while rejecting as many negative
sub-windows as possible. In this way the early
stage classifiers, that are more efficient to run, filter out the majority of the
state space (the state space is all possible sub-windows in the image) so that
the more complex classifiers in later stages, which are more computationally
expensive, can focus on the task of classifying a much smaller
set of sub-windows.
The classifier in each
stage of the cascade is trained with Adaboost, on  training data constructed
from sub-windows that have passed the previous stages. Since the sub-windows that
pass earlier stages are harder to classify than the typical instance the
classifiers further along the cascade have a more difficult task. As a result
they give higher false-positive rate for a fixed value of recall (\ie true-positive
rate). To train the detector each stage in the cascade is trained by
progressively increasing the number of features the classifier in a stage uses until
the desired minimum false-positive rate and maximum true-positive rate are
achieved on a held out set. To illustrate, Viola and Jones train
a cascade with an initial stage consisting of a two feature classifier that remove $50\%$ of the non-face
sub-windows while retaining $99\%$ of the sub-windows containing faces, at a
$50\%$ false-positive rate. The next stage classifier uses ten features and
filters $80\%$ of the non-face sub-windows while retaining $100\%$ of the faces.
Further stages use more and more features until the false-positive rate is
virtually zero while maintaining a high true-positive rate. The result is a high
performance face detector that is efficient to run at multiple-scales over the
image.

\subsubsection{Structured prediction cascades}
The Structured Prediction Cascades of Weiss
\etal~\cite{weiss2010,weiss2010structured} provide a general framework for
tractable and efficient MAP inference in structured output models (\eg graphical
models) with either high
tree-width or large output state space. The idea is to reduce the state space by
removing clique assignments that do not correspond to the MAP assignment. The
Structured Prediction Cascade consists of sequential stages that take as input a set of
possible assignments to cliques in the model and prune some of the clique assignments
before passing the remaining to the next stage. Subsequent stages consist of
increasingly more complex models. Model complexity can be achieved either by
considering higher-order cliques, using more complex features, or starting with
a coarsened state space and successively refining it in subsequent stages. In
each stage pruning is done by running inference on the stage's
model and identifying states that will be pruned. Pruning occurs based on
the max-marginals of the model. Recall that the max-marginal on a clique $c$ of
a model capturing some joint distribution $f(\cdot)$ (alternatively an energy or
score) is defined as,
\begin{flalign}
    f^*(\bx_c)\doteq \max\limits_{x'\in\calX^n} \{f(\bx')\;:\; \bx'_c=\bx_c\}
\end{flalign}
where $\bx_c$ is the portion of complete assignment $\bx$ that is associated with variables in
clique $c$.
The max-marginal is the maximum probability/score of any assignment that agrees
on the clique assignment $\bx_c$. The cascade stages filter out any clique
assignment $\bx_c$ for which $f^*(\bx_c)\leq t_{\alpha}$ for some tuned stage
specific threshold $t_{\alpha}$. Consider the score, $f(\bx)$, of a joint
assignment $\bx$. Note that if $f(\bx)>f^*(\bx'_c)$ then $\bx\neq \bx'_c$, which
follows from the definition of $f^*(\bx'_c)$. This implies a \emph{safe
filtering property}~\cite{weiss2010}: if $f(\bx)>t_{\alpha}$ for some $t_{\alpha}$, then for all
$\bx_c$, $f^*(\bx_c)> t_{\alpha}$ (since $f^*(\bx_c)\geq f(\bx)>t_{\alpha}$).
Therefore as long as $f(\bx)>t_{\alpha}$ pruning clique assignments for which
$f^*(\bx_c)<t_{\alpha}$ will not remove the optimal clique assignments.
Threshold $t_{\alpha}$ is defined as a convex combination of the MAP assignment
score and the mean of the max-marginal score with combination weight $\alpha$
($0\leq \alpha \leq 1$), set to minimize the filtering error (number of correct
clique states pruned by a stage). The threshold is similar to a quantile of the
max-marginal values. Adjusting $\alpha$ is a trade-off between efficiency (\ie
aggressive pruning) and accuracy. Each stage is learned independently and
sequentially using stochastic sub-gradient descent on the model parameters
$\btheta$ (recall $f(\bx)=\langle\btheta,\bphi(\bx)\rangle$) in order to minimize the filtering error. The threshold for each stage
is set via cross-validation on $\alpha$.

If inference over the models in each stage is intractable an ensemble
method~\cite{weiss2010} is proposed that breaks the model into a collection of
sub-models (\eg graph reduced to sub-graphs collectively covering all nodes and
edges) for which exact inference is tractable. A similar analysis to the case of
a single model leads to an equivalent approach for filtering by thresholding the
\emph{sum} of max-marginals computed over the indivual sub-models
(cf.~\cite{weiss2010}), and analogous \emph{joint} safe filtering property.
Significant performance improvement on a number of vision tasks where the
structured output has very high state-space, such as articulated pose
estimation~\cite{sapp2010}, show the performance benefits of the cascade approach.

\subsection{Relation to proposal-generation methods}
Similar to the \divrank pipeline are a number of segmentation methods which produce an initial pool of
segmentations of the image that are subsequently ranked according to how well
they segment the objects in the scene. We mention a few noteworthy methods.

The category-independent object proposal and diverse ranking method of
Endres \etal~\cite{endres2010} that was review in \cref{sec:segmethods} produces a diverse set of object
segmentations which align well with object boundaries. The final region proposal
can be used to automatically localize object in the image for further processing
for recognition, or can alternatively be used to provide improved spatial
support, compared to bounding boxes, for detection tasks.

Russell \etal~\cite{russell2006} use a pool of global segmentations of images to
learn object categories and their segmentations. Their approach can be viewed as
a two-stage strategy of producing a diverse set of segments in the first stage
followed by ranking the segments according to how well they match each of the
discovered categories. More specifically, the approach (cf. \cite{russell2006})
uses the normalized-cut segmentation algorithm (cf. \cref{sec:segmethods}) to produce a pool
of candidate global segmentations of the image. For each segment in the pool
of segmentations a histogram of visual words is computed. To
concisely represent the segments a dictionary of visual words is constructed by
computing SIFT~\cite{lowe1999} descriptors over the images of a held-out dataset which
are clustered using k-means clustering. Each cluster center represent a
visual word. Visual words can be used to represent an image or a region in an
image. By using the dictionary to quantize the SIFT descriptors over an image
into visual words a representation for a segment is built by computing the
histogram of visual words contained in the segment. Given this \emph{bag of
visual words} representation for each segment statistical text analysis models
are used to learn \emph{topics} (\ie object categories) from the pool of
segments. A topic is thus a visual word histogram as well. Thus a segment can then be represented by a mixture of the discovered
topics.  For each of the discovered object categories the
segments are subsequently ranked by how well a segment matches the visual word
(KL-divergence between corresponding visual word histograms).

Another related approach is used to perform object segmentation with
category-independent shape priors~\cite{kim2012}. Multiple segmentations of
an image are considered since an image might
contain multiple objects, coupled with the fact that shape priors are imperfect
so multiple competing hypotheses might exist.
The shape priors are constructed by aggregating multiple learned
category-independent shape priors. Each segmentation problem is initialized with
one of the aggregated shape priors and a binary-labeling of the image is
inferred (via graph-cut inference). Producing a pool of object
segmentations increases the chance that at least one of the shape priors is
useful in producing a high-quality object segmentation. Similar
to CPMC (cf.~\cref{sec:segmethods}) classifiers can be trained to rank the object proposals.

\subsection{Discriminative re-ranking in other domains.} Discriminative
re-ranking of multiple solutions is also a common approach found in domains such
as speech~\cite{collins2005,dinarelli2012} and natural language
processing~\cite{collins2005nlp,shen2004,raymond2006}.
\section{DivMBest + Re-rank}\footnotemark\footnotetext{The contributions to the thesis presented in this section are found in \protect\cite{yadollahpour2013}, and are in collaboration with Gregory Shakhnarovich and Dhruv Batra.}
In~\cref{sec:divmbest} we presented the \divmbest algorithm for producing a diverse set of $m$ highly probable segmentations from a discrete probabilistic graphical model, such as a CRF. Often we would like to return the single best segmentation of the image from this diverse set --- that is we'd like an algorithm that can perform a $1$-out-of-$m$ inference task.
This section presents a novel two-stage approach to ranking the diverse segmentations
produced by the \divmbest algorithm presented in
chapter~\ref{ch:introduction}, called \divrank. In the first stage a
probabilistic model generates a set of diverse plausible segmentations. In the
second stage, a discriminatively trained re-ranking model selects the best
segmentation from this set. The re-ranking stage can use much more complex
features than what could be tractably used in the probabilistic model, allowing a
better exploration of the solution space than possible by simply producing the
most probable solution from the probabilistic model.

\subsection{Notation}\label{sec:rerankernotation}
In chapter~\ref{ch:divmbest} we denoted a segmentation (equivalently an
assignment of labels to $n$ (super)pixels or regions) by a vector $\bx$ where
$\bx=\{x_1,\dots,x_n\}\in\calX^n$, $\calX^n\doteq \times_{s\in [n]} \calX_s$ --- where $\calX_s$ is the set of labels for region $s$.

In this chapter we will make a change of variable for the assignment vectors.
For the first stage (which produces \divmbest candidate segmentations) let
$\by=\{y_1,\dots,y_n\}\in\calY^n$ be a segmentation of the image, where the
space of labellings is $\calY^n\doteq \times_{s\in [n]} \calY_s$, and $\calY_s$ is the set of labels for region $s$. Recall the \divmbest formulation for finding the $m$-th best diverse segmentation (\ie \textsc{MModes} problem),
\begin{subequations}
    \begin{flalign}
        \bmu^{(m)} = \argmin\limits_{\bmu\in \mathbb{L}(G),\;
        \bmu_A(\by_A)\in\{0,1\}} &\sum\limits_{A\in V\cup E} \btheta_A\cdot \bmu_A \\
        \mbox{s.t.} &\quad \Delta(\bmu,\bmu^{(i)})\geq k_i &\forall i\in[m-1].
    \end{flalign}
\end{subequations}
Note that the indicator vectors, $\bmu$, encode the label assignment to vectors
$\by$, \ie $(\bmu_j(\ell)=1) \implies (\by_j=\ell)$. Therefore we can define a
mapping $v:\; \{0,1\}^d\rightarrow \calY^n$ from $d$-dimensional indicator
vector $\bmu$ to labelling $\by$: $\by=v(\bmu)$. Given $\bmu^{(m)}$
the corresponding segmentation is $\by^{(m)}=v(\bmu^{(m)})$. Let
$\bY_i=\{\by_i^{(1)},\dots,\by_i^{(m)}\}$ denote the set of $m$ diverse
segmentations of the $i$-th image. At training time, the input to the second
stage is a set of (image, ground-truth, segmentation-set) triples ---
$\{x_i,\by_i^{gt},\bY_i \;|\; i\in [N]\}$, where $x_i$ is the i-th image and
$\by_i^{gt}$ is the corresponding ground-truth segmentation. The quality of a
segmentation is measured by a loss function, $\ell(\by_i^{gt},\widehat{\by})$,
that returns the cost of predicting $\widehat{\by}$ when the ground-truth is
$\by_i^{gt}$.

Let $\by_i^{(*)}$ denote the most accurate segmentation in the set $\bY_i$ --- that
is,
\begin{flalign}\label{eqn:oraclesol}
    \by^{(*)}_i = \argmin\limits_{\by\in \bY_i} \ell(\by_i^{gt},\by).
\end{flalign}
The re-ranker uses features $\bpsi(x,\by)\;:\;
\mathbb{R}^{3\times w\times h} \times \calY^n
\rightarrow \mathbb{R}^p$ that are computed on
the image $x$, and corresponding segmentation $\by$.
The score of the re-ranker on segmentation $\by_i$ of image $x_i$ is
denoted by,
\begin{flalign}\label{eqn:rankerscore}
    S_r(\by_i)=\balpha^T\bpsi(x_i,\by_i),
\end{flalign}
where $\balpha$ are the $p$-dimensional re-ranker parameters.

%\subsection{Re-ranking diverse segmentations}

\subsection{Re-ranker model}\label{sec:rerankermodel}
As mentioned above the re-ranker is modelled as a linear combination of
features, $\bpsi(x,\by)$ computed on the image $x$ and corresponding
segmentation $\by$, which assigns a score to each segmentation:
$S_r(\by)=\balpha^T\bpsi(x,\by)$. Inferring the best segmentation under the
re-ranker corresponds to computing the highest score,
\begin{flalign}\label{eqn:rankerpred}
\widehat{\by}_i =
\argmax_{\by\in \bY_i} S_r(\by).
\end{flalign}

Re-scoring the segmentations using a ranker has a couple of benefits,
\begin{enumerate}
\item \emph{Can use more complex features than segmentation model:} The
    re-ranker features $\bpsi$ can be different from the features used in the
    model that generated the segmentations. In fact they can be quite complex
    and expensive to compute. The reason for this is that the re-ranker only
    needs to compute features on a relatively small set of candidate
    segmentations in contrast to the exponential number of
    segmentations that have been pruned by the first stage. Inference in the
    second stage is simply taking a dot product of the features with the
    re-ranker parameters $\balpha$ and sorting the resulting scores. Hence we
    can afford to compute computationally expensive re-ranker features.
\item \emph{Can incorporate features that are intractable to include in the segmentation
    model:} In the first stage the segmentation model can only compute features
    on the image or segmentation that are tractable. Incorporating higher-order
    interactions between regions into the model would result in potentials that
    could make inference over the model intractable. Hence incorporating
    performance limiting dependencies between variables are typically avoided in
    segmentation models. In contrast,the re-ranker features are a
    function of both the image $x_i$ and segmentation $\by_i$. That means we can
    compute features like size of various categories, connectivity of the label
    masks, relative location of the label masks, and other such quantities that
    are functions of global statistics of the segmentation and thus intractable
    to include in the first stage.
\end{enumerate}

\subsection{Re-ranker loss}

To train the re-ranker we need a measure of performance. Let
$\calL(\by_i^{gt},\by)$ be the re-ranker loss. Earlier we mentioned that the
quality of a segmentation $\widehat{\by}$ predicted by the re-ranker as being
captured by the task loss $\ell(\by_i^{gt},\widehat{\by})$. Thus we could use the
task loss as the re-ranker loss, \ie $\calL(\by_i^{gt},\widehat{\by})\doteq
\ell(\by_i^{gt},\widehat{\by}_i)$. However , using
$\ell(\by_i^{gt},\widehat{\by})$ has a drawback. Consider the following case: we
are given two images $i$, $j$ with two segmentations each, and corresponding
accuracies $Acc(\bY_i)=\{95\%,\; 75\%\}$ and $Acc(\bY_j)=\{40\%,\; 35\%\}$. When the
re-ranker loss is set to the task loss, for $\bY_i$ we have that the loss on the
two segmentations are $\{100-95\%,\; 100-75\%\} = \{5\%,\; 25\%\}$ whereas for
$\bY_j$ the re-ranker incurs much higher losses
$\{100-40\%,\;100-35\%\}=\{60\%,\;65\%\}$. This means that the re-ranker will
focus on picking the best segmentation in set $j$ and ignore how well it does on set $i$. This is undesirable because set
$j$ segmentations are all of relatively the same (albeit poor) quality. Given
that we are committed to the set, if the re-ranker makes a poor selection for
the best segmentation in the set the cost incurred is only $5\%$ compared to if
the re-ranker had made the correct choice. On the other hand set $i$ contains
segmentations that are of very different qualities --- the re-ranker will
incur a $20\%$ cost if it makes the wrong choice. Clearly it would be better for
the re-ranker to focus attention on making the correct choice on set $i$ instead
of set $j$.

In order to shift the re-ranker to focus its effort on training instances where
it is under performing relative to the set the following relative re-ranker loss
is proposed,
\begin{flalign}\label{eqn:rankerloss}
    \calL(\by_i^{gt},\widehat{\by}_i)=\ell(\by_i^{gt},\widehat{\by}_i) -
    \ell(\by_i^{gt},\by_i^{(*)}).
\end{flalign}
Using the relative loss in eqn.~\ref{eqn:rankerloss} gives losses:
$\{5-5\%,\;25-5\%\}=\{0\%,\; 20\%\}$ for set $i$ and
$\{60-60\%,\;65-60\%\}=\{0\%,\;5\%\}$ for set $j$ --- this shifts the focus to
set $i$ because an incorrect choice in that set is much costlier (difference of
$20\%$) than an incorrect choice in set $j$ (difference of $5\%$). Using the
relative loss compared to the task loss was found empirically to play an
important role in the performance of the re-ranker.

\subsubsection{Re-ranker Training}
Note that it is not necessary for the re-ranker to produce a scoring that
induces a total ordering of the segmentations in set $\bY_i$. We only desire
the re-ranker to assign the best segmentation in $\bY_i$ a higher score than the
other segmentations in the set, \ie we desire $Acc(\widehat{\by}) > Acc(\by)$
for all $\by \in \bY_i \setminus \widehat{\by}$, where $\widehat{\by}$ is defined
in eqn.~\ref{eqn:rankerpred}. Thus we want to learn parsimonious re-ranker
parameters $\balpha$ such that for image $i$,
\begin{flalign}\label{eqn:scoremargin}
    \balpha^T \bpsi(x_i,\by_i^{(*)}) -\balpha^T \bpsi(x_i,\by) > \gamma \qquad
    \forall \by\in\bY_i \setminus \by_i^{(*)},
\end{flalign}
where $\by^{(*)}_i$ is the best segmentation in the set $\bY_i$,
and $\gamma \geq 0$ is some margin.

Then object in eqn.~\ref{eqn:scoremargin} coincides with the following quadratic
program (QP),
\begin{flalign}\label{prob:ranker}
    \max\limits_{\balpha\in \mathbb{R^p}}\quad C\cdot\sum\limits_{i\in[N]}\sum\limits_{\by \in
    \bY_i\setminus \by_i^{(*)}} \left[
    \balpha^T(\bpsi(x_i,\by_i^{(*)})-\bpsi(x_i,\by)) - \gamma \right] -
    ||\balpha||_1,
\end{flalign}
where the first term encourages the best segmentations to be scored higher than
the other segmentations for each image, and the $\ell_1$-penalty term is a regularization on $\balpha$ in order to reduce
over-fitting on the training set by producing a parsimonious (\ie sparse)
representation of the features. The scalar value $C$ balances the importance
of the two terms. Introducing scalar \emph{slack variables} for each image in the first
term in the objective of problem~\ref{prob:ranker} we can write it as,
\begin{subequations}\label{prob:rankerslack}
\begin{flalign}
    \max\limits_{\balpha,\;\xi_i}\quad &C\cdot\sum\limits_{i\in[N]} \xi_i -
    ||\balpha||_1\\
    \mbox{s.t.}\quad &\balpha^T(\bpsi(x_i,\by_i^{(*)})-\bpsi(x_i,\by))\geq
    \gamma + \xi_i &\forall i\in[N],\; \forall \by\in \bY_i \setminus
    \by_i^{(*)}, \\
                   & \xi_i \geq 0 &\forall i\in[N].
\end{flalign}
\end{subequations}
Re-writing problem~\ref{prob:rankerslack} as a minimization and replacing the
$\ell_1$-loss (because it's not differentiable) with the $\ell_2$-loss we get,
\begin{subequations}\label{prob:rankerslackminl2}
\begin{flalign}
    \min\limits_{\balpha,\;\xi_i}\quad &\frac{1}{2}||\balpha||_2^2 + C\cdot\sum\limits_{i\in[N]} \xi_i \\
    \mbox{s.t.}\quad &\balpha^T(\bpsi(x_i,\by_i^{(*)})-\bpsi(x_i,\by))\geq
    \gamma - \xi_i &\forall i\in[N],\; \forall \by\in \bY_i \setminus
    \by_i^{(*)}, \\
                   & \xi_i \geq 0 &\forall i\in[N].
\end{flalign}
\end{subequations}
If we let $\gamma=1$ (any choice of $\gamma\neq 0$ can be incorporated by the
magnitude of $\balpha$) and rescaling the slack variables by
$\calL(\by_i^{gt},\by)$ gives the familiar \emph{Structured SVM
QP}~\cite{joachims2009},
\begin{subequations}\label{prob:ssvm}
\begin{flalign}
    \min\limits_{\balpha,\;\xi_i}\quad &\frac{1}{2}||\balpha||_2^2 + C\cdot\sum\limits_{i\in[N]} \xi_i \\
    \mbox{s.t.}\quad &\balpha^T(\bpsi(x_i,\by_i^{(*)})-\bpsi(x_i,\by))\geq
    1 - \frac{\xi_i}{\calL(\by_i^{gt},\by)} &\forall i\in[N],\; \forall \by\in \bY_i \setminus
    \by_i^{(*)}, \label{constr:slackrescale}\\
                   & \xi_i \geq 0 &\forall i\in[N].
\end{flalign}
\end{subequations}
Intuitively we can see that constraint~\ref{constr:slackrescale} tries to
maximize the (soft)margin between the score of the oracle solution and all other
solutions in the set. Importantly, the slack (or violation in the margin) is
scaled by the loss of the solution. Thus if in addition to $\by_i^{(*)}$ there
are other good solutions in the set, the margin for such solutions will not be
tightly enforced. On the other hand, the margin between $\by_i^{(*)}$ and bad
solutions will be very strictly enforced. We solve problem~\ref{prob:ssvm} via
the $1$-slack cutting-plane algorithm of Joachims~\cite{joachims2009} which we
re-produce in alg~\ref{alg:1-slackcuttingplane} for reference.
\begin{algorithm}[!h]\caption{1-slack cutting-plane algorithm for training
    Structural SVM with slack-rescaling~\cite{joachims2009}}\label{alg:1-slackcuttingplane}
    \begin{algorithmic}[1]
        \State Input: $D=\{(x_i,\by_i^{gt},\bY_i)\;|\;i\in[N]\},\;C,\;\epsilon$
        \State $\calW \gets \emptyset$
        \Comment{initialize working set of constraints}
        \Repeat
        \State
        \vspace{-2.5em}
        \begin{subequations}\label{prob:1slackcuttingplane}
        \begin{align}(\balpha,\xi) \gets &\quad \argmin_{\balpha,\xi\geq 0}
            \frac{1}{2}\balpha^T\balpha + C\xi&\\
        \mbox{s.t.} &\quad \forall (\bar{\by}_1,\dots,\bar{\by}_N)\in \calW:\\
                    &\quad \frac{1}{N}\alpha^T \sum\limits_{i\in[N]}
        \calL(\by_i^{gt},\bar{\by})
        (\bpsi(x_i,\by_i^{(*)})-\bpsi(x_i,\bar{\by}_i))\geq
    \frac{1}{N}\sum\limits_{i\in[N]} \calL(\by_i^{gt},\bar{\by}_i)-
        \xi \label{constr:cuttingplane}
        \end{align}
        \end{subequations}\label{ln:cpprob}
        \For{$i=1,\dots,N$}\label{ln:cpws}
        \Comment{find most violated constraints}
        \State $\widehat{\by}_i \gets
        \argmax\limits_{\widehat{\by}\in\bY_i}\left\{\calL(\by_i^{gt},\widehat{\by}_i)\left(1-\balpha^T\left[\bpsi(x_i,\by_i^{(*)})-\bpsi(x_i,\widehat{\by}_i)\right]\right)\right\}$
        \EndFor\label{ln:cpwsend}
        \State $\calW \gets \calW \cup \label{ln:ws}
        \{\widehat{\by}_1,\dots,\widehat{\by}_N\}$
        \Comment{{\small add corresponding segmentations to constraint set}}
        \Until {$\frac{1}{N}\balpha^T\sum\limits_{i\in[N]}
            \calL(\by_i^{gt},\bar{\by}_i)(\bpsi(x_i,\by_i^{(*)})-\bpsi(x_i,\bar{\by}_i))-\frac{1}{N}\sum\limits_{i\in[N]}
    \calL(\by_i^{gt},\bar{\by}_i)\leq \xi + \epsilon$}
    \State\Return{$(\balpha,\xi)$}
    \end{algorithmic}
\end{algorithm}
In each iteration the cutting-plane algorithm finds the segmentation that most
violates the margin constraint on each image, \ie
lines~\ref{ln:cpws}-~\ref{ln:cpwsend} of
alg.~\ref{alg:1-slackcuttingplane} and add it to the working set
(line~\ref{ln:ws}).
In the 1-slack formulation we add a single constraint in each iteration
(constraint~\ref{constr:cuttingplane}) consisting of the average loss
re-weighted margin constraints. Notice that in problem~\ref{prob:1slackcuttingplane}
of line~\ref{ln:cpprob} in
alg.\ref{alg:1-slackcuttingplane} there is a single slack variable $\xi$ instead
of an $\xi_i$ for each image as
in the original SSVM QP (\ie problem~\ref{prob:ssvm}).

At test time we compute stage-1 features $\bphi$ on an image and segmentation
model potential functions $\btheta_A$, which we use to run the \divmbest
algorithm to produce a set of diverse segmentations $\bY$. We compute re-ranker
features $\bpsi$ on this set and score each segmentation
using~\ref{eqn:rankerscore}, returning
the highest scoring solution (\ie perform re-ranker inference in
eqn.~\ref{eqn:rankerpred}).
\subsection{Summary}
This chapter has presented a two-stage approach to segmentation: produce a set
of diverse segmentations from a discrete probabilistic model, then re-rank them
using a discriminative re-ranker formulated as a structural SVM. The re-ranking
stage can use arbitrarily complex features, such as global features that are
computed over the entire image or solution, in order to evaluate the best
segmentation in the set. The first-stage filters the exponential space of
possible segmentations to a small set of highly plausible solutions that are not
merely minor perturbations of each other. The second-stage can focus on the
best-out-of-$m$ inference task on a much reduced space, and thus only needs to
compute features that are relevant in discrimination within the set. In
chapter~\ref{ch:divrankexp} we evaluate the performance of this approach on a
number of semantic segmentation tasks.
\printbibliography

%% file: Chapters/Chapter04.tex
\chapter{DivMBest Experiments}\label{ch:divmbestexp}
%************************************************

\subfile{\main/Sections/DivmbestExp}

\printbibliography

%% file: Sections/DivmbestExp.tex
\subfile{\main/Sections/ExpOverview.tex}

\subfile{\main/Sections/InteractiveSeg.tex}

\subfile{\main/Sections/Figureground.tex}

\subfile{\main/Sections/CategorySeg.tex}

%% file: Sections/ExpOverview.tex
\section{Evaluating DivMBest segmentations}
We look at a number segmentation tasks and investigate the quality of the
\divmbest segmentations against the MAP solution produced by the respective
underlying segmentation models. 
\subsection{Baselines}\label{sec:divmbestblines}
We evaluate the \divmbest segmentations against a number of baselines:
\begin{itemize}
    \item{\textbf{M-Best MAP ---}} The method of Yanover and
        Weiss~\cite{yanover2004} is used to produce a set of low energy (\ie high
        probability) solutions, where the solutions are only constrained to be
        different on at least a \emph{single} label assignment. There is no
        additional characterization of diversity between solutions.
    \item{\textbf{Random ---}} Multiple solutions can be generated without any
        optimization as well. A new solution is created by taking a subset of
        the nodes in the MAP solution at \emph{random} and changing their label
        assignment to the next best label according to the node min-marginals.
        Repeating this process produces a set of segmentations.  
    \item{\textbf{Confidence ---}} Similar to how the random
        segmentations were produced except the nodes are selected based on a
        \emph{confidence} measure. A subset of the nodes with highest entropy
        (according to their min-marginals) are selected and their label set to the next best
        value.
\end{itemize}

For each solution $\bmu$ of the \divmbest algorithm, let $d(\bmu,\bmu^{(1)})$ denote the
number of places that $\bmu$ differs from the MAP solution $\bmu^{(1)}$.  
In order to have a fair comparison between the perturbation based
baselines and the \divmbest solutions, for each solution $\bmu$ produced by \divmbest
we generate a perturbation based solution that differs from the MAP solution in
exactly $d(\bmu,\bmu^{(1)})$ locations. This ensures that the
solutions generated using \emph{random} or \emph{confidence}
perturbations have an equal measure of diversity compared to
the \divmbest solutions.   

\subsection{Oracle Solution}\label{sec:oracledef}
In order to evaluate the upper-bound on the quality of the segmentations in the
\divmbest set we can compute the \emph{oracle} solution. Given a set of
segmentations for the $i^{\text{th}}$ image,
$\{\bx_i^{(1)},\bx_i^{(2)},\dots,\bx_i^{(m)}\}$, and corresponding ground-truth
segmentation $\bx^{gt}_i$, let the segmentation
accuracy w.r.t ground-truth be denoted as
$Acc(\bx_i^{(k)},\bx^{gt}_i) $\sidenote{$Acc(\cdot,\cdot)$ can be intersection-over-union
score between two segmentations or any relevant measure on accuracy of a predicted segmentation with respect to
ground-truth.}. The \emph{oracle} segmentation, $\bx_i^{(*)}$, is defined to be the segmentation
\emph{within the set} that achieves maximum segmentation accuracy, \ie,
\begin{flalign}
\bx_i^{(*)} = \argmin\limits_{\bx_i \in \{\bx_i^{(1)},\dots,\bx_i^{(m)}\}}
Acc(\bx_i,\by_i).
\end{flalign}

%% file: Sections/InteractiveSeg.tex
\section{Interactive Segmentation}\footnotemark\footnotetext{The contributions to the thesis presented in this section are found in \protect\cite{batra2012}, and are in collaboration with Gregory Shakhnarovich and Dhruv Batra.}
Recall from~\cref{sec:segmentation} that in interactive segmentation the user
is interested in cutting out the foreground object from the rest of the image
via annotations like scribbles~\cite{boykov2001} or bounding
boxes~\cite{rother2004}. The problem is typically formulated as a figure-ground
segmentation task where some model variables are fixed according to the user
annotations. In each round the MAP solution is computed and presented to the user, at which
point the user provides additional supervision and the MAP solution is updated.
This process is repeated until the MAP solution is acceptable. Instead of showing
a single cutout each round the number of user interactions could be minimized by
having the interface show a set of possible cutouts for the user to pick from.
Ideally we'd like an algorithm that can \emph{efficiently} produce a \emph{small} set of \emph{diverse} solutions.

\subsection{CRF Model}
Consider the image-scribble pair $(X,\calS)$, where each image is a
collection of $n$ superpixels $X\doteq\{X_s\;|\; s\in [n]\}$. Let
$\bx=\{x_s\;|\; s\in [n]\}$ be a corresponding label assignment to all the
superpixels in the image, where each superpixel takes on either
foreground or background label, \ie $x_s\in\{fg,\;bg\}$. A subset $\calS\subset [n]$ of the
superpixels have known label according to the scribbles, \ie the
superpixel labels $\{x_s\;|\; s\in \calS\}$ are assigned according to the
manually provided scribbles. Alternatively pixels could have been used but
superpixels were preferred for computational efficiency reasons, and better alignment of
segmentations to internal image boundaries. We chose the SLIC
algorithm~\cite{achanta2010} to generate superpixels, with the desired number of
superpixels in an image set to $3000$. The average image in our dataset
contains $\simeq 150K-200K$ pixels. To model the figure-ground
segmentation problem we build a graph $G=(V,E)$ over
the superpixels and define a pairwise CRF with the following
energy,
\begin{flalign}\label{eqn:intsegenrgy}
    E(\bx;\;\calA)\doteq \sum\limits_{s\in V} \left[\theta_s(x_s;\;\calA) +
    \lambda\cdot \sum\limits_{t\in \calN(s)}
\theta_{st}(x_s,x_t)\right],
\end{flalign}
where $\calN(s)$ are the superpixels adjacent to superpixel $s$ in the
image. The \emph{data term} is the cost of assigning a superpixel to
foreground or background, and it depends on the appearance model $\calA$
that's learned from the user scribbles $\calS$. The pairwise \emph{smoothness term}
penalizes neighboring superpixels being assigned different labels.
\begin{table}[!t]
    \begin{tabular}{|l|}
        \hline
        \textbf{Features}\\
        \hline
        \hline
        \textbf{Color~\cite{hoiem2005}}\\
        C1: RGB mean values \\
        C2: C1 in HSV colorspace \\
        C3: Hue histogram and entropy \\
        C4: Saturation histogram and entropy \\
        \hline
        \textbf{Texture}\\
        T1: Histogram of gradients (HOG)~\cite{dalal2005}\\
        \hline
        \textbf{Local}\\
        L1: Histogram of SIFT~\cite{lowe1999} codewords\\
        \hline
    \end{tabular}
    \caption{Superpixel features used to learn the appearance model for the
    interactive segmentation figure-ground cutout model.}
    \label{table:intsegfeat}
\end{table}

\subsubsection{Data term}
The data term depends on an appearance model $\calA$ that is based on the output of a
\emph{Transductive SVM} (TSVM). The appearance model is learned by
extracting features $\bphi(X_s)$from labelled and unlabelled superpixels and training a
TSVM~\cite{sindhwani2006} to predict if a superpixel belongs to foreground or
background. The features include the low level color, texture, and local cues
listed in table~\ref{table:intsegfeat}. Let the superpixel \emph{score} for belonging to
foreground be $score(X_s)=\bw^T\bphi(X_s)$, where $\bw$ is the learnt weight vector of
the TSVM. We define the foreground data term energy as,
\begin{flalign}
    \theta(x_s=fg)\doteq\left\{\begin{array}{ll}
            \eta & \mbox{if $score(X_s)\geq 0$,}\\
            \eta-1 & \mbox{otherwise}
        \end{array}\right.
\end{flalign}
where $\eta=0.5e^{\frac{-|score(X_s)|^2}{\alpha \sigma^2}}$,
$\sigma^2=\var(\{score(X_s)\;|\; s\in \calS\})$, and $\alpha$ is a
constant that is set via cross-validation on a held out set that is kept the
same for all images. The background energy is then
simply,
\begin{flalign}
    \theta(x_s=bg)\doteq 1-\theta_s(x_s=fg).
\end{flalign}

\subsubsection{Smoothness term}
The smoothness term is a contrast sensitive Potts energy~\cite{boykov2001} that penalizes adjacent labels taking different labels. The
penalty is proportional to how similar the two superpixels are in feature
space. The more similar the features are the more penalty is paid,
\begin{flalign}
    \theta_{st}(x_s,x_t)\doteq \delta_{x_s\neq x_t} \cdot
    \beta_1\cdot e^{-\beta_2 d_{st}}
\end{flalign}
where $d_{st}$ is the distance between the feature vectors of superpixels $s$
and $t$, and the scale parameters are set to, $\beta_1=2$,
$\beta_2=\sqrt{\frac{\max\{d_{st}\}}{20}}$.

\subsubsection{Inference}
The figure-ground interactive segmentation problem amounts to finding the
assignment to unlabelled superpixels that minimizes the
energy in eqn.~\ref{eqn:intsegenrgy}. Note that this is a
binary (\ie two-label) contrast sensitive Potts model with submodular pairwise
energy terms, for which efficient graph-cut algorithms exist to
compute the \emph{exact} MAP solution in polynomial time~\cite{boykov2004,kohli2005}.
\subsubsection{Data + training}
To evaluate the interactive segmentation model, and its \divmbest
extension described next, evaluation was performed on 100 images from Pascal VOC2010. For each
image scribbles marking foreground objects and background regions were manually
provided. Fifty of the images were used for tuning the parameters and the rest
were used for reporting test accuracy. The weight on the smoothness term
($\lambda$) was tuned by doing grid search in the range $[0,\;1]$. The best setting on
training images was achieved with $\lambda=.18$, and used on the test set
experiments.

\subsection{Interactive segmentation + \divmbest}
Using the \divmbest framework the underlying interactive segmentation model can be extended to generate a set
of plausible segmentations each round, instead of just a single MAP solution.
The \divmbest formulation will encourage high quality segmentations under the
model that are diverse. Experimental results are provided for the
dissimilarity functions described below.
\subsubsection{Hamming dissimilarity}
The negative dot-product distance function $\Delta(\bmu,\bmu^{(i)})\doteq
-\sum\limits_{s\in V}\bmu_s^T\bmu_s^{(i)}$ captures the Hamming dissimilarity
between two solutions (see~\cref{sec:divmbestrelax}). Recall that since the dot-product function
decomposes over nodes in the graph, the \divmbest formulation under Hamming
dissimilarity is equivalent to a $\Delta$-augmented energy minimization problem were the
unary terms in the energy have been perturbed in a certain way
(cf.~\cref{sec:divmbestrelax}). The
pairwise interaction terms remain unaffected, so if the smoothness term is
submodular in the original problem then it remains submodular. This means that
the $m$-modes can be computed using the same efficient inference algorithm used
to compute the MAP solution. In the case of our binary pairwise energy
in eqn.~\ref{eqn:intsegenrgy}, we can compute the $m$-modes using the same efficient graph-cuts
algorithm that was used to optimally compute the MAP solution.

\subsubsection{Higher-order potential (HOP) dissimilarity}
Let $\#\bmu \doteq \sum\limits_{s\in V} \bmu_s(1)$ denote the number of
nodes in the solution that are set to the foreground label. The value
$\#\bmu$ represents the size of the foreground region. The HOP dissimilarity is
defined as,
\begin{flalign}
    \Delta(\bmu,\bmu^{(1)})=\left\{\begin{array}{ll}
            (\#\bmu-\#\bmu^{(1)})^2 & \mbox{if $\#\bmu \geq \#\bmu^{(1)}$,}\\
            0 & \mbox{otherwise}
        \end{array}\right.
\end{flalign}
Intuitively the HOP dissimilarity is zero if the foreground size of the current
segmentation is smaller than the foreground size of the MAP solution. Otherwise
the dissimilarity grows quadratically. Since the \textsc{2Modes} constraint on the
MAP solution is $\Delta(\bmu,\bmu^{(1)})\geq k$, for some $k>0$, the HOP
dissimilarity encourages foreground size of the current solution to be larger
than the MAP. By changing the sign on $\Delta$ we can alternatively encourage smaller
solutions. Note that $\#\bmu$ is a global measure of the solution so it
cannot be decomposed over subsets of variables. However, the $\Delta$-augmented
energy minimization problem that we get from the Lagrange relaxation of the
\divmbest problem, under HOP dissimilarity, results in a \emph{cardinality
potential}~\cite{gupta2010,tarlow2010} for which efficient approximate
solutions exist. To solve this $\Delta$-augmented energy minimization problem
we can use the HOPMAP algorithm of Tarlow \etal~\cite{tarlow2010}.

\begin{figure}[!t]
\begin{minipage}[c]{\linewidth}
    \centering
    $\begin{array}{c}
      \includegraphics[width=\linewidth]{\main/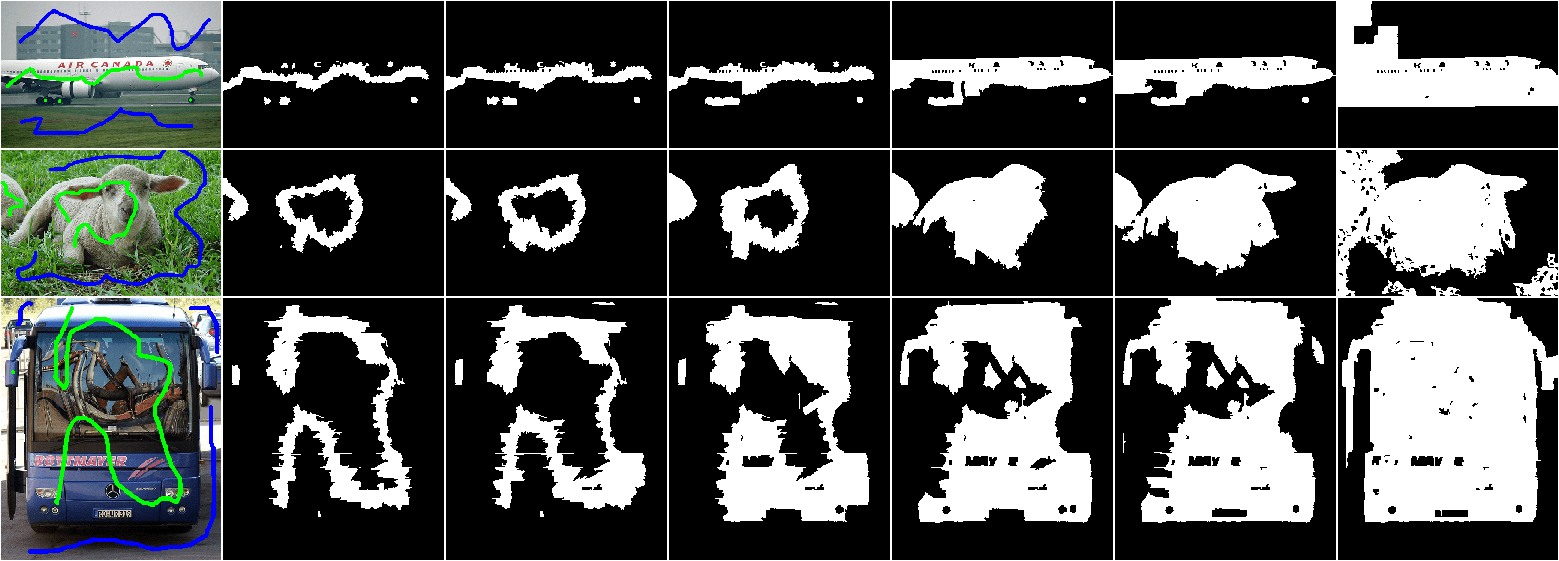}
   \end{array}$
\end{minipage}\hspace{1em}%\vspace{\captionReduceTop}
%\begin{minipage}[c]{.35\linewidth}
\caption{\divmbest \emph{modes} under cardinality-based HOP. \emph{From left-to-right}:
    image-scribble pair $(X,\calS)$, MAP solution,
\nth{2}-\emph{mode},\dots,\nth{6}-\emph{mode}. The modes are ordered in
increasing size of foreground object.} %\vspace{\captionReduceBot}
\label{fig:hop}
%\end{minipage}
\end{figure}

Figure~\ref{fig:hop} shows a few \divmbest \emph{modes} using HOP dissimilarity on the
interactive segmentation problem.

\subsection{Experiments}
In this section we evaluate the quality of the object cutouts generated in the
first round of interactive segmentation. Specifically we compare the MAP
segmentation against the alternative approach of generating a set of \divmbest
solutions. We also evaluate some other baseline approaches, which we describe next.
\subsubsection{Baselines}
Since eqn.~\ref{eqn:intsegenrgy} is a binary contrast sensitive Potts
energy~\cite{boykov2001} that is submodular its exact MAP solution can be computed using the
graph-cut implementation of~\cite{kolmogorov2004}. The \divmbest solutions under the Hamming
dissimilarity can also be computed using the same graph-cut implementation. We
also evaluate the \divmbest solutions under HOP dissimilarity generated using
HOPMAP~\cite{tarlow2010}.

The first baseline is the $M$-best MAP algorithm~\cite{yanover2004} which we reviewed
in~\cref{sec:mbestmap}. We also compare against the \emph{random} and
\emph{confidence} baselines
(see~\cref{sec:divmbestblines})
where the exact min-marginals are computed using dynamic
graph-cuts~\cite{kohli2008}.
\subsubsection{Results}
\subimport{\main/data/intseg/}{statmeans.tex}
For each of the 50 test images in PascalVOC10 the MAP segmentation cutout plus five
additional modes was generated using the approaches described above.
Table~\ref{table:stats} shows the best-out-of-6 cutout accuracies
averaged over 50 images for the different methods. Note that the \divmbest
cutouts (column two) achieves the best results. Some example cutouts are
shown in figure~\ref{fig:interactiveExamples}. Notice how Yanover and
Weiss'~\cite{yanover2004} \nth{2}-best MAP cutout is almost identical to the MAP cutout.
In contrast the \nth{2}-mode of \divmbest is qualitatively different and more
likely the cutout the user intended. The \nth{2}-mode corrects where the MAP
solution has likely made a mistake, for example the arm of the person in the
first image is completed and a second instance of an object category is found
in other image. The MAP solution suggests that the interactive segmentation model
(TSVM + contrast sensitive Potts) does not perfectly capture the most probable
cutout the user intended and yet we see that other modes of the underlying model
distribution correspond to good segmentations. With \divmbest we
have a framework that provides a principled way of extracting these other segmentations.

%% file: data/intseg/statmeans.tex
% latex table generated in R 2.13.2 by xtable 1.6-0 package
% Wed Nov  9 15:44:29 2011
%\begin{table}[t]
%\begin{center}
%\small
%\begin{tabular}{rccccc}
%  \hline
% & Map & M-Modes & M-Best & Rand & Conf Ranked \\ 
%  \hline
%  Mean & 91.542 & \bf{95.157} & 91.590 & 91.676 & 93.174 \\ 
%   \hline
%\end{tabular}
%\end{center} \vspace{\captionReduceTop}
%\caption{\small{Absolute pixel accuracies averaged over 50 test images.}} \vspace{\captionReduceBot}
%\label{table:stats}
%\end{table}

\begin{table}[!t]
%\begin{center}
{\footnotesize
\setlength{\tabcolsep}{1pt}
\begin{tabular*}{\columnwidth}{@{\extracolsep{\fill}}p{1cm}cccccc}
%\begin{tabular*}{1\textwidth}{c|ccccccccccccccccccccc|c}
\toprule
& MAP
& $MModes$-dot prod.
& $MModes$-HOP
& M-Best
& Random
& Confidence\\
%& \rotatebox{0}{MAP}
%& \rotatebox{0}{\begin{tabular}{c}$MModes$-\\Dot\end{tabular}}
%& \rotatebox{0}{\begin{tabular}{c}$MModes$-\\HOP\end{tabular}}
%& \rotatebox{0}{M-Best}
%& \rotatebox{0}{Random}
%& \rotatebox{0}{\begin{tabular}{c}Confidence-\\Ranked\end{tabular}}\\
\midrule
Acc.(\%) & 91.542 & \textbf{95.16} & 93.82 & 91.59 & 91.68 & 93.17 \\ 
\bottomrule
\end{tabular*}
%\end{center} 
%\vspace{\captionReduceTop}
}
\caption{\small{Interactive segmentation: pixel accuracies averaged over 50 test
images.}} %\vspace{\captionReduceBot}\vspace{-20pt}
\label{table:stats}
\end{table}

%% file: Sections/Figureground.tex
\section{Figure-ground Segmentation}\footnotemark\footnotetext{The contributions to the thesis presented in this section are found in \protect\cite{yadollahpour2014}, and are in collaboration with Gregory Shakhnarovich.}\label{sec:divmbestfg}
This section presents the application of the \divmbest framework to the
figure-ground image segmentation task. Instead of relying on a complex model for
foreground and background, the approach uses a simple binary pairwise CRF which
relies on features computed over superpixels. The CRF can be learned efficiently
using Structured SVM formulation.

Figure-ground segmentation can be used as input to multi-category segmentation
models, or used in feed-forward approaches as a way to generate a set of
candidate masks that are processed further~\cite{carreira2010,rosenfeld2011}. As
such generating high-quality figure-ground segmentations of an image is an
important task.

A common approach to category-level image segmentation relies on building
structured probabilistic models with low-order interactions (such as pairwise
CRFs). Such models are appealing because inference over them tends to be
tractable and is often guaranteed to be optimal. On the other hand the
simplifying independence assumptions of these models lead to exact MAP
assignments that are highly inaccurate. In contrast, more complex models have
been introduced that incorporate different types of higher-order interactions
over the image such as cardinality and co-occurrence
potentials~\cite{kohli2009,ladicky2010b,tarlow2010} and hierarchical
CRFs~\cite{ladicky2009}. However, even though these models may better capture
complex statistics of natural scenes they can be inefficient and slow to train.
Therefore approximate inference algorithms are often needed to make inference
tractable.

In contrast, this section presents a fairly simple probabilistic model for
figure-ground segmentation where inference is efficient. While the inferred MAP
solution is often not good enough, the set of solutions generated from this
model using the \divmbest method tend to contain highly accurate segmentations.
This suggests that even though the model does not accurately model the most
likely figure-ground segmentation, the underlying CRF distribution tends to have
high quality solutions as one of its \emph{modes}. The next section introduces
our simple figure-ground CRF model.
\subsection{CRF Model}
We represent the image, $I$, as as set of $n$ disjoint superpixels,
$I=\{X_1,\ldots,X_n\}$, and $\mathcal{V} =[n]$.  We define a graph $G=(V,E)$
where the vertices correspond to superpixels and edges connect adjacent
superpixels in the image. A segmentation corresponds to an
assignment $\bx=\{x_1,\ldots, x_n\}\in\{0,1\}^{n}$, where $x_s=1$
indicates assignment of $X_s$ to foreground. The binary CRF energy is define to
be,
%\begin{flalign}
\begin{multline}
  \label{eqn:fgcrf}
  E(\bx;\; \bu)=\bu_1^T \sum_{s\in V}\btheta_1(x_s) +
  \bu_2^T\sum_{(s,t)\in
  E}\btheta_2(x_s,x_t) +\\
  u_c\sum\limits_{s\in V}\theta_c(x_s) + \sum\limits_{(s,t)\in E} \theta_e(x_s,x_t).
\end{multline}
%\end{flalign}
The unary and pairwise potentials are $\btheta_1(\cdot)$ and $\btheta_2(\cdot)$
respectively. Potential functions $\btheta_c(\cdot)$ and $\btheta_e(\cdot)$ are
cardinality potentials on the nodes and edges respectively.

Intuitively, the unary $\btheta_1$ captures
characteristic properties of superpixels in figure vs. background classes, while
the pairwise $\btheta_2$ captures the likelihood of neighboring regions to be
assigned the same class.

\subsection{CRF potentials}
The details of each type of potential are described below.
\paragraph{Unary:}{
    The unary potential consists of $p$-channels, $\btheta_1 =
    (\theta_{11},\dots,\theta_{1p})$. The $j^{\text{th}}$ channel is defined to
    be,
    \begin{flalign}
        \theta_{1j}(X_s=x_s)\doteq P(X_s=x_s\; | \; \bphi_{1j}(X_s);\; \bw_{1j}).
    \end{flalign}
    The channel captures the likelihood that superpixel $s$ belongs to a
    foreground/background object, given the $j^{\text{th}}$ unary features computed on the
    superpixel, $\bphi_{1j}(X_s)$, and learned parameters $\bw_{1j}$. The background score is simply
    $\theta_{1j}(X_s=0)= 1-\theta_{1j}(X_s=1)$. The foreground probability
    of a superpixel is modelled as a logistic regression classifier,
    \begin{flalign}
        P(X_s=1\; | \; \bphi_{1j}(X_s);\; \bw_{1j}) \doteq
        \sigma\left(\langle\bw_{1j},\;\bphi_{1j}(X_s)\rangle\right),
    \end{flalign}
    where $\sigma(z)=1/(1+e^{-z})$.
    Since foreground is less common that background, we use asymmetric logistic
    loss, tuned to provide $90\%$ recall for foreground on training data.
}

\paragraph{Pairwise:}{ The pairwise potential consist of $q$-channels,
    $\btheta_2=(\theta_{21},\dots,\theta_{2q})$. The $k^{\text{th}}$ pairwise
    channel is defined as,
    \begin{flalign}
        \btheta_{2k}(X_s=x_s,X_t=x_t)\doteq P(x_s \neq x_t\; | \;
        \bphi_{2k}(X_s,X_t); \; \bw_{2k}),
    \end{flalign}
    and captures the likelihood that adjacent superpixels $X_s$ and $X_t$ should have consistent
    labels, according to the $k^{\text{th}}$ pairwise features
    $\bphi_{2k}(X_s,X_t)$ and learned parameters $\bw_{2k}$. The pairwise
    potentials are modelled as a logistic regression classifiers similar to the
    unary case, and trained with asymmetric logistic loss tuned to achieve
    $90\%$ recall for neighboring regions with different labels.

}
\paragraph{Higher Order:} {
The third and fourth terms in the CRF energy of eqn.~\ref{eqn:fgcrf} are unary and pairwise
\emph{cardinality} potentials~\cite{tarlow2012fast} respectively. They are a function of the global solution
and not local image evidence. The unary potentials are defined as,
\begin{flalign}
    \theta_c(x_s)\doteq [\![x_s=1]\!],
\end{flalign}
and count the number of superpixels labelled foreground, capturing what portion
of the image is assigned to foreground.

The pairwise cardinality potentials are,
\begin{flalign}
    \theta_e(x_s)\doteq [\![x_s=x_t]\!],
\end{flalign}
and measure the length of the boundary between foreground and background regions
in the solution.
}

\subsection{Features}

\subsubsection{Superpixel (unary) features}\label{sec:crf-unary}
The segmentation model relies on basic appearance features described below.

\vspace{-1em}
\paragraph{Intensity histograms} We bin the intensity into $n_c$ equally spaced
bins. Since color/intensity distribution within an image may be
skewed, this may be an inefficient binning scheme, and so we also use
adaptive binning according to $1/{n_c}$ quantiles of the intensity in
the given image. For graylevel images this produces four
histograms: fixed and adaptive
binning schemes, each with $n_c=8$ and with $n_c=32$ bins. For color
images there are twelve histograms, four per each dimension in the L*a*b space.

\vspace{-1em}
\paragraph{Texton histogram} We compute a dictionary of 32
textons~\cite{malik1999} on all training images, using a bank of
12 filters. Histogram of texton assignments within a region forms a
single 32-dimensional histogram.

\vspace{-1em}
\paragraph{Gradient features} We compute the histogram of oriented
gradient~\cite{dalal2005} within the region, binned into four directional
bins. Furthermore, we compute the statistics of the gradient magnitude: sum of $L_2$ and of $L_1$ norms of the
gradients within region, as well as the ratio of the two sums, known as blur
index~\cite{kohlberger2012}.

\vspace{-1em}
\paragraph{Entropy features} Finally, we set up one-dimensional
features computed as measures of entropy of histogram-based features.
This is intended to
capture how homogeneous a region is. There is one entropy value for
each intensity, texton and gradient histogram; total of 6 for
graylevel and 14 for color images.

Note that in contrast to many other models, we do not employ HoG/SIFT
descriptors or shape features in this model. This is because at the
level of small superpixels we do not expect such features to be informative.

\subsubsection{Boundary (pairwise) features}

\paragraph{Histogram differences} For each of the histogram-based unary
features we compute the $\chi^2$ difference between the two
regions. We also compute the earth mover's distance (EMD) between the
histograms. High values of these features here indicate different color/intensity
content between the two regions.

\vspace{-1em}
\paragraph{Entropy differences} For each entropy feature, we compute
the absolute value of the difference in entropies. High value here
indicates one region is more homogeneous than the other in the
respective feature.

\vspace{-1em}
\paragraph{Boundary strength} We compute the integral of the
boundary probability according to $gPb$~\cite{maire2008} along the
boundary between the two regions. High value corresponds to pronounced
boundary evidence according to $gPb$.

\vspace{-1em}
\paragraph{MSER correlation} We extract a set of maximally stable extremal
regions (MSERs,~\cite{matas2004}), and for each MSER compute
the percentage of the superpixel covered by that MSER. With $M$ MSERs,
this produces an $M$-dimensional vector for each
superpixels. The correlation coefficient of these vectors is a pairwise
feature; high value indicates that the two superpixels tend to belong
 to the same MSERs. This and the next feature were inspired by ideas
 in~\cite{mostajabi2012}.

\vspace{-1em}
\paragraph{MSER overlap} Another pairwise feature is the largest
overlap of any of the MSERs and the union of the two superpixels. Higher
value of this feature indicates that ``merging'' the two superpixels
in the same mask is better supported by MSERs.

All of these features could be used directly in the
model (eqn.~\ref{eqn:fgcrf}). However, this would lead to a fairly
high-dimensional parameterization making learning more
challenging. Instead we proceed in two stages. First we train for each
unary feature group $\bphi_{1j}(X_s)$ (each histogram, each
entropy value, etc.) a
logistic regression classifier $\sigma(\langle \bw_{1j},\; \bphi_{1j}(X_s)\rangle)$ predicting FG/BG label. Similarly, for
each pairwise feature group $\bphi_{2k}(X_s,X_t)$ we train a classifier
predicting whether the two superpixels are in the same class or not.

\subsection{CRF learning}\label{sec:fgcrflearn}
Instead of using max-likelihood training to learn the CRF weights
$\bu=(\bu_1,\bu_2,u_c,u_e)$ in eqn.~\ref{eqn:fgcrf} we optimize
the following Structured SVM~\cite{tsochantaridis2005} objective,
\begin{subequations}\label{eqn:fgssvm}
\begin{flalign}
    \bu &= \quad\argmin\limits_{\bu, \xi\geq 0} \frac{1}{2}||\bu||_2^2 + C\cdot\xi \\
        &\mbox{s.t.}  \quad \frac{1}{N}\sum\limits_{i\in[N]}
    \max\limits_{\bar{\bx}_i\in \calX}\left[\calL(\bx_i,\bar{\bx}_i)-\langle
    \bu,\bpsi(I_i,\bar{\bx}_i)\rangle + \langle
    \bu,\bpsi(I_i,\bx_i^*)\rangle\right] \leq
    \xi, \label{cnstr:fgcrf}
\end{flalign}
\end{subequations}
where $I_i$ represents the $i^{\text{th}}$ image and $\bx^*_i$ is the best
segmentation achievable  given a particular superpixel
partitioning of image $I_i$. For an image with $n$ superpixels the potential values are pooled across all nodes and edges,
\begin{flalign}
    \bpsi(I^i,\bar{\bx}_i) = \left[ \begin{array}{l}
            \sum\limits_{s\in[n]} \btheta_1(x_{is})\\
            \sum\limits_{(s,t)\in E} \btheta_2(x_{is},x_{it}) \\
            \sum\limits_{s\in[n]} \theta_c(x_{is}) \\
            \sum\limits_{(s,t)\in E} \theta_e(x_{is},x_{it})
        \end{array}
    \right],
\end{flalign}
and the task loss $\calL(\bx_i,\bar{\bx}_i)$ is discussed in the next section.
The quadratic program in eqn.~\ref{eqn:fgssvm} is a one-slack, margin-rescaled,
structural SVM~\cite{joachims2009}.

\subsection{Task loss}
The task loss we use for the binary segmentation problem is the
intersection-over-union score (IoU) for predicting segmentation $\bar{\bx}_i$ with
respect to the ground-truth for image $I_i$. It can be written as,
\begin{flalign}
    \text{IoU}(\bar{\bx}) = \frac{1}{2} \sum\limits_{\ell\in\{0,1\}} \frac{\sum_j
    [\![\bar{p}_j=\ell \wedge p^*_j = \ell]\!]}{\sum_j [\![\bar{p}_j=\ell \vee
    p^*_j = \ell]\!]}\quad,
\end{flalign}
where $\bar{p}_j\doteq g(\bar{\bx})$, $g(\cdot)$ is a function mapping a figure-ground segmentation
over superpixels to the label assignment of the underlying pixels in the image,
$g:\; \calX \rightarrow \{0,1\}^{w\times h}$, and $j\in\{1,\dots,wh\}$. The
ground-truth label pixel label assignment is $p^*_j$.

Since superpixels are the underlying image elements used in the model it is
unlikely that a perfect image segmentation can be achieved, \ie
$\text{IoU}(\bx^*)< 1$. This is because the superpixels may not have perfect
alignment with foreground objects. Instead, we use a task loss
that measures performance \emph{relative} to the best achievable segmentation when committed to a
specific set of superpixels,
\begin{flalign}\label{eqn:fgtaskloss}
    \calL(\bx^{*},\bar{\bx}) \doteq \text{IoU}(\bx^{*}) - \text{IoU}(\bar{\bx}).
\end{flalign}
where $\bx^*$ is the best segmentation achievable and $\calL(\cdot,\cdot)\in [0,1]$.

\subsection{Loss-augmented inference}
The MAP solution to the binary CRF in eqn~\ref{eqn:fgcrf} given
parameters $\bu$, as well as the loss-augmented inference in
eqn~\ref{cnstr:fgcrf} were
solved using graph-cuts~\cite{kolmogorov2004}. The pairwise smoothness term in
eqn~\ref{eqn:fgcrf} is
not submodular so the graph-cut algorithm is not guaranteed to return an optimal
solution, however we consistently attained good performance using this approach.

The constraint in eqn~\ref{cnstr:fgcrf} uses the relative task loss defined in
eqn~\ref{eqn:fgtaskloss}.
Since the intersection-over-union score doesn't decompose over image elements,
the loss-augmented inference problem is more difficult to solve. We can
approximately solve it by solving the simpler problem,
\begin{flalign}
    \min\limits_{\bar{\bx}_i\in\calX} \langle \bu,\bpsi(I^i,\bar{\bx}_i)\rangle,
\end{flalign}
using graph-cuts and applying a greedy hill climbing procedure that sequentially flips the label of each
superpixel in image $i$, $\bx_{ij}$, in order to maximize the loss adjusted score of the
predicted solution relative to the ground-truth, until no more improvement can
be attained. Alternatively a message-passing inference algorithm designed to
handle high-order potentials~\cite{tarlow2012} could have been used to approximately
solve the loss-augmented inference problem.

\subsection{\divmbest inference with Hamming dissimilarity}
Given a fixed $C$ and model parameters $\bu$ from training the CRF
(\cref{sec:fgcrflearn}), we can
generate a diverse set of plausible segmentations using the \divmbest framework.
In order to compute \divmbest solutions (\cref{sec:divmbestrelax}
) we using Hamming dissimilarity for the
$\Delta$-function. The $m^{\text{th}}$-mode is generated by solving the following
minimization problem,
\begin{multline}
    \bx_i^{(m)} = \argmin\limits_{\bar{\bx}_i\in \calX} \sum\limits_{j\in[n]}
    \left[\bu_1^T\btheta_1(\bar{x}_{ij}) + \sum\limits_{k\in [m-1]} \lambda\cdot
    [\![\bar{x}_{ij}=x_{ij}^{(k)}]\!]\right]\\ + \sum\limits_{(s,t)\in E}
    \bu_2^T\btheta_2(\bar{x}_{is},\bar{x}_{it}) + \sum\limits_{j\in
    [n]}u_c\theta_c(\bar{x}_{ij}) + \sum\limits_{(s,t)\in
    E}\theta_e(\bar{x}_{is},\bar{x}_{it}).
\end{multline}
We use the same \emph{s-t} graph-cut implementation~\cite{kolmogorov2004} to solve this problem as we use to
compute the MAP solution. The value of $\lambda$ is set using cross-validation
on the training set.

\subsection{Superpixels}
In order to have a computationally efficient model that produces segmentations
with good alignment to internal image boundaries superpixels are used as
opposed to the image pixels.
To produce the superpixels the SLIC superpixel
segmentation algorithm~\cite{achanta2010} is employed. For each image the desired number of
superpixels is set to 400.

\subsection{Experiments}
The purpose of the experiments in this section is to evaluate the quality of
the segmentations from our figure-ground model produced by the \divmbest method.
\subsubsection{Data sets}
We experimented with four benchmark data sets of natural images designed for evaluation
of figure ground segmentation, and an additional data set of
radiological images.

\vspace{-1em}
\paragraph{Weizmann horses}~\cite{borenstein2008} 328 color
images of horses. This is the easiest of the five data sets, with
large prominent foreground (horses in a variety of scenes).

\vspace{-1em}
\paragraph{Graz bikes, cars, people}~\cite{marszalek2007} Each
set containes 300 color images, generally harder than horses: bikes
and cars in a variety of orientations and locations, some partially
occluded, and people in a variety of locations/poses and with varying
degree of occlusion.

\vspace{-1em}
\paragraph{Ultrasound}
This medical dataset contains 416 ultrasound (graylevel) images collected from
five hospitals with different acquisition devices, varying image
quality, noise levels and resolutions. Each image in the set contains
a single lesion with validated pathology diagnosis.
The ground truth segmentation for each image was
created manually by a radiologist, who marked the boundary of
lesions. We include this data set to evaluate the performance of the
proposed, very general, segmentation approach on images very different
from the natural scenes in the other sets. We plan to make the data
set including annotations public.

\subsubsection{Evaluation}

The Weizmann horses and the three Graz data sets are split into a single
train/test split with 1/3 of the horses and 1/2 of the Graz data sets used
as test sets~\cite{kuettel2012} on which we report performance. For cross-validation
purposes the ultrasound data set is split into five equal folds. The average of
the five evaluations is reported, where one of the five folds is used as the
test set and the remaining four folds used for training. In all the experiments
the CRF learnable parameters and diversity weight $\lambda$ were tuned using
cross-validation on the training set.

The goal of the evaluation in this section is to determine whether generating a
diverse set of plausible segmentations for the figure-ground problem can be
beneficial. Therefore we evaluate the \emph{oracle} performance (\ie best
segmentation in the \divmbest set) against the MAP solutions.

The model MAP and \emph{oracle} performance is evaluated in the context of
state-of-the-art among the published work at the time of these experiments. For
this purpose we compare our results to those in~\cite{kuettel2012}, which were
shown to be competitive for the state-of-the-art title.
\paragraph{Performance measures}{
    Performance is measured by intersection-over-union (IoU) score which is
    most common measure used in semantic segmentation. IoU is also the task loss
    that is used in learning the CRF. For each experiment the results are reported in terms of
    \emph{average IoU} over the images in the test set(s).
}

\paragraph{Running time}
For a typical image the CRF model has approximately 400 variables.
Once the bottom-up segmentation CRF is trained, producing a diverse set of segmentations
for a new images involves the following stages. SLIC superpixels are extracted
and the CRF features are computed ($\simeq$15 sec/image on a 6-core machine).
Bottom-up inference of 10 diverse solutions using graph cuts takes approximately
5 seconds.
\subsubsection{Results}
The oracle performance of the \divmbest segmentations for all data sets is summarized in
table~\ref{tab:fgoracle}. A single MAP segmentation gives reasonable accuracy
compared to results from a state-of-the-art figure-ground segmentation model~\cite{kuettel2012} at the
time of these experiments. But the MAP solution is  inferior to the
of the oracle performance over just a small set of 10 \divmbest solutions. The
oracle is in fact superior to a state-of-the-art method that does not generate
multiple segmentations. In later chapter we'll explore the
1-out-of-10 inference problem using the \divrank framework to automatically pick
the likely best segmentation from the set.

\begin{table}[!t]
  \centering
  \begin{tabular}{|l|c|c|c|c|}%c|c|}
\hline
& MAP & Oracle & \%gap & \cite{kuettel2012}\\
\hline
Weizman horses & 75.4 & {\textbf 83.0}  & 51.3\% & 79.1\\
\hline
Graz bikes &53.1 & {\textbf 61.4}  & 36.1\% & 45.0\\
\hline
Graz cars &50.0 & {\textbf 66.3}  &  59.5\% & 58.8\\
\hline
Graz people & 44.1& {\textbf 57.0}  & 26.4\% & 47.5\\
\hline
Ultrasound &39.7 & {\textbf 57.3}  & 58.0\% & 26.6\\
\hline
  \end{tabular}
  \caption{Segmentation performance on all data sets, in IoU values $\times$100. MAP: single solution from
      the bottom-up CRF model. \emph{Oracle}: (hindsight) best of 10 diverse solutions from
    the CRF. Third column: percentage of gap (oracle-MAP) recovered
    by the ranking. Last column: Figure-ground segmentation model of Kuettel
\etal.~\cite{kuettel2012}.}
  \label{tab:fgoracle}
\end{table}
Examples of the MAP vs. \divmbest \emph{oracle} figure-ground segmentations using the proposed CRF
are shown in figures~\ref{fig:fggraz} and~\ref{fig:fgweiz} along with the
ground-truth segmentations. Note how the MAP foreground regions often "bleed"
into the surrounding regions whereas the \emph{oracle} results show that there is
typically a \emph{mode} of the solution space distribution that can accurately
recover the foreground boundaries.

Samples of the \divmbest foreground segmentations for the Horses and Graz
datasets can be found in Appendix~\ref{sec:divmbestfgsample}.
\subfile{\main/gfx/figureground/graz_divmbest.tex}

\subfile{\main/gfx/figureground/weiz_ultra_divmbest.tex}

%% file: gfx/figureground/graz_divmbest.tex
\begin{figure*}[!ht]
  \centering
\vspace{-1em}
\includegraphics[width=.8\textwidth,trim=0 0 10cm 0, clip]{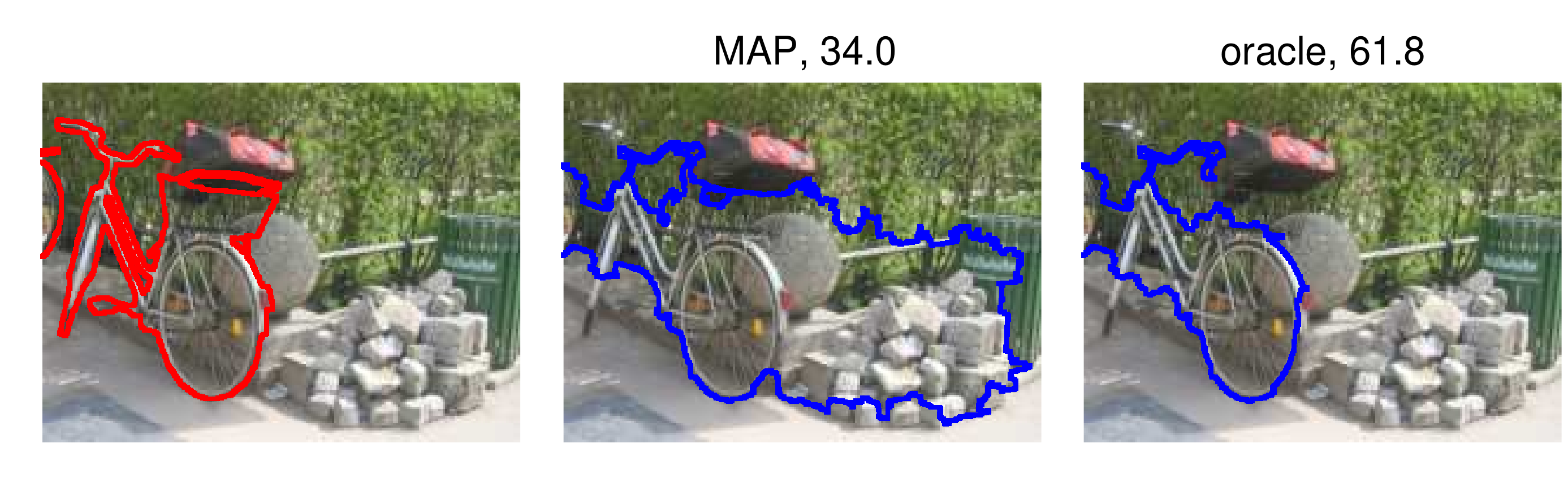}
\vspace{-1em}  \includegraphics[width=.8\textwidth,trim=0 0 10cm 0, clip]{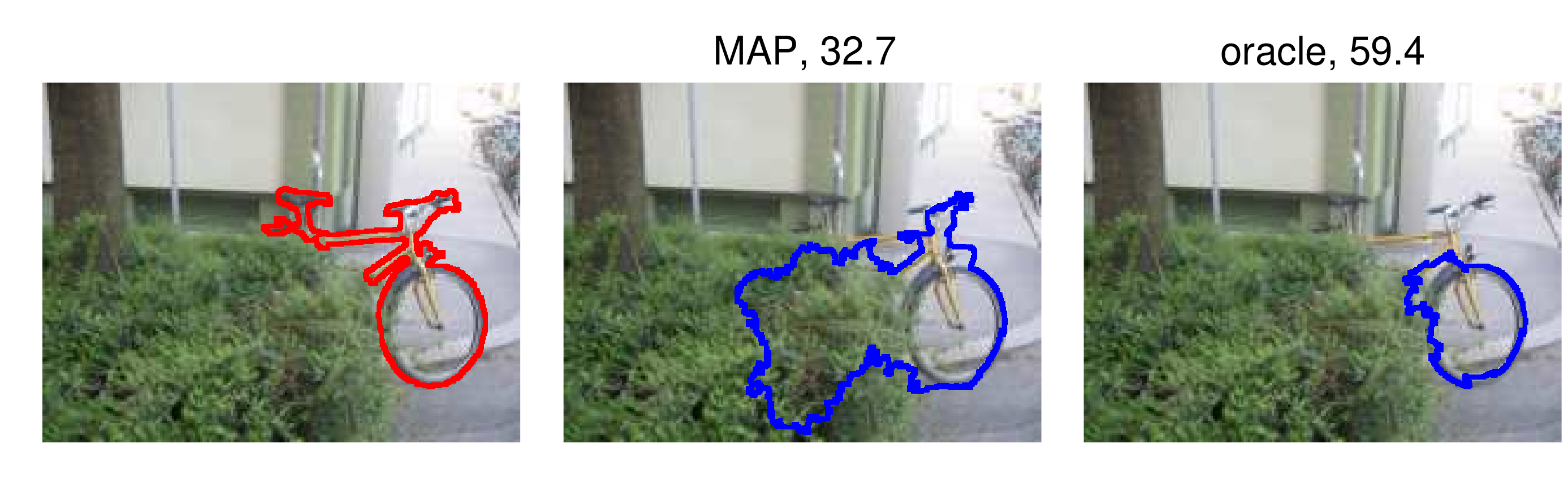}\\
 \vspace{-1em}  \includegraphics[width=.8\textwidth,trim=0 0 10cm 0, clip]{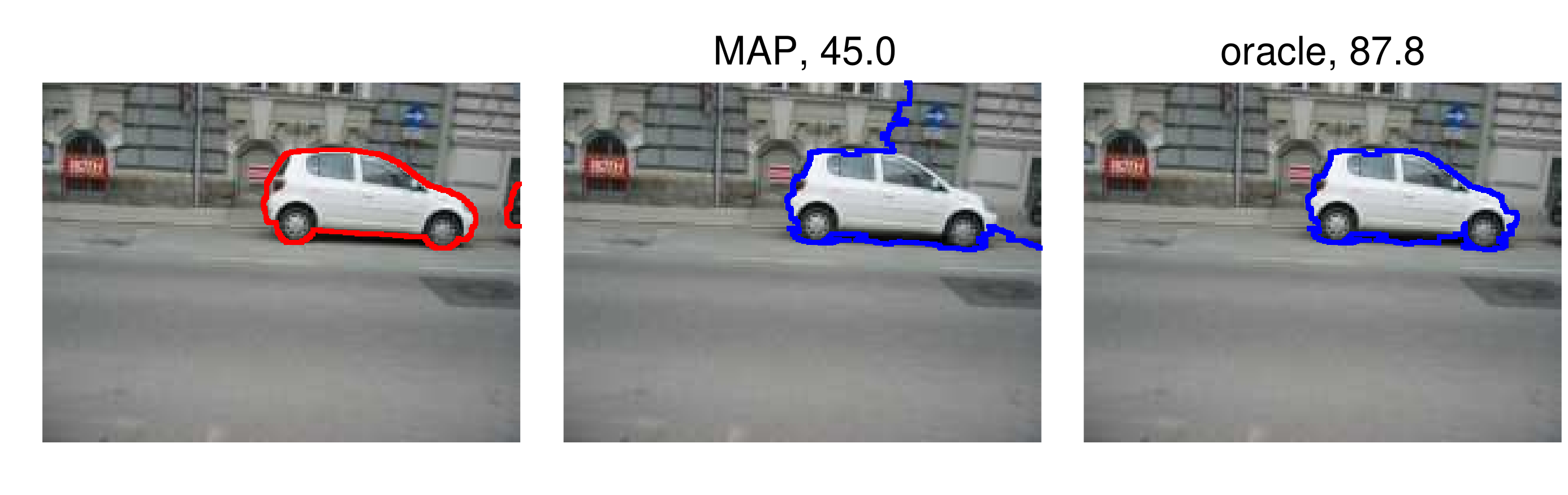}\\
 \vspace{-1em}  \includegraphics[width=.8\textwidth,trim=0 0 10cm 0, clip]{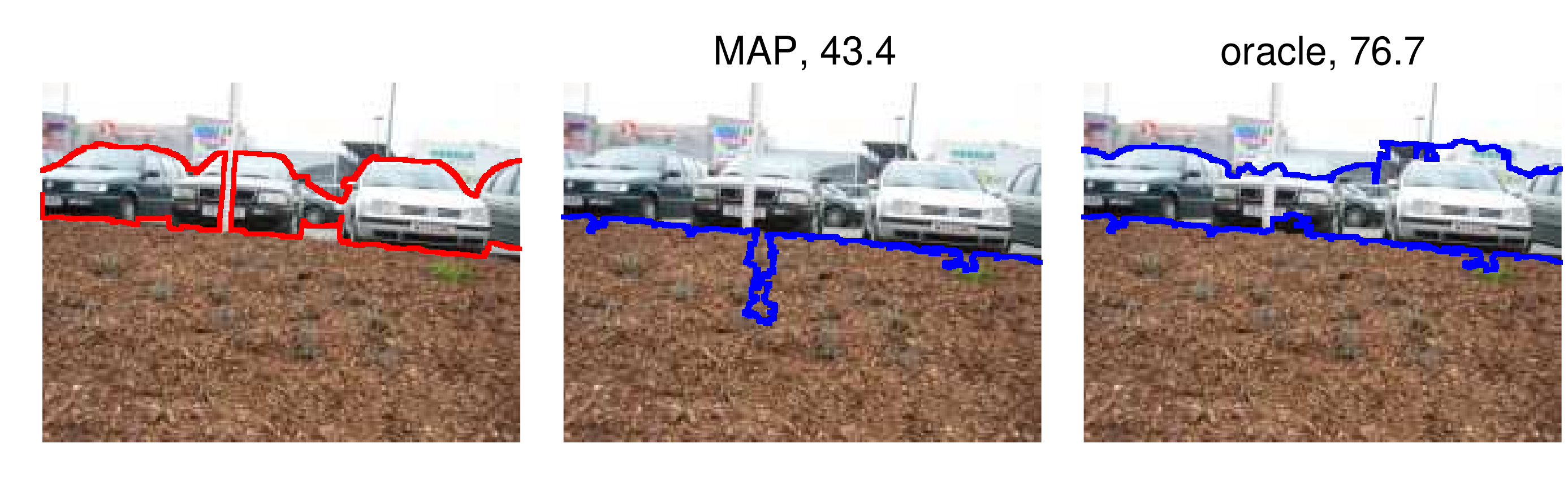}\\
 \vspace{-1em}  \includegraphics[width=.8\textwidth,trim=0 0 10cm 0, clip]{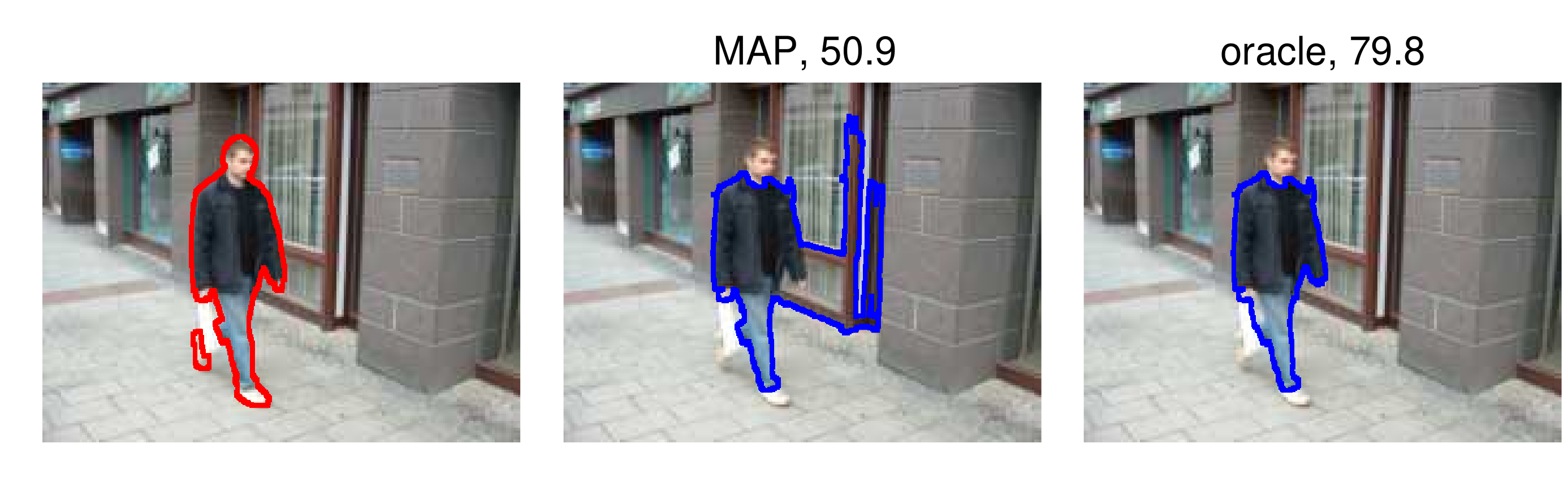}\\
 \vspace{-1em}  \includegraphics[width=.8\textwidth,trim=0 0 10cm 0, clip]{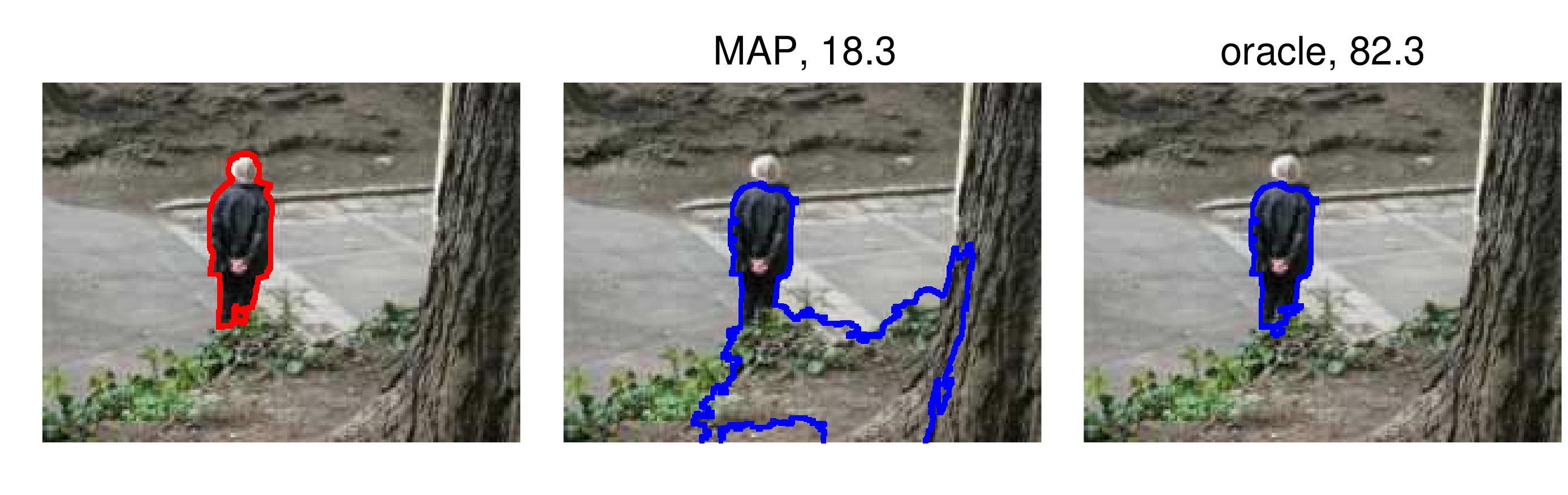}\\
 \caption{Examples of (left to right) input image with ground truth, MAP from the bottom-up CRF model, \emph{oracle} out of 10 diverse solutions. All examples are from the test portions of Graz data sets.}
\label{fig:fggraz}
\end{figure*}

%% file: gfx/figureground/weiz_ultra_divmbest.tex
\begin{figure*}[!ht]
  \centering
  \includegraphics[width=.8\textwidth, trim=0 0 10cm 0, clip]{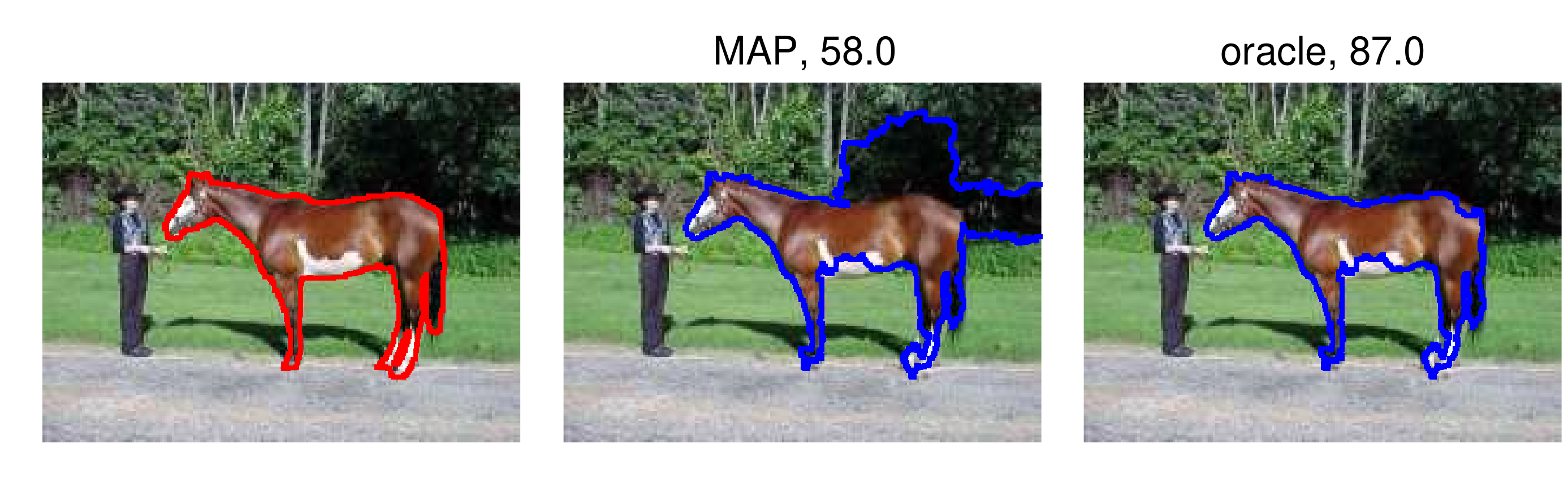}\\
  \vspace{-1em} \includegraphics[width=.8\textwidth, trim=0 0 10cm 0, clip]{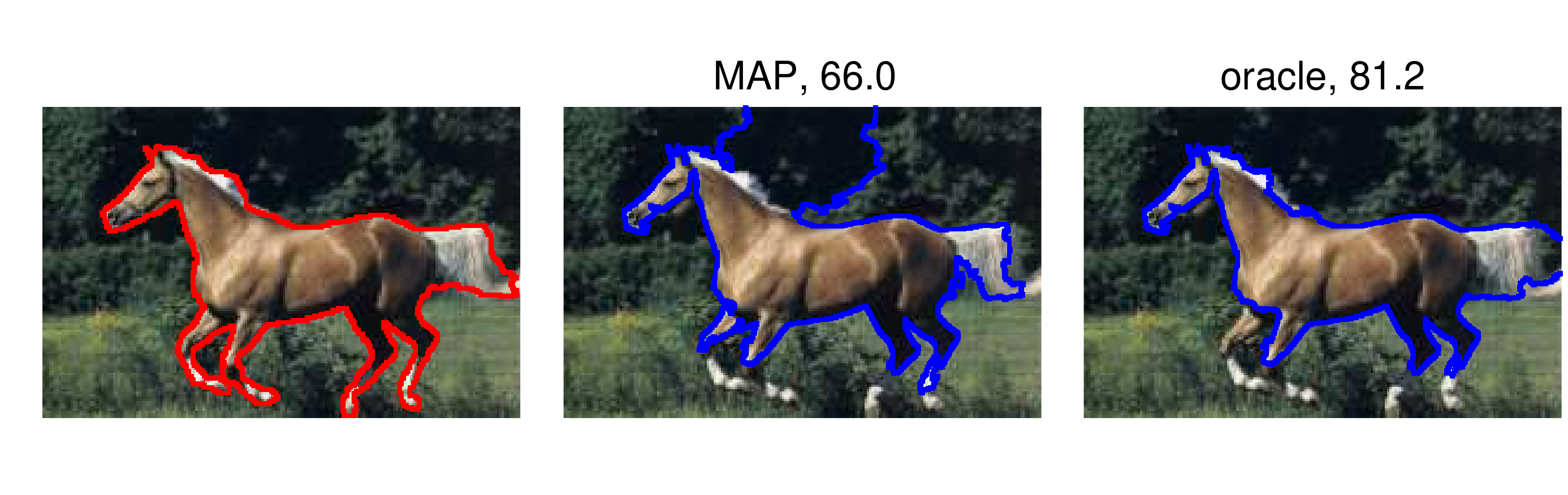}\\
  \includegraphics[width=.45\textwidth, trim=0 0 11cm 0, clip]{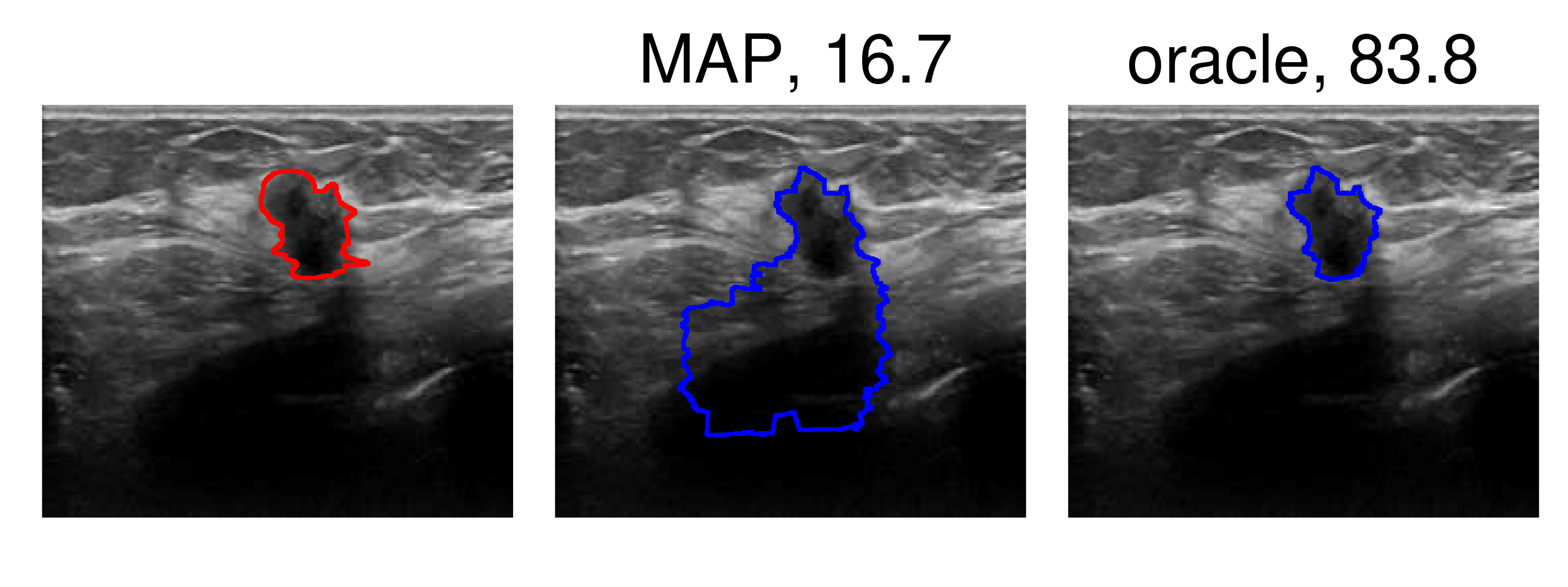}\hspace{2em}
  \vspace{-.3em} \includegraphics[width=.45\textwidth, trim=0 0 11cm 0, clip]{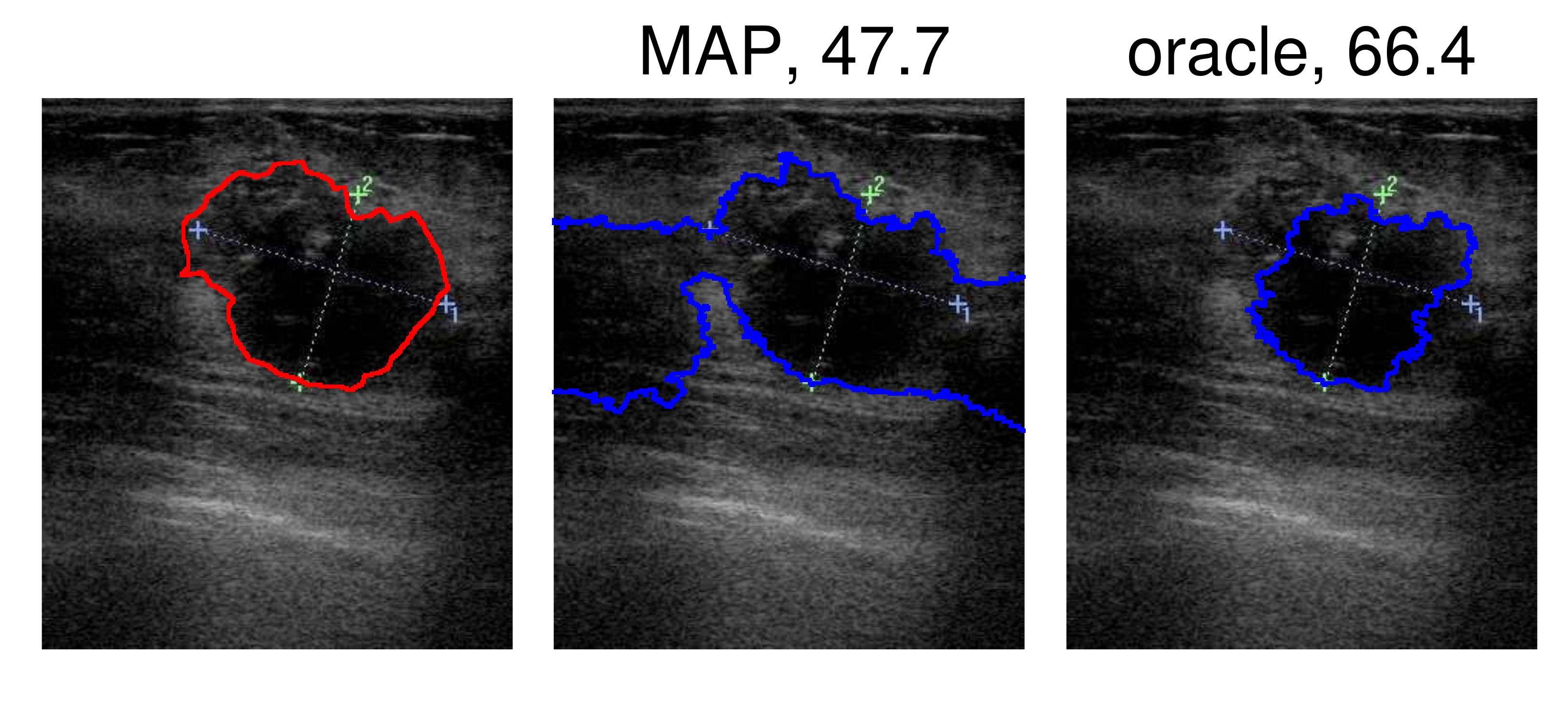}\\
  \includegraphics[width=.45\textwidth, trim=0 0 11cm 0, clip]{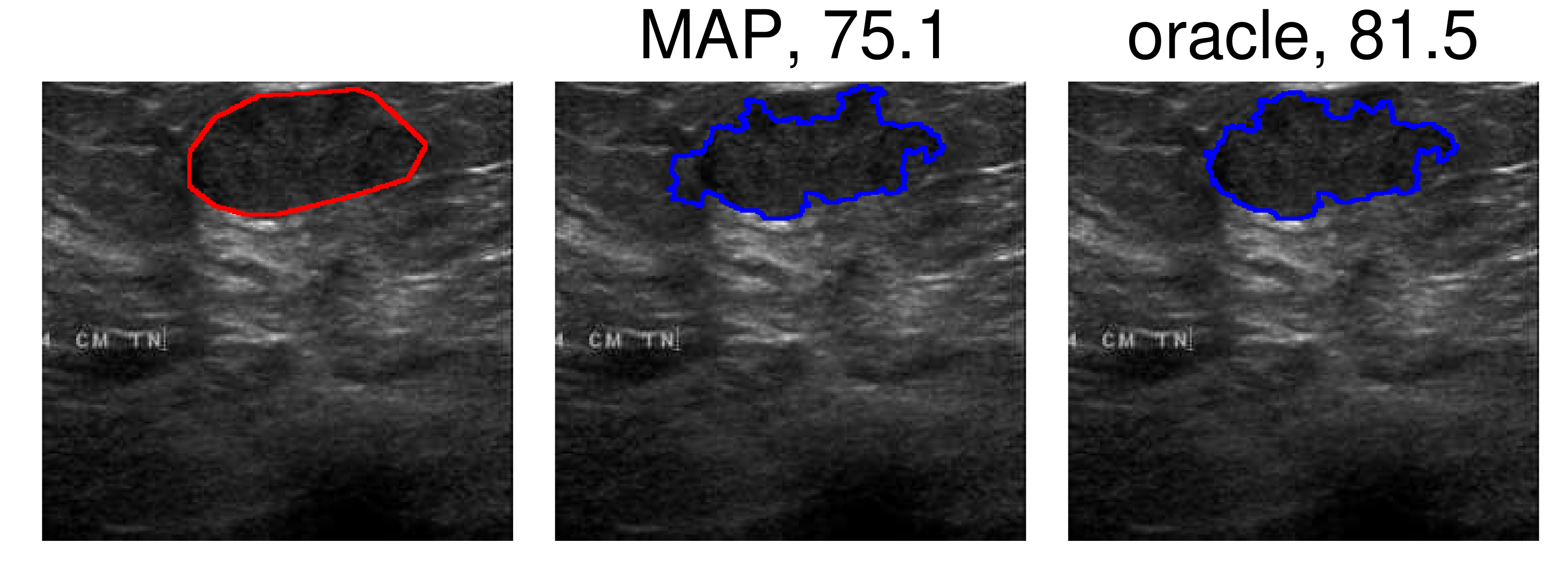}\hspace{2em}
  \vspace{-.3em} \includegraphics[width=.45\textwidth, trim=0 0 11cm 0, clip]{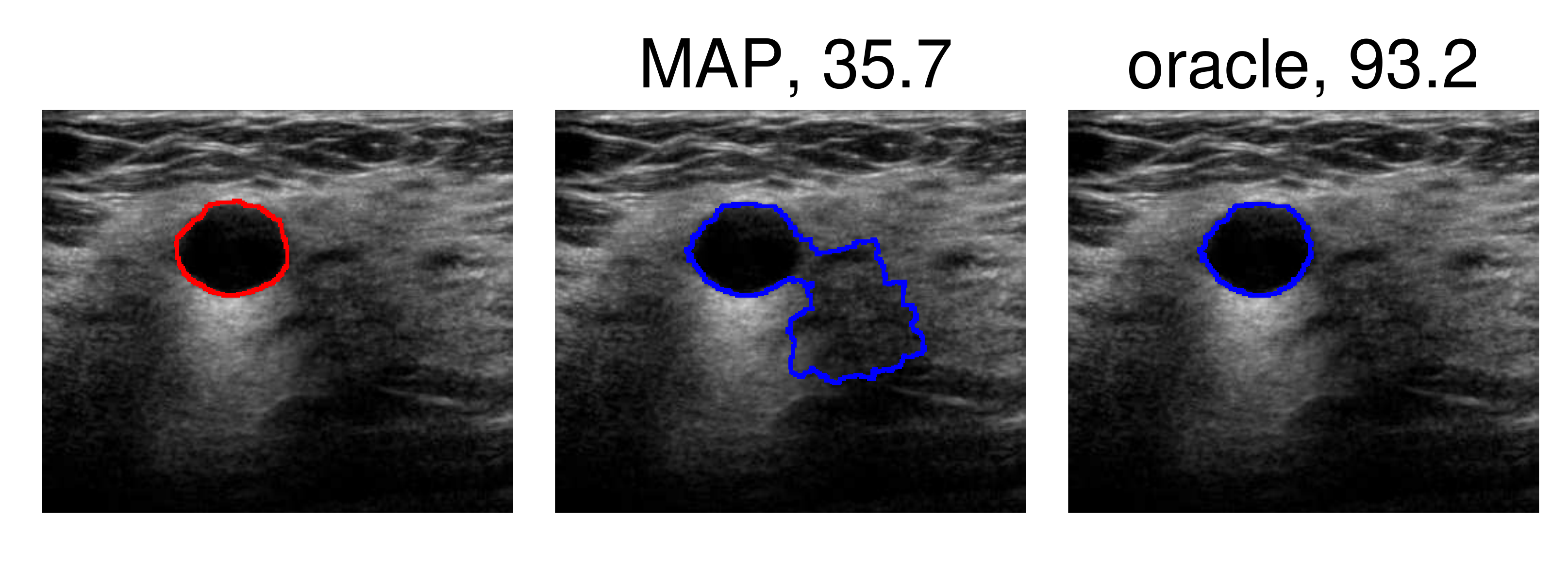}\\
  %\includegraphics[width=.45\textwidth]{figs/us37res}\hspace{2em}
  %\vspace{-.3em} \includegraphics[width=.45\textwidth]{figs/us55res}\\
  \includegraphics[width=.45\textwidth, trim=0 0 11cm 0, clip]{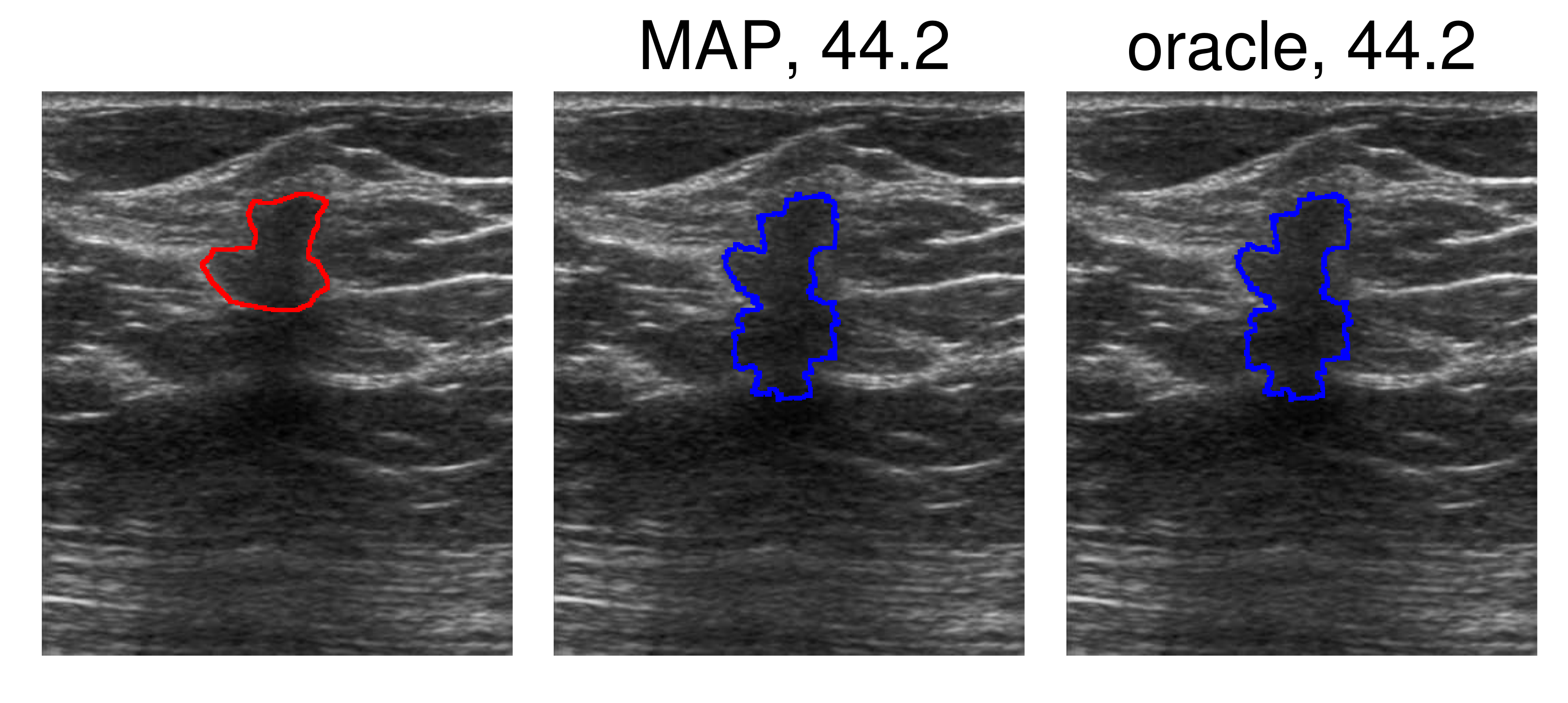}\hspace{2em}
  \vspace{-.3em} \includegraphics[width=.45\textwidth, trim=0 0 11cm 0, clip]{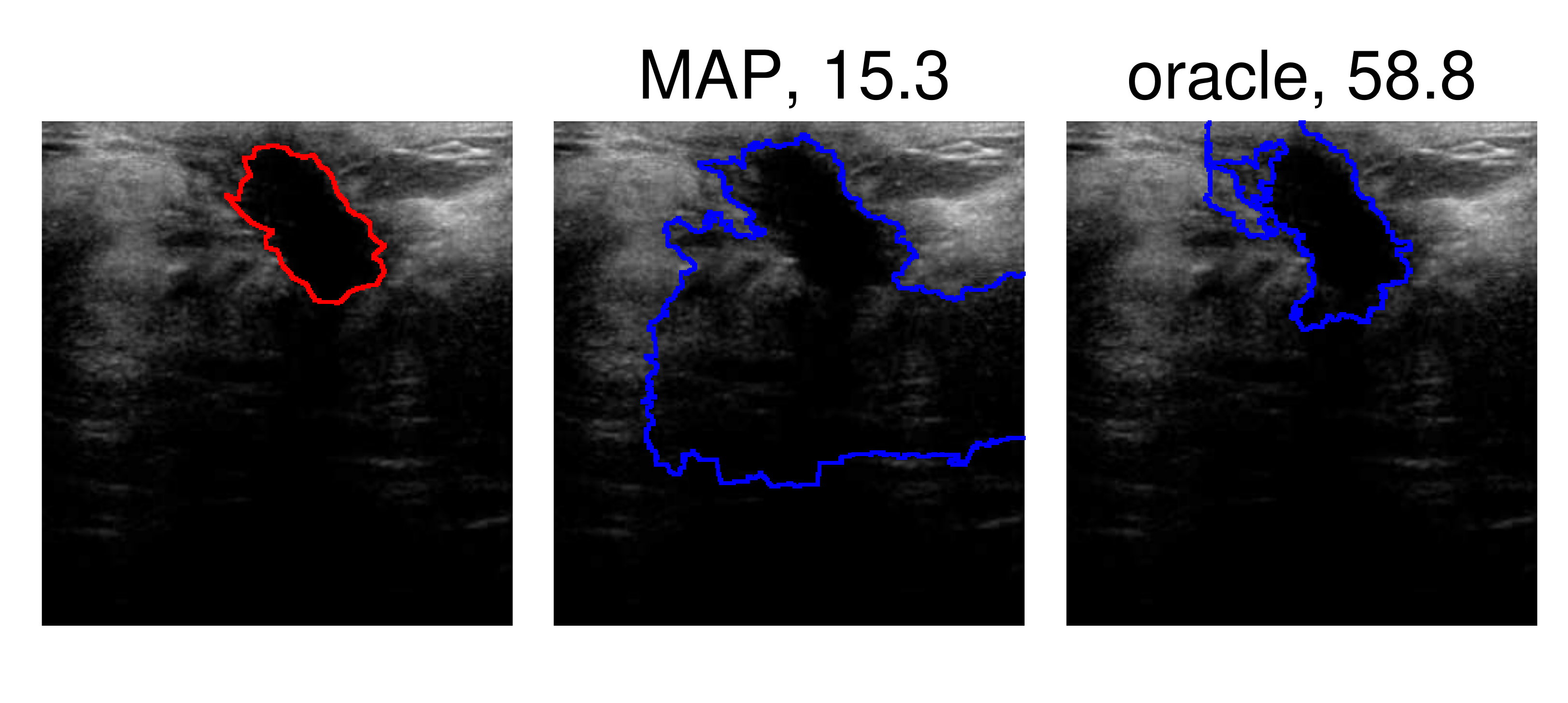}\\
  \caption{Examples of (left to right) input image with ground truth, MAP from the bottom-up CRF model, \emph{oracle} out of 10 diverse solutions. Examples are from
    the test portion of Weizmann horses data set, and from one of the
    test folds of the Ultrasound data set. Last row shows some failures.}
  \label{fig:fgweiz}
\end{figure*}

%% file: Sections/CategorySeg.tex
\section{Multi-category segmentation}\footnotemark\footnotetext{Part of the contributions to the thesis presented in this section are found in \protect\cite{batra2012}, and are in collaboration with Gregory Shakhnarovich and Dhruv Batra.}\label{sec:divmbestmulticat}
So far we have seen that the \divmbest framework can generate sets which often
contain segmentations better than the MAP solution for a number of segmentation
tasks. In this section we further apply the \divmbest algorithm to the
multi-category segmentation problem. The \divmbest solutions are evaluated
against the MAP segmentation produced by three different discrete probabilistic
models.

\subsection{Hierarchical model}
The first model we consider is the Associative Hierarchical CRF of Ladicky
\etal~\cite{ladicky2009} (see~\cref{sec:segmethods}). At the time of evaluation this model
gave competitive performance on the multi-category segmentation task.
Experiments use the Automatic Labeling Environment
(ALE)~\cite{ladicky2012} which is an implementation of the hierarchical CRF model by the authors. The
model incorporates a number of potentials including unary potentials that look
at local texture (based on textonboost features~\cite{shotton2006,shotton2009}), low-level pairwise $P^n$ Potts
potentials~\cite{kohli2007} between pixels and similar mid-level potentials between superpixels,
as well as a global co-occurrence potential\cite{ladicky2010}.

The graph-cut inference algorithm of Ladicky \etal~\cite{ladicky2009,ladicky2010} is used to compute the MAP
solution over this hierarchical model. Assuming a Hamming dissimilarity function
$\Delta(\cdot,\cdot)$, we can compute the subsequent \emph{modes} (\ie \divmbest
segmentations) by
appropriately
modifying the unary potentials according to the \divmbest formulation and
rerunning
the same inference algorithm.
\subsubsection{Baselines}
The same baselines as in the figure-ground experiments are used
(see~\cref{sec:divmbestfg}). The \emph{random} baseline
averaged over ten runs as well as the \emph{confidence} baseline are reported.
The M-Best MAP algorithm~\cite{yanover2004} (cf.~\cref{sec:mbestmap}) is infeasibly slow to run for
this model. Since the energy is not sub-modular computing min-marginals cannot
be done efficiently. Even with an implementation that re-uses search trees and
caches $\alpha$-expansion graphs it would take 10 years to compute each
additional solution for each image. Computing \divmbest solutions, however,
takes the same amount of time as computing the MAP solution.
\subsubsection{Dataset}
Multi-category level experiments were carried out on the PASCAL Visual Object
Classes (VOC ) 2010 segmentation benchmark~\cite{everingham2010}. PASCAL VOC 2010 contains 21 semantic
categories (20 object categories (\emph{$\{\texttt{aeroplane, bicycle, bird, bottle,
car,}\dots\}$} + \emph{\texttt{background}}), and the task is to label every pixel
in the image with one of the 20 object categories or the background. This task
is part of the PASCAL VOC challenge. The dataset contains
\train, \val, and \test splits that contain 964 images each. The object
categories appear in natural scenes under varying appearance, lighting, and pose.
Many images contain multiple instances of the same category and more than
one category can appear in the same image.

Segmentation accuracy is scored using the standard PASCAL VOC
intersection-over-union (IoU) measure (\ie pixelwise
$\frac{\text{intersection}}{\text{union}}$ measure averaged over masks of all
categories). The relevant parameters, such as the multiplier on the diversity
term ($\lambda$) in the \divmbest formulation, are tuned on the \val set,
after the model has been trained on \train. Ground-truth segmentations are not
provided for \test but test set accuracies can be obtained by submitting a single
segmentation prediction per image to the VOC evaluation server.

\subsubsection{Results}
\subfile{\main/data/pascal/valaccuracies}

\subfile{\main/data/pascal/testaccuracies}

To measure the upper-bound on segmentation accuracy achievable with the
\divmbest solutions we evaluate on the \val set for which we have ground-truth.
\emph{Oracle} accuracy is computed as follows: for each image the segmentation in the
\divmbest set that has highest pixel-wise IoU w.r.t ground-truth (averaged over
all category masks) is selected. The results on \val for 10 \divmbest
\emph{modes} for \ALE are reported in table~\ref{table:val_accuracies} along with the MAP accuracy and
baselines that generate multiple solutions using different perturbation
strategies on the MAP segmentation. The result on \test using 10-\emph{modes} is
summarized in table~\ref{table:test_accuracies}. To illustrate how the upper-bound on segmentation
accuracy grows as the number of solutions increases,
figure~\ref{fig:aleoracleacc} shows a plot
of \emph{oracle} accuracy versus number of \divmbest solutions. The MAP accuracy
is shown as the dashed horizonal line, and the accuracy of the \emph{confidence}
based solution is only slightly better than MAP. With $m=30$
solutions the \emph{oracle} accuracy reaches $48\%$ on \val. Though \val and
\test set performance aren't directly comparable the \emph{oracle} performance
on \test
is likely better --- by a significant margin --- than state-of-the-art methods
at the time of experiments.\sidenote{the winning entry of VOC2010 \texttt{comp5}
challenge achieved $40.1\%$ on \test}
\subfile{\main/gfx/pascal/modestats.tex}
Sample \divmbest segmentations on PASCAL VOC 2010 \val set images are shown in
Appendix~\ref{sec:divmbestsample}.

\subsubsection{Evaluating \divmbest modes}
Figure~\ref{fig:modehamdist} shows a plot of the distance of the \divmbest
solutions (aka \emph{modes}) to the MAP solution and previous modes. The
normalized (w.r.t image size) Hamming distance between solutions monotonically
increases with each additional solution. This show that the \divmbest Hamming
dissimilarity constraints (\ie $\Delta(\bmu,\bmu^{(i)})\geq k$) are encouraging
diversity between solutions. The energy of the modes as a percentage of the MAP
energy is shown in figures~\ref{fig:energymode6} and~\ref{fig:energymode31}. A
majority of \emph{modes} have higher energy than the MAP
solution\sidenote{guaranteed if using exact inference}. A small proportion of
\emph{modes} have less energy than MAP due to the fact that the model uses
\emph{approximate} inference.
\subsection{Feed-forward model}
In \ALE complex interactions between image elements is captured by the
hierarchical structure of the graph and the higher-order graph cliques.
An alternative approach is presented by the Second-Order Pooling (\otwop) approach of
Carreira \etal~\cite{carreira2012o2p}. In \otwop, complex interactions between
regions in the image are captured by global region descriptors that are
constructed by second-order pooling of local
descriptors such as SIFT and local binary patterns (LBP)~\cite{ojala1996,ojala2002}. Carreira \etal present
a simple inference algorithm for the \otwop model, which can be applied directly
to generating the diverse segmentations of the
\divmbest algorithm with Hamming dissimilarity. The details of the model and
evaluation are deferred to Chapter~\ref{ch:divrankexp} where we also evaluate an
approach to re-ranking the \divmbest solutions.

\subsection{Convolutional neural network + dense CRF model}\label{sec:cnn_divmbest}

Current state-of-the-art semantic segmentation is done by combining very deep
convolutional neural networks (CNNs) or residual neural networks (RNNs) with fully
connected dense pairwise
CRFs~\cite{chen2014}~\cite{he2016}. Deep networks are superior at
building local features that capture information at multiple spatial scales of
the image, however the output suffers from a decrease in resolution compared to the
input image. Dense pairwise CRFs can introduce low-order
dependencies between image elements, that are not constrained to be local.
Additionally local pairwise potentials can provide spatial smoothness
constraints on the solution and improve alignment with image boundaries. Piggy-backing dense
CRFs on top of deep segmentation networks combines the benefits of both
approaches - efficient computation of complex features that incorporate both
local and global interactions in the image along with constraints on local smoothness.
We can view the CNN + dense CRF pipeline as a discrete probabilistic model on a
dense pairwise graph where the unary potentials are defined by the output of the
CNN at each image element (superpixel or pixel).

The \divmbest framework is agnostic to the underlying discrete probabilistic
model so we can apply it to this model in a similar manner as previous
models.
Given a CNN trained on the semantic segmentation task, the above deep network
approaches use the features from the last layer of the network to initialize the
unary potentials of a fully-connected CRF. For a fully-connected CRF where the
pairwise edge potentials are defined by a linear combination of Gaussian kernels
an efficient approximate inference algorithm exists~\cite{krahenbuhl2011} for computing the MAP
solution. Approximate inference is based on an iterative message passing algorithm where messages are computed using efficient Gaussian
filtering in feature space. Assuming pixel-wise Hamming diversity constraints
between solutions the \divmbest
algorithm amounts to modifying the unary potentials and rerunning the message
passing algorithm to compute successive solutions.

We investigate the benefit of applying \divmbest to one such deep neural network +
dense CRF pipeline -- we use the Zoom-out network~\cite{mostajabi2015} with the
DeepLab dense CRF implementation~\cite{chen2014}\cite{krahenbuhl2011}. The
Zoom-out network (cf.~\cref{sec:segmethods}) is first pre-trained to perform
multi-category image classification on the
ImageNet dataset~\cite{russakovsky2015}. Subsequently the final
fully-connected layer of the CNN is modified to a fully-convolutional layer with
output feature map depth set to 21\sidenote{corresponding to the 21 PASCAL VOC
categories} and spatial extents up-sampled to be the same size as the input
image. The CNN is then fine-tuned in a end-to-end manner
for the semantic segmentation task
using the PASCAL VOC 2012 \train set + 11.3K annotated PASCAL VOC 2011 images from the
Semantic Boundaries Dataset~\cite{bharath2011}. Given a trained network, the dense CRF unary
potentials are initialized with the features from the last fully-convolutional
layer of the network, and the CRF hyper-parameters are fixed to the
defaults set by the implementation of Chen \etal~\cite{chen2014}. On the PASCAL VOC 2012
\val set this pipeline achieves 72\% MAP accuracy\sidenote{current
state-of-the-art methods~\cite{wu2016,zhao2016} that use additional training data currently achieve
$\sim 85\%$ accuracy on PASCAL VOC 2012} (IoU accuracy averaged over
all categories/images).

\subsubsection{Dataset}
The segmentation results in this section are reported on images from the PASCAL VOC
2012 benchmark~\cite{everingham2010}. It contains 4,369 images split into
\train (1,464 images), \val (1,449 images), and \test (1,456 images) sets. The
CNN is pre-trained for the 1000-category classification task from the ImageNet Large Scale Visual Recognition
Challenge (ILSVRC)~\cite{russakovsky2015}, using $\simeq 1.2$M ILSVRC2014 images.
An additional 11.3K images from the Semantic Boundaries
Dataset~\cite{bharath2011} are used along with PASCAL VOC 2012 \train set to
fine-tune the CNN to the 21-category PASCAL VOC segmentation task.

\subsubsection{Results}\label{sec:compresults}
We can explore the maximum accuracy achievable when using the \divmbest
algorithm to generate segmentations with the CNN + dense CRF model.
To this end we evaluate two approaches to producing oracle segmentations for each image, (1)
selecting the best-out-of-$m$ segmentations based on accuracy relative to
ground-truth and, (2) constructing full image labellings from connected components
found in the $m$ segmentations using a greedy inference approach.

The oracle accuracy versus MAP for the first approach can be seen in
figure~\ref{fig:oraclecnn}. This suggest that if we pick the best-out-of-$40$
solutions we can achieve more than 5\%-point improvement over MAP in overall segmentation
accuracy.
\subfile{\main/gfx/cnn/oracle_comp}
The second approach relies on a greedy inference algorithm over connected
components (\ie contiguous image regions taking the same label) found across the
$m$ segmentations. We first compute a bag containing tuples, $(R_j,\ell_j)$ of
connected components $R_j$ extracted from the $m$ segmentations with
corresponding category
labels $\ell_j$. Note that duplicate tuples can exist if a contiguous region
with corresponding label is found in more than one image segmentations. We
assign each connected component in the bag a score defined to be its highest IoU
with all connected components taking the same label in the
ground-truth segmentation. Starting with an empty labeling, the full image
labeling for the image is constructed using a greedy strategy of \emph{pasting} the connected
components in order of decreasing IoU score until a prescribed
score threshold is reached, at which point the algorithm stops. During the
pasting procedure if the current connected component overlaps with a region in
the image that has already been assigned a label then the previous label for
pixels in that
region is retained. A non-maxima suppression step is applied at each iteration: after
pasting a connected component from the bag we cull the bag of all connected
components that have intersection greater than a fixed threshold. The result of
this greedy construction, as $m$ varies from 1 to 40 solutions, is shown in
figure~\ref{fig:oraclecomp}. Notice that the accuracy of greedy inference on
connected components from the first solution (\ie MAP) is higher
than the accuracy of original MAP solution computed on the model. This is
because greedy inference on connected components culls from the final image
labeling those connected components that do not align well with the ground truth
segmentation. With this approach we get a significant increase in oracle
accuracy compared to the best-out-of-$m$ results -- an almost 13\% point
increase over MAP. An example result of composing the oracle segmentation from the \divmbest segmentations is shown in figure~\ref{fig:comp}.

The MAP solution from current state-of-the-art models, that combine deep neural
networks with dense CRFs,
achieve comparable accuracy\sidenote{$\sim 85\%$ on PASCAL VOC 2012
\texttt{comp6} challenge~\cite{wu2016,zhao2016}} to this greedy-inference approach on the less
accurate Zoom-out
CNN + dense CRF model. The \emph{oracle} results illustrate that it's plausible to
leverage the \divmbest algorithm to generate a set of diverse segmentations that
often contain highly accurate solutions even when the MAP solution from the underlying segmentation
model is inaccurate. This suggests that near state-of-the-art results can be had
in the multi-category image segmentation problem\sidenote{as well as the
segmentation tasks presented earlier} by devising methods with the goal of
picking the best solution from the \divmbest set. We explore one such method in
Chapter~\ref{ch:divrank}
that is learned with the goal of re-ranking the segmentations in set so that the
best segmentation is top ranking.

\subsection{Summary}
The results on a number of semantic segmentation datasets show the utility of
using the \divmbest formulation to produce a diverse set of highly plausible
segmentations. Specifically, the \emph{oracle} accuracies show that across the
segmentation tasks the \divmbest set often contains much higher quality
segmentations than MAP. This validates the alternative approach of leveraging
models, in which exact of provable approximate inference is tractable, by efficiently producing a diverse set of
segmentations as opposed to devising more complex models where inference becomes
intractable. The \emph{oracle} results highlight the importance of being able to
pick the best segmentation from the \divmbest set via ranking and we investigate
this in chapter~\ref{ch:divrankexp}.
\printbibliography

%% file: data/pascal/valaccuracies.tex
%\begin{table}[t]
%%\setlength{\tabcolsep}{ 1.5pt}
%\begin{minipage}[b]{2\linewidth}
%\begin{center}
%\resizebox{!}{1 cm}{
% %\setlength{\tabcolsep}{1.5pt}
% \begin{tabular}{r| m{1cm} m{1cm} m{1cm} m{1cm} m{1cm} m{1cm} m{1cm} m{1cm} m{1cm} m{1cm} m{1cm} m{1cm} m{1cm} m{1cm} m{1cm} m{1cm} m{1cm} m{1cm} m{1cm} m{1cm} m{1cm} m{1cm}} 
%\hline\\[.25pt]
%&bg & plane & bicycle & bird & boat & bottle & bus & car & cat & chair & cow & dtable & dog & horse & motorbike & person & plant & sheep & sofa & train & tvmonitor & Avg\\[1pt]
%\hline\\[.5pt]
%MAP & 78.5 & 35.1 & 5.2 & 20.3 & 20.8 & 11.8 & 39.4 & 38.2 & 25.8 & 8.9 & 14.1 & 30.2 & 10.0 & 12.3 & 37.6 & 33.5 & 10.3 & 24.2 & 16.2 & 28.7 & 20.5 & 24.8\\
%\hline\\[.5pt]
%MBM + H & 84.0 & 48.4 & 10.0 & 28.4 & 30.4 & 25.1 & 54.2 & 55.1 & 34.2 & 18.1 & 25.0 & 39.9 & 18.8 & 22.0 & 49.8 & 46.9 & 19.7 & 44.4 & 21.9 & 43.6 & 33.5 & 35.9\\
%\hline\\[.5pt]
%Unary + H & 78.5 & 35.1 & 5.3 & 20.1 & 20.7 & 12.6 & 39.4 & 37.9 & 26.8 & 8.9 & 14.1 & 30.2 & 10.3 & 12.2 & 39.5 & 33.4 & 10.6 & 24.2 & 17.3 & 28.4 & 20.5 & 25.1\\
%\hline\\[.5pt]
%Random &74.9 & 32.4 & 6.4 & 16.1 & 14.7 & 12.3 & 34.3 & 32.6 & 22.6 & 8.0 & 13.2 & 21.1 & 8.7 & 10.4 & 32.9 & 28.9 & 7.8 & 20.6 & 10.8 & 23.5 & 17.3  & 21.4\\[.5pt]
%%MBM + ranking &
%%Unary Conf + ranking &
%\end{tabular}}
%\end{center}
%\end{minipage}
%\end{table}
%
%
\begin{table*}[t]
\begin{center}
%{\footnotesize
{\tiny
\setlength{\tabcolsep}{1pt}
\begin{tabular*}{1.04\textwidth}{@{\extracolsep{\fill}}p{1.2cm}ccccccccccccccccccccc|c}
%\begin{tabular*}{1\textwidth}{c|ccccccccccccccccccccc|c}
\toprule
& \rotatebox{90}{Backgr.}
& \rotatebox{90}{Plane}
& \rotatebox{90}{Bicycle}
& \rotatebox{90}{Bird}
& \rotatebox{90}{Boat}
& \rotatebox{90}{Bottle}
& \rotatebox{90}{Bus}
& \rotatebox{90}{Car}
& \rotatebox{90}{Cat}
& \rotatebox{90}{Chair}
& \rotatebox{90}{Cow}
& \rotatebox{90}{D.Table}
& \rotatebox{90}{Dog}
& \rotatebox{90}{Horse}
& \rotatebox{90}{M.bike}
& \rotatebox{90}{Person}
& \rotatebox{90}{Plant}
& \rotatebox{90}{Sheep}
& \rotatebox{90}{Sofa}
& \rotatebox{90}{Train}
& \rotatebox{90}{TV.Mo.}
& \rotatebox{90}{Average}\\
\midrule
MAP & 78.5 & 35.1 & 5.2 & 20.3 & 20.8 & 11.8 & 39.4 & 38.2 & 25.8 & 8.9 & 14.1 & 30.2 & 10.0 & 12.3 & 37.6 & 33.5 & 10.3 & 24.2 & 16.2 & 28.7 & 20.5 & 24.8\\
\emph{Confidence} & 78.5 & 35.1 & 5.3 & 20.1 & 20.7 & 12.6 & 39.4 & 37.9 & 26.8 & 8.9 & 14.1 & 30.2 & 10.3 & 12.2 & 39.5 & 33.4 & 10.6 & 24.2 & 17.3 & 28.4 & 20.5 & 25.1\\
\emph{Random} &74.9 & 32.4 & 6.4 & 16.1 & 14.7 & 12.3 & 34.3 & 32.6 & 22.6 & 8.0 & 13.2 & 21.1 & 8.7 & 10.4 & 32.9 & 28.9 & 7.8 & 20.6 & 10.8 & 23.5 & 17.3  & 21.4\\
\textcolor{red}{10\textsc{Modes}} & \textbf{85.6} & \textbf{53.9} &
\textbf{14.6} & \textbf{36.9} & \textbf{33.6} &
\textbf{33.2} & \textbf{64.2} & \textbf{56.3} &
\textbf{47.7} & \textbf{16.1} & \textbf{30.3} & \textbf{46.8} & \textbf{29.1} &
\textbf{28.7} & \textbf{59.0} & \textbf{50.0} & \textbf{32.5} & \textbf{46.7} & 
\textbf{31.2} & \textbf{52.9} & \textbf{39.0} & \textbf{42.3}\\
%$MModes$+rank \\
%Unary+rank \\
\bottomrule
\end{tabular*}
}
\end{center}
%\vspace{\captionReduceTop}
\caption{Pascal VOC 2010 \val set accuracies for \ALE model.}
\label{table:val_accuracies}%\vspace{\captionReduceBot}\vspace{-15pt}
\end{table*}

%% file: data/pascal/testaccuracies.tex
\begin{table*}[t]
\begin{center}
%{\footnotesize
{\tiny
\setlength{\tabcolsep}{1pt}
\begin{tabular*}{1.03\textwidth}{@{\extracolsep{\fill}}p{1.2cm}ccccccccccccccccccccc|c}
%\begin{tabular*}{1\textwidth}{c|ccccccccccccccccccccc|c}
\toprule
& \rotatebox{90}{Backgr.}
& \rotatebox{90}{Plane}
& \rotatebox{90}{Bicycle}
& \rotatebox{90}{Bird}
& \rotatebox{90}{Boat}
& \rotatebox{90}{Bottle}
& \rotatebox{90}{Bus}
& \rotatebox{90}{Car}
& \rotatebox{90}{Cat}
& \rotatebox{90}{Chair}
& \rotatebox{90}{Cow}
& \rotatebox{90}{D.Table}
& \rotatebox{90}{Dog}
& \rotatebox{90}{Horse}
& \rotatebox{90}{M.bike}
& \rotatebox{90}{Person}
& \rotatebox{90}{Plant}
& \rotatebox{90}{Sheep}
& \rotatebox{90}{Sofa}
& \rotatebox{90}{Train}
& \rotatebox{90}{TV.Mo.}
& \rotatebox{90}{Average}\\
\midrule
MAP & 73.7 & 44.0 & 14.2 & 15.3 & 21.0 & \textbf{23.2} & 41.3 & 37.0 & 27.6 & 6.1 & 23.9
    & 25.2 & 12.8 & 24.3 & 51.0 & 27.8 & 20.0 & 28.2 & 17.1 & 36.5 & 23.9 &
28.3\\
\textcolor{red}{10\textsc{Modes}} &\textbf{83.4} & \textbf{54.4} & \textbf{19.6}
                                  & \textbf{22.4} & \textbf{34.5} & 22.2 &
\textbf{60.8} & \textbf{55.5} & \textbf{45.8} & \textbf{14.0} & \textbf{45.5} &
\textbf{35.1} & \textbf{34.8} & \textbf{40.1} & \textbf{53.6} & \textbf{48.7} &
\textbf{28.0} &
\textbf{48.7} & \textbf{31.2} & \textbf{50.5} & \textbf{33.9} & \textbf{41.1}\\
\bottomrule
\end{tabular*}
}
\end{center}
%\vspace{\captionReduceTop}
\caption{Pascal VOC 2010 \test set accuracies for \ALE model.}
\label{table:test_accuracies}%\vspace{\captionReduceBot}\vspace{-15pt}
\end{table*}

%% file: gfx/pascal/modestats.tex
\begin{figure}[!p]
    \centering
    \subfloat[]
    {\includegraphics[width=0.5\textwidth]{\main/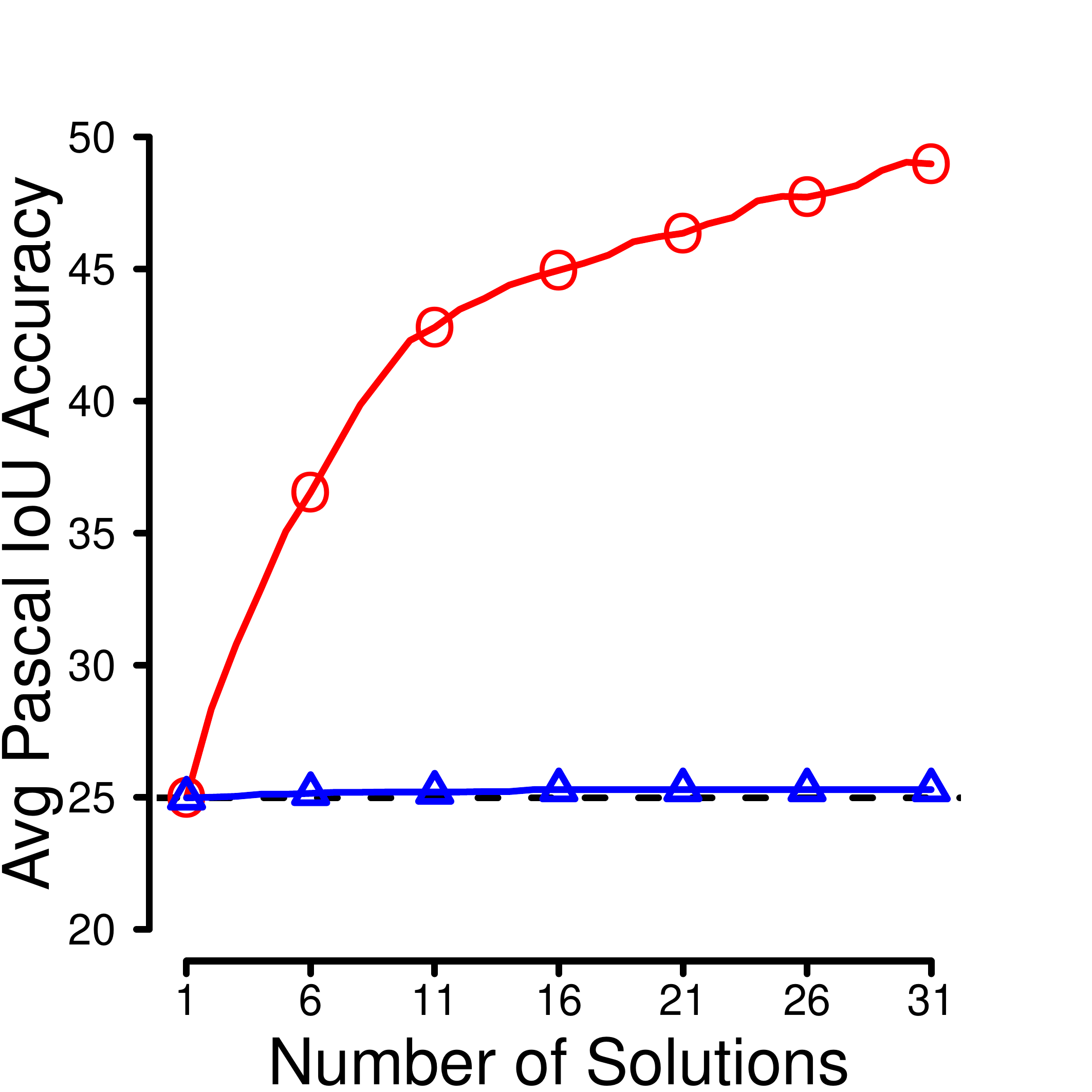}\label{fig:aleoracleacc}}
    \subfloat[]
    {\includegraphics[width=0.5\textwidth]{\main/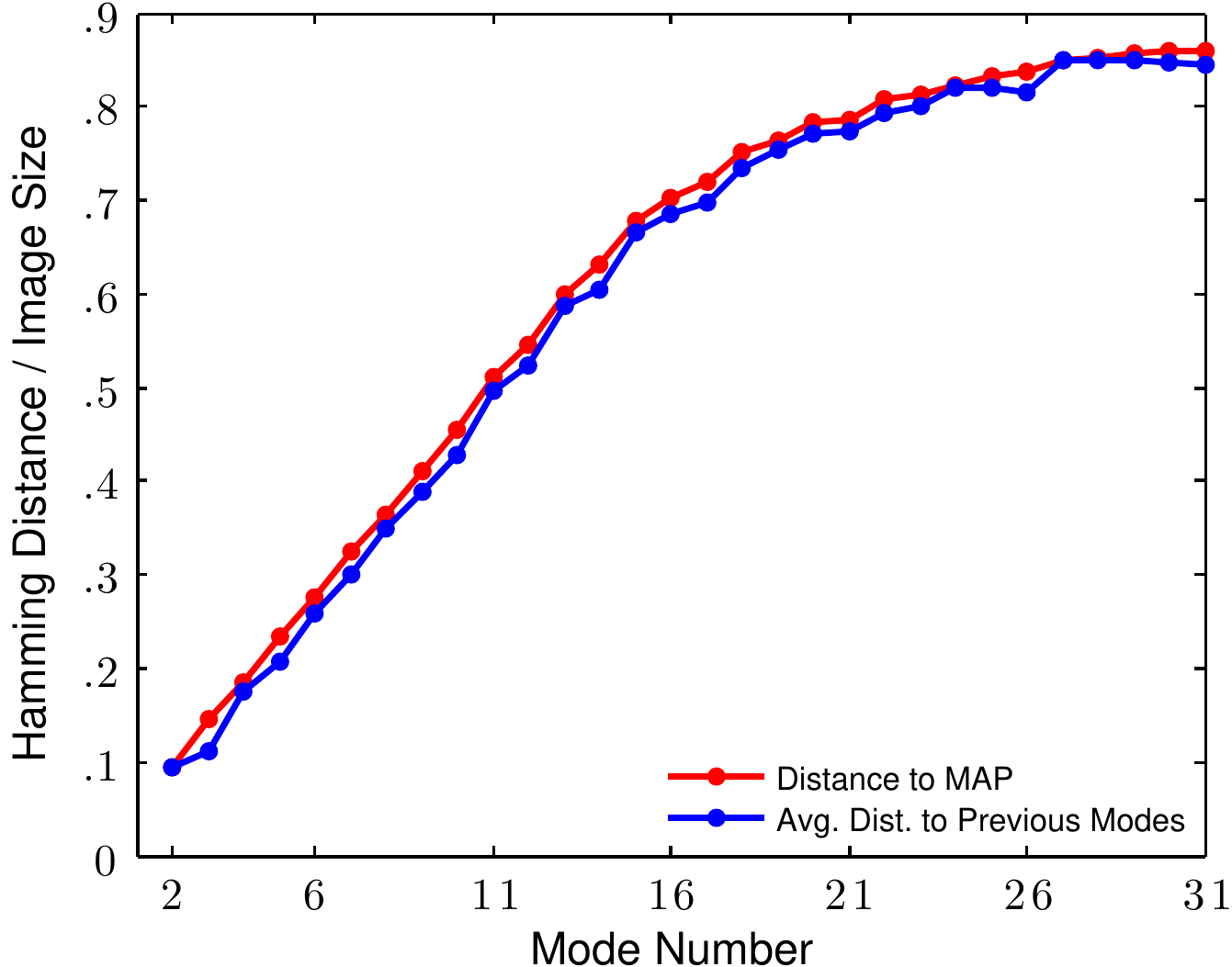}\label{fig:modehamdist} }
	\hspace{2em}
    \subfloat[]
	{\includegraphics[width=0.5\textwidth]{\main/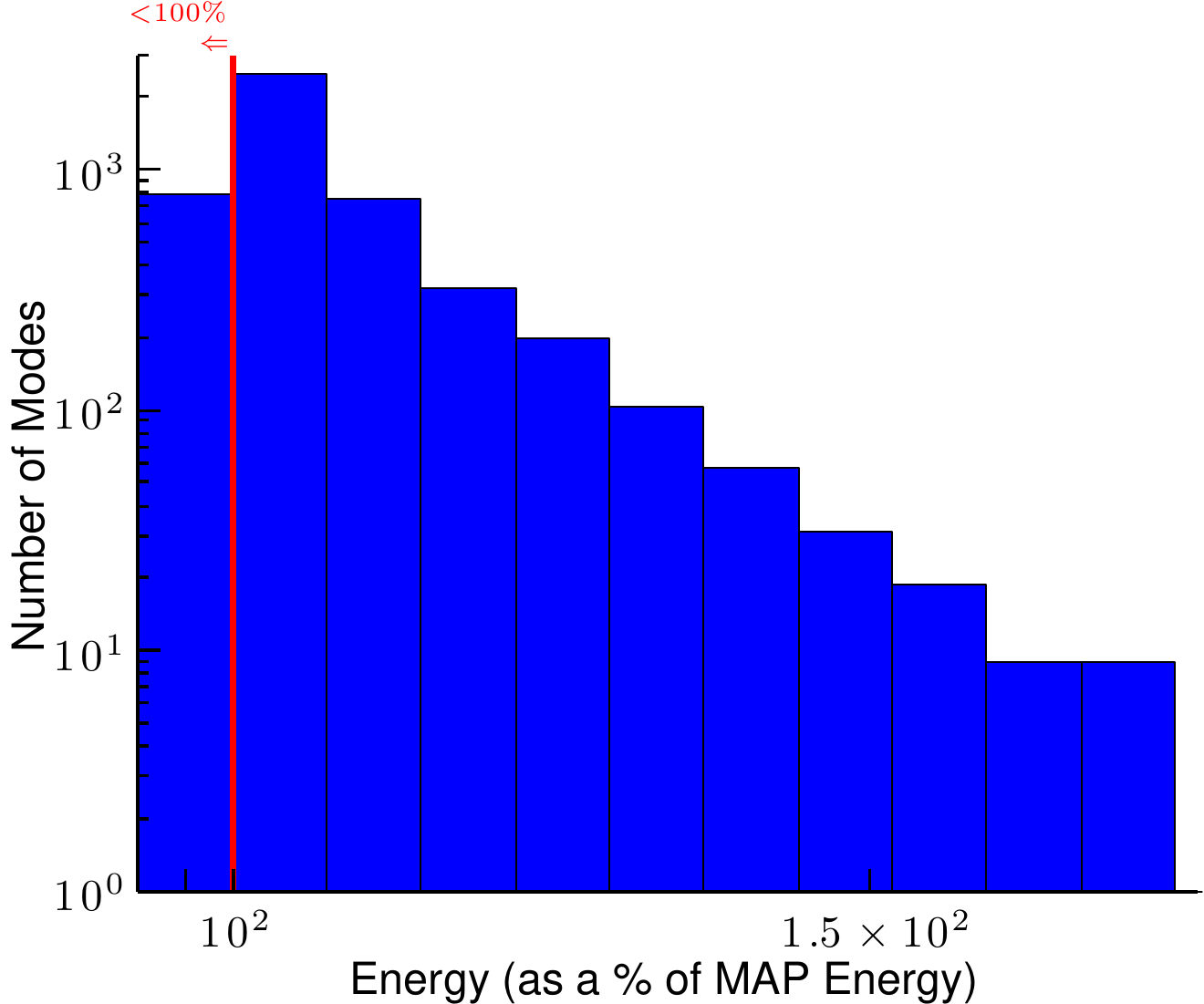}\label{fig:energymode6} }
    %\hspace{2em}
    \subfloat[]
	{\includegraphics[width=0.5\textwidth]{\main/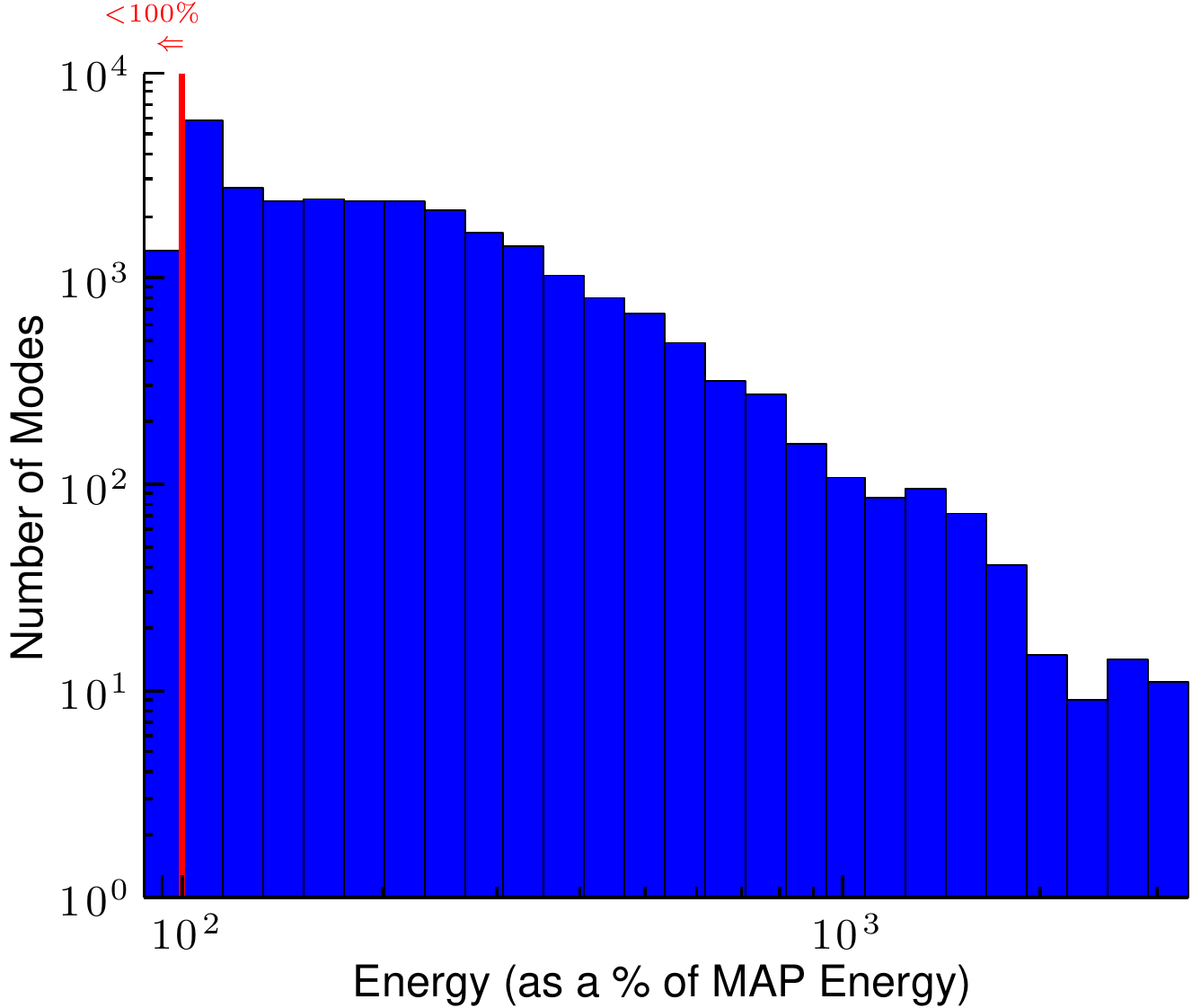}\label{fig:energymode31} }

%\vspace{\captionReduceTop}
%\begin{minipage}[c]{.35\linewidth}
    \caption{\small{(\emph{a}) \emph{Oracle} accuracy vs. number solutions on
            VOC2010 \val for \divmbest (red) and \emph{confidence} based
            perturbations (blue), along with MAP performance (black dashed).
            (\emph{b}) Mean hamming distances between
            each \emph{mode} (\divmbest solution) and the MAP solution (red),
            and average to previous \emph{modes}
            (blue), normalized by image size on PASCAL VOC 2010 \val set.
            Also show, histogram of energies (as $\%$ of MAP) over (\emph{c}) 6
            \emph{modes},
            (\emph{d}) 31 \emph{modes}, on validation set. The bar to the left of red vertical
            lines indicate number of \emph{modes}
            with energy less than or equal to MAP.}} %\vspace{\captionReduceBot}
%\end{minipage}
\label{fig:modestats}
\end{figure}

%% file: gfx/cnn/oracle_comp.tex
\begin{figure*}[!th]
    \begin{minipage}[c]{\textwidth}
        \subfloat[]
        {\includegraphics[width=0.5\textwidth]{\main/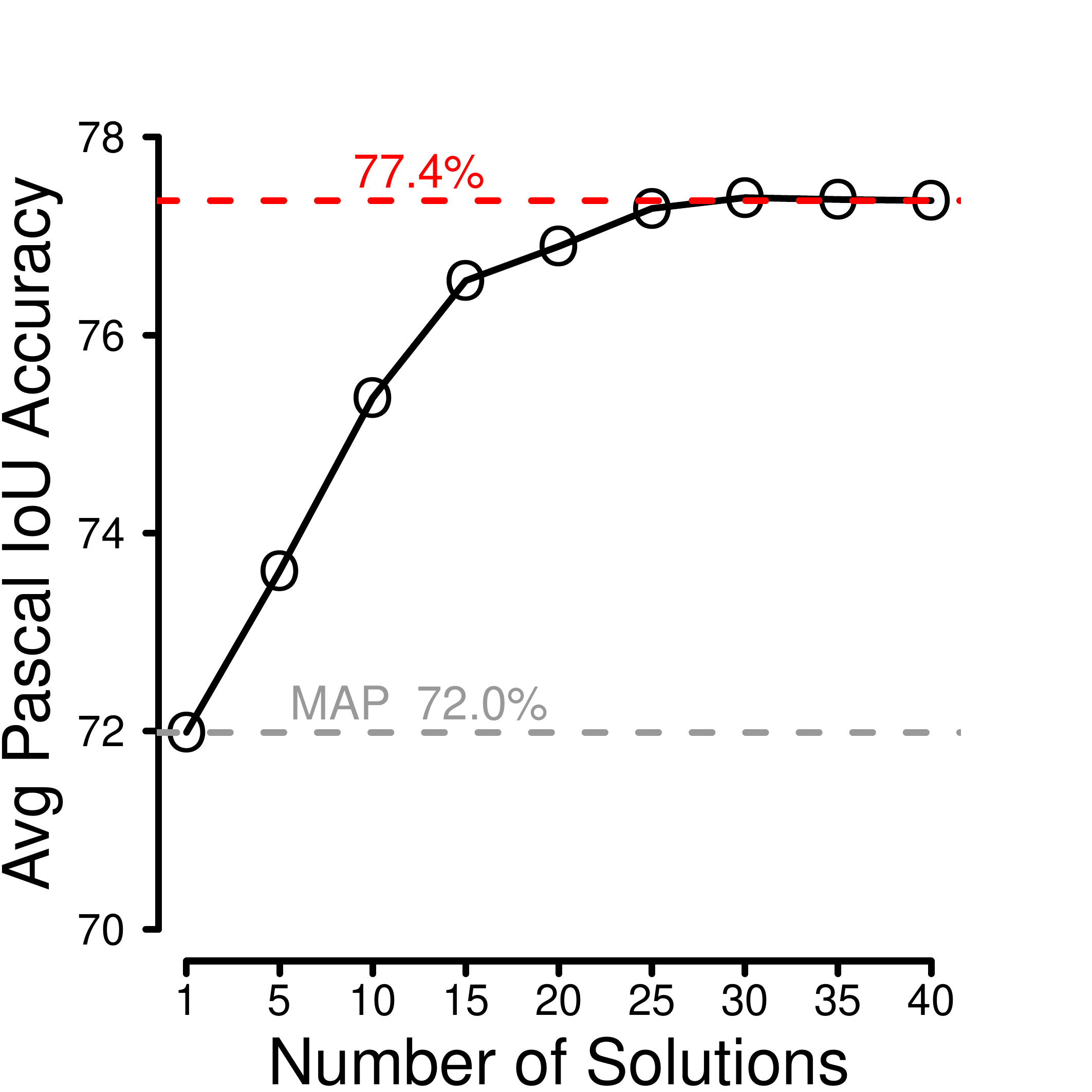}\label{fig:oraclecnn}}
        \subfloat[]
        {\includegraphics[width=0.5\textwidth]{\main/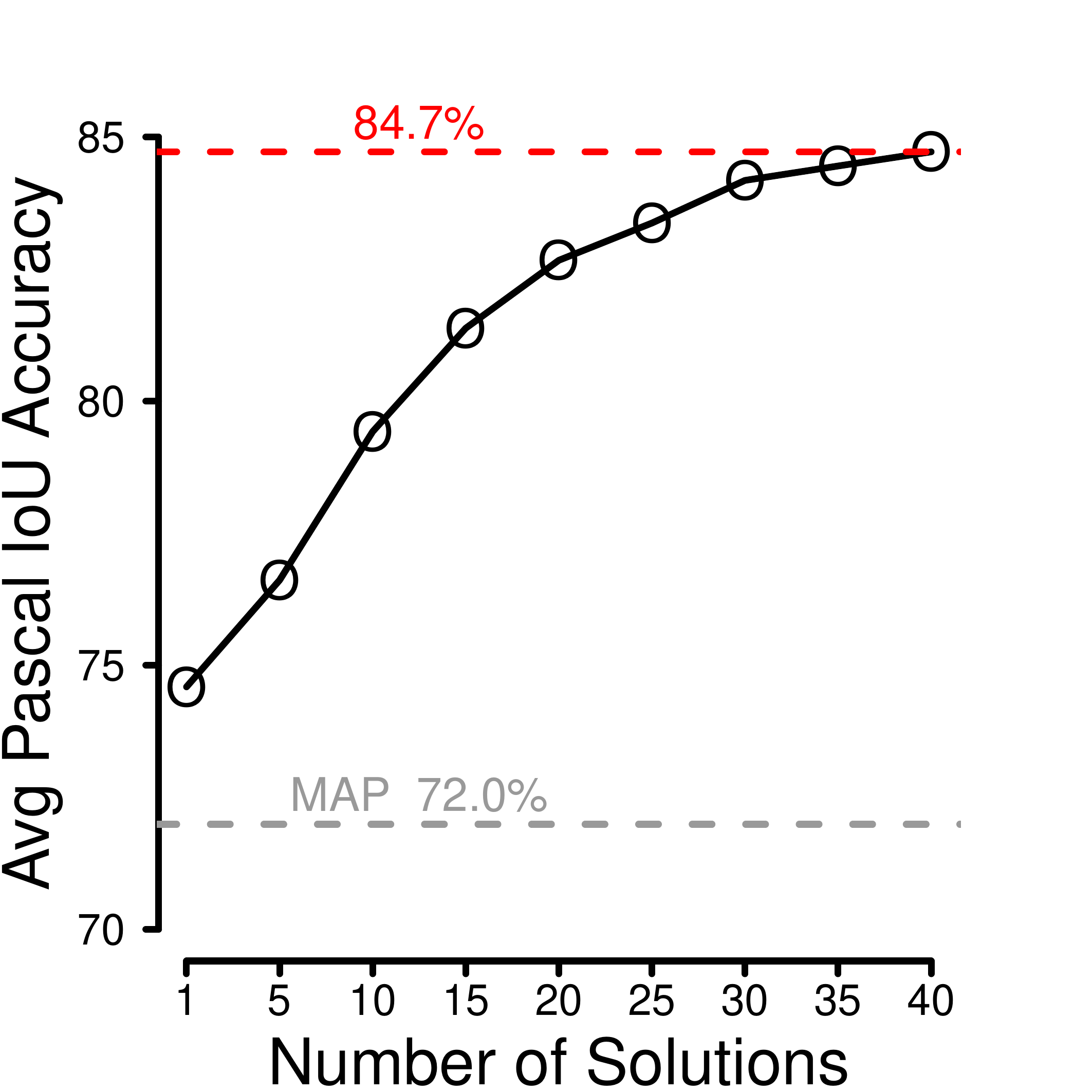}\label{fig:oraclecomp}}
        %\vspace{\captionReduceTop}
    \end{minipage}%
    \hspace{1em}\begin{minipage}[c]{.32\textwidth}
        \caption{\small{Oracle performance (IoU accuracy against ground-truth)
        on PASCAL VOC 2012 \texttt{val}, when (a) selecting best-out-of-$m$
solutions, and (b) composing full image labellings from connected components
found among $m$ solutions.}}
        \label{fig:cnnoracleacc}%\vspace{\captionReduceBot}\vspace{-3pt}
    \end{minipage}
\end{figure*}
\begin{figure*}[!h]
    \begin{minipage}[c]{\textwidth}
        \begin{tabular}{l}
            \includegraphics[width=0.15\textwidth]{\main/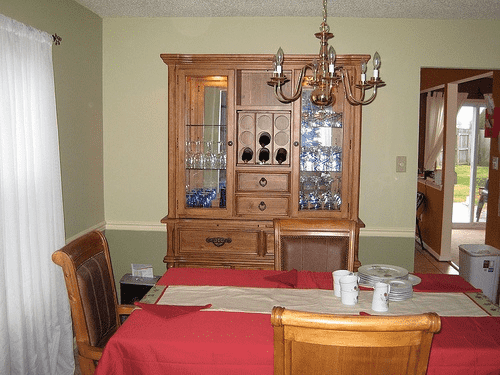}
            \includegraphics[width=0.15\textwidth]{\main/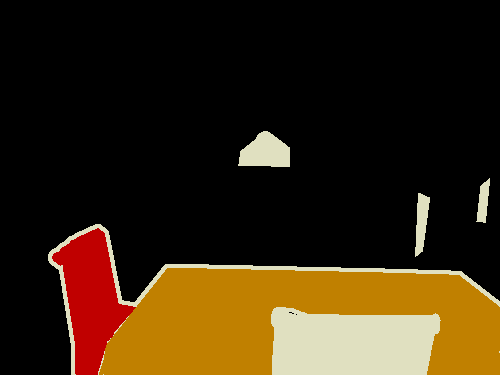}
            \includegraphics[width=0.15\textwidth]{\main/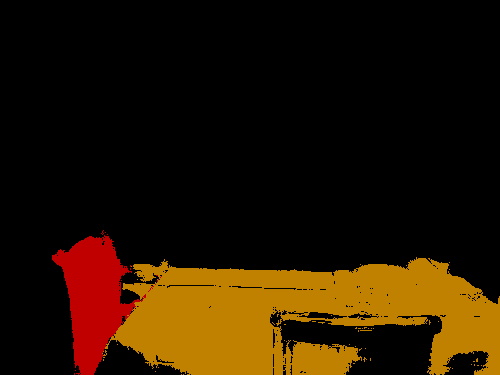}\\
            \includegraphics[width=0.15\textwidth]{\main/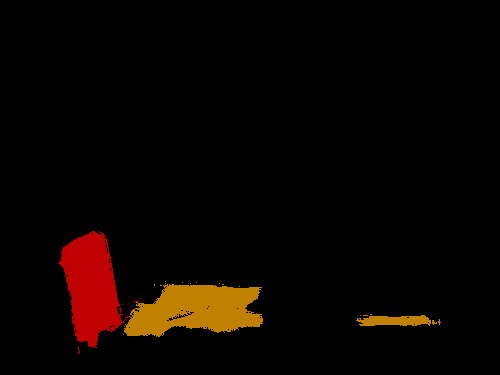}
            \includegraphics[width=0.15\textwidth]{\main/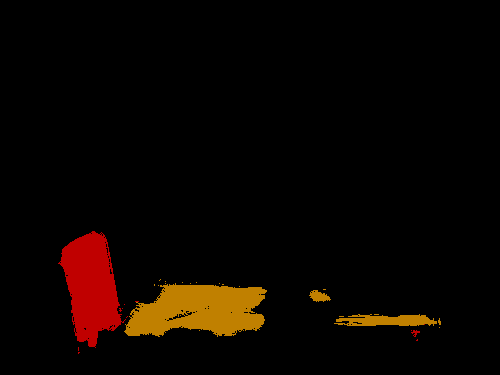}
            \includegraphics[width=0.15\textwidth]{\main/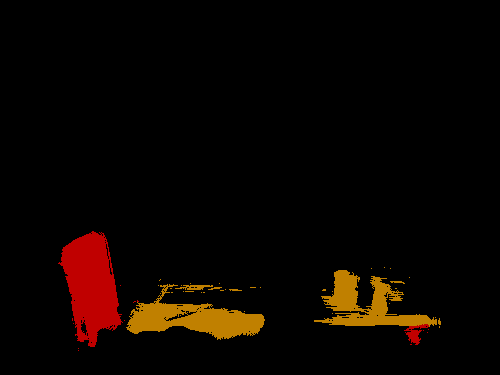}
            $\cdots$
            \includegraphics[width=0.15\textwidth]{\main/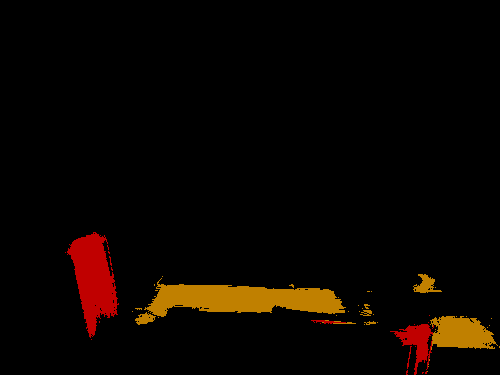}
            \includegraphics[width=0.15\textwidth]{\main/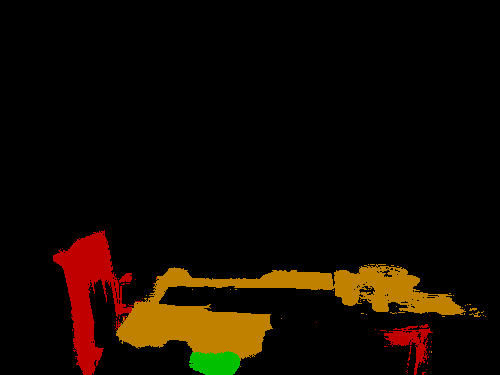}
            \includegraphics[width=0.15\textwidth]{\main/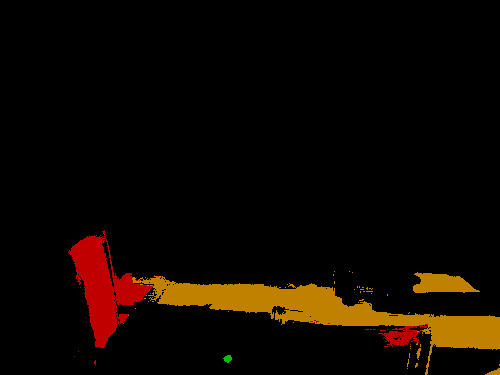}
            $\cdots$\\
            \includegraphics[width=0.15\textwidth]{\main/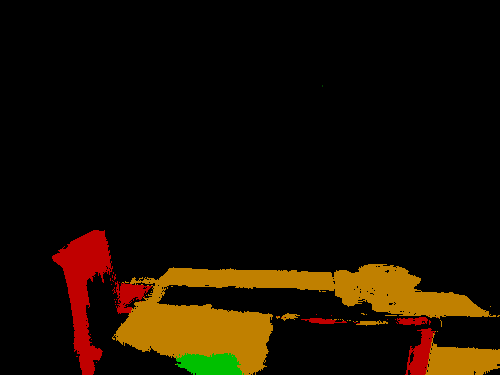}
            \includegraphics[width=0.15\textwidth]{\main/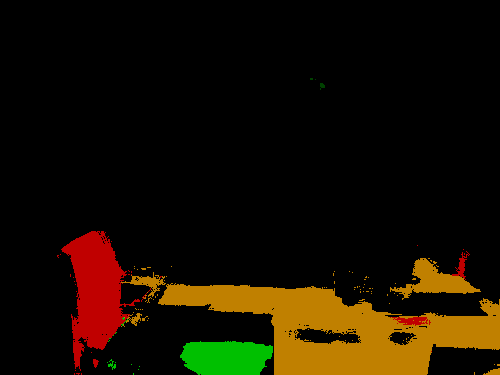}
            \includegraphics[width=0.15\textwidth]{\main/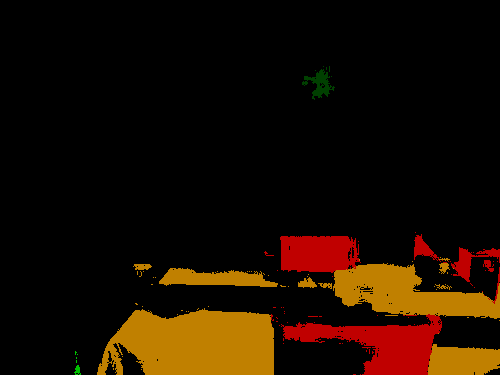}
            \includegraphics[width=0.15\textwidth]{\main/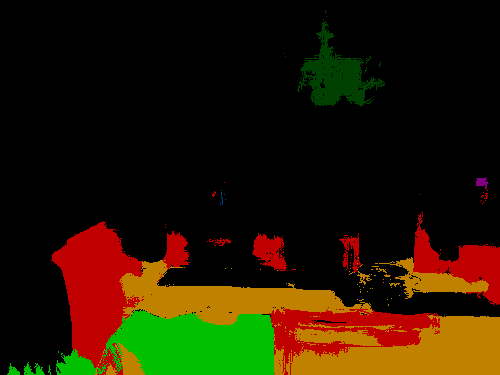}
            $\cdots$%\hspace{-2pt}%/comp
            \includegraphics[width=0.15\textwidth]{\main/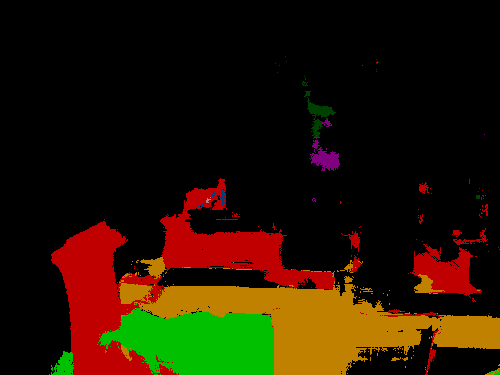}
            \includegraphics[width=0.15\textwidth]{\main/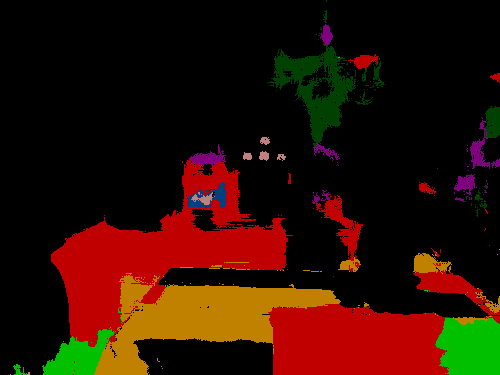}
        \end{tabular}
    \end{minipage}
    \hspace{1em}\begin{minipage}[c]{.32\textwidth}
        \caption{\small{Result of composing solutions from \divmbest segments. Second and third row show a subset of 40 \divmbest segmentations generated from the CNN+CRF model of~\cref{sec:cnn_divmbest}. First row shows in order the image, ground-truth segmentation, and composed segmentation oracle using the second approach of \cref{sec:compresults}. Note how the composed segmentation oracle is a much better segmentation of the image than the MAP solution (first segmentation in the second row).}}
        \label{fig:comp}.%\vspace{\captionReduceBot}\vspace{-3pt}
    \end{minipage}
\end{figure*}

%% file: Chapters/Chapter05.tex
\chapter{DivMBest+ReRank Experiments}\label{ch:divrankexp}
%************************************************

\subfile{\main/Sections/DivrankExp}

\printbibliography

%% file: Sections/DivrankExp.tex
\subfile{\main/Sections/DivRankExpOverview.tex}

\subfile{\main/Sections/DivRankFigureground.tex}

\subfile{\main/Sections/DivRankCategorySeg.tex}

%% file: Sections/DivRankExpOverview.tex
\section{Evaluating DivMBest+ReRerank pipeline}
To evaluate how well the proposed method in Chapter~\ref{ch:divrank} re-ranks the
\divmbest solutions the following experiments are carried out,
\begin{enumerate}
    \item Re-ranking of the object cut-outs generated from applying the \divmbest
        algorithm to the figure-ground model in~\cref{sec:divmbestfg}.
    \item Evaluate joint recognition and segmentation performance using the
        \divrank pipeline on two multi-category segmentation models: the \otwop
        model by Carreirra \etal~\cite{carreira2012o2p}, and the hierarchical \ALE model by
        Ladicky \etal~\cite{ladicky2009,ladicky2012}.
\end{enumerate}

%% file: Sections/DivRankFigureground.tex
\section{Figure-ground Segmentation}\label{sec:divrankfg}
Recall that in ~\cref{sec:divmbestfg} we presented a model for figure-ground segmentation and
evaluated \emph{oracle} accuracy of \divmbest solutions against MAP. We now
evaluate the proposed \divrank (cf. chapter~\ref{ch:divrank}) approach to segmentation where we
take the \divmbest segmentations from chapter~\ref{ch:divmbest} and rank them, returning the
highest ranking segmentation in the set as the final solution.

\subsection{Re-ranking segmentations}\footnotemark\footnotetext{The contributions to the thesis presented in this section are found in \protect\cite{yadollahpour2014}, and are in collaboration with Gregory Shakhnarovich.}
Using the notation from~\cref{sec:rerankernotation}, let the $i^{\text{th}}$ image be denoted as
$x_i$ and $\bY_i = \{\by_i^{(1)},\dots,\by_i^{(m)}\}$ be the set of predicted
foreground masks. A foreground mask, $\by_i$, is a labelling of all the $n$ (super)pixels
in the image where (super)pixel $i$ can take labels from the set $\{0,1\}$, \ie
$\by_i\in \{0,1\}^n$. Given a feature-function that computes a $p$-length feature vector on an
image/foreground mask pair, $\bpsi(x_i,\by^{(j)}_i)\;:\; \mathbb{R}^{3\times w \times
h} \times \{0,1\}^n \rightarrow \mathbb{R}^p$, we learn a linear re-ranker
model\sidenote{refer to \cref{sec:rerankermodel}},
\begin{flalign}
    S_r(\by_i) = \balpha^T \bpsi(x_i,\by^{(j)}_i).
\end{flalign}

\subsection{Ranking features}
There are a few considerations that guide the design of region ranking
features for our approach. These features  need to be evaluated only
on a small number of segmentations, hence we can afford fairly
complex/expensive computation. Furthermore, these features will be
deployed to evaluate entire hypothesized foreground masks,
assumed to include high-quality ones. Therefore we can use image
features that would be hard to incorporate into the bottom-up model,
in particular, shape properties of the mask and its position in the
image.
In addition, we can incorporate properties of the hypothesized regions
that are less meaningful for the small, regular superpixels; for
instance, measurements of homogeneity of the regions.
These considerations led us to design the following features.

\vspace{-1em}
\paragraph{Shape and position} We extract the following properties of
the hypothesized foreground mask: area; perimeter; location of
centroid; minor/major axis length, eccentricity and orientation for an ellipse fit
to the region; area of the convex hull of the region and its ratio to
the area; Euler number (number of holes); diameter of the disk with
area equal to that of the region; extent (fraction of the bounding box
occupied by the mask); and finally size and location of the bounding
box.
All of
these quantities are normalized per image (e.g., area is expressed in
percentage of image area).

\vspace{-1em}
\paragraph{Color} We compute histogram with 32 fixed bins for each
color channel; we do not use adaptive binning as in the bottom-up
model, since color distribution for entire objects is more stable than
for small parts (superpixels).

\vspace{-1em}
\paragraph{Texture} We compute histogram of assignment to 32 textons~\cite{shotton2006},
computed \emph{for the entire training set} and not per image as in
the bottom-up model.

\vspace{-1em}
\paragraph{Entropy} For each histogram feature (color and texton) we
compute its entropy. This measures the homogeneity of hypothesized
object.

\subsection{Re-ranker training}
For each of the five experiments re-ranking parameters were learned on the
training set of each dataset. The parameters were tuned by cross-validation on
the training set.

\subsection{Results}
Table~\ref{tab:fgreranking} summarizes the results of re-ranking the \divmbest solutions. In
four out of five benchmarks region ranking closes the gap between \emph{oracle}
and MAP performance by up to or more than 50\%. On Graz \texttt{people} $\sim
30\%$ of the gap is closed.

\subfile{\main/data/figground/reranking}

In all the datasets the \divrank approach to figure-ground segmentation achieves
accuracy equal to or better than~\cite{kuettel2012}.

\subfile{\main/data/figground/rerankingfeats}

The contribution of each of the re-ranking features to the re-ranker performance
is summarized in Table~\ref{tab:fgrerankfeats}. All the features contribute to the overall
performance of the re-ranker across all datasets, however the amount of
contribution per feature differs between datasets. For example, removing color
features from the re-ranker drastically reduces performance on the
Horses dataset, probably because the color distribution of background --- which
is often grass or foliage --- is very different from foreground in these images.
On Ultrasound dataset the color intensity is important because lesions usually
appear dark on the ultrasound but the importance of color features is less
important than on Horses.

%% file: data/figground/reranking.tex
\begin{table}[!t]
  \centering
  \begin{tabular}{|l|c|c|c|c|c|}
\hline
&\cite{kuettel2012}& MAP & \emph{Oracle} & Ranking & \%gap\\
\hline
weizman horses&79.1  & 75.4 & 83.0& 79.3 & 51.3\%\\
\hline
graz bikes &45.0&53.1 & 61.4& 56.1 & 36.1\%\\
\hline
graz cars &58.8 &50.0 & 66.3& 59.7 &  59.5\%\\
\hline
graz people &47.5& 44.1& 57.0& 47.5 & 26.4\%\\
\hline
ultrasound &26.6&39.7 & 57.3& 49.9 & 58.0\%\\
\hline
  \end{tabular}
  \caption{Segmentation performance on all data sets, in IoU values $\times$100. MAP: single solution from
      the bottom-up CRF model. \emph{Oracle}: (hindsight) best of 10 diverse solutions from
    the CRF. Ranking: full ranking model (all
    features). last column: percentage of gap (oracle-map) recovered
    by the ranking.}
  \label{tab:fgreranking}
\end{table}

%% file: data/figground/rerankingfeats.tex
\begin{table}[!t]
  \centering
  \begin{tabular}{|l|c|c|c|c|c|}
\hline
   Method & horses & bikes & cars & people & ultrasound\\
\hline
\cite{kuettel2012} & 79.1 &45.0 &  58.8 & {\textbf 47.5} & 26.6 \\
\hline
ours(full) & {\textbf 79.3} & {\textbf 55.4} & {\textbf 59.7} & {\textbf 47.5} & {\textbf 49.9} \\
\hline
shape& 73.7 & 54.8 & 58.6& 44.7 &  48.2\\
textons &74.0  &53.4 & 54.3 & 45.6 & 48.1\\
color & 69.5& 53.2& 53.1& 43.5 & 44.1\\
full-entropy  & 76.9 & 54.4 & 57.6& 44.7& 47.3\\
\hline 
  \end{tabular}
  \caption{Comparative results between methods and feature sets for
    region ranking. All numbers are IoU$\times100$. Shape: only shape
    and position. Textons: only textons. Color: only color
    histograms. full-entropy: shape, color and textons, but not their entropies.}
  \label{tab:fgrerankfeats}
\end{table}

%% file: Sections/DivRankCategorySeg.tex
\section{Multi-category segmentation}\footnotemark\footnotetext{Part of the contributions to the thesis presented in this section are found in \protect\cite{batra2012}, and are in collaboration with Gregory Shakhnarovich and Dhruv Batra.}
We revisit the two multi-category segmentation models that we introduced in the
\divmbest experiments (cf.~\cref{sec:divmbestmulticat}), and evaluate segmentation performance of
these models when predicting the best solution in the \divmbest set for each
image.  To do this we use the \divrank pipeline introduced in~\cref{sec:rerankermodel}.

\subsection{Dataset}
Evaluation of the \divrank pipeline applied to the models below is carried out
on the PASCAL VOC 2012 dataset~\cite{everingham2010}, which contains the same 20 categories as VOC
2010 but with additional images for each category. There's a total of
4,369 images, split into \train (1,464), \val (1,449), and \test (1,456 images) subsets.

\subsection{Hierarchical model}
We reviewed the Associative Hierarchical CRF of Ladicky \etal~\cite{ladicky2009}
in~\cref{sec:segmethods} and corresponding \ALE implementation, and showed that the
\emph{oracle} performance on \divmbest
segmentations improved by more than 20\% over MAP on the \val subset of PASCAL VOC
2010. To evaluate how good the \divrank implementation is at returning a high
quality segmentation of the image we produce diverse segmentations and rank them
on images from the PASCAL VOC 2012 dataset.

\subsection{Feed-forward model}
We also compare against the Second-Order Pooling (\otwop) implementation of Carreira \etal~\cite{carreira2012o2p} contrast with
the Hierarchical CRF model above, in the way higher-order dependencies are
incorporated in the model. Whereas is \ALE the higher-order interactions are due
to the hierarchical structure of the graph and high-order cliques,  \otwop
incorporates high-order dependencies using second-order pooling of local
descriptors over regions in the image.

In \otwop, for each image location local descriptors such as SIFT and local binary
patterns (LBP)~\cite{ojala1996,ojala2002} as well as color and location are densely computed. Given an
initial set of $\sim 150$ candidate figure-ground masks for the image ---
produced using the bottom-up CMPC segmentation algorithm~\cite{carreira2010} --- the
second-order statistics (\ie vector outer-products) of local
descriptors that fall within each region are pooled (\eg average/max) to give
\emph{global} features capturing higher-order interactions between image
elements. These \emph{region-level} features are fed to support-vector regressors (SVR) for each
category, that are trained to predict how well the region overlaps objects of
that category.

The implementation uses a simple and efficient greedy inference strategy to produce the final multi-category
segmentation. Starting with an initial background threshold, in decreasing
order, the segment and
category with highest score above the threshold is pasted in the image. Segments
with higher score are laid on top of segments with lower score. Each time a
segment is pasted the background threshold is increased, and the process stops
when there are no more segments with category score above the threshold. The
initial background threshold is set such that the average number of segments
with score above the threshold is roughly the same as the number of objects per
image in the training set. Note that we can reformulate this approach as a CRF
constructed on overlapping CPMC segments in the image with unary potentials
defined by the SVR category scores of each segment.

Through the use of a number of tricks to speedup computation, like caching
pooling results and dimensionality reduction on features (cf.~\cite{carreira2012o2p}), the \otwop
implementation is faster to train and run inference over than the \ALE model.

\subsection{Diversity and Oracles}

For the analysis reported in this subsection, we used the VOC 2012
\train and \val sets. \ALE and \otwop models were trained
on VOC 2012 \train,
and the models were used to produce 10 segmentations for each image
in \val. The Lagrangian multiplier in the \divmbest formulation (cf.~\cref{sec:divmbest}) was tuned via cross-val ($\lambda_{ALE}=1.25$ and
$\lambda_{O_2P}=0.08$).
%\gscomment{are both set by cv on train?}

\iffalse
\begin{figure*}[t!]\centering
  \begin{tabular}{ccc}
    \subfloat[]{\includegraphics[width=0.3\linewidth]{../figures/solmincover.pdf}\label{fig:sol-div:a}} &
    \subfloat[]{\includegraphics[width=0.3\linewidth]{../figures/oracle_labelinMAP.pdf}\label{fig:sol-div:b}} &
    \subfloat[]{\includegraphics[width=0.3\linewidth]{../figures/oracle_maskinMAP.pdf}\label{fig:sol-div:c}}
  \end{tabular}
  \vspace{\captionReduceTop}
  \caption{\small{(a)~Average minimum-covering~\eqref{eq:mincover} of MAP in the first $M$ solutions vs. $M$.
  (b) Accuracy of an oracle restricted to labels present in the MAP, or (c) restricted to masks present in MAP. See text for details.}}
  \label{fig:sol-div}\vspace{\captionReduceBot}\vspace{-3pt}
\end{figure*}
\fi
\subfile{\main/plots/divrank/rankeracc}
\textbf{Oracle Accuracies.}
Since ground-truth is known for VOC \val images, we
can find the \emph{oracle} accuracy, \ie the accuracy of the best solution in
the set, as described in~\cref{sec:oracledef}.
This accuracy is shown in figure~\ref{fig:rankeracc} (lines with circles):
with $10$ solutions on \otwop, it reaches 60.12\%, which is 15\%-points higher
the accuracy of MAP.
\emph{Oracle} accuracy with \ALE solutions show a similar increase \wrt \ALE's MAP.

To put these \emph{oracle} numbers in context, we can try to find what is the best segmentation accuracy achievable using the 150 CPMC segments for each image. To find a good approximation of the best segmentation we can achieve,
 we can consider a greedy algorithm that tries to find the subset of CPMC segments that best cover
ground-truth segments and then simply copies labels over from the ground-truth. This achieves
an accuracy of 80.78\%. Notice that this procedure takes the supremum of accuracy of \emph{exponentially} many
solutions, whereas \divmbest with 10 solutions reaches $60.12\%$, closing the
gap to within 21\% points.

\textbf{Diversity of solutions.} We now turn to empirical analysis that quantifies the
amount of diversity in these solutions, and how that affects the oracle performance.

\emph{The first question to address is: how much diversity do the
\divmbest solutions contain over MAP?} To answer this, we can look at the
solution in the set that is most different from MAP, as measured by average
region overlap.

Let $\{s^{(m)}_{i,1}, \ldots,s^{(m)}_{i,K}\}$ denote the set of $K$ segments in the $m^{th}$ solution
for image $i$ and $\{s^{(1)}_{i,1}, \ldots,s^{(1)}_{i,K'}\}$ denote the set of segments in MAP.
We can define a category-independent
\emph{covering measure}, which for a given image $i$ captures how much of the
MAP segmentation is covered by one of the subsequent solutions,
\vspace{-5pt}
{\small
\begin{align}
D_1(\by^{(j)}_i) = \frac{1}{\sum\limits_{k'} |s^{(1)}_{i,k'}|} \sum\limits_{k'=1}^{K'} |s^{(1)}_{i,k'}| \max\limits_{k\in[K]}
				O(s^{(1)}_{i,k'},s^{(j)}_{i,k}),\label{eq:D1}
\end{align}
}
where $|s^{(m)}_{i,k}|$ denotes the size of the segment and $O(\cdot,\cdot)$ is the intersection-over-union
measure of the two segments.
For \otwop these segments correspond to CPMC segments \cite{carreira2010}, while in \ALE the
segments are connected components in the segmentation.

To get an idea of how different the \emph{most} diverse solution is, we can define
the minimum cover of the MAP solution by the $M$ segmentations for image $i$ as:
\vspace{-5pt}
{\small
\begin{align}
D_1^{(i,m)}\doteq \min\limits_{j=1,\dots,m} \{D_1(\by_i^{(j)})\}. \label{eq:mincover}
\end{align}
}
A plot of average minimum diversity in the dataset, \ie
$\sum_i D_1^{(i,m)}/n$ for $m=1,\ldots,10$ is shown in figure~\ref{fig:sol-div:a}.
We can see that both models produce at least one solution that is significantly different from the MAP.
With 10 solutions, the minimum covering of MAP drops to about 0.3
for \otwop and 0.1 for \ALE. Thus, on average at least one out of 10 \divmbest
solutions for \otwop overlaps MAP by only 10\%.

\textbf{Diversity of Oracle.}
Of course, diversity is useful only if it brings improved quality.
The previous measure simply captures diversity and can be easily affected by poor quality solutions that are
different from MAP.
We can also try to characterize the diversity in the oracle solutions.
This measure tells us how different the oracle solution is from the MAP solution on average.
Analogous to eqn.~\ref{eq:D1} we can compute $D_1(\by^{(*)}_i)$ to measure by how much the segments in the
oracle segmentation cover the MAP segments.
% irrespective of their labels.
We can also use a category-specific covering measure which takes into account label agreement to get a measure of how much the MAP segments are covered with same labelled segments in the oracle solution,
\vspace{-5pt}
{\small
\begin{align}
    D_2(\by^{(*)}_i) = \frac{1}{\sum\limits_{k'} |s^{(1)}_{i,k'}|} \sum\limits_{k'=1}^{K'}
|s^{(1)}_{i,k'}| \max\limits_{{k\in[K]:\atop y^{(*)}_{i,k}=y^{(1)}_{i,k'}}}
				O(s^{(1)}_{i,k'},s^{*}_{i,k}),
\end{align}
}
where $y^{(*)}_{i,k}$ and $y^{(1)}_{i,k'}$ are the labels of the oracle and MAP
segments respectively. Table~\ref{table:oraclediv} summarizes these results which show that the oracle segmentations are not simply minor perturbations of the MAP segmentations.
\begin{table}[t]
{\footnotesize
\setlength{\tabcolsep}{1pt}
\begin{tabular*}{\columnwidth}{@{\extracolsep{\fill}}p{1cm}ccc}
%\toprule
&
\textbf{$\mathbf{D_1}$ Oracle Covering} &
\textbf{$\mathbf{D_2}$ Oracle Covering}\\
\midrule
\textbf{ALE} & 0.55 & 0.45\\
\botwop & 0.61 & 0.58\\
\bottomrule
\end{tabular*}
}
\caption{\small{Average covering score between oracle solutions and MAP: (\emph{left}) show
the category-independent measure and (\emph{right}) shows the category-specific measure.}}
\label{table:oraclediv}
\end{table}
On average the MAP covering by oracle irrespective of segment label is less
than 61\% for \otwop and 55\% for \ALE. If we constrain the covering
to be category-consistent, these numbers drop to 58\% and 45\%
respectively. Thus, we can conclude that the
oracle segmentations are not simply minor perturbations of the MAP.

\textbf{Gain from diversity.}

The previous measure tells us that the oracle solution is indeed quite different from the MAP. We now try to
study \emph{how} it is different -- do the additional solutions introduce new categories or new masks or both?
In order to answer this question, we measure the performance of a restricted
\emph{oracle} that
chooses in each additional solution the best label possible for all segments, albeit restricted to the
set of labels found in MAP.
Specifically, if a segment $s^{(j)}_{i,k}$
overlaps with the ground-truth background by more than $50\%$, then we set its label to background.
Otherwise if there is a segment $g_{i,l}\in\by^{gt}_i$, where $y(g_{i,l})\in y_{MAP}$
(where $y_{MAP}\doteq \{y(s^{(1)}_{i,k})|k\in[K]\}$), with $s^{(j)}_{i,k}\cap
g_{i,l}\neq\emptyset$,
we set $y(s^{(j)}_{i,k})=y(g_{i,l})$. If there is no such $g_{i,l}$ then
$y(s^{(j)}_{i,k})$ is assigned a
random label from $y_{MAP}$.
Figure~\ref{fig:sol-div:b} shows that such a restricted oracle (\texttt{\small \otwop-oracle-label} and \texttt{\small \ALE-oracle-label})
performs worse than the unrestricted oracle, indicating that the additional solutions do in fact introduce
categories present in ground-truth but not in MAP.

\subfile{\main/plots/divmbest/soldiv}

Similarly, we can restrict the
oracle to only take segment masks found in MAP and
assign to them the best possible labels found in $\by^{(m)}_i$.
Again figure~\ref{fig:sol-div:c} shows that such a restricted oracle significantly under-performs, indicating the
MAP masks are not ideal and that the additional solutions do in fact introduce useful masks.

Thus, we can conclude that there are clear differences in both the labels and segments of the oracle
segmentations compared to the MAP.

\subsection{\Reranker features}
Our \reranker uses a number of features that we separate into a few groups.
In the discussion, below we
say a label $c$ is present in \soln if at least one pixel in \soln is labeled $c$.

\textbf{Model features} rely on
properties derived from the model that produced segmentation \soln --
model score of \soln,
average pixel score, number of CPMC masks used to construct
foreground, the final background threshold at the end of the greedy
foreground assembly, and the rank of \soln among the $m$
diverse hypotheses for the given input image. (5 dimensions)

\textbf{Diversity features} measure average per pixel
agreement of \soln with the majority vote
by the diverse set (weighted or unweighted by the model scores). (2 dimensions)

\textbf{Recognition features.} We
use outputs of object detectors from~\cite{mottaghi2012} to get detector-based segmentations
$\mathcal{D}_1, \mathcal{D}_2$,
where each pixel is assigned by majority vote on detection scores (thresholded \& un-thresholded).
Then we compute the agreement matrix: for every $c_1,c_2$ we count pixels assigned to
$c_1$ by \soln and to $c_2$ by $\mathcal{D}_1$, yielding
a 441-dimensional feature. We compute max/median/min of
the detection score (with and without thresholding) for every category in \soln (120 dims);
the average overlap between category masks in $\mathcal{D}_1, \mathcal{D}_2$ and in \soln (2 dims);
and pixelwise average detector scores for categories in \soln (2 dims).
We also use the
the estimated posterior for each category present in \soln,
using the classifier from~\cite{uijlings2011} (20 dimensions).

\textbf{Segment features} measure
the geometric properties of the segments in \soln: perimeter, area,
and the ratio of the two; computed separately for segments in every
class and for the entire foreground (63 dimensions).
Relative location of the centroids of masks for each category pair (420 dimensions).
\subfile{\main/data/divrank/testaccuracies}

\textbf{Label features} rely
on information regarding the labels assigned to masks in \soln,
but not the geometry of these masks. For every pair of
labels $c_1,c_2$ we compute the binary co-occurrence (1 if both
categories are present in \soln) and the percentage
of pixels assigned to $c_1$ \& $c_2$. (420 dimensions)

All the features above are independent of the image \inim; the
following features rely on image measurements as well as properties of the solution \soln.

\textbf{Boundary features.}
We compute the
total globalPb probability of boundary response~\cite{arbelaez2011}
in a band along the category boundaries;
for 3 widths of the band, this produces
a 3-dimensional feature (with 3 more for normalized versions).
We also compute
recall by the globalPb map of the category boundaries
in the \soln; this produces a 10 dimensional feature for ten equally spaced
precision values. Finally, we compute the histogram (6 bins) of
Chamfer distance between the boundaries in \soln and the thresholded
globalPb, and vice versa; with 10 thresholds this produces
a 120 dimensional feature.
For each category, we also computed normalized histogram of globalPb responses in the non-boundary regions (210 dims).

\textbf{Entropy features.} For every category (and the combined
foreground) we measure the entropy of color histograms, computed per
color channel with two binning resolutions, yielding 126
dimensions. We do the same for textons~\cite{shotton2006,shotton2009}, with a single
binning, for another 21 features.

We stress that most of these features rely on higher-order information that would be
intractable to incorporate into the CRF model used in stage 1.
For instance, using features that refer to segment boundaries is hard in
CRF. However, evaluating these features on $m$
segmentations is easy, which allows us to use them at the \reranking stage.

\subsection{Re-ranker training}

The combined feature vector per solution \soln has 1988
dimensions. The only hyper-parameter for the \reranker is the regularization
parameter $C$~\eqref{prob:ssvm}, which is chosen via
cross-validation on the \texttt{val} set\sidenote{We also used
  cross-validation to evaluate the feature set, rejecting some
  additional features not listed here that did not contribute to
  \reranking accuracy.}.

One important practical
question is how many diverse solutions to use. While we have seen
above that the oracle accuracy increases through $M=30$ solutions, it
is possible that too many solutions make it hard to train an effective
\reranker. Indeed, we found that the best results in cross-validation
are obtained when training on 10 solutions per image; we use the same
number of diverse solutions per image when \reranking the test segmentations.

\subsection{Re-ranker results}
The performance of the \divrank pipeline on the \otwop and \ALE models
(\texttt{Rerank}), as the
size of the \divmbest set grows, is reported in figure~\ref{fig:rankeracc} along
with MAP and \emph{oracle} accuracies. The plot also shows results of a binary
classifier baseline (\texttt{Classifier}) that is trained to discriminate
between the best and worst segmentations in the set, and used at test time to
re-rank according to classification score. As a second baseline, we compare
against randomly picking one out of the $j\leq m$ segmentations (\texttt{Rand}).

On PASCAL VOC 2012 \val, the MAP segmentation IoU accuracy is 24.3\% on \ALE and 45.1\% on
\otwop. In contrast \divrank achieves 29.27\% on ALE and 48.2\% on \otwop, an
increase of $>5\%$ and $>3\%$-points respectively.
Table~\ref{table:rerankertest_accuracies} shows VOC 2012 \test set performance
of the \otwop ranker when trained on the \val set. \otwop--\divrank achieves a
1.6\%-point performance improvement over \otwop--MAP\sidenote{this was
state-of-the-art results on PASCAL VOC 2012 \texttt{comp6} challenge at time of
experiments}. A few examples where \otwop--\divrank beats \otwop--MAP are shown
in figure~\ref{fig:rankerbtmap}. More examples can be found in Appendix~\ref{sec:rankerbtmap}.

\subsection{Re-ranker Analysis}
We can consider how the re-ranker behaves in picking a solution from the
\divmbest set. In the original ranking of \divmbest solutions (\ie order in
which they were generated) figure~\ref{rankerstats_oraclehist} shows the number
of images in which the \emph{oracle} solution is at rank $j$, for
$j\in\{1,\dots,10\}$. The \emph{oracle} distribution has a heavy tail, indicating
that high-quality solutions are often found at the bottom of the list.
Figure~\ref{rankerstats_rerankhist} shows the number of images where the top
re-ranked solution was originally at rank $j$. The re-ranked distribution has a
much lighter tail which suggest that the re-ranker "plays it safe" and often
predicts MAP. The correlation between segmentation quality and re-ranker score
is shown in figure~\ref{rankerstats_corr}, which indicates that the re-ranker
score is well correlated with solution quality.
\subfile{\main/plots/divrank/rankerstats}

\subfile{\main/gfx/divrank/rankerbtmap/rankerbtmap}
\subsection{Human Ranking Experiments}
\subfile{\main/gfx/humanexp/humanexp}

\subfile{\main/data/humanexp/humanexp}
To evaluate and characterize the difficulty of the re-ranking problem we can
investigate how well people perform the task of picking a good segmentation ---
which symbolizes the gold standard. 150 images were chosen from PASCAL VOC 2012
\val set where the MAP segmentation was neither the worst nor best segmentation.
On Amazon Mechanical Turk (AMT), subjects were presented with three different
types of binary comparison tasks for each image: comparing \emph{Best}-vs-MAP,
\emph{Best}-vs-\emph{Worst}, and MAP-vs-\emph{Worst }segmentations in the \divmbest set. The subjects
had to make the choice using only the labellings (with category names annotated)
and \emph{not} the image. The subjects were also presented with the option to
provide feedback on reasons for their choice. Figure~\ref{fig:humanexp} shows
the interface with actual examples of results from AMT workers. The workers'
comments illustrate that people are very good at discriminating good versus bad
segmentations using cues such as category co-occurrence
(figure~\ref{fig:humanexp:a}), category shape (figure~\ref{fig:humanexp:b}), and
part-vs-whole relashionships (figure~\ref{fig:humanexp:c}). These cues provide
evidence for our choice of re-ranker features. A summary of how well subjects
did on the three tasks is shown in Table~\ref{table:humanexp}. The most
difficult binary task for subjects was choosing between \emph{Best} and MAP
segmentations. In the case of the \otwop model picking between MAP and
\emph{Best} is even more difficult that for \ALE because the MAP solutions are
better for the \otwop model. Note that the segmentations picked by humans (HR)
achieve substantial improvement over MAP, which is significant given that the
choice is made without seeing the original image.
\subsection{Summary}
The analysis in this section on a number of segmentation tasks shows that the
set of solutions obtained from the \divmbest stage are of significantly higher
quality than the MAP solutions. The source of diversity between solutions is
also non-trivial. Re-ranking the \divmbest set results in significant
performance
improvement over MAP. The results also highlight the importance of choosing
re-ranking features that can discriminate good versus bad solutions within the
\divmbest set. Learning rich features for the within set discrimination task is
an area for future research.

%% file: plots/divrank/rankeracc.tex
\begin{figure*}[!p]
%\begin{minipage}[c]{.65\textwidth}
\subfloat[]
	{\includegraphics[width=0.7\linewidth]{\main/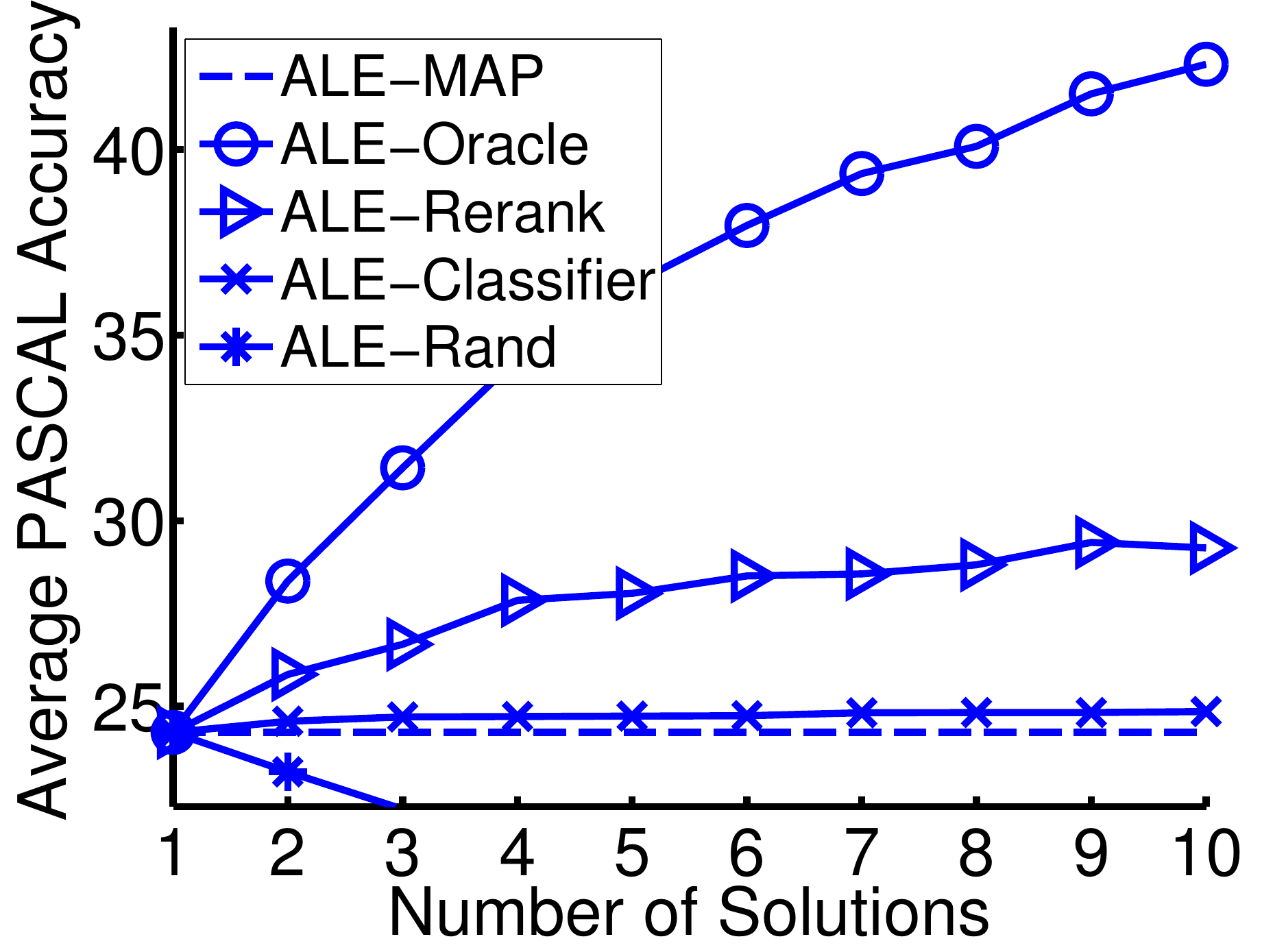}\label{fig:rankeracc:o2p}}
    \vspace{3em}
\subfloat[]
	{\includegraphics[width=0.7\linewidth]{\main/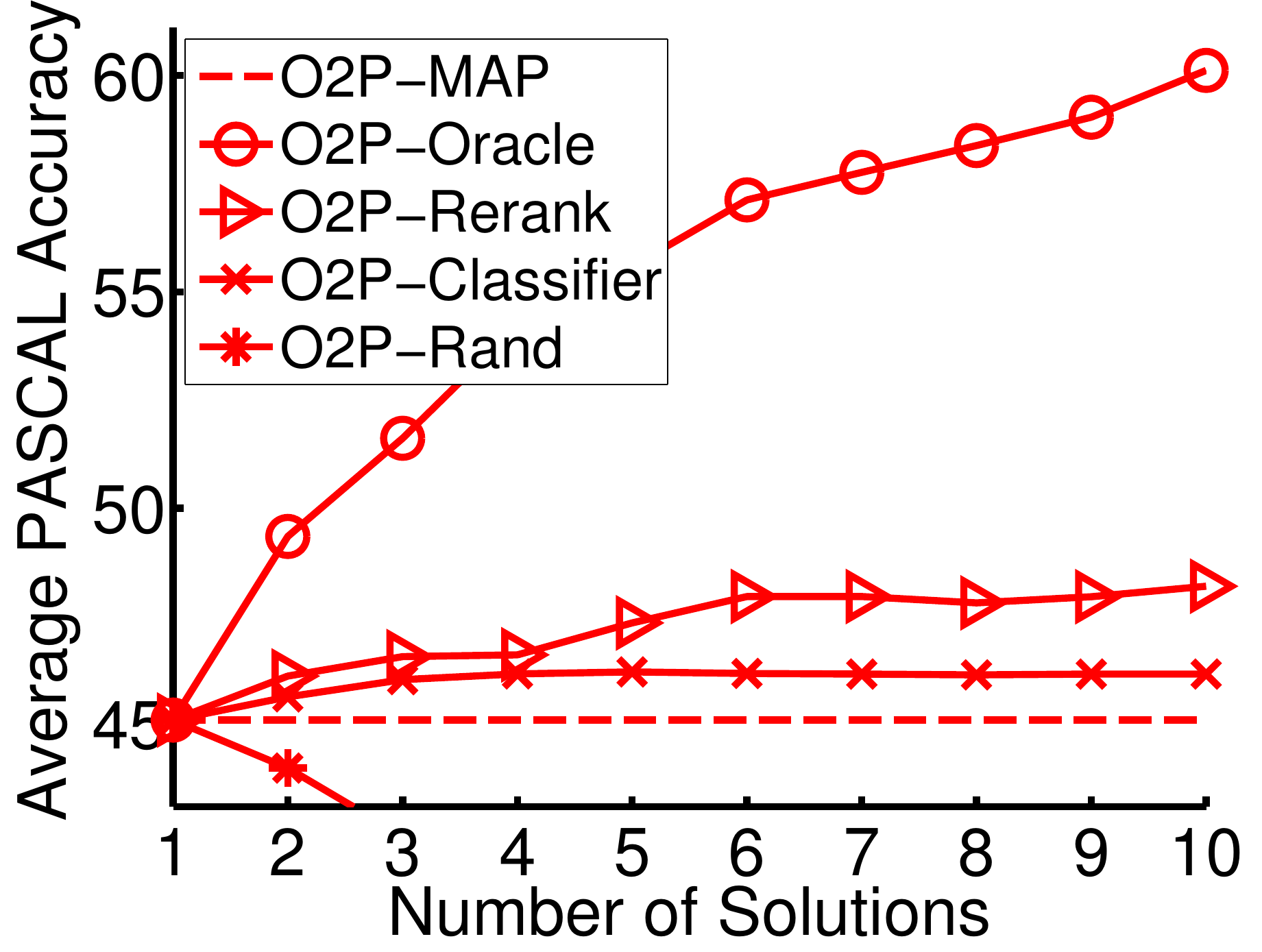}\label{fig:rankeracc:ale}} 
%\vspace{\captionReduceTop}
%\end{minipage}%
%\hspace{1em}\begin{minipage}[c]{.32\textwidth}
\caption{\small{\divrank performance on PASCAL VOC 2012 \val using (a) \ALE and (b) \otwop models 
vs.~the number of solutions.}
}
  \label{fig:rankeracc}%\vspace{\captionReduceBot}\vspace{-3pt}
%\end{minipage}
\end{figure*}

%% file: plots/divmbest/soldiv.tex
\begin{figure*}[p!]\centering
%  \begin{tabular}{ccc}
    \subfloat[]{\includegraphics[width=0.5\linewidth]{\main/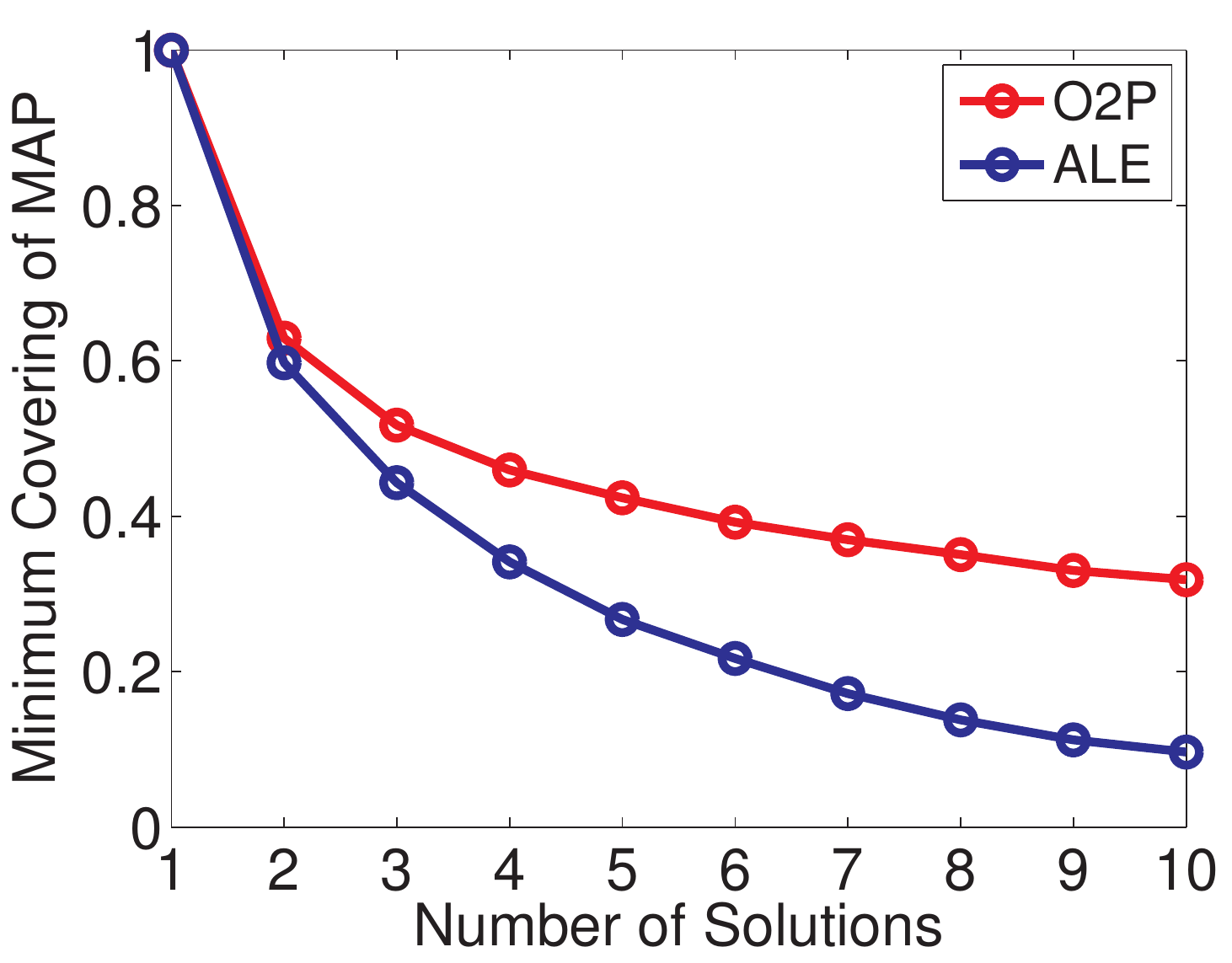}\label{fig:sol-div:a}}%&
    \hspace{1em}
    \subfloat[]{\includegraphics[width=0.5\linewidth]{\main/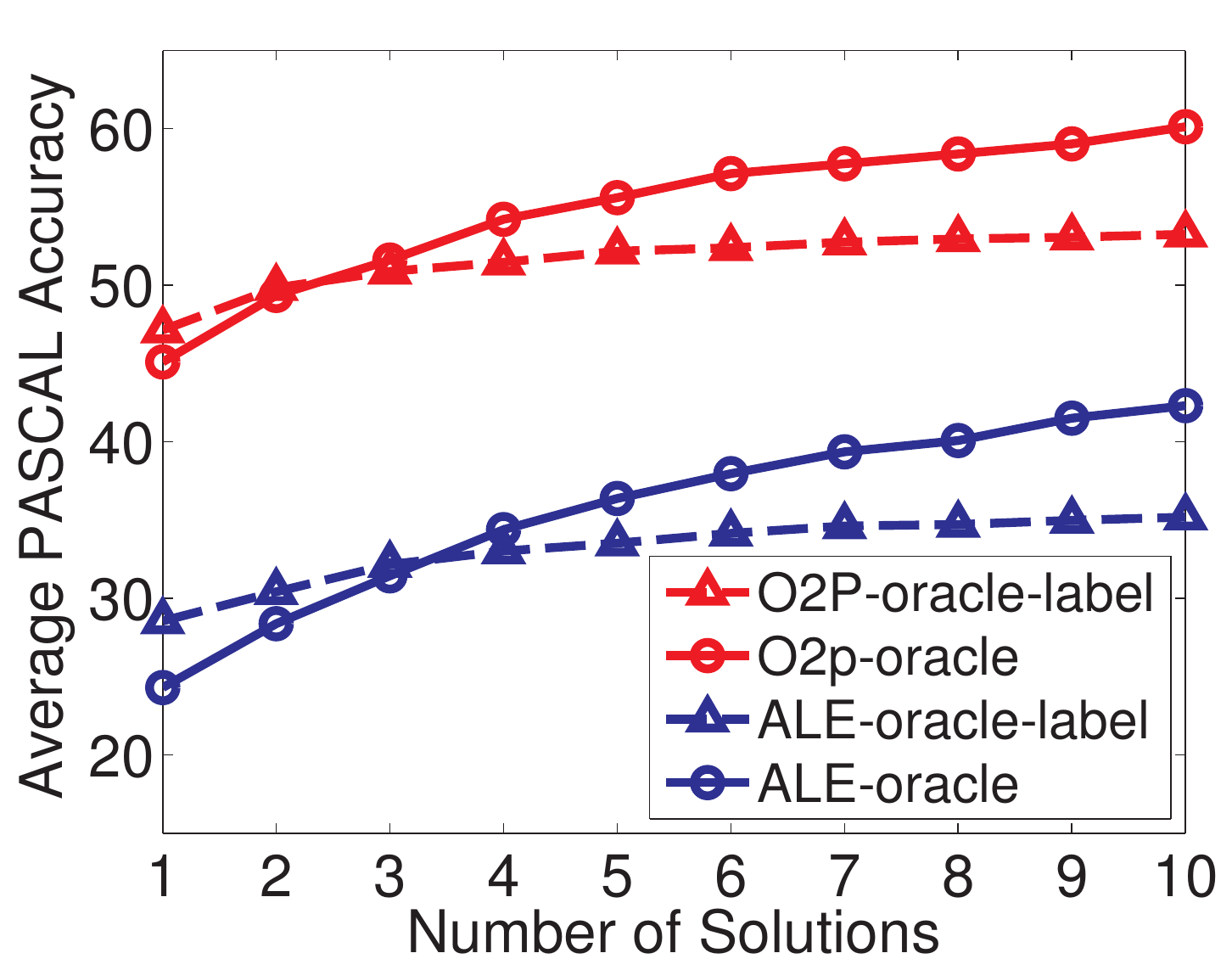}\label{fig:sol-div:b}}%&
    \hspace{1em}
    \subfloat[]{\includegraphics[width=0.5\linewidth]{\main/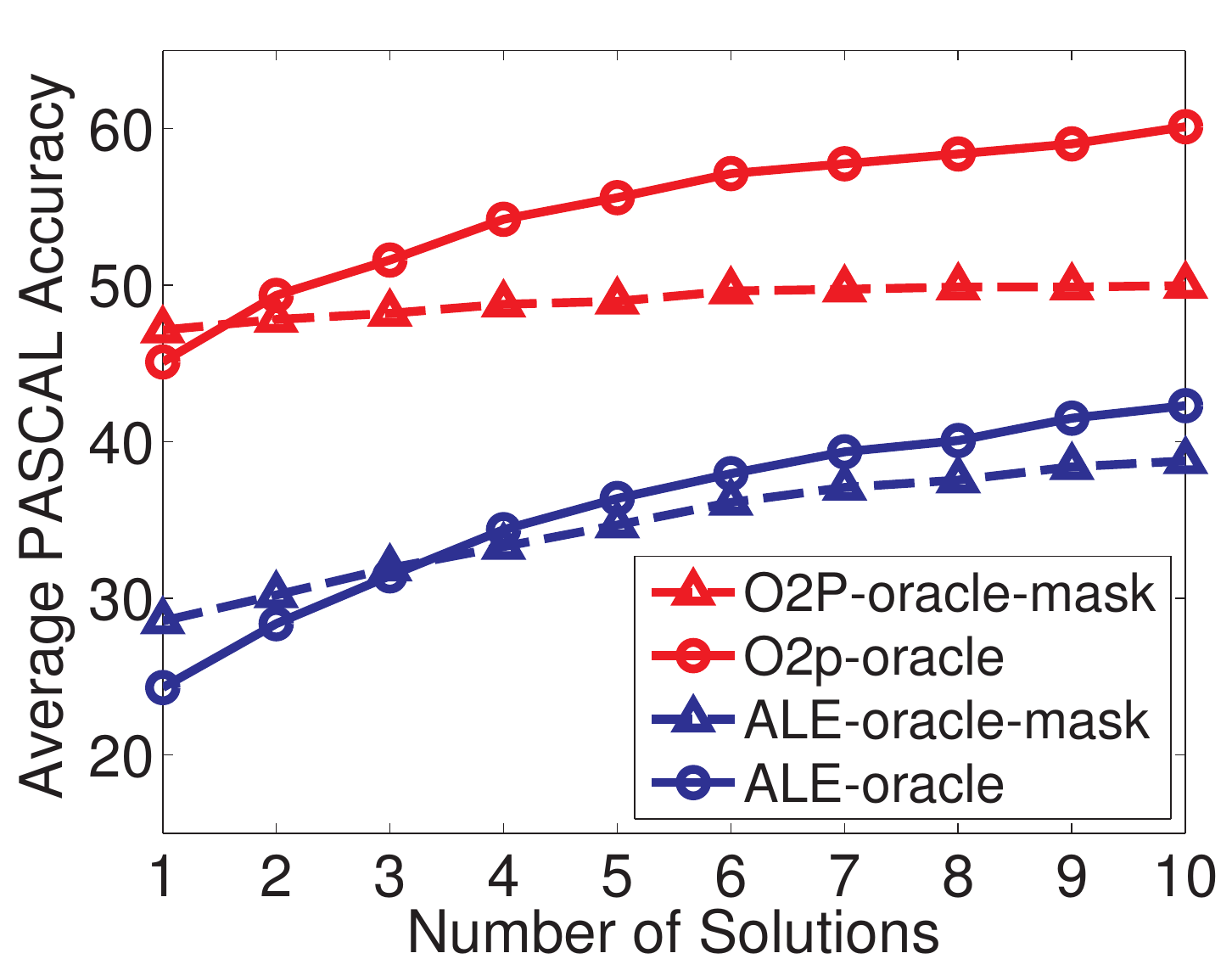}\label{fig:sol-div:c}}
%  \end{tabular}
%  \vspace{\captionReduceTop}
  \caption{\small{(a)~Average minimum-covering~\eqref{eq:mincover} of MAP in the
          first $j\leq 10$ solutions vs. $j$.
  (b) Accuracy of an oracle restricted to labels present in the MAP, or (c) restricted to masks present in MAP. See text for details.}}
  \label{fig:sol-div}%\vspace{\captionReduceBot}\vspace{0pt}
\end{figure*}

%% file: data/divrank/testaccuracies.tex
\begin{table*}[t]
\begin{center}
%{\footnotesize
{\tiny
\setlength{\tabcolsep}{1pt}
\begin{tabular*}{1.00\textwidth}{@{\extracolsep{\fill}}p{2.6cm}ccccccccccccccccccccc|c}
%\begin{tabular*}{1\textwidth}{c|ccccccccccccccccccccc|c}
\toprule
& \rotatebox{90}{Backgr.}
& \rotatebox{90}{Plane}
& \rotatebox{90}{Bicycle}
& \rotatebox{90}{Bird}
& \rotatebox{90}{Boat}
& \rotatebox{90}{Bottle}
& \rotatebox{90}{Bus}
& \rotatebox{90}{Car}
& \rotatebox{90}{Cat}
& \rotatebox{90}{Chair}
& \rotatebox{90}{Cow}
& \rotatebox{90}{D.Table}
& \rotatebox{90}{Dog}
& \rotatebox{90}{Horse}
& \rotatebox{90}{M.bike}
& \rotatebox{90}{Person}
& \rotatebox{90}{Plant}
& \rotatebox{90}{Sheep}
& \rotatebox{90}{Sofa}
& \rotatebox{90}{Train}
& \rotatebox{90}{TV.Mo.}
& \rotatebox{90}{Average}\\
\midrule
\otwop-MAP & 84.8 & 63.7 & 23.4 & 44.9 & 40.8 & 45.1 & 58.0 & 58.8 & 57.6 & 12.1 & 43.8 & 31.0 
& 44.8 & 56.2 & 56.8 & 52.3 & 37.1 & 44.0 & 29.5 & 48.6 & 42.9 & 46.5\\
%\textcolor{red}{\divrank} & \textbf{85.7} & 62.7 & \textbf{25.6} & \textbf{46.9} & \textbf{43.0} 
%& \textbf{54.8} & \textbf{58.4} & 58.6 & 55.6 & \textbf{14.6} & \textbf{47.5} & \textbf{31.2} 
%& 44.7 & 51.0 & \textbf{60.9} & \textbf{53.5} & 36.6 & \textbf{50.9} & \textbf{30.1} & \textbf{50.2} & \textbf{46.8} & \textbf{48.1}\\
\textcolor{red}{\divrank} & \textbf{85.7} & 62.7 & \textbf{25.6} & \textbf{46.9} & \textbf{43.0} 
& \textbf{54.8} & \textbf{58.4} & 58.6 & 55.6 & \textbf{14.6} & \textbf{47.5} & \textbf{31.2} 
& 44.7 & 51.0 & \textbf{60.9} & \textbf{53.5} & 36.6 & \textbf{50.9} & \textbf{30.1} & \textbf{50.2} & \textbf{46.8} & \textbf{48.1}\\
\bottomrule
\end{tabular*}
}
\end{center}
%\vspace{\captionReduceTop}\vspace{-5pt}
\caption{PASCAL VOC 2012 \test set accuracies.}
\label{table:rerankertest_accuracies}%\vspace{\captionReduceBot}\vspace{-9pt}
\end{table*}

%% file: plots/divrank/rankerstats.tex
\begin{figure*}[t]
%\vspace{\captionReduceTop}
\centering
\subfloat[Oracle Solution Rank Histogram.]
{\includegraphics[width=0.5\linewidth]{\main/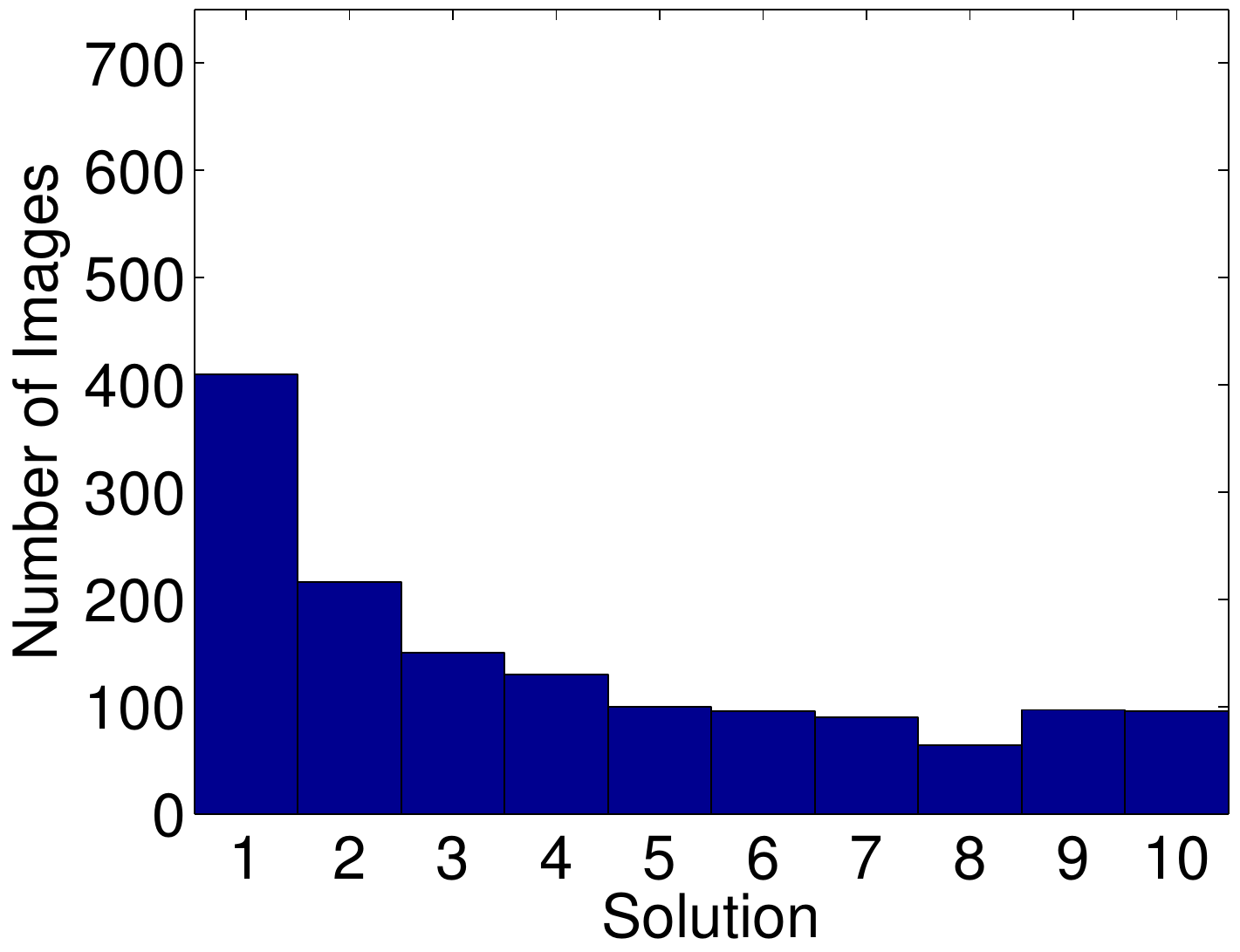}\label{rankerstats_oraclehist}}% \qquad \,
\vspace{1em}
\subfloat[Predicted Solution Rank Histogram.]
{\includegraphics[width=0.5\linewidth]{\main/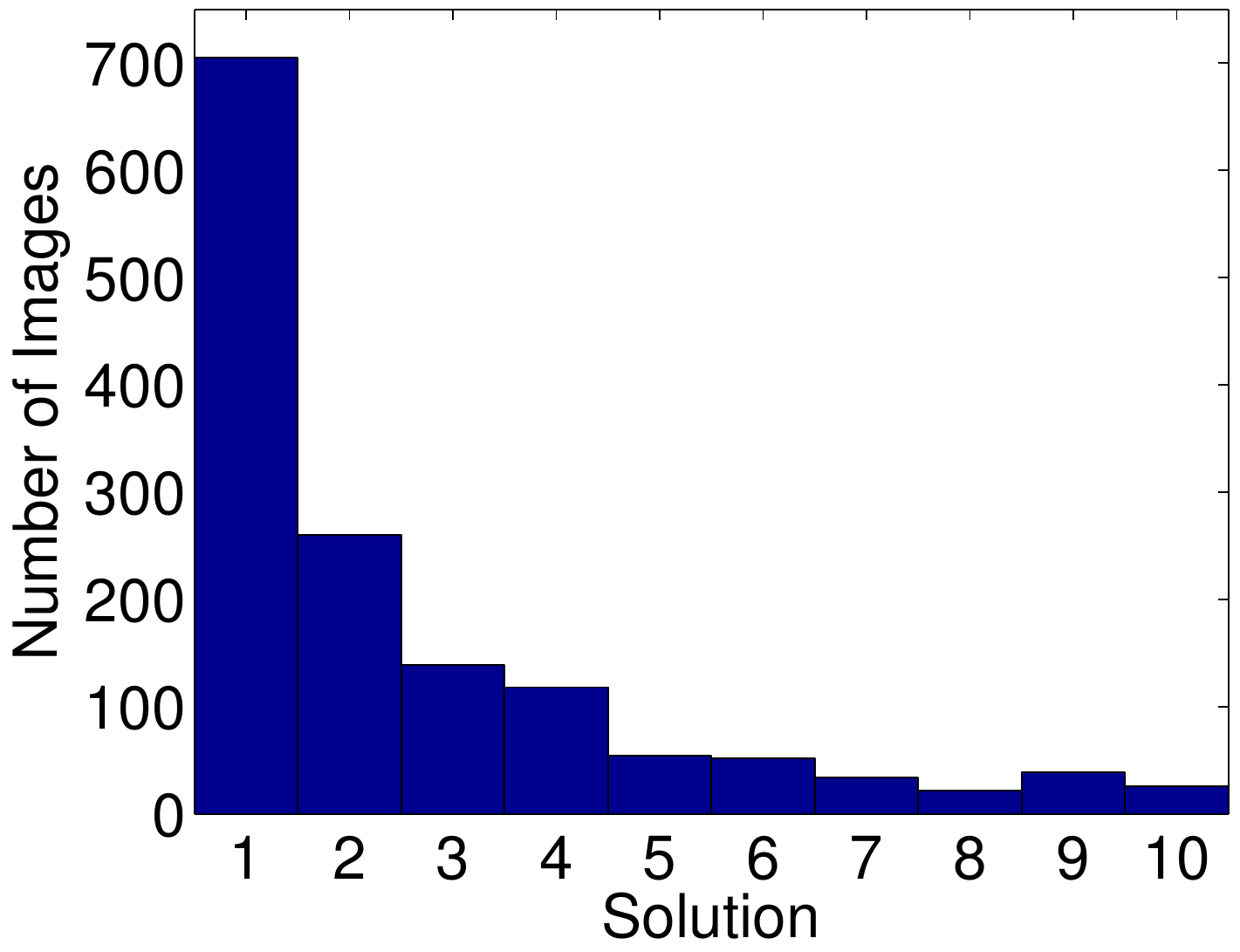}\label{rankerstats_rerankhist}}% \qquad \,
\vspace{1em}
\subfloat[\Reranker Score vs Solution Accuracy.]
{\includegraphics[width=0.5\linewidth]{\main/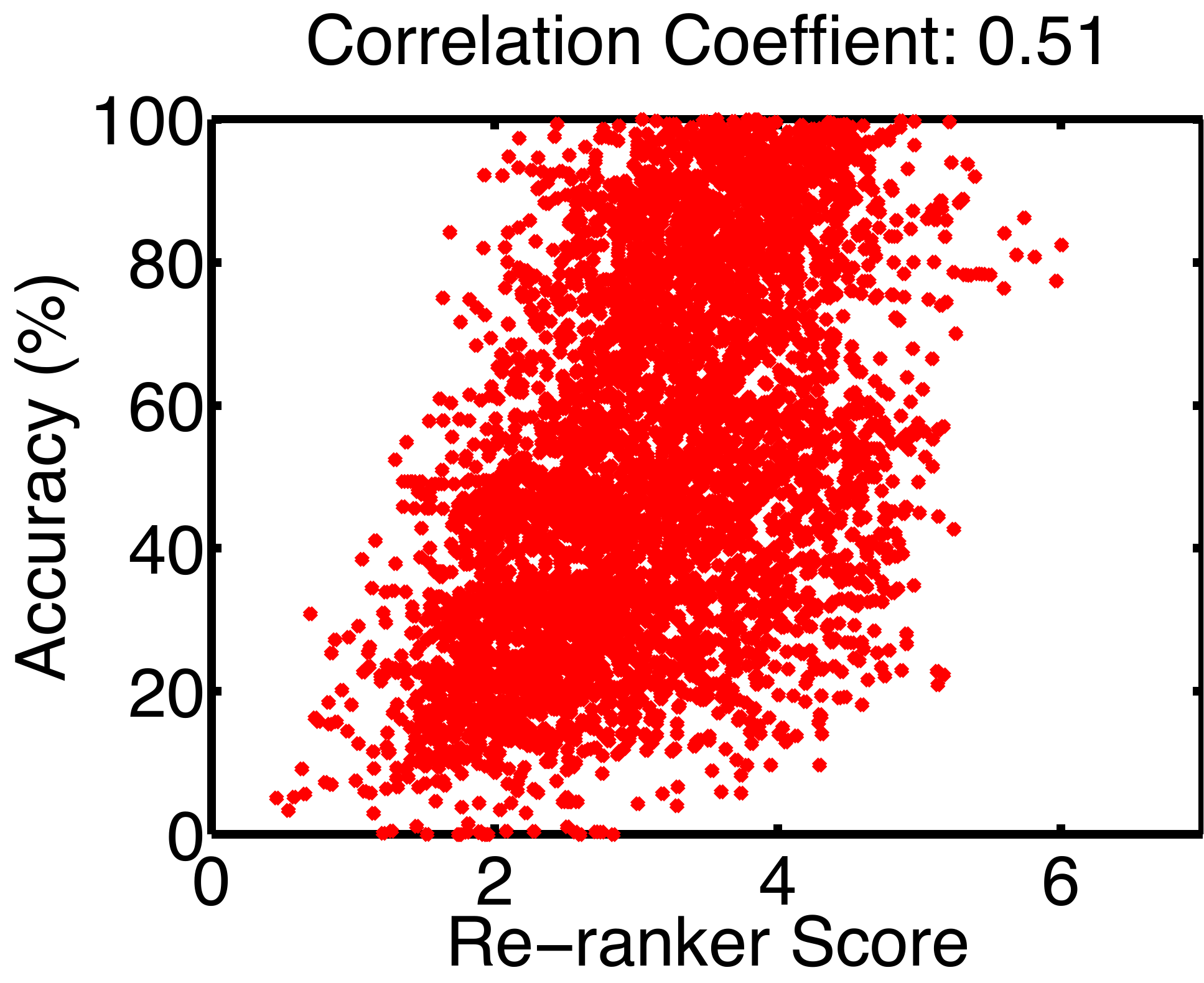}\label{rankerstats_corr}}
\vspace{-2pt}
\caption{Statistics on PASCAL VOC 2012 \val with \otwop model: (a),(b) show the number of images in which the 
    \emph{oracle} / top-\reranked solution was originally at rank $j\leq 10$. 
We can see that there is a heavy tail in the oracle distribution, but a much lighter tail in the \reranker, 
suggesting that the \reranker ``plays it safe'' and predicts MAP very frequently;
(c) shows a scatter plot of \reranker score vs solution accuracy.}
  \label{fig:rankerstats}%
%\vspace{\captionReduceBot}
\end{figure*}

%% file: gfx/divrank/rankerbtmap/rankerbtmap.tex
\begin{figure}[t!]
\centering
{\includegraphics[width=.8\linewidth]{\main/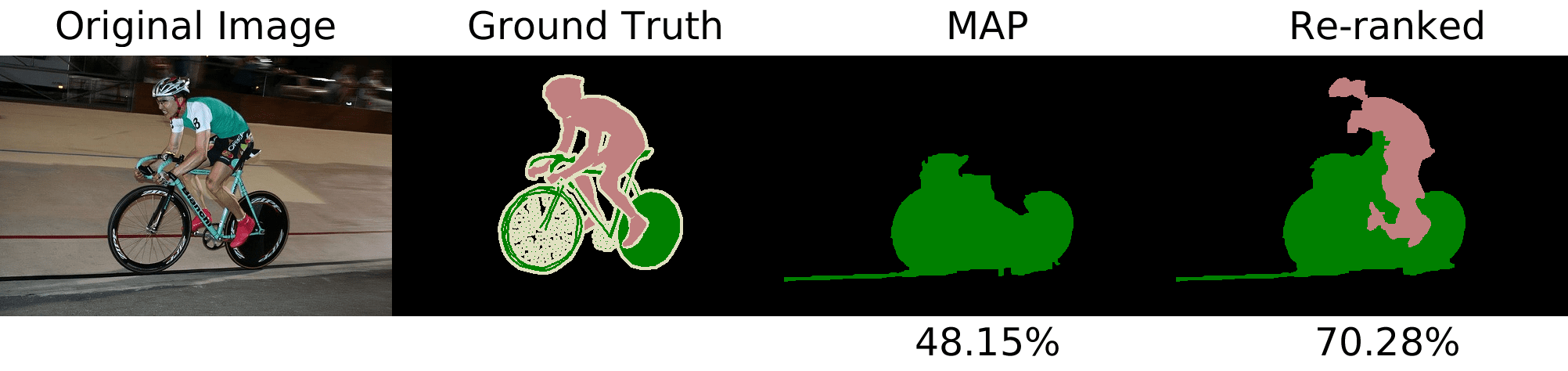}}\\\vspace{5pt}
{\includegraphics[width=.8\linewidth]{\main/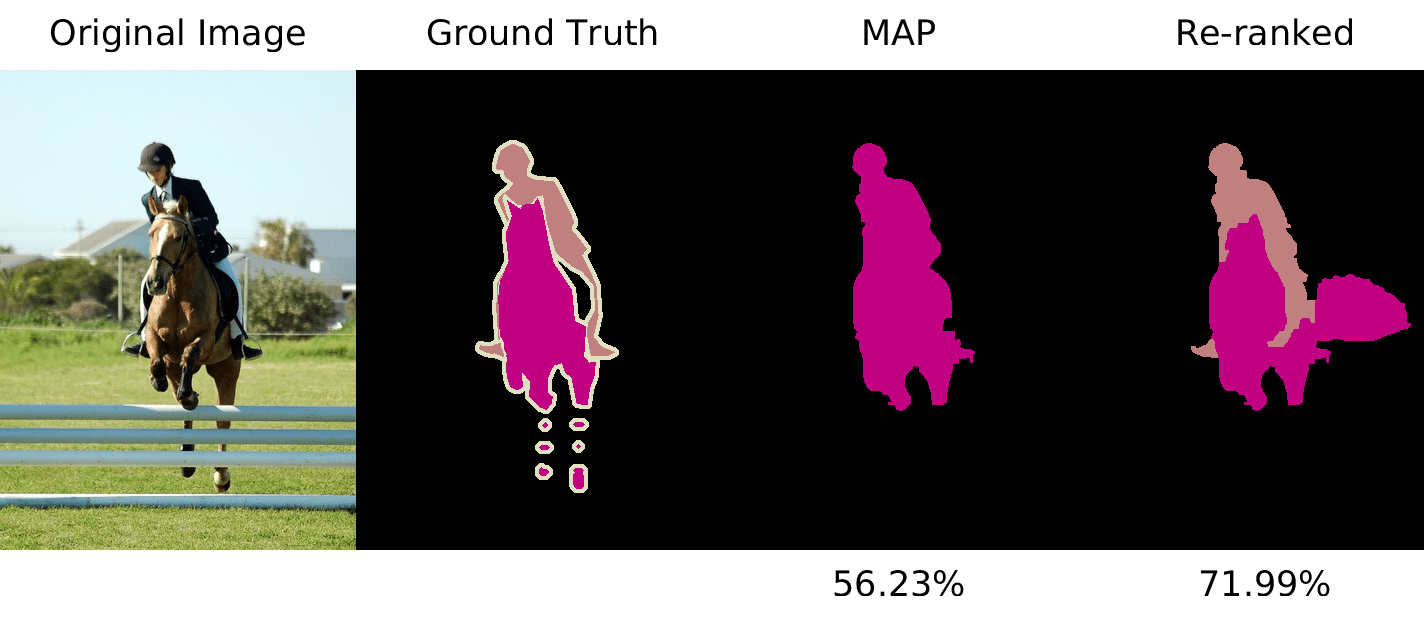}}\\\vspace{5pt}
{\includegraphics[width=.8\linewidth]{\main/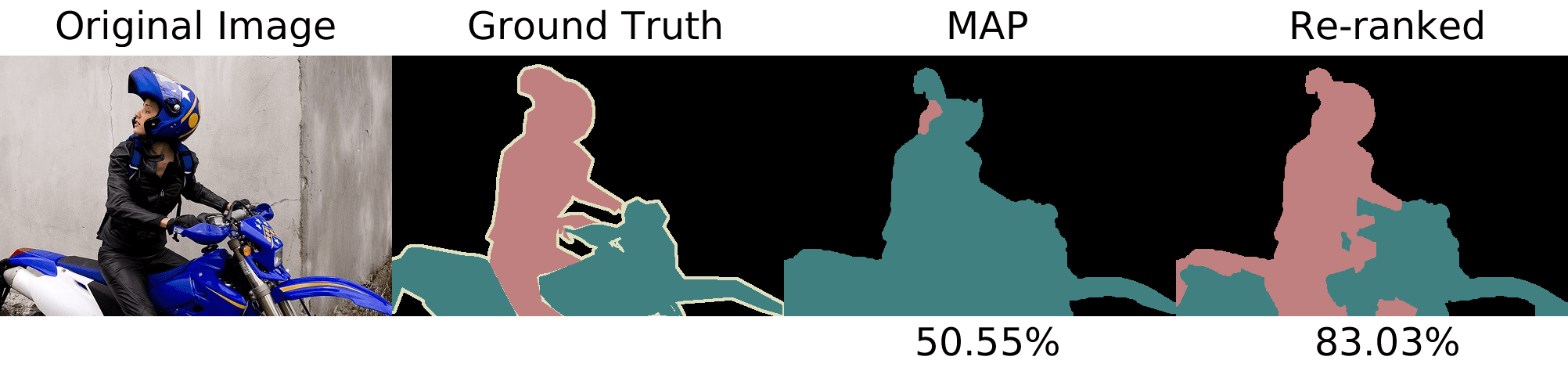}}\\\vspace{5pt}
{\includegraphics[width=.8\linewidth]{\main/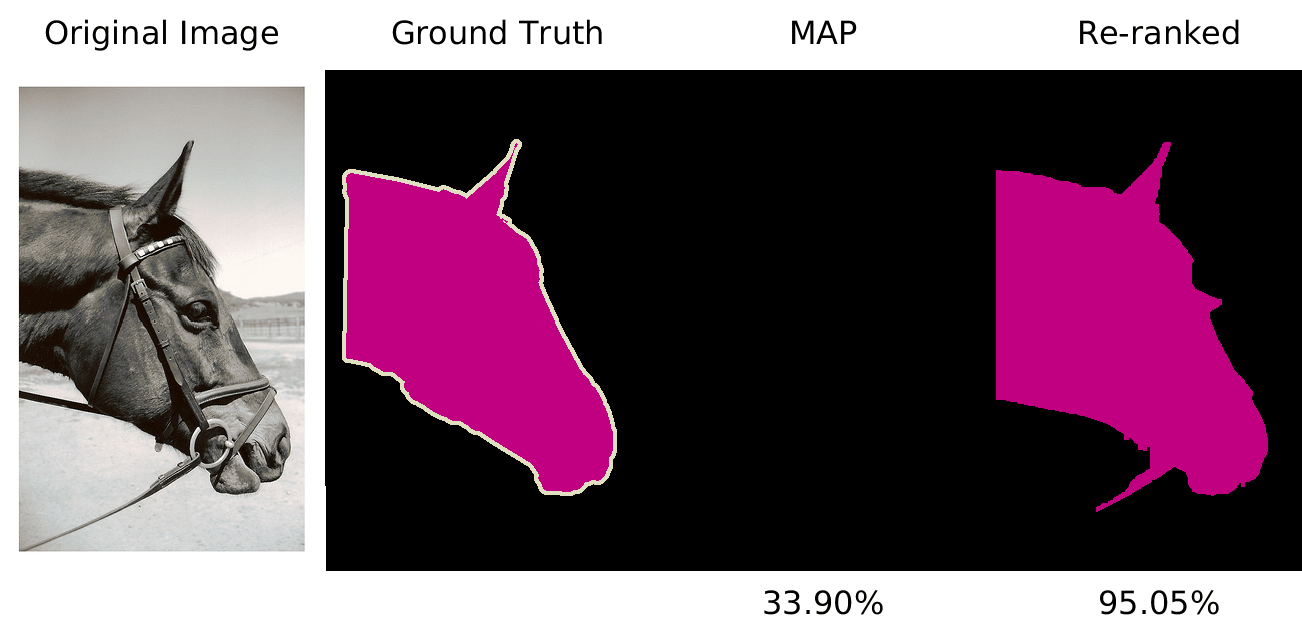}}\\\vspace{5pt}
\caption{\small{Cases where \otwop-\divrank outperforms \otwop-MAP. In each group of images, the first column shows the original image followed by the ground-truth, MAP, and top \reranked solution returned by \divrank. PASCAL intersection-over-union accuracy is shown below the segmentations.}}
  \label{fig:rankerbtmap}
  %\vspace{\captionReduceBot}
\end{figure}

%% file: gfx/humanexp/humanexp.tex
\begin{figure*}[p!]
\centering
\subfloat[]
	{\includegraphics[width=0.8\linewidth]{\main/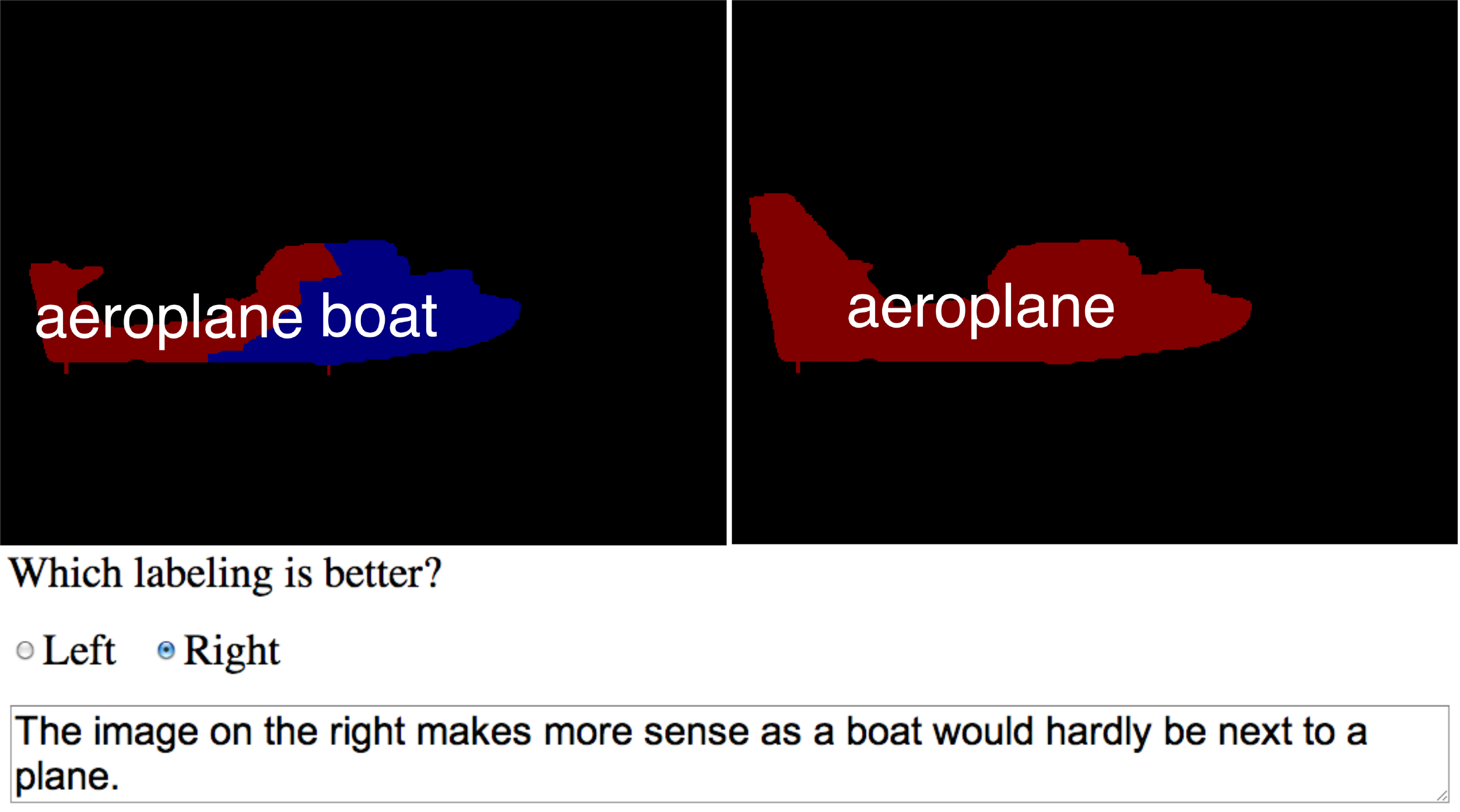}\label{fig:humanexp:a}} 
\vspace{2em}
\subfloat[]
	{\includegraphics[width=0.8\linewidth]{\main/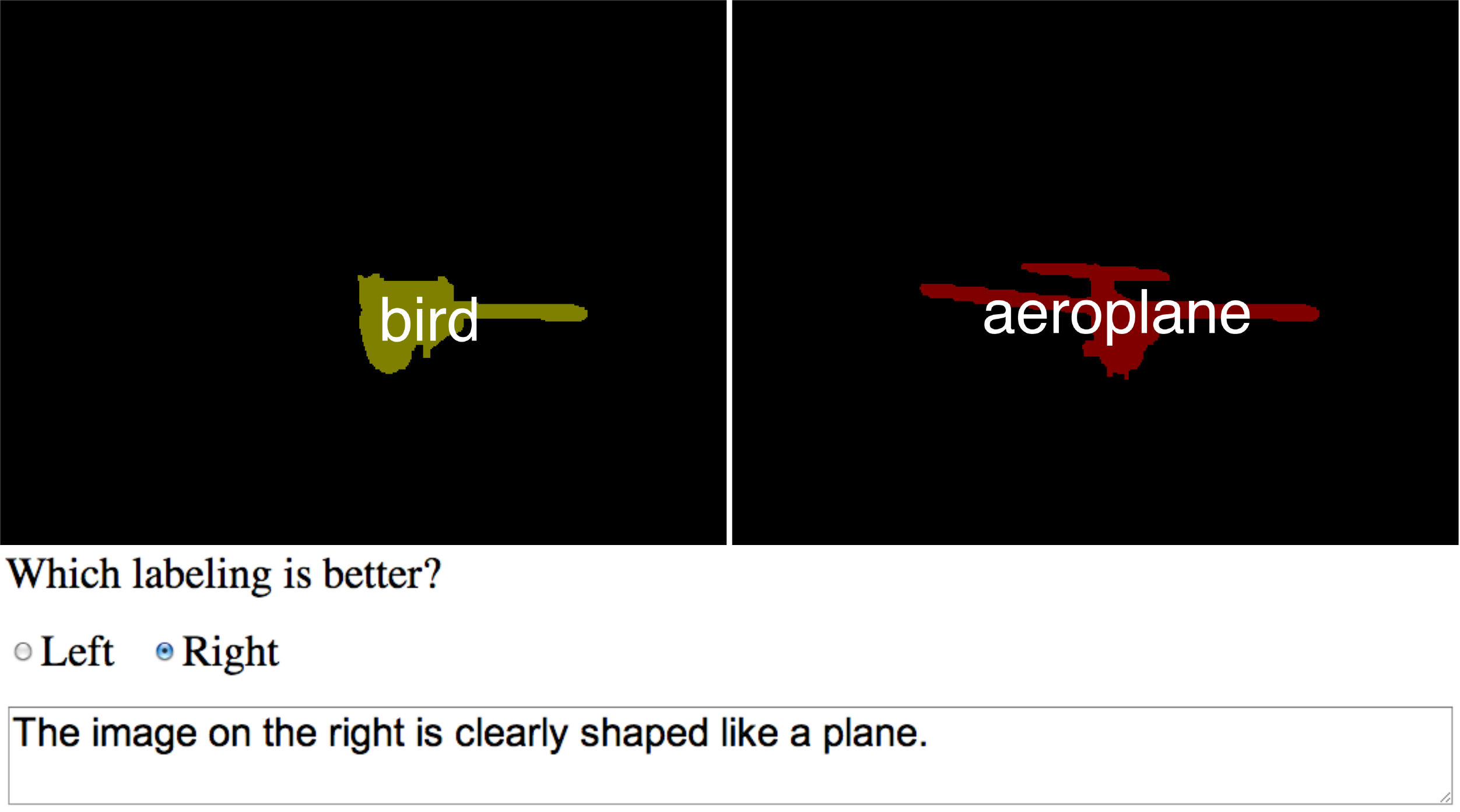}\label{fig:humanexp:b}}
\vspace{2em}
\subfloat[]
	{\includegraphics[width=0.8\linewidth]{\main/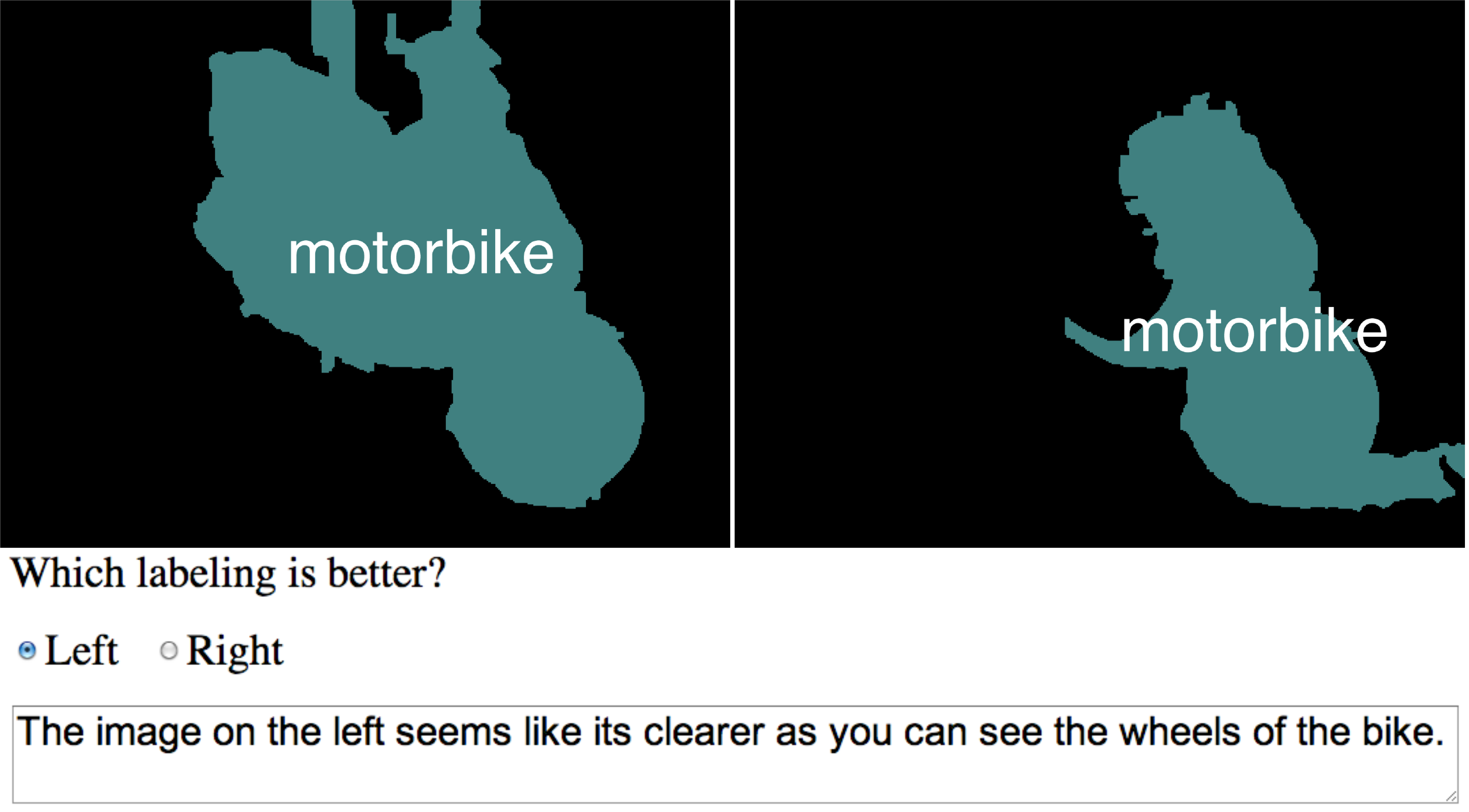}\label{fig:humanexp:c}}
%\vspace{\captionReduceTop}
\caption{\small{Example MTurk tasks along with user-provided responses which were instructive in the creation of segmentation-specific features.}}
\label{fig:humanexp}%\vspace{\captionReduceBot}\vspace{-3pt}
\end{figure*}

%% file: data/humanexp/humanexp.tex
\begin{table}[t]
{\footnotesize
\setlength{\tabcolsep}{1pt}
\begin{tabular*}{\columnwidth}{@{\extracolsep{\fill}}p{1cm}ccc@{\quad}|ccccc}
&\multicolumn{3}{c}{\textbf{Binary Task Accuracies}}&\multicolumn{4}{c}{\textbf{Pascal VOC Avg. Acc.}}\\ [.5ex]
\midrule
%\midrule
&B-vs-W
&M-vs-W
&B-vs-M
&\emph{Best}
&MAP
&\emph{Worst}
&HR
\\
\midrule
\textbf{ALE} & 71.9 &  64.4 &  61.7  &  38.0  &  19.1  &  3.2  &  20.5\\
\botwop  &  73.9  &  73.1  &  56.3  &  62.8  &  43.6  &  24.5  &  49.0\\
\bottomrule
\end{tabular*}
}
\caption{\small{(\emph{left})~Human accuracy in predicting
\emph{(B)est}-vs-\emph{(W)orst}, (M)AP-vs-\emph{(W)orst}, and \emph{(B)est}-vs-(M)AP solutions. (\emph{right})~Pascal VOC accuracies over 150 images for best, MAP, worst, and human response (HR) solutions.}}
%\vspace{\captionReduceBot}%\vspace{-5pt}
\label{table:humanexp}
\end{table}

%% file: Chapters/Chapter06.tex
\chapter{Conclusion}\label{ch:conclusion}
%************************************************
\subfile{\main/Sections/Conclusion.tex}

\printbibliography

%% file: Sections/Conclusion.tex
%\section{Conclusions}
In summary, this thesis presents an approach to obtaining performance gains from a semantic segmentation model. Instead of achieving gains by opting for a more complex model which would be more expensive or possibly intractable to optimize over the gains are achieved through a careful redesign of the inference procedure that leverages diversity between output labellings. The thesis also outlines an approach to ranking these \divmbest segmentations to automatically pick the best from the set. It contains the following contributions:
\begin{itemize}
        \item A framework (\divmbest) for inferring multiple highly probable yet
            diverse segmentations from a probabilistic structured output model.
            It is motivated and derived from the integer programming problem for
            solving inference on probabilistic graphical models with discrete
            output space. The \divmbest framework is generally applicable in any
            setting where you have an inference model over a structured output
            label distribution. What results is an elegant and practical
            iterative algorithm for inference that is more akin to finding
            \emph{modes} of the output space distribution. It reuses the inference procedure from the original model, providing means for improved prediction without incurring any cost in tractability.
        \item The \divmbest formulation can accept different measures of diversity between
            ``modes``, such as Hamming and cardinality distance. Finding
            \emph{modes} of a CRF under Hamming dissimilarity amounts to only modifying
            the unary energy terms and reusing the same MAP inference machinery
            to compute subsequent solutions, yielding an approach that is as
            efficient as the underlying MAP inference algorithm.
        \item A discriminative large margin approach to ranking the \divmbest
            segmentations (\divrank) is introduced that allows for arbitrarily complex
            features to evaluate each segmentation.
        \item Evaluation of the \divmbest and \divrank algorithms on a number of
            semantic image labelling problems including interactive,
            figure-ground, and multi-category segmentation. The results provide
            evidence of the benefits these approaches offer for the segmentation task.
        \item Oracle experiments on semantic segmentation show that the
            \divmbest approach has the potential to achieve results comparable
            or better than even existing state-of-the-art CNN models for segmentation.
        \item Application of the \divmbest framework to present CNN+dense CRF
            models for segmentation which shows that the \divmbest and \divrank
            algorithms are very much relevant to the current trends in semantic segmentation.
\end{itemize}

%% file: Chapters/Chapter04A.tex
%\providecommand{\main}{..}
%\documentclass[\main/ClassicThesis.tex]{subfiles}
%\begin{document}
\chapter{Appendix A}\label{ch:ch4app}
%************************************************
\section{Sample of DivMBest solutions from figure-ground model}\label{sec:divmbestfgsample}

For overall document size considerations the results in this appendix section have been moved to~\url{http://ttic.uchicago.edu/~pyadolla/papers/thesis.pdf}.
\section{Sample of results when DivMBest is applied to muli-category
segmentation}\label{sec:divmbestsample}

For overall document size considerations the results in this appendix section have been moved to~\url{http://ttic.uchicago.edu/~pyadolla/papers/thesis.pdf}.

%% file: Chapters/Chapter05A.tex
\chapter{Appendix B}\label{ch:ch5app}
%\subfile
\input{\main/gfx/divrank/examples/ranker_examples}
\pagebreak
%\subfile
\input{\main/gfx/divrank/rankerbtmap/rankerbtmapsamples}

%% file: gfx/divrank/examples/ranker_examples.tex
%\providecommand{\main}{../../..}
%\documentclass[\main/ClassicThesis.tex]{subfiles}
%\begin{document}
%%%%%%%%%%%%%%%%%%%%%%%%%%%%%%%%%%%%%%%%%%%%%%%%%%%%%%%%%%%
\section{Example \Reranking Results}
\label{sec:divrankexamples}
%%%%%%%%%%%%%%%%%%%%%%%%%%%%%%%%%%%%%%%%%%%%%%%%%%%%%%%%%%%

For overall document size considerations the results in this appendix section have been moved to~\url{http://ttic.uchicago.edu/~pyadolla/papers/thesis.pdf}.

%% file: gfx/divrank/rankerbtmap/rankerbtmapsamples.tex
%\providecommand{\main}{../../..}
%\documentclass[\main/ClassicThesis.tex]{subfiles}
%\begin{document}
\section{Highest ranked vs. MAP}\label{sec:rankerbtmap}
\label{sec:rankerbtmapsamples}\vspace{-2em}

For overall document size considerations the results in this appendix section have been moved to~\url{http://ttic.uchicago.edu/~pyadolla/papers/thesis.pdf}.